%% file: levy.tex
\documentclass{article}

\usepackage[utf8]{inputenc}
\usepackage[T1]{fontenc}
\usepackage{lmodern}
\usepackage{graphicx}
\usepackage[figurename=Fig.,labelfont=bf,labelsep=period]{caption}
\usepackage{subcaption}
\usepackage{amsmath}
\usepackage{amsfonts}
\usepackage{amssymb}
\usepackage{amsbsy}
\usepackage[colorlinks=true,citecolor=red,linkcolor=black]{hyperref}
\usepackage[margin=1in]{geometry}

\usepackage{todonotes}
\usepackage{floatrow}

\usepackage{pgfplots,algorithmic,algorithm}
\usepackage[toc,page]{appendix}
\usepackage{float}
\usepackage{booktabs}
\usepackage{bm}
\newtheorem{theorem}{Theorem}

\usepackage[parfill]{parskip}

\newcommand{\RR}[0]{\mathbb{R}}

\newcommand{\bx}{\mathbf{x}}
\newcommand{\ii}{\mathrm{i}}
\newcommand{\bxi}{\bm{\xi}}

\newcommand{\bb}{\mathbf{b}}
\newcommand{\bA}{\mathbf{A}}

\newcommand{\bX}{\mathbf{X}}

\newcommand{\bs}{\mathbf{s}}

\newcommand{\bt}[0]{\bm{\theta}}

\usepackage[utf8]{inputenc}
\usepackage{cleveref}
\usepgfplotslibrary{groupplots}
\pgfplotsset{compat=newest}

\usepackage{tgtermes,tgheros,tgcursor}
\usepackage[T1]{fontenc}
\usepackage{indentfirst}
\date{}
\usepackage{sectsty,textcase}
\usepackage{authblk}

\begin{document}

\title{Calibrating Multivariate L\'evy Processes with Neural Networks}

\author[1]{Kailai Xu}
\author[1,2]{Eric Darve}

\affil[1]{Institute for Computational and Mathematical Engineering, Stanford University, Stanford, CA, 94305}
\affil[2]{Mechanical Engineering, Stanford University, Stanford, CA, 94305}

\maketitle


\begin{abstract}
Calibrating a L\'evy process usually requires characterizing its jump distribution. Traditionally this problem can be solved with nonparametric estimation using the empirical characteristic functions~(ECF), assuming certain regularity, and results to date are mostly in 1D. For multivariate L\'evy processes and less smooth L\'evy densities, the problem becomes challenging as ECFs decay slowly and have large uncertainty because of limited observations. We solve this problem by approximating the L\'evy density with a parametrized functional form; the characteristic function is then estimated using numerical integration. In our benchmarks, we used deep neural networks and found that they are robust and can capture sharp transitions in the L\'evy density. They perform favorably compared to piecewise linear functions and radial basis functions. The methods and techniques developed here apply to many other problems that involve nonparametric estimation of functions embedded in a system model. 
\end{abstract}

\section{Introduction}

L\'evy processes generalize the Gaussian processes by allowing the jump-diffusion. Because of their ability to allow continuous evolution and abrupt jumps of random variables~\cite{chen2010nonparametric}, many models in finance, physics or biology have been built based on L\'evy processes. For example, in the classical Black-Scholes model for risky assets, the price $S_t$ of an asset at time $t$ is governed by~\cite{figueroa2004nonparametric}
\begin{equation}
    S_t = S_0e^{\sigma B_t +\mu t}
\end{equation}
where $B_t$ is the standard Brownian motion, $\sigma$ and $\mu$ are the standard deviation and the drift mean. To account for the excessive skewness and kurtosis in the log return distributions in empirical financial data, the model has been generalized to the exponential L\'evy process 
\begin{equation}
    S_t = S_0e^{X_t}
\end{equation}
where $X_t$ is a L\'evy process. Yet because of the lack of analytical closed-form density functions for general L\'evy processes, an exact maximum likelihood estimator is not feasible. This leads to the difficulty of calibrating L\'evy processes in the presence of jumps. 

The multivariate L\'evy process can be described by three parameters~\cite{menn2006calibrated}: a positive semi-definite matrix $\bA = \bm{\Sigma}\bm{\Sigma}^T \in \RR^{d\times d}$, where $\bm{\Sigma}\in \RR^{d\times d}$, a vector $\bb\in \RR^d$ and a measure $\nu\in \RR^d\backslash\{\mathbf{0}\}$. The L\'evy process $\bX_t$ is a superposition of a Wiener process $\bm{\Sigma} \mathbf{B}_t + \bb t$, where $\mathbf{B}_t$ is the standard i.i.d. Brownian motion, and a pure-jump L\'evy process with the L\'evy measure
\begin{equation}
    \nu(A) = \frac{1}{t}\mathbb{E}\left( \sum_{s\leq t} \mathbf{1}_{A}(\mathbf{X}_{s}-\bX_{s-}) \right)
\end{equation}
where $\mathbf{1}_A$ is an indicator for $A$, i.e., $\mathbf{1}_A(\bx)=1$ for $\bx\in A$ and 0 otherwise. The corresponding characteristic function is given by the L\'evy-Khintchine representation~\cite{papapantoleon2008introduction}
\begin{equation}\label{equ:CF}
        \phi(\bm{\bxi}) = \mathbb{E}[e^{\ii \langle \bxi, \bX_t \rangle}] =\exp\left[t\left( \ii \langle \bb, \bxi \rangle - \frac{1}{2}\langle \bxi, \bA\bxi\rangle  +\int_{\RR^d} \left( e^{\ii \langle \bxi, \bx\rangle} - 1 - \ii \langle \bxi, \bx\rangle \mathbf{1}_{\|\bx\|\leq 1}\right)\nu(d\bx)\right) \right]
    \end{equation}
The subject of this paper is to study the nonparametric calibration of pure jump processes $\nu(\bx)$ and thus we assume $\bA=\mathbf{0}$ and $\bb=\mathbf{0}$ throughout the paper. In addition, we assume that $\nu$ is determined by a density function such that $\nu(d\bs)=\nu(\bs)d\bs$, and we call $\nu(\bs)$ the \textit{L\'evy density}. 

The traditional nonparametric estimation for L\'evy processes in 1D has two regimes~\cite{neumann2009nonparametric}: (1) the L\'evy process $X_t$ is observed at high frequency at times $t_i$, i.e., $\max_i (t_i-t_{i-1})$ is small. In this case, a large increment $X_{t_i}-X_{t_{i-1}}$ indicates that a jump occurred. For example \cite{comte2009nonparametric} proposed nonparametric inference methods for L\'evy process in this case. (2) in the low-frequency observation regime, there are zero or several jumps present within the increment $X_{t_i}-X_{t_{i-1}}$. In this case, \cite{neumann2009nonparametric} applied a deconvolution algorithm to estimate $\nu(x)$ from the empirical characteristic function. \cite{cont2004nonparametric} discretized the L\'evy density $\nu(x)$ on a grid and  applied relative entropy minimization to find the optimal $\nu(x)$. We consider the latter regime and assume that the data are given at equispaced time intervals, i.e., $t_1 = {\Delta t}$, $t_2 ={2\Delta t}$, $t_3 = {3\Delta t}$, $\ldots$

However, much of the attention in the literature has been restricted to 1D case and well-behaved $\nu(x)$. In the case where $\nu(x)$ is discontinuous, the decay of the characteristic function is very slow so accurate deconvolution in \cite{neumann2009nonparametric} requires large computational domains. Besides, \cite{cont2004nonparametric} assigned one degree of freedom~(DOF) to the discretized L\'evy density $\nu(x)$ per grid point, which partially contributed to the ill-posedness of the nonlinear optimization problem. The ill-posedness problem becomes more severe in higher dimensions since DOFs grow exponentially.

In this paper, we tackle those challenges by proposing a novel approach for 2D nonparametric estimation of the L\'evy density $\nu(x)$. This approach proceeds in four stages: 
\begin{enumerate}
    \item The L\'evy density is approximated by a parametric functional form---such as piecewise linear functions---with parameters $\bt$,
\begin{equation}\label{equ:1}
    \nu(\bx) \approx \nu_{\bt}(\bx)
\end{equation}

\item The characteristic function is approximated by numerical integration 
\begin{equation}\label{equ:2}
    \phi(\bxi)\approx    \phi_{\bt}(\bxi) := \exp\left[ \Delta t \sum_{i=1}^{n_q} \left(e^{\ii \langle\bxi, \bx_i \rangle}-1-\ii\langle\bxi, \bx_i \rangle\mathbf{1}_{\|\bx_i\|\leq 1}  \right)\nu_{\bt}(\bx_i) w_i \right]
\end{equation}
where $\{(\bx_i, w_i)\}_{i=1}^{n_q}$ are quadrature nodes and weights.

\item The empirical characteristic functions are computed given observations $\{\bX_{i\Delta t}\}_{i=0}^n$
\begin{equation}\label{equ:ECF}
    \hat\phi_n(\bxi) := \frac{1}{n}\sum_{i=1}^n \exp(\ii\langle \bxi, \bX_{i\Delta t}-\bX_{(i-1)\Delta t}\rangle ),\  \bxi \in \RR^d
\end{equation}
\item Solve the following optimization problem with a gradient based method. Here $\{\bxi_i \}_{i=1}^m$ are collocation points depending on the data. 
\begin{equation}\label{equ:4}
    \min_{\bt}\frac{1}{m} \sum_{i=1}^m \|\hat\phi_n(\bxi_i)-\phi_{\bt}(\bxi_i)  \|^2
\end{equation}
\end{enumerate}

One challenge for this approach is the error $\|\hat\phi_n(\bxi)-\phi(\bxi)\|$ in computing the empirical characteristic function. In theory, the empirical characteristic function converges to the exact one given infinite observations. However, in practice the observations are limited and thus the empirical characteristic function is not exact. Another challenge is the discontinuity of L\'evy densities. This occurs when the jump distribution experiences sudden changes in some domains. 

The choice of approximation functional form $\nu_{\bt}(\bx)$ is essential. From the previous discussion, a potential form must have the following properties: (1) universal approximation, i.e., the capability of approximating any continuous functions given sufficient computing budget; (2) robustness to noise; (3) ability to handle discontinuity. In this paper, we apply and benchmark three popular parametric functional forms: neural networks~(NN), piecewise linear functions~(PL) and radial basis functions~(RBF).

\begin{figure}[htpb]
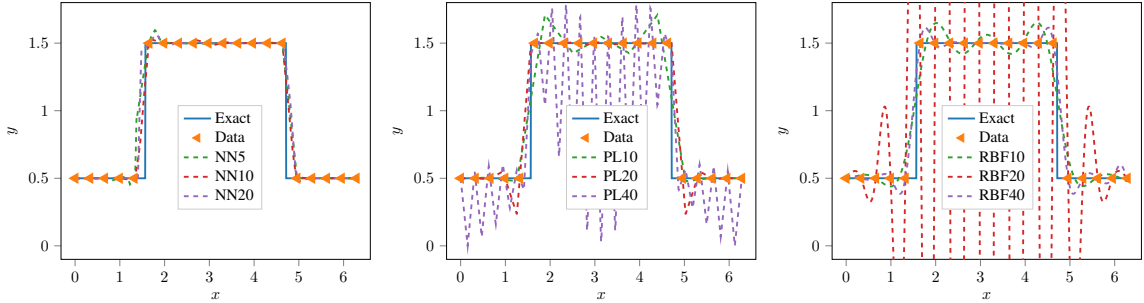

\centering
\scalebox{0.6}{\input{figures/compareNN.tex}}~
\scalebox{0.6}{\input{figures/comparePL.tex}}~
\scalebox{0.6}{\input{figures/compareRBF.tex}}
\caption{Training with 20 sample points from a step function. NN$x$ stands for neural network model with $x$ layers, $20$ neurons per hidden layer and ReLU as activation function. PL$x$ stands for piecewise linear function with $x$ equispaced distributed nodes. RBF$x$ stands for radial basis functions with $x$ equispaced distributed centers.}
\label{fig:comparenn}
\end{figure}

The neural network enjoys many favorable properties and we demonstrate empirically that it outperforms the others in several situations. On the one hand, PL consists of local basis functions and therefore DOFs with no data points nearby around are not optimized. On the other hand, although the basis functions in RBF are global such that it suffers less from the problem PL struggles with, it is well known that RBF is susceptible to noise and discontinuity. Besides, the choice of centers and shape parameters can be tricky. However, the problems are alleviated for NN, partially because it is adaptive to non-uniform data~\cite{huang2019predictive}, robust to noise and can overcome Gibbs phenomenon~\cite{xu2019neural}. This is also demonstrated in \Cref{fig:comparenn}, where the basis functions are trained on 20 data points in a step function. NN honors the sharp transitions and does not oscillate as severely as others. 


With the re-parametrization technique, we show that the method can also be applied to multivariate symmetric $\alpha$-stable processes~\cite{gulian2018machine}, a subclass of L\'evy processes. In this case, $\nu(\bx)$ is singular at $\bx=\mathbf{0}$, but we can re-parametrize the characteristic function as 
\begin{equation}\label{equ:problem2}
\phi(\bxi) = \mathbb{E}\left( \exp(\ii t \langle\bt, \bxi\rangle )\right) = \exp\left[ t\left(- \frac{1}{2}\langle \bxi, \bA\bxi\rangle + \ii \langle \bxi, \bb\rangle - \int_{\mathbb{S}^d} | \langle\bt, \bxi\rangle|^\alpha\Gamma(\bs)d\bs    \right) \right]
\end{equation}
where $\Gamma(\bs)$ is a function defined on $\mathbb{S}^d$. Here we can substitute $\Gamma(\bs)$ by a parametrized functional form $\Gamma_{\bt}(\bs)$, apply the quadrature rule on the unit circle and minimize the discrepancy between $\hat\phi_n(\bxi)$ and $\phi_{\bt}(\bxi)$.

Finally, we built a toolset \texttt{LevyNN} for calibrating L\'evy processes based on the open source library \texttt{ADCME.jl}. The latter is an automatic differentiation library with \texttt{TensorFlow} and \texttt{PyTorch} backends and is specially designed for scientific computing.
 The library automates the gradient computation and integrates the optimization workflow.

\section{ Nonparametric Estimation of the L\'evy processes}

\subsection{Characteristic Function Matching Method}

The characteristic function matching method~\cite{yu2004empirical} minimizes the discrepancy between the empirical characteristic function $\hat\phi_n(\bxi)$ (\Cref{equ:ECF}) and the characteristic function $\phi(\bxi)$ (\Cref{equ:CF}). The rationales are: (1) As $n\rightarrow \infty$, $\hat\phi_n(\bxi)\rightarrow \phi(\bxi)$ because of the large number law; (2) there is a one-to-one correspondence between the characteristic function $\phi(\bx)$ and the density function for $\bX_{i\Delta t}-\bX_{(i-1)\Delta t}$, a.k.a., $\nu(\bx)$. Consequently, we can estimate $\nu(\bx)$ from $\hat\phi_n(\bxi)$.  


\subsection{ Approximation to the L\'evy Density}

\begin{figure}[hbt]
  \includegraphics[width=0.8\textwidth]{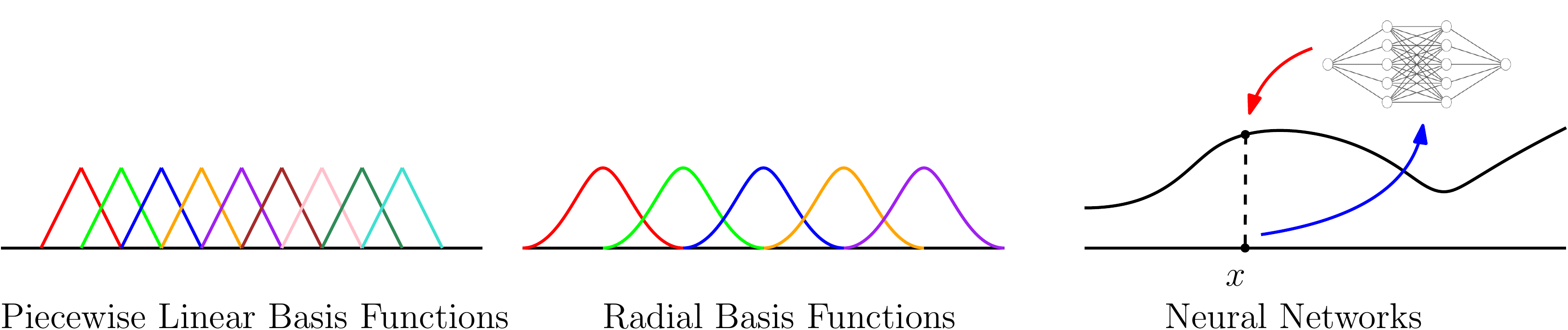}
  \caption{Different functional forms in 1D. In PL and RBF, the target function are approximated by linear combination of basis functions; in NN, it is approximated by composing linear transformations and nonlinear activation functions.}
  \label{fig:compare basis}
\end{figure}

The L\'evy density $\nu(\bx)$ is a mapping from the coordinates $\bx\in \RR^2$ to $\RR$. We first truncate the infinite computational domain to $\bx\in [-M,M]^2$ and then approximate $\nu(\bx)$ with $\nu_{\bt}(\bx)$. In the following we discuss three functional forms for $\nu_{\bt}(\bx)$ (\Cref{fig:compare basis}). 

One type of neural networks~(NN) is a composition of linear operations followed by a nonlinear activation function. In this paper, we consider ReLU dense neural networks, where 
\begin{equation}
    \nu_{\bt}(\bx) = \mathbf{W}_L\mathrm{ReLU}(\mathbf{W}_{L-1} \mathrm{ReLU}(\cdots \mathrm{ReLU}(\mathbf{W}_1\bx + \mathbf{b}_1)\cdots  ) + \bb_{L-1}) + \bb_L
\end{equation}
here $\mathrm{ReLU}(x) = \max(x,0)$ and it is applied elementwise, $L$ is the number of layers and $\bt=\{ (\mathbf{W}_i, \bb_i)\}_{i=1}^L$ are the weights and biases. For all the hidden layers, we use 20 neurons. NN is special because information at each data point is not represented by linear combination of predetermined basis functions but composing linear and nonlinear mappings. 


For piecewise linear functions~(PL), the computational domain is first triangulated and each vertex is associated with one DOF. The value $\nu_{\bt}(\bx)$ is linearly interpolated from the nodal values of the triangle where $\bx$ is located. $\bt$ consists of all those DOFs. In this paper, we obtain the triangulation by splitting each square cell into two triangles on a uniform grid. One disadvantage of PL is the local DOF problem, where the DOFs with no data points nearby are not trained.  

For radial basis functions~(RBF), we have
\begin{equation}
    \nu_{\bt}(\bx) = \sum_{i=1}^M a_i \frac{1}{\sqrt{(\bx-\bx_i)^2 + c^2}}
\end{equation}
where $\{a_i\}_{i=1}^M$ are coefficients, $\{\bx_i\}_{i=1}^M$ are centers, $c$ is the shape parameter. In this paper, the centers are chosen as the grid points on a uniform grids. $c$ is given by the grid step size, suggested by \cite{wu2012using}. Although the basis functions are global, the coefficients in RBF are more affected by data points that are closer to the corresponding centers. Hence, we expect RBF also suffers from the local DOF problem like PL. 

\subsection{ Numerical Approximation to the Characteristic Function}

\begin{figure}[hbt]
\scalebox{0.6}{\input{figures/quad}}
  \caption{Quadrature points used for $\RR^2$ and the unit circle. To avoid bloated plots, we show fewer quadrature points with order $n_{q}=100$, $n_q=20$ respectively.}
  \label{fig:quad}
\end{figure}
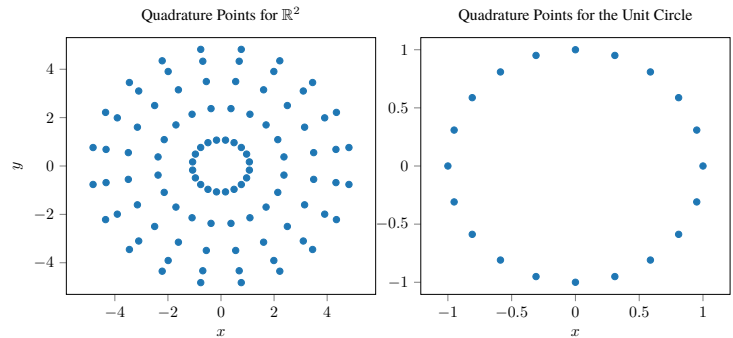

We assume that $\nu(\bx)$ decays as $\|\bx\|\rightarrow \infty$, we use the quadrature rule on the truncated domain $\{(x,y): x^2+y^2<M^2\}$ for approximating the integral in \Cref{equ:CF}
$$\int_{\RR^2} \left( e^{\ii \langle \bxi, \bx\rangle} - 1 - \ii \langle \bxi, \bx\rangle \mathbf{1}_{\|\bx\|\leq 1}\right)\nu(\bx)d\bx \approx \sum_{i=1}^{n_q} \left(e^{\ii \langle\bxi, \bx_i \rangle}-1-\ii\langle\bxi, \bx_i \rangle\mathbf{1}_{\|\bx_i\|\leq 1}  \right)\nu_{\bt}(\bx_i) w_i  $$
The quadrature points and weights $\{(\bx_i, w_i)\}_{i=1}^{n_q}$ are obtained according to \cite{cools2000survey}. For multivariate stable processes in the following, we use quadrature rules on the unit circle. \Cref{fig:quad} shows examples of quadrature points with order $n_{q}=100$ and $n_q=20$.

Consequently, we obtain the expression for the approximation to the characteristic function
\begin{equation}
    \phi(\bxi)\approx    \phi_{\bt}(\bxi) := \exp\left[ \Delta t \sum_{i=1}^{n} \left(e^{\ii \langle\bxi, \bx_i \rangle}-1-\ii\langle\bxi, \bx_i \rangle\mathbf{1}_{\|\bx_i\|\leq 1}  \right)\nu_{\bt}(\bx_i) w_i \right]
\end{equation}

\subsection{Optimization}

The characteristic function matching method requires minimizing the discrepancy between $\hat\phi_n(\bxi)$ and $\phi_{\bt}(\bxi)$. For computation, we consider a set of collocation points $\{\bxi_i \}_{i=1}^m$ uniformly drawn from $[-M',M']^2$ and solve the nonlinear least square problem
\begin{equation}\label{equ:Lbt}
    \min_{\bt} L(\bt) := \frac{1}{m}\sum_{i=1}^m \|\hat\phi_n(\bxi_i)-\phi_{\bt}(\bxi_i)  \|^2
\end{equation}
The choice of $M'$ is based on data. For example, we can choose $M'$ such that $|\hat\phi_n(\bxi)|$ is smaller than a certain value for $\bxi\in \RR^2\backslash [-M',M']^2$. 

The optimization problem \Cref{equ:Lbt} is solved with \texttt{ADCME}. It computes the gradient $\nabla L(\bt)$ using automatic differentiation~\cite{baydin2018automatic} and applies a gradient-based optimizer such as \texttt{L-BFGS-B}~\cite{dai2013perfect} for minimization. The considerable flexibility makes it easy to test different approximation functional forms $\nu_{\bt}(\bx)$ without deriving and implementing new gradients or optimization procedures. 

\subsection{Multivariate $\alpha$-Stable Process: Re-parametrization}

For the multivariate symmetric $\alpha$-stable distribution, the characteristic function of the increment $X_{i\Delta t}-X_{(i-1)\Delta t} $ is given by the following theorem~\cite{samorodnitsky1994levy}
\begin{theorem}\label{thm:alpha}
    $\bX$ is a symmetric $\alpha$-stable vector in $\RR^d$ with $0<\alpha<2$ if and only if there exists a unique symmetric finite measure $\Gamma$ on the unit sphere $\mathbb{S}^d$ such that 
    \begin{equation}\label{equ:alphadis}
        \phi(\bxi) = \mathbb{E}\left( \exp(\ii\Delta t \langle\bX, \bxi\rangle )\right) = \exp\left(-\Delta t\int_{\mathbb{S}^d} | \langle\bs, \bxi\rangle|^\alpha\Gamma(d\bs)     \right)    \end{equation}
    $\Gamma$ is the spectral measure of the symmetric $\alpha$-stable vector $\bX$. 
\end{theorem}
 We assume that $\Gamma$ is determined by a density function such that $\Gamma(d\bs)=\Gamma(\bs)d\bs$.
The previous procedure will fail because $\nu(\bx)$ is singular at $\bx=\mathbf{0}$  thus the given quadrature rule is unable to handle. For example, when $\Gamma(\bs)=1$, the corresponding L\'evy density satisfies~\cite{nolan2008overview}
 \begin{equation}
     \nu(\bx) \propto \frac{1}{\|\bx\|^{\alpha+2}}, \quad \|\bx \|\rightarrow 0
 \end{equation} 

Instead of working with $\nu(\bx)$, we approximate \Cref{equ:alphadis} directly. For calibrating the multivariate symmetric $\alpha$-stable process, we apply the quadrature rule $\{(\bs_i, w_i) \}_{i=1}^{n_q}$ on a unit circle instead of $\RR^2$ and we obtain
\begin{equation}
    \phi(\bxi) \approx \phi_{\bt}(\bxi):=  \exp(\ii\Delta t \langle\bt, \bxi\rangle ) = \exp\left(-\Delta t\sum_{i=1}^{n_q} | \langle\bxi, \bs_i\rangle|^\alpha\Gamma_{\bt}(\bs_i)w_i \right)  
\end{equation}
We have the additional constraint $\Gamma(\bs)=\Gamma(-\bs)$ according to \Cref{thm:alpha}. This is enforced directly by the functional form $\Gamma_{\bt}$. For example, we assume $\Gamma_{\bt}(\bs)=\Gamma'_{\bt}(\bs)+\Gamma'_{\bt}(-\bs)$, where $\Gamma'_{\bt}(\bs)$ is NN, PL or RBF in the 1D domain $[0, 2\pi)$~(since there exists a one-to-one correspondence between $\mathbb{S}^2$ and $[0, 2\pi)$). 

\section{ Numerical Results}

We now present the results of numerical experiments. We first compare the accuracy of three functional forms based on exact characteristic function, ignoring the uncertainty from observations. Then we apply and compare the functional forms to symmetric $\alpha$-stable processes and general L\'evy processes in the presence of uncertainty from observations. We show that NN has very favorable properties in terms of being robust and capturing sharp transitions. 

\subsection{Multivariate $\alpha$-Stable Processes: Estimation from Exact Empirical Characteristic Functions}

\begin{figure}[htpb]
\centering
\scalebox{0.6}{\input{figures/benchmark/benchmarkNN}}~
\scalebox{0.6}{\input{figures/benchmark/benchmarkPL}}~
\scalebox{0.6}{\input{figures/benchmark/benchmarkRBF}}
\scalebox{0.6}{\input{figures/benchmark/benchmarkNN2}}~
\scalebox{0.6}{\input{figures/benchmark/benchmarkPL2}}~
\scalebox{0.6}{\input{figures/benchmark/benchmarkRBF2}}
\caption{Estimated $\Gamma_{\bt}(\bs)$ from exact characteristic functions with different methods. The $x\in[0,2\pi)$ axis corresponds to the angle of $\bs$. For details about legend abbreviations, see \Cref{fig:compare}.}
\label{fig:compare}
\end{figure}
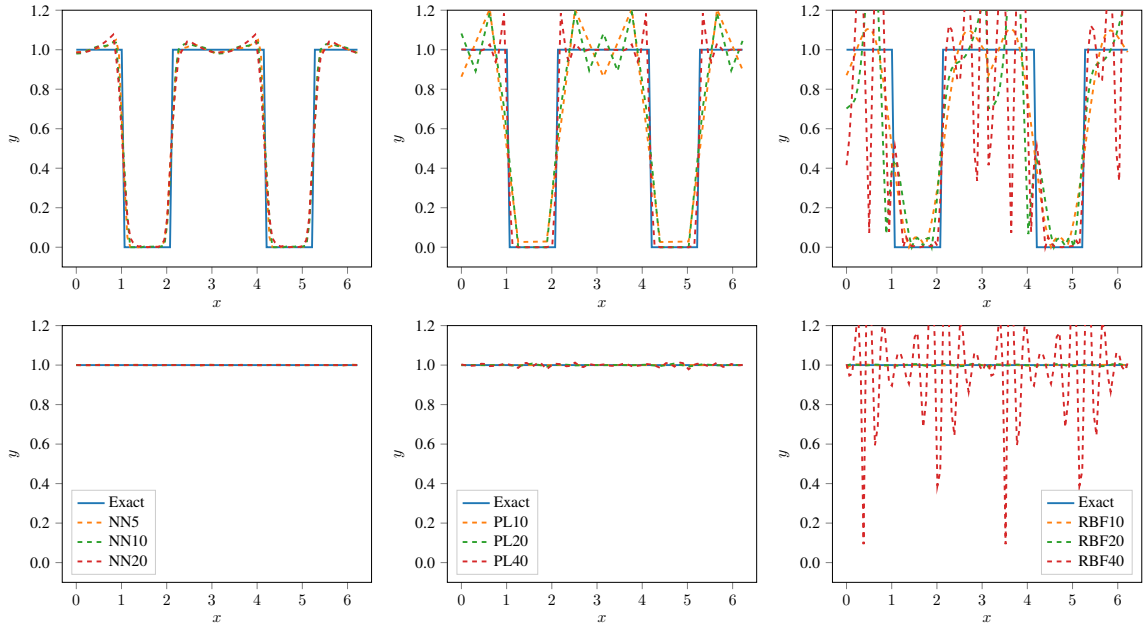

\begin{table}[]
\begin{tabular}{@{}llllllllll@{}}
\toprule
Function & NN5    & NN10   & NN20   & PL10   & PL20   & PL40   & RBF10  & RBF20  & RBF40  \\ \midrule
Step          & 0.7500 & 0.7499 & 0.7498 & 0.7493 & 0.7494 & 0.7500 & 0.7482 & 0.7483 & 0.7504 \\
Constant      & 0.7499 & 0.7500 & 0.7500 & 0.7500 & 0.7500 & 0.7499 & 0.7500 & 0.7500 & 0.7499 \\ \bottomrule
\end{tabular}
\caption{Estimated $\alpha$ for different methods and test functions. The exact fractional index $\alpha$ is $0.75$. We can see that the current method is able to learn $\alpha$ quite accurately, regardless of the choices of basis functions.  }
\label{tab:my-alpha}
\end{table}

In this example, we assume that $\phi(\bxi)$ is computed  with accurate numerical quadrature rules $n_q=10000$ for
\begin{equation}
     \Gamma(\bs) = \mathbf{1}_{|s_1|>0.5}(\bs),\ \mbox{and }\Gamma(\bs) = 1,\ \bs=(s_1, s_2),\ \bs\in \mathbb{S}^2
\end{equation}
hence the error is negligible for estimating $\phi(\bxi)$. We assume $\Delta t = 0.5$, $\alpha=0.75$, and $n_q=100$ for approximating $\phi_{\bt}(\bxi)$. The results in \Cref{fig:compare} indicate that NN can capture the sharp transition better than others. For PL, if DOFs are too few, it is unable to capture the transition; however, too many DOFs results in that some of them are not trained. For RBF, results for RBF40 implies that too few data points compared to the number of centers make the optimization problem ill-posed. Besides, RBF fails to capture the sharp transitions. 

The fractional indices are estimated quite accurately~(\Cref{tab:my-alpha}). This implies that compared to the ``directional'' information of the jump, the heavy tail information is easier to capture. 

\subsection{Multivariate $\alpha$-Stable Processes: Estimation from Observations}

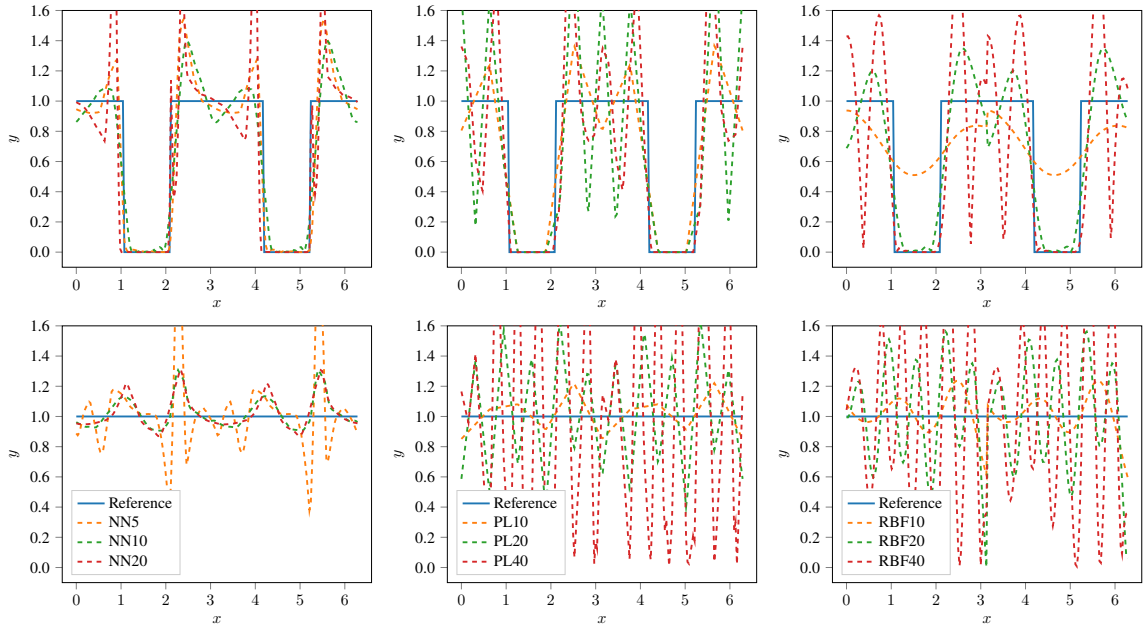
\begin{figure}[htpb]
\centering
\scalebox{0.6}{\input{figures/experiment/experimentNN}}~
\scalebox{0.6}{\input{figures/experiment/experimentPL}}~
\scalebox{0.6}{\input{figures/experiment/experimentRBF}}
\scalebox{0.6}{\input{figures/experiment/experimentNN2}}~
\scalebox{0.6}{\input{figures/experiment/experimentPL2}}~
\scalebox{0.6}{\input{figures/experiment/experimentRBF2}}
\caption{Estimated $\Gamma_{\bt}(\bs)$ from observations with different methods. The $x\in[0,2\pi)$ axis corresponds to the angle of $\bs$. For details about legend abbreviations, see \Cref{fig:compare}.}
\label{fig:compare2}
\end{figure}

\begin{table}[]
\caption{Estimated $\alpha$ for different methods and test functions. The reference fractional index $\alpha$ is $1.5$. For step functions, if we use too few centers for radial basis functions, the estimation is not accurate~(RBF10). This is also demonstrated in \Cref{fig:compare2}.}
\label{tab:ex-alpha}
\begin{tabular}{@{}llllllllll@{}}
\toprule
Function & NN5    & NN10   & NN20   & PL10   & PL20   & PL40   & RBF10  & RBF20  & RBF40  \\ \midrule
Step          & 1.5164 & 1.5156 & 1.5162 & 1.5151 & 1.5166 & 1.5169 & \textbf{3.2155} & 1.5154 & 1.5171 \\
Constant      & 1.5331 & 1.5329 & 1.5329 & 1.5330 & 1.5329 & 1.5330 & 1.5329 & 1.5329 & 1.5330 \\ \bottomrule
\end{tabular}
\end{table}

Now we consider estimating the multivariate $\alpha$-stable process from $m=1000$ observations. Different from last section, $\phi(\bxi)$ is unknown and is estimated with $\hat\phi_n(\bxi)$. The difference $|\hat\phi_n(\bxi)-\phi(\bxi)|$ introduces additional uncertainty, which can also be interpreted as ``noise'' in the nonlinear optimization problem. 

The results in \Cref{fig:compare2} implies that NN is most robust in either case. The $\alpha$ indices are properly estimated as expected in \Cref{tab:ex-alpha}, except for the step function and RBF10 case because of the noise.

\subsection{Multivariate L\'evy Processes.}

\begin{figure}[hbt]
  \includegraphics[width=0.33\textwidth]{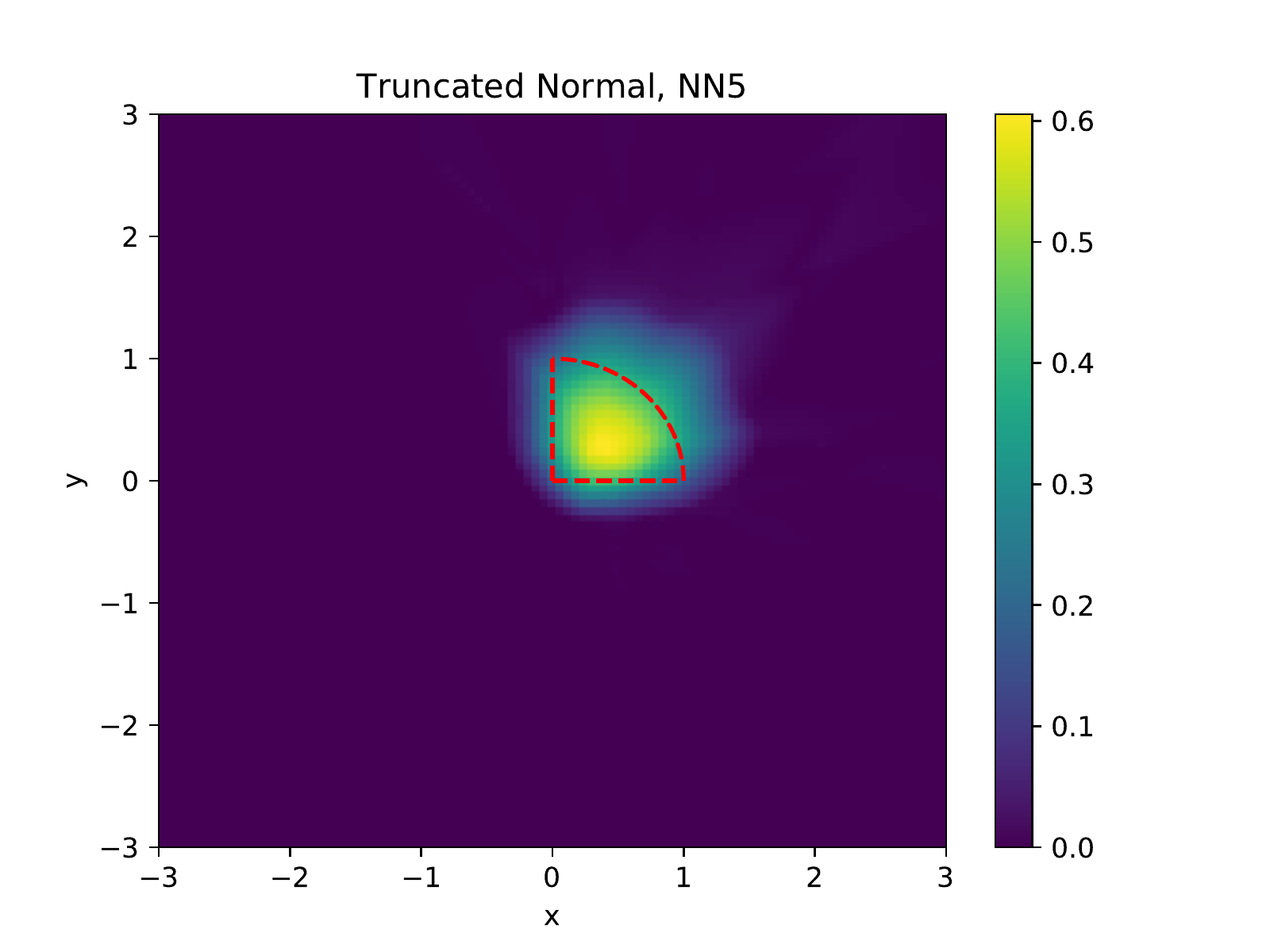}~
  \includegraphics[width=0.33\textwidth]{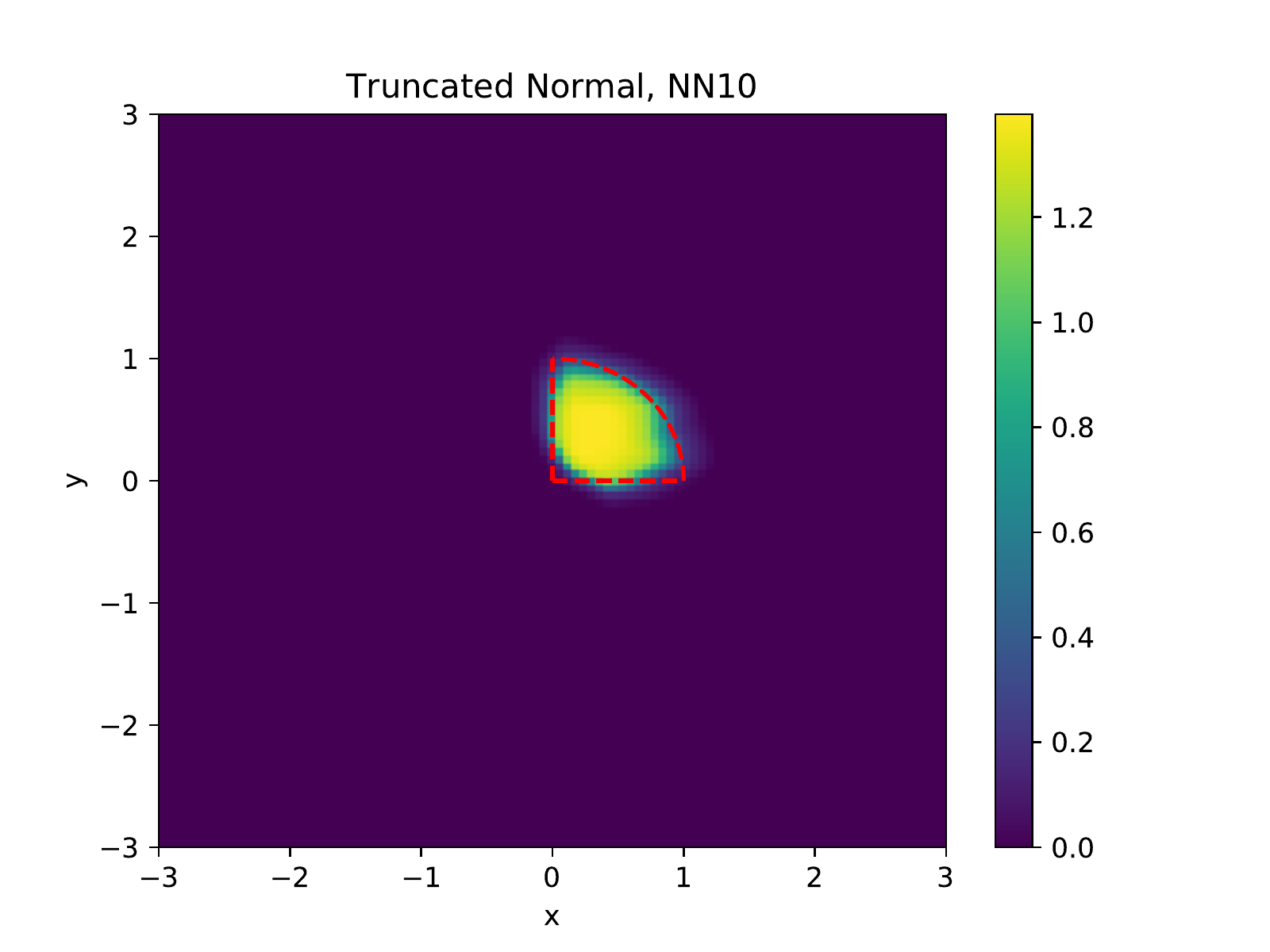}~
  \includegraphics[width=0.33\textwidth]{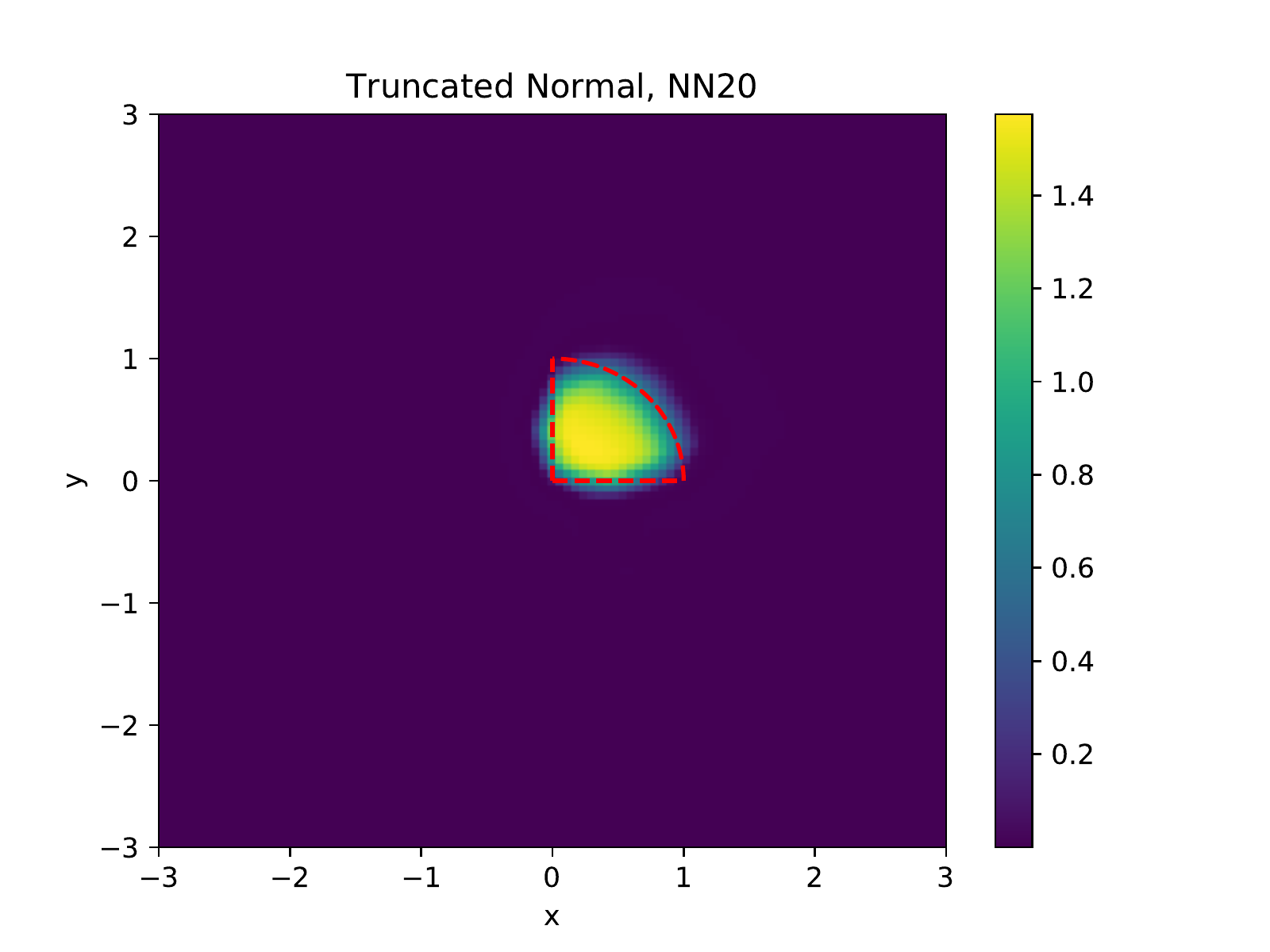}
  \includegraphics[width=0.33\textwidth]{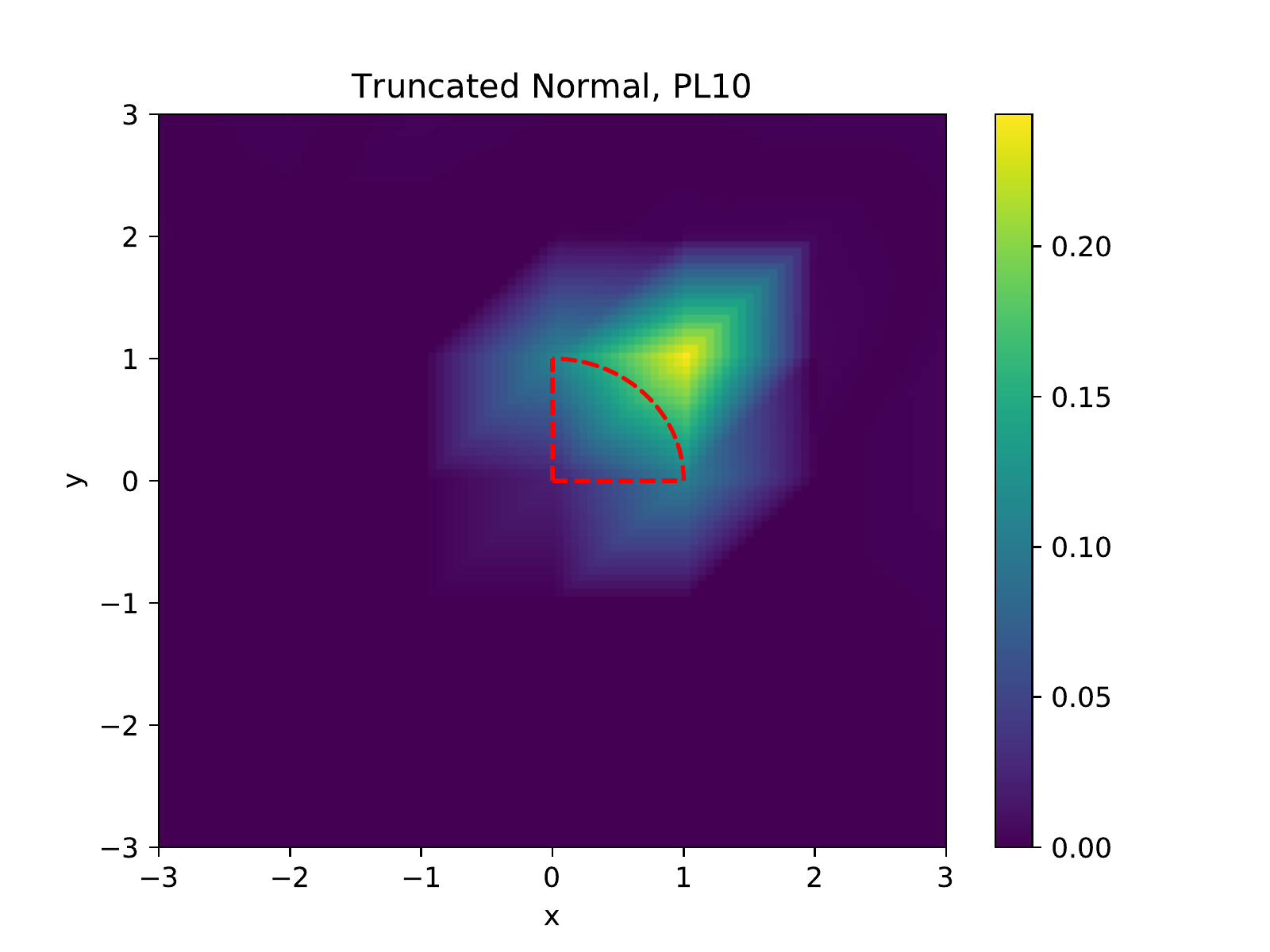}~
  \includegraphics[width=0.33\textwidth]{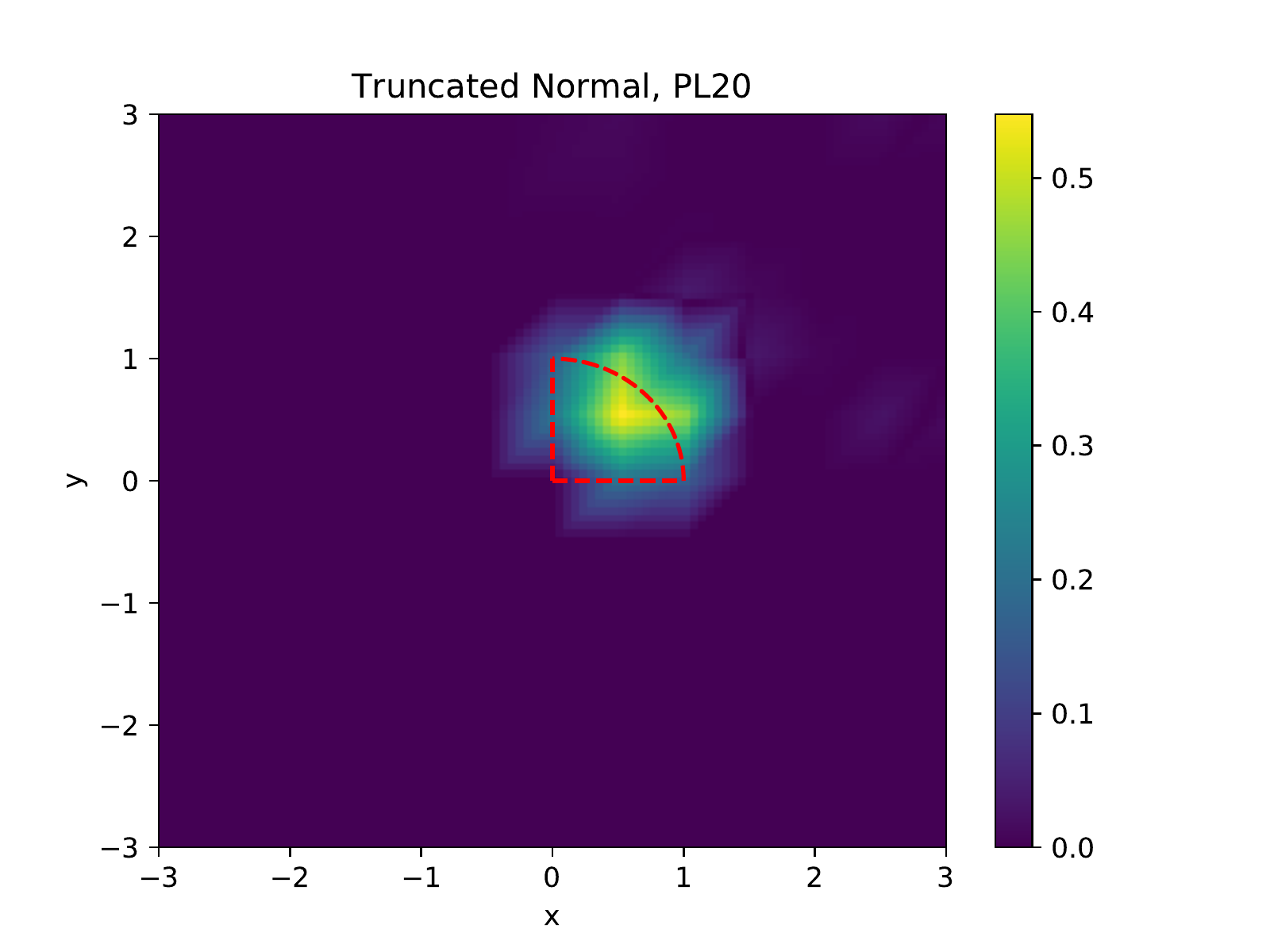}~
  \includegraphics[width=0.33\textwidth]{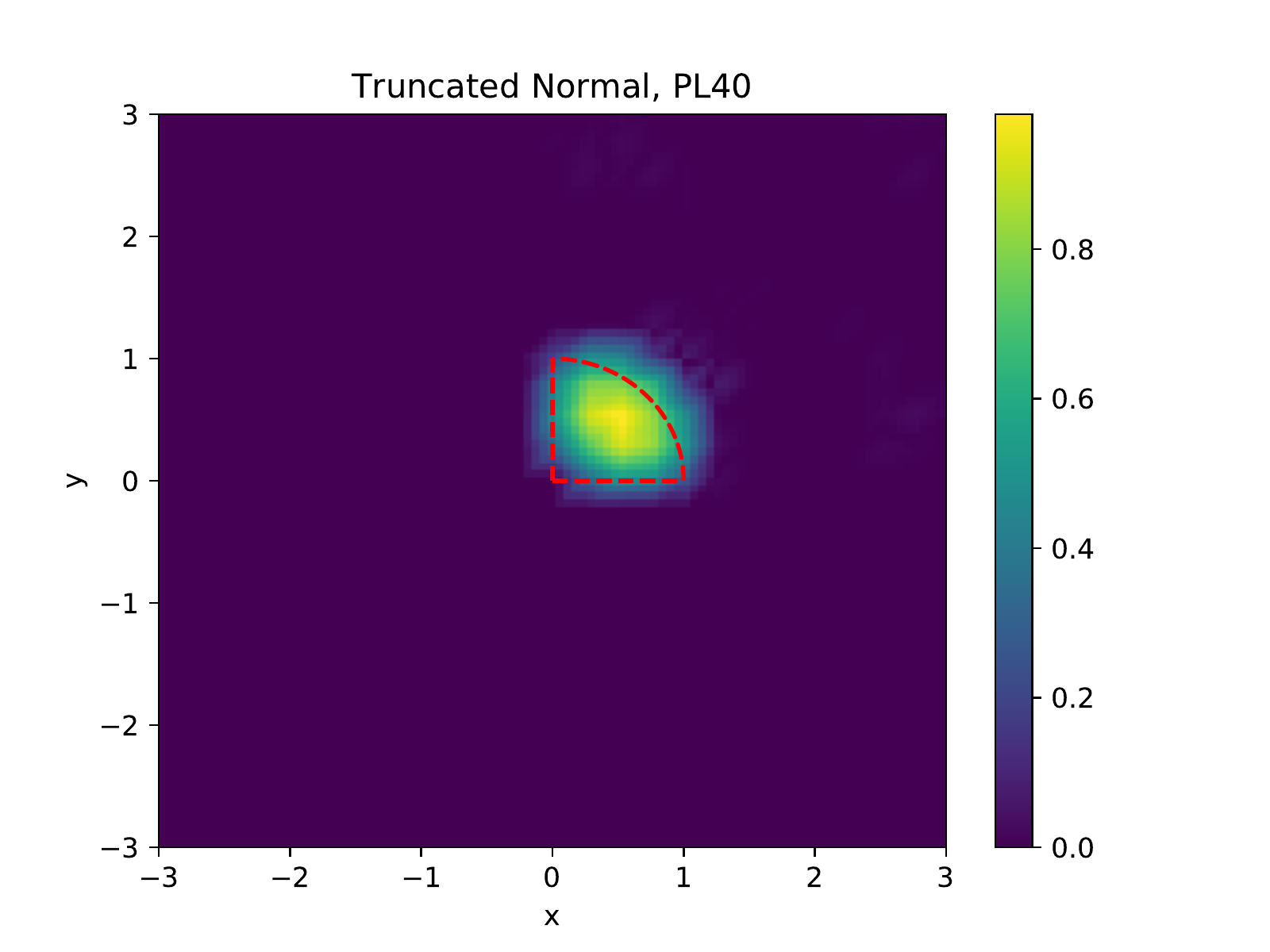}
  \includegraphics[width=0.33\textwidth]{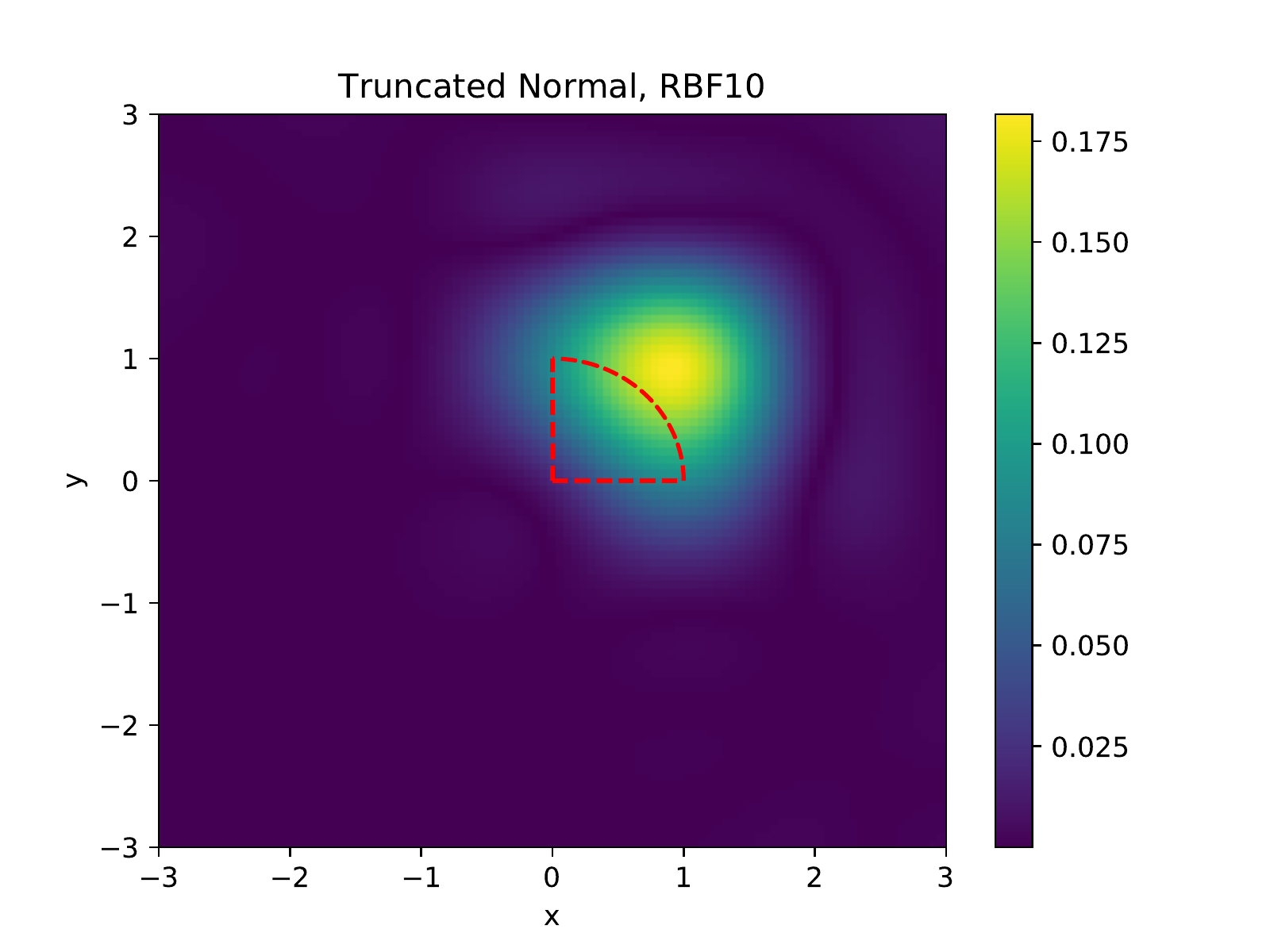}~
  \includegraphics[width=0.33\textwidth]{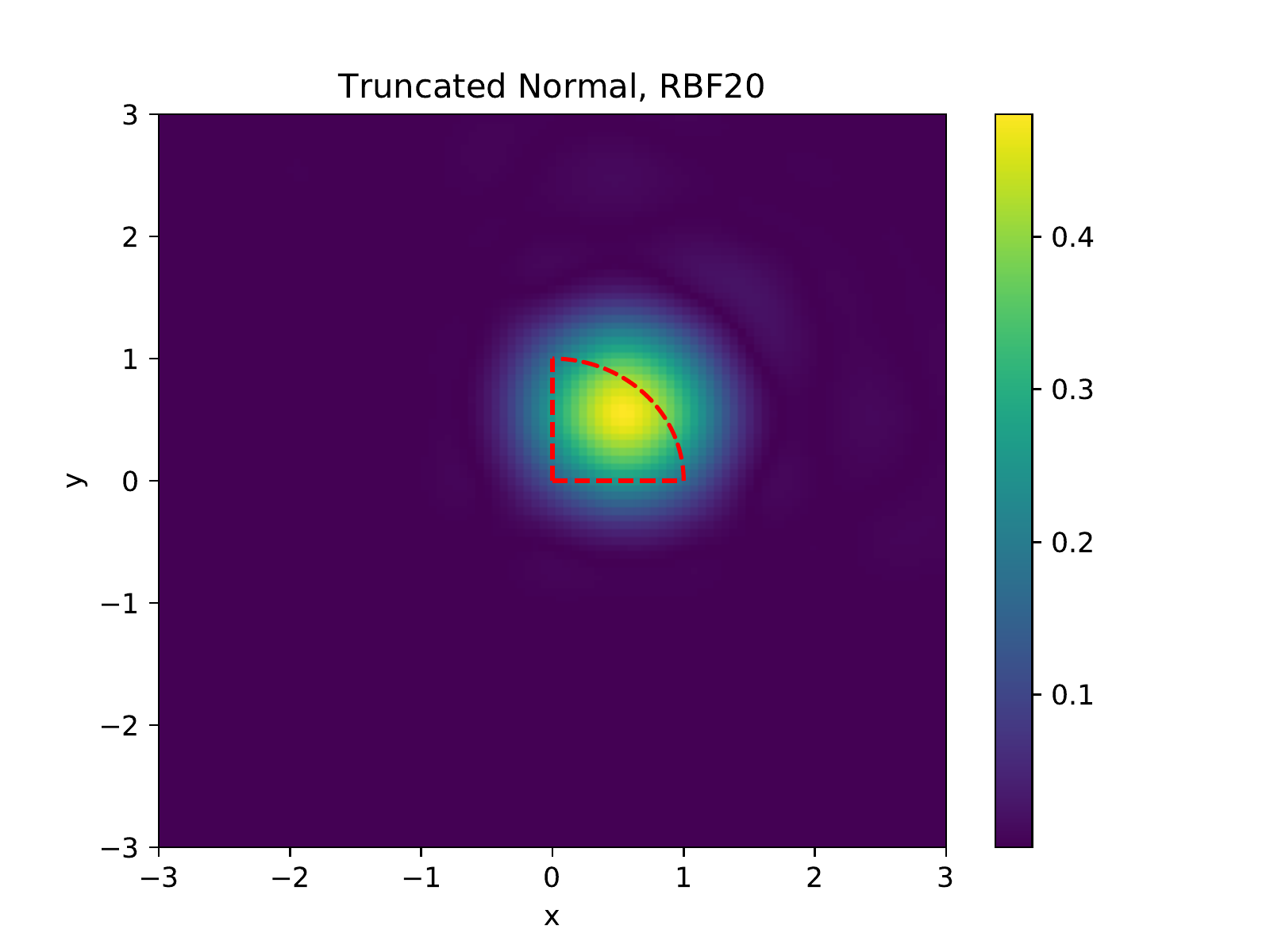}~
  \includegraphics[width=0.33\textwidth]{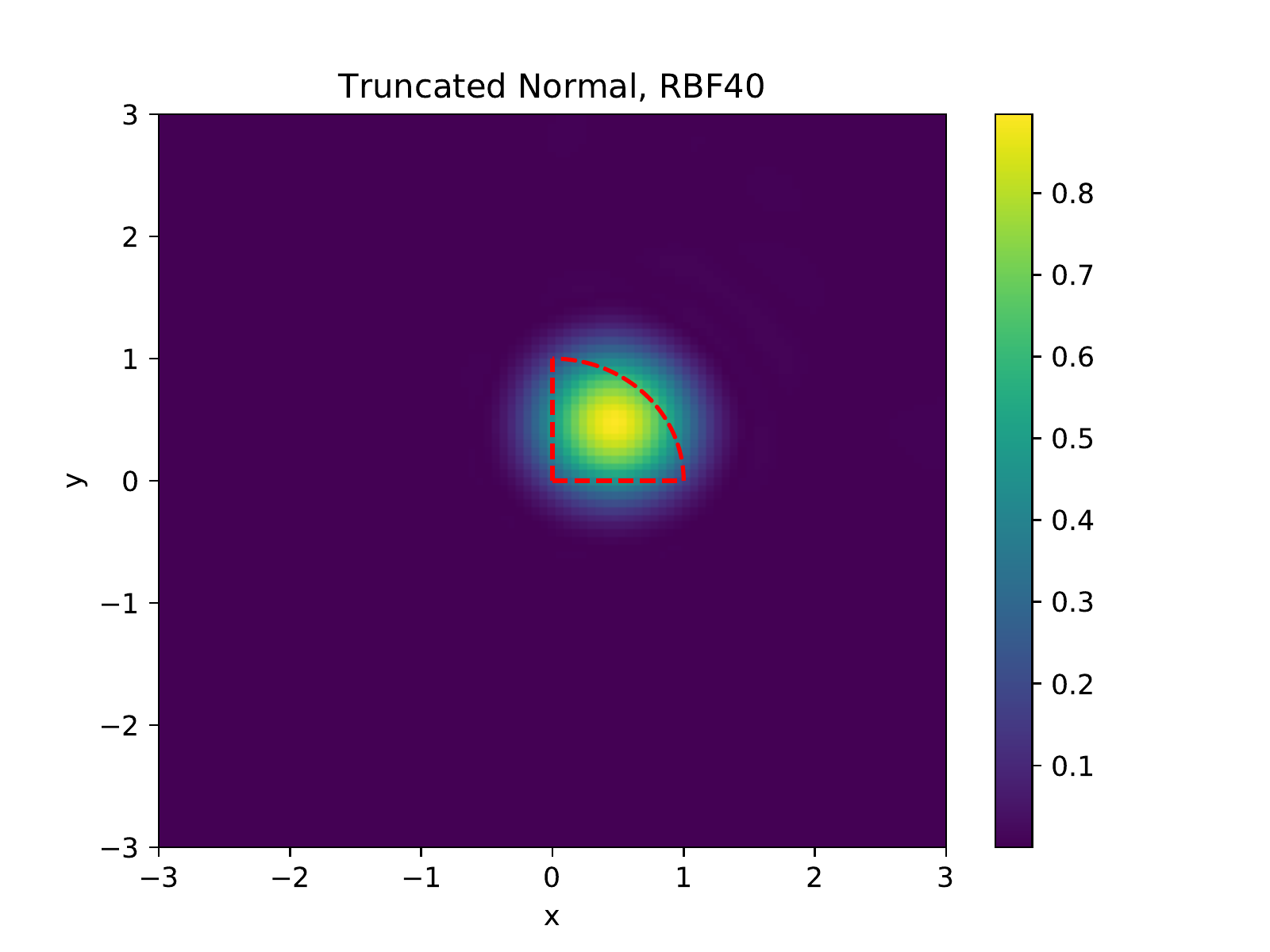}
  \caption{Estimated $\nu_{\bt}(\bx)$ from observations with different methods. The intercepted arc of the dashed red sector is $\{(x,y):x^2+y^2= 1, x\geq 0, y\geq 0 \}$. The sector indicates the area with most of the density for the reference $\nu(\bx)$. For details about subtitle abbreviations, see \Cref{fig:compare}.}
  \label{fig:levy}
\end{figure}

In this example, we consider the L\'evy process where the jump distributions are truncated normal distributions. The L\'evy density has the expression
\begin{equation}
    \nu(\bx) = \frac{2}{\pi}\exp\left( -\frac{\|\bx\|^2}{2} \right)\mathbf{1}_{\bx\in \RR^2_{+}}
\end{equation}
The density $\nu(\bx)$ is only nonzero for $\bx\in \RR^2_{+}$ and has sharp transition at axes $x=0$ and $y=0$ in the first quadrant. In our experiment, we assume $\Delta t = 0.5$, $m=10000$, $n_q = 4096$. The data $\{\bX_{i\Delta t}\}_{i=1}^n$ are simulated according to \cite{nolan2008overview}. Notably, NN captures the sharp transition~(\Cref{fig:levy}). The results from PL shows artifacts because of the localized DOFs. Despite properly indicating the location of the major density mass, RBF creates a smooth profile of the density distribution.

\subsection{Application to Stock Markets.}

\begin{figure}[hbt]
  \includegraphics[width=0.35\textwidth]{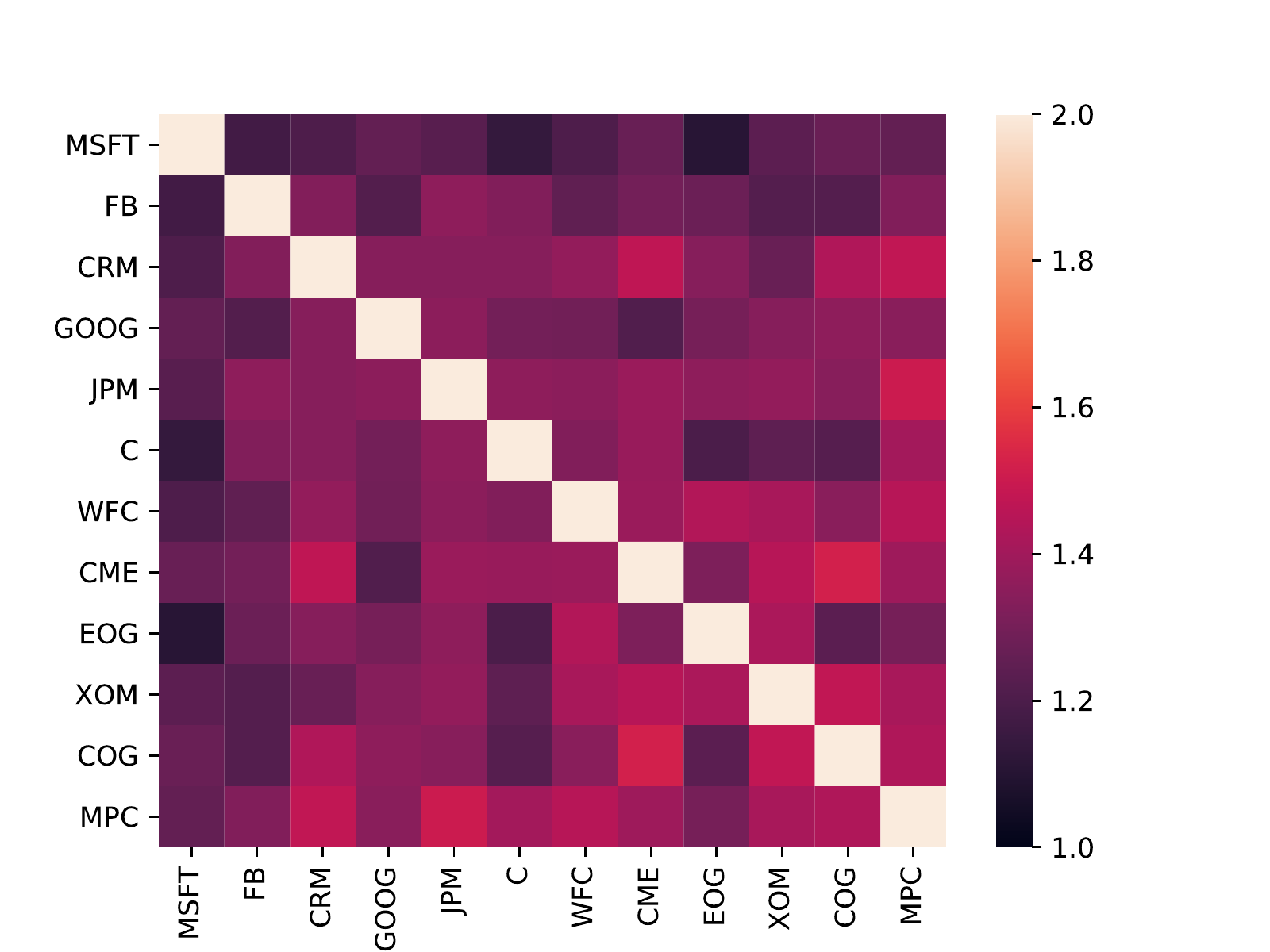}~
  \scalebox{0.55}{\input{figures/stock/91/theta}}~
\scalebox{0.55}{\input{figures/stock/91/x0}}
  \caption{The first plot shows the pairwise $\alpha$ indices. For each stock, the $\alpha$ index against itself is not computed. The right two plots show the calibrated $\Gamma_{\bt}(\bs)$ and shifted log return for EOG and MSFT. Here $\bs=(\cos(\Theta), \sin(\Theta))$}
  \label{fig:stock}
\end{figure}

Finally, we apply the developed procedure to a stock market example. We investigate 12 stocks from 01/01/2016 to 08/01/2019, which are from the technology sector~(MSFT, AAPL, AMZN, GOOG), the financial sector~(JPM, C, WFC, CME) and the energy sector~(EOG, XOM, COG, MPC). The $\alpha$ index is computed for each pair of stocks. The stock prices are turned into the log return and then shifted such that data for each stock are unbiased.

We model the pairwise shifted log return of the stocks by a 2D symmetric $\alpha$-stable process with unknown $\alpha$ and $\Gamma(\bs)$. \Cref{fig:stock} shows the estimated pairwise $\alpha$ indices. Most of the indices are between $1.1$ and $1.5$. This implies that there does exist jumps in the pairwise log return changes. We also show $\Gamma(\bs)$ for EOG vs. MSFT. We identify 4 peaks in the plot, which indicate that there is a larger tendency for price to jump in those 4 directions compared to nearby directions.

\section{ Conclusion}

We have proposed a novel nonparametric estimation approach for L\'evy processes and compared three approximation functional forms: (1) neural network; (2) piecewise linear functions; (3) radial basis functions. We found that for the tested cases the neural network performed best for being robust to noise and capturing sharp transitions. However, one should not expect that neural networks are always superior to others. Most likely, a certain functional form may be more suitable to a class of problems, since the performance highly depends on the characteristics of the training data. 

Besides L\'evy processes, the same idea---approximating an unknown function in a system model with the neural network, and training by matching the model outputs with observations---can be applied to many other fields as well. For example, in mechanical engineering, constitutive laws have been reconstructed from observed displacement data~\cite{huang2019predictive}; in general, coupled partial differential equation systems,  closure relations are discovered from observations~\cite{xu2019neural}. In the future, a deeper understanding of the neural network approximation properties and improvement of the training algorithm will broaden the applications of the nonparametric estimation approach. 

\bibliographystyle{unsrt}
\bibliography{levy}

\end{document}

%% file: figures/quad.tex
\begin{tikzpicture}

\definecolor{color0}{rgb}{0.12156862745098,0.466666666666667,0.705882352941177}

\begin{groupplot}[group style={group size=2 by 1}]
\nextgroupplot[
tick align=outside,
tick pos=left,
title={Quadrature Points for $\mathbb{R}^2$},
x grid style={white!69.01960784313725!black},
xlabel={$x$},
xmin=-5.82773657002982, xmax=5.82773657002982,
y grid style={white!69.01960784313725!black},
ylabel={$y$},
ymin=-5.30762674711318, ymax=5.30762674711318
]
\addplot [only marks, draw=color0, fill=color0, colormap/viridis]
table [row sep=\\]{%
x                      y\\ 
-7.636060921222967e-01 +4.821219120752241e+00\\ 
-2.216071191713472e+00 +4.349284601804179e+00\\ 
-3.451611802783832e+00 +3.451611802783832e+00\\ 
-4.349284601804179e+00 +2.216071191713472e+00\\ 
-4.821219120752241e+00 +7.636060921222967e-01\\ 
-4.821219120752241e+00 -7.636060921222967e-01\\ 
-4.349284601804179e+00 -2.216071191713472e+00\\ 
-3.451611802783832e+00 -3.451611802783832e+00\\ 
-2.216071191713472e+00 -4.349284601804179e+00\\ 
-7.636060921222967e-01 -4.821219120752241e+00\\ 
-6.860122430101184e-01 +4.331310838390760e+00\\ 
-1.990885070956174e+00 +3.907331955511954e+00\\ 
-3.100876196845098e+00 +3.100876196845098e+00\\ 
-3.907331955511954e+00 +1.990885070956174e+00\\ 
-4.331310838390760e+00 +6.860122430101184e-01\\ 
-4.331310838390760e+00 -6.860122430101184e-01\\ 
-3.907331955511954e+00 -1.990885070956174e+00\\ 
-3.100876196845098e+00 -3.100876196845098e+00\\ 
-1.990885070956174e+00 -3.907331955511954e+00\\ 
-6.860122430101184e-01 -4.331310838390760e+00\\ 
-5.530793552061858e-01 +3.492005616668552e+00\\ 
-1.605098804800515e+00 +3.150183776675253e+00\\ 
-2.500000000000000e+00 +2.500000000000000e+00\\ 
-3.150183776675253e+00 +1.605098804800515e+00\\ 
-3.492005616668552e+00 +5.530793552061858e-01\\ 
-3.492005616668552e+00 -5.530793552061858e-01\\ 
-3.150183776675253e+00 -1.605098804800515e+00\\ 
-2.500000000000000e+00 -2.500000000000000e+00\\ 
-1.605098804800515e+00 -3.150183776675253e+00\\ 
-5.530793552061858e-01 -3.492005616668552e+00\\ 
-3.757402676727721e-01 +2.372330684143371e+00\\ 
-1.090440727682122e+00 +2.140110427779614e+00\\ 
-1.698401251718651e+00 +1.698401251718651e+00\\ 
-2.140110427779614e+00 +1.090440727682122e+00\\ 
-2.372330684143371e+00 +3.757402676727721e-01\\ 
-2.372330684143371e+00 -3.757402676727721e-01\\ 
-2.140110427779614e+00 -1.090440727682122e+00\\ 
-1.698401251718651e+00 -1.698401251718651e+00\\ 
-1.090440727682122e+00 -2.140110427779614e+00\\ 
-3.757402676727721e-01 -2.372330684143371e+00\\ 
-1.694086254719484e-01 +1.069603965672531e+00\\ 
-4.916429798153517e-01 +9.649036771434659e-01\\ 
-7.657518938163654e-01 +7.657518938163654e-01\\ 
-9.649036771434659e-01 +4.916429798153517e-01\\ 
-1.069603965672531e+00 +1.694086254719484e-01\\ 
-1.069603965672531e+00 -1.694086254719484e-01\\ 
-9.649036771434659e-01 -4.916429798153517e-01\\ 
-7.657518938163654e-01 -7.657518938163654e-01\\ 
-4.916429798153517e-01 -9.649036771434659e-01\\ 
-1.694086254719484e-01 -1.069603965672531e+00\\ 
+1.694086254719484e-01 -1.069603965672531e+00\\ 
+4.916429798153517e-01 -9.649036771434659e-01\\ 
+7.657518938163654e-01 -7.657518938163654e-01\\ 
+9.649036771434659e-01 -4.916429798153517e-01\\ 
+1.069603965672531e+00 -1.694086254719484e-01\\ 
+1.069603965672531e+00 +1.694086254719484e-01\\ 
+9.649036771434659e-01 +4.916429798153517e-01\\ 
+7.657518938163654e-01 +7.657518938163654e-01\\ 
+4.916429798153517e-01 +9.649036771434659e-01\\ 
+1.694086254719484e-01 +1.069603965672531e+00\\ 
+3.757402676727721e-01 -2.372330684143371e+00\\ 
+1.090440727682122e+00 -2.140110427779614e+00\\ 
+1.698401251718651e+00 -1.698401251718651e+00\\ 
+2.140110427779614e+00 -1.090440727682122e+00\\ 
+2.372330684143371e+00 -3.757402676727721e-01\\ 
+2.372330684143371e+00 +3.757402676727721e-01\\ 
+2.140110427779614e+00 +1.090440727682122e+00\\ 
+1.698401251718651e+00 +1.698401251718651e+00\\ 
+1.090440727682122e+00 +2.140110427779614e+00\\ 
+3.757402676727721e-01 +2.372330684143371e+00\\ 
+5.530793552061858e-01 -3.492005616668552e+00\\ 
+1.605098804800515e+00 -3.150183776675253e+00\\ 
+2.500000000000000e+00 -2.500000000000000e+00\\ 
+3.150183776675253e+00 -1.605098804800515e+00\\ 
+3.492005616668552e+00 -5.530793552061858e-01\\ 
+3.492005616668552e+00 +5.530793552061858e-01\\ 
+3.150183776675253e+00 +1.605098804800515e+00\\ 
+2.500000000000000e+00 +2.500000000000000e+00\\ 
+1.605098804800515e+00 +3.150183776675253e+00\\ 
+5.530793552061858e-01 +3.492005616668552e+00\\ 
+6.860122430101184e-01 -4.331310838390760e+00\\ 
+1.990885070956174e+00 -3.907331955511954e+00\\ 
+3.100876196845098e+00 -3.100876196845098e+00\\ 
+3.907331955511954e+00 -1.990885070956174e+00\\ 
+4.331310838390760e+00 -6.860122430101184e-01\\ 
+4.331310838390760e+00 +6.860122430101184e-01\\ 
+3.907331955511954e+00 +1.990885070956174e+00\\ 
+3.100876196845098e+00 +3.100876196845098e+00\\ 
+1.990885070956174e+00 +3.907331955511954e+00\\ 
+6.860122430101184e-01 +4.331310838390760e+00\\ 
+7.636060921222967e-01 -4.821219120752241e+00\\ 
+2.216071191713472e+00 -4.349284601804179e+00\\ 
+3.451611802783832e+00 -3.451611802783832e+00\\ 
+4.349284601804179e+00 -2.216071191713472e+00\\ 
+4.821219120752241e+00 -7.636060921222967e-01\\ 
+4.821219120752241e+00 +7.636060921222967e-01\\ 
+4.349284601804179e+00 +2.216071191713472e+00\\ 
+3.451611802783832e+00 +3.451611802783832e+00\\ 
+2.216071191713472e+00 +4.349284601804179e+00\\ 
+7.636060921222967e-01 +4.821219120752241e+00\\ 
};

\ \ \ \

\nextgroupplot[
tick align=outside,
tick pos=left,
title={Quadrature Points for the Unit Circle},
x grid style={white!69.01960784313725!black},
xlabel={$x$},
xmin=-1.21249789171867, xmax=1.21249789171867,
y grid style={white!69.01960784313725!black},
ymin=-1.10428571428571, ymax=1.10428571428571
]
\addplot [only marks, draw=color0, fill=color0, colormap/viridis]
table [row sep=\\]{%
x                      y\\ 
+1.000000000000000e+00 +0.000000000000000e+00\\ 
+9.510565162951535e-01 +3.090169943749474e-01\\ 
+8.090169943749475e-01 +5.877852522924731e-01\\ 
+5.877852522924731e-01 +8.090169943749475e-01\\ 
+3.090169943749475e-01 +9.510565162951535e-01\\ 
+6.123233995736766e-17 +1.000000000000000e+00\\ 
-3.090169943749473e-01 +9.510565162951536e-01\\ 
-5.877852522924730e-01 +8.090169943749475e-01\\ 
-8.090169943749473e-01 +5.877852522924732e-01\\ 
-9.510565162951535e-01 +3.090169943749475e-01\\ 
-1.000000000000000e+00 +1.224646799147353e-16\\ 
-9.510565162951536e-01 -3.090169943749473e-01\\ 
-8.090169943749475e-01 -5.877852522924730e-01\\ 
-5.877852522924732e-01 -8.090169943749473e-01\\ 
-3.090169943749476e-01 -9.510565162951535e-01\\ 
-1.836970198721030e-16 -1.000000000000000e+00\\ 
+3.090169943749472e-01 -9.510565162951536e-01\\ 
+5.877852522924729e-01 -8.090169943749476e-01\\ 
+8.090169943749473e-01 -5.877852522924734e-01\\ 
+9.510565162951535e-01 -3.090169943749476e-01\\ 
};

\end{groupplot}

\end{tikzpicture}

%% file: figures/benchmark/benchmarkNN.tex
\begin{tikzpicture}

\definecolor{color0}{rgb}{0.12156862745098,0.466666666666667,0.705882352941177}
\definecolor{color1}{rgb}{1,0.498039215686275,0.0549019607843137}
\definecolor{color2}{rgb}{0.172549019607843,0.627450980392157,0.172549019607843}
\definecolor{color3}{rgb}{0.83921568627451,0.152941176470588,0.156862745098039}

\begin{axis}[
legend cell align={left},
tick align=outside,
tick pos=left,
x grid style={white!69.01960784313725!black},
xlabel={$x$},
xmin=-0.31101767270539, xmax=6.53137112681318,
y grid style={white!69.01960784313725!black},
ylabel={$y$},
ymin=-0.1, ymax=1.2,
ytick={-0.2,0,0.2,0.4,0.6,0.8,1,1.2},
yticklabels={−0.2,0.0,0.2,0.4,0.6,0.8,1.0,1.2}
]
\addplot [very thick, color0]
table [row sep=\\]{%
0	1 \\
0.0628318530717959	1 \\
0.125663706143592	1 \\
0.188495559215388	1 \\
0.251327412287183	1 \\
0.314159265358979	1 \\
0.376991118430775	1 \\
0.439822971502571	1 \\
0.502654824574367	1 \\
0.565486677646163	1 \\
0.628318530717959	1 \\
0.691150383789754	1 \\
0.75398223686155	1 \\
0.816814089933346	1 \\
0.879645943005142	1 \\
0.942477796076938	1 \\
1.00530964914873	1 \\
1.06814150222053	0 \\
1.13097335529233	0 \\
1.19380520836412	0 \\
1.25663706143592	0 \\
1.31946891450771	0 \\
1.38230076757951	0 \\
1.4451326206513	0 \\
1.5079644737231	0 \\
1.5707963267949	0 \\
1.63362817986669	0 \\
1.69646003293849	0 \\
1.75929188601028	0 \\
1.82212373908208	0 \\
1.88495559215388	0 \\
1.94778744522567	0 \\
2.01061929829747	0 \\
2.07345115136926	0 \\
2.13628300444106	1 \\
2.19911485751286	1 \\
2.26194671058465	1 \\
2.32477856365645	1 \\
2.38761041672824	1 \\
2.45044226980004	1 \\
2.51327412287183	1 \\
2.57610597594363	1 \\
2.63893782901543	1 \\
2.70176968208722	1 \\
2.76460153515902	1 \\
2.82743338823081	1 \\
2.89026524130261	1 \\
2.95309709437441	1 \\
3.0159289474462	1 \\
3.078760800518	1 \\
3.14159265358979	1 \\
3.20442450666159	1 \\
3.26725635973339	1 \\
3.33008821280518	1 \\
3.39292006587698	1 \\
3.45575191894877	1 \\
3.51858377202057	1 \\
3.58141562509236	1 \\
3.64424747816416	1 \\
3.70707933123596	1 \\
3.76991118430775	1 \\
3.83274303737955	1 \\
3.89557489045134	1 \\
3.95840674352314	1 \\
4.02123859659494	1 \\
4.08407044966673	1 \\
4.14690230273853	1 \\
4.20973415581032	0 \\
4.27256600888212	0 \\
4.33539786195391	0 \\
4.39822971502571	0 \\
4.46106156809751	0 \\
4.5238934211693	0 \\
4.5867252742411	0 \\
4.64955712731289	0 \\
4.71238898038469	0 \\
4.77522083345649	0 \\
4.83805268652828	0 \\
4.90088453960008	0 \\
4.96371639267187	0 \\
5.02654824574367	0 \\
5.08938009881547	0 \\
5.15221195188726	0 \\
5.21504380495906	0 \\
5.27787565803085	1 \\
5.34070751110265	1 \\
5.40353936417444	1 \\
5.46637121724624	1 \\
5.52920307031804	1 \\
5.59203492338983	1 \\
5.65486677646163	1 \\
5.71769862953342	1 \\
5.78053048260522	1 \\
5.84336233567702	1 \\
5.90619418874881	1 \\
5.96902604182061	1 \\
6.0318578948924	1 \\
6.0946897479642	1 \\
6.157521601036	1 \\
6.22035345410779	1 \\
};
\addplot [very thick, color1, dashed]
table [row sep=\\]{%
0	0.986759082552423 \\
0.0628318530717959	0.988303755600533 \\
0.125663706143592	0.989291534367183 \\
0.188495559215388	0.993127278911961 \\
0.251327412287183	0.996804851021916 \\
0.314159265358979	1.00013937083451 \\
0.376991118430775	1.00311767852345 \\
0.439822971502571	1.00572802006894 \\
0.502654824574367	1.00794141998881 \\
0.565486677646163	1.00967732208178 \\
0.628318530717959	1.0137717436977 \\
0.691150383789754	1.02102822337654 \\
0.75398223686155	1.0291718276431 \\
0.816814089933346	1.04003847597807 \\
0.879645943005142	1.04961699488551 \\
0.942477796076938	0.982523302905834 \\
1.00530964914873	0.650433511773122 \\
1.06814150222053	0.30459475452395 \\
1.13097335529233	0.0515169577999326 \\
1.19380520836412	0.00138970515184095 \\
1.25663706143592	0.0002543819994128 \\
1.31946891450771	0.000172804704689389 \\
1.38230076757951	1.36788120918552e-05 \\
1.4451326206513	0.00266067358845179 \\
1.5079644737231	2.25616198503142e-05 \\
1.5707963267949	0.00140124478877884 \\
1.63362817986669	0.000715195677814862 \\
1.69646003293849	0.00131645495344956 \\
1.75929188601028	0.0030929628294894 \\
1.82212373908208	0.00248276403377015 \\
1.88495559215388	0.00226418672709366 \\
1.94778744522567	0.0309991721769841 \\
2.01061929829747	0.135851885251004 \\
2.07345115136926	0.396173741157711 \\
2.13628300444106	0.648432505245685 \\
2.19911485751286	0.88841093789421 \\
2.26194671058465	0.988568786669953 \\
2.32477856365645	0.99829955782025 \\
2.38761041672824	1.00932863416307 \\
2.45044226980004	1.01273036144372 \\
2.51327412287183	1.02030409915959 \\
2.57610597594363	1.02897816577691 \\
2.63893782901543	1.02496914629278 \\
2.70176968208722	1.01945946699171 \\
2.76460153515902	1.01334645072953 \\
2.82743338823081	1.00665422278867 \\
2.89026524130261	0.999409194335408 \\
2.95309709437441	0.991639958187121 \\
3.0159289474462	0.987052541579939 \\
3.078760800518	0.985853214168728 \\
3.14159265358979	0.986759082552423 \\
3.20442450666159	0.988303755600533 \\
3.26725635973339	0.989291534367183 \\
3.33008821280518	0.993127278911961 \\
3.39292006587698	0.996804851021916 \\
3.45575191894877	1.00013937083451 \\
3.51858377202057	1.00311767852345 \\
3.58141562509236	1.00572802006894 \\
3.64424747816416	1.00794141998881 \\
3.70707933123596	1.00967732208178 \\
3.76991118430775	1.0137717436977 \\
3.83274303737955	1.02102822337654 \\
3.89557489045134	1.0291718276431 \\
3.95840674352314	1.04003847597807 \\
4.02123859659494	1.04961699488551 \\
4.08407044966673	0.982523302905835 \\
4.14690230273853	0.650433511773126 \\
4.20973415581032	0.304594754523949 \\
4.27256600888212	0.0515169577999337 \\
4.33539786195391	0.00138970515184078 \\
4.39822971502571	0.000254381999412689 \\
4.46106156809751	0.000172804704689333 \\
4.5238934211693	1.36788120917442e-05 \\
4.5867252742411	0.00266067358845168 \\
4.64955712731289	2.25616198502587e-05 \\
4.71238898038469	0.00140124478877873 \\
4.77522083345649	0.000715195677815084 \\
4.83805268652828	0.00131645495344956 \\
4.90088453960008	0.0030929628294892 \\
4.96371639267187	0.00248276403377004 \\
5.02654824574367	0.00226418672709353 \\
5.08938009881547	0.0309991721769843 \\
5.15221195188726	0.135851885251004 \\
5.21504380495906	0.39617374115771 \\
5.27787565803085	0.648432505245686 \\
5.34070751110265	0.88841093789421 \\
5.40353936417444	0.988568786669953 \\
5.46637121724624	0.99829955782025 \\
5.52920307031804	1.00932863416307 \\
5.59203492338983	1.01273036144372 \\
5.65486677646163	1.02030409915959 \\
5.71769862953342	1.02897816577691 \\
5.78053048260522	1.02496914629278 \\
5.84336233567702	1.01945946699171 \\
5.90619418874881	1.01334645072953 \\
5.96902604182061	1.00665422278867 \\
6.0318578948924	0.999409194335409 \\
6.0946897479642	0.991639958187121 \\
6.157521601036	0.987052541579939 \\
6.22035345410779	0.985853214168728 \\
};
\addplot [very thick, color2, dashed]
table [row sep=\\]{%
0	0.980402178946657 \\
0.0628318530717959	0.981337268056434 \\
0.125663706143592	0.982386171802597 \\
0.188495559215388	0.987677034288272 \\
0.251327412287183	0.995267694266534 \\
0.314159265358979	1.00187306833252 \\
0.376991118430775	1.00809318543536 \\
0.439822971502571	1.01368678972052 \\
0.502654824574367	1.01846754869524 \\
0.565486677646163	1.01949628084926 \\
0.628318530717959	1.01915537856404 \\
0.691150383789754	1.02184552395701 \\
0.75398223686155	1.03059952235088 \\
0.816814089933346	1.03769880757187 \\
0.879645943005142	1.03583220839844 \\
0.942477796076938	0.8491194171094 \\
1.00530964914873	0.6198902788014 \\
1.06814150222053	0.384777954049187 \\
1.13097335529233	0.149778698568561 \\
1.19380520836412	0.000179643974416943 \\
1.25663706143592	1.14012046055212e-05 \\
1.31946891450771	0.000110099726553903 \\
1.38230076757951	0.000742638126661199 \\
1.4451326206513	4.75644510111728e-07 \\
1.5079644737231	0.00364973257143941 \\
1.5707963267949	0.0006098291097546 \\
1.63362817986669	0.00267996426369443 \\
1.69646003293849	0.00302687374038424 \\
1.75929188601028	0.00765553817268555 \\
1.82212373908208	3.6362290956049e-10 \\
1.88495559215388	0.00512419658332175 \\
1.94778744522567	0.00839114395008961 \\
2.01061929829747	0.116601510343989 \\
2.07345115136926	0.364112869999059 \\
2.13628300444106	0.629591924714259 \\
2.19911485751286	0.89159212351293 \\
2.26194671058465	1.00131160688525 \\
2.32477856365645	1.01476822847766 \\
2.38761041672824	1.02833878560434 \\
2.45044226980004	1.02826536379342 \\
2.51327412287183	1.02631095801424 \\
2.57610597594363	1.02317193922764 \\
2.63893782901543	1.01898514991205 \\
2.70176968208722	1.01651183249548 \\
2.76460153515902	1.01389419292263 \\
2.82743338823081	1.01028723394764 \\
2.89026524130261	1.00506161701333 \\
2.95309709437441	0.998435137125754 \\
3.0159289474462	0.989005448741333 \\
3.078760800518	0.979791585487984 \\
3.14159265358979	0.980402178946657 \\
3.20442450666159	0.981337268056434 \\
3.26725635973339	0.982386171802597 \\
3.33008821280518	0.987677034288272 \\
3.39292006587698	0.995267694266533 \\
3.45575191894877	1.00187306833252 \\
3.51858377202057	1.00809318543536 \\
3.58141562509236	1.01368678972052 \\
3.64424747816416	1.01846754869524 \\
3.70707933123596	1.01949628084926 \\
3.76991118430775	1.01915537856404 \\
3.83274303737955	1.02184552395701 \\
3.89557489045134	1.03059952235088 \\
3.95840674352314	1.03769880757187 \\
4.02123859659494	1.03583220839843 \\
4.08407044966673	0.8491194171094 \\
4.14690230273853	0.619890278801402 \\
4.20973415581032	0.384777954049187 \\
4.27256600888212	0.149778698568562 \\
4.33539786195391	0.000179643974417221 \\
4.39822971502571	1.14012046052436e-05 \\
4.46106156809751	0.000110099726554042 \\
4.5238934211693	0.000742638126660977 \\
4.5867252742411	4.75644510000706e-07 \\
4.64955712731289	0.00364973257143902 \\
4.71238898038469	0.000609829109754656 \\
4.77522083345649	0.00267996426369477 \\
4.83805268652828	0.00302687374038441 \\
4.90088453960008	0.00765553817268588 \\
4.96371639267187	3.63622632004734e-10 \\
5.02654824574367	0.00512419658332203 \\
5.08938009881547	0.00839114395008955 \\
5.15221195188726	0.11660151034399 \\
5.21504380495906	0.36411286999906 \\
5.27787565803085	0.62959192471426 \\
5.34070751110265	0.89159212351293 \\
5.40353936417444	1.00131160688525 \\
5.46637121724624	1.01476822847766 \\
5.52920307031804	1.02833878560434 \\
5.59203492338983	1.02826536379342 \\
5.65486677646163	1.02631095801424 \\
5.71769862953342	1.02317193922764 \\
5.78053048260522	1.01898514991205 \\
5.84336233567702	1.01651183249548 \\
5.90619418874881	1.01389419292263 \\
5.96902604182061	1.01028723394764 \\
6.0318578948924	1.00506161701333 \\
6.0946897479642	0.998435137125754 \\
6.157521601036	0.989005448741333 \\
6.22035345410779	0.979791585487984 \\
};
\addplot [very thick, color3, dashed]
table [row sep=\\]{%
0	0.986163772596366 \\
0.0628318530717959	0.9864393045449 \\
0.125663706143592	0.986415929224335 \\
0.188495559215388	0.987956177092845 \\
0.251327412287183	0.991016809721569 \\
0.314159265358979	0.995627649903408 \\
0.376991118430775	1.00164923941306 \\
0.439822971502571	1.00905781378773 \\
0.502654824574367	1.01771331183568 \\
0.565486677646163	1.0273101893607 \\
0.628318530717959	1.03782165678837 \\
0.691150383789754	1.04884839952228 \\
0.75398223686155	1.06259516955531 \\
0.816814089933346	1.07605601349895 \\
0.879645943005142	0.964824947358186 \\
0.942477796076938	0.817022219468989 \\
1.00530964914873	0.569167872825692 \\
1.06814150222053	0.335851912941846 \\
1.13097335529233	0.15079865662884 \\
1.19380520836412	0.0471436034685297 \\
1.25663706143592	0.0229817672528589 \\
1.31946891450771	0.00909352376615441 \\
1.38230076757951	0.00369982051708856 \\
1.4451326206513	0.00627384202590076 \\
1.5079644737231	0.00357248390304643 \\
1.5707963267949	1.63957736276643e-12 \\
1.63362817986669	0.00123293774840605 \\
1.69646003293849	0.00145463643554766 \\
1.75929188601028	0.00103145045876774 \\
1.82212373908208	0.00180112110516017 \\
1.88495559215388	0.0142856243227693 \\
1.94778744522567	0.0367467208209087 \\
2.01061929829747	0.149185536693362 \\
2.07345115136926	0.401816973110539 \\
2.13628300444106	0.641789639344179 \\
2.19911485751286	0.825016054530435 \\
2.26194671058465	0.95735677445413 \\
2.32477856365645	0.990132562358134 \\
2.38761041672824	1.01478823838754 \\
2.45044226980004	1.04158723026504 \\
2.51327412287183	1.0437698663917 \\
2.57610597594363	1.03349797055091 \\
2.63893782901543	1.02445060190717 \\
2.70176968208722	1.01666274798761 \\
2.76460153515902	1.00999901655709 \\
2.82743338823081	1.00274738692003 \\
2.89026524130261	0.996781191997676 \\
2.95309709437441	0.992123977635698 \\
3.0159289474462	0.988794123731501 \\
3.078760800518	0.98680477169718 \\
3.14159265358979	0.986163772596365 \\
3.20442450666159	0.9864393045449 \\
3.26725635973339	0.986415929224336 \\
3.33008821280518	0.987956177092846 \\
3.39292006587698	0.991016809721569 \\
3.45575191894877	0.995627649903408 \\
3.51858377202057	1.00164923941306 \\
3.58141562509236	1.00905781378773 \\
3.64424747816416	1.01771331183568 \\
3.70707933123596	1.0273101893607 \\
3.76991118430775	1.03782165678837 \\
3.83274303737955	1.04884839952228 \\
3.89557489045134	1.06259516955531 \\
3.95840674352314	1.07605601349895 \\
4.02123859659494	0.964824947358186 \\
4.08407044966673	0.817022219468989 \\
4.14690230273853	0.569167872825693 \\
4.20973415581032	0.335851912941848 \\
4.27256600888212	0.150798656628842 \\
4.33539786195391	0.0471436034685273 \\
4.39822971502571	0.0229817672528588 \\
4.46106156809751	0.00909352376615474 \\
4.5238934211693	0.00369982051708956 \\
4.5867252742411	0.00627384202590153 \\
4.64955712731289	0.00357248390304643 \\
4.71238898038469	1.63979940737136e-12 \\
4.77522083345649	0.00123293774840527 \\
4.83805268652828	0.00145463643554722 \\
4.90088453960008	0.00103145045876751 \\
4.96371639267187	0.00180112110516117 \\
5.02654824574367	0.01428562432277 \\
5.08938009881547	0.0367467208209077 \\
5.15221195188726	0.149185536693359 \\
5.21504380495906	0.401816973110537 \\
5.27787565803085	0.641789639344181 \\
5.34070751110265	0.825016054530435 \\
5.40353936417444	0.957356774454129 \\
5.46637121724624	0.990132562358134 \\
5.52920307031804	1.01478823838754 \\
5.59203492338983	1.04158723026504 \\
5.65486677646163	1.0437698663917 \\
5.71769862953342	1.03349797055091 \\
5.78053048260522	1.02445060190717 \\
5.84336233567702	1.01666274798761 \\
5.90619418874881	1.00999901655709 \\
5.96902604182061	1.00274738692003 \\
6.0318578948924	0.996781191997677 \\
6.0946897479642	0.992123977635698 \\
6.157521601036	0.988794123731501 \\
6.22035345410779	0.98680477169718 \\
};

\end{axis}

\end{tikzpicture}

%% file: figures/benchmark/benchmarkPL.tex
\begin{tikzpicture}

\definecolor{color0}{rgb}{0.12156862745098,0.466666666666667,0.705882352941177}
\definecolor{color1}{rgb}{1,0.498039215686275,0.0549019607843137}
\definecolor{color2}{rgb}{0.172549019607843,0.627450980392157,0.172549019607843}
\definecolor{color3}{rgb}{0.83921568627451,0.152941176470588,0.156862745098039}

\begin{axis}[
legend cell align={left},
tick align=outside,
tick pos=left,
x grid style={white!69.01960784313725!black},
xlabel={$x$},
xmin=-0.31101767270539, xmax=6.53137112681318,
y grid style={white!69.01960784313725!black},
ylabel={$y$},
ymin=-0.1, ymax=1.2,
ytick={-0.2,0,0.2,0.4,0.6,0.8,1,1.2},
yticklabels={−0.2,0.0,0.2,0.4,0.6,0.8,1.0,1.2}
]
\addplot [very thick, color0]
table [row sep=\\]{%
0	1 \\
0.0628318530717959	1 \\
0.125663706143592	1 \\
0.188495559215388	1 \\
0.251327412287183	1 \\
0.314159265358979	1 \\
0.376991118430775	1 \\
0.439822971502571	1 \\
0.502654824574367	1 \\
0.565486677646163	1 \\
0.628318530717959	1 \\
0.691150383789754	1 \\
0.75398223686155	1 \\
0.816814089933346	1 \\
0.879645943005142	1 \\
0.942477796076938	1 \\
1.00530964914873	1 \\
1.06814150222053	0 \\
1.13097335529233	0 \\
1.19380520836412	0 \\
1.25663706143592	0 \\
1.31946891450771	0 \\
1.38230076757951	0 \\
1.4451326206513	0 \\
1.5079644737231	0 \\
1.5707963267949	0 \\
1.63362817986669	0 \\
1.69646003293849	0 \\
1.75929188601028	0 \\
1.82212373908208	0 \\
1.88495559215388	0 \\
1.94778744522567	0 \\
2.01061929829747	0 \\
2.07345115136926	0 \\
2.13628300444106	1 \\
2.19911485751286	1 \\
2.26194671058465	1 \\
2.32477856365645	1 \\
2.38761041672824	1 \\
2.45044226980004	1 \\
2.51327412287183	1 \\
2.57610597594363	1 \\
2.63893782901543	1 \\
2.70176968208722	1 \\
2.76460153515902	1 \\
2.82743338823081	1 \\
2.89026524130261	1 \\
2.95309709437441	1 \\
3.0159289474462	1 \\
3.078760800518	1 \\
3.14159265358979	1 \\
3.20442450666159	1 \\
3.26725635973339	1 \\
3.33008821280518	1 \\
3.39292006587698	1 \\
3.45575191894877	1 \\
3.51858377202057	1 \\
3.58141562509236	1 \\
3.64424747816416	1 \\
3.70707933123596	1 \\
3.76991118430775	1 \\
3.83274303737955	1 \\
3.89557489045134	1 \\
3.95840674352314	1 \\
4.02123859659494	1 \\
4.08407044966673	1 \\
4.14690230273853	1 \\
4.20973415581032	0 \\
4.27256600888212	0 \\
4.33539786195391	0 \\
4.39822971502571	0 \\
4.46106156809751	0 \\
4.5238934211693	0 \\
4.5867252742411	0 \\
4.64955712731289	0 \\
4.71238898038469	0 \\
4.77522083345649	0 \\
4.83805268652828	0 \\
4.90088453960008	0 \\
4.96371639267187	0 \\
5.02654824574367	0 \\
5.08938009881547	0 \\
5.15221195188726	0 \\
5.21504380495906	0 \\
5.27787565803085	1 \\
5.34070751110265	1 \\
5.40353936417444	1 \\
5.46637121724624	1 \\
5.52920307031804	1 \\
5.59203492338983	1 \\
5.65486677646163	1 \\
5.71769862953342	1 \\
5.78053048260522	1 \\
5.84336233567702	1 \\
5.90619418874881	1 \\
5.96902604182061	1 \\
6.0318578948924	1 \\
6.0946897479642	1 \\
6.157521601036	1 \\
6.22035345410779	1 \\
};
\addplot [very thick, color1, dashed]
table [row sep=\\]{%
0	0.863174747820741 \\
0.0628318530717959	0.897562724557258 \\
0.125663706143592	0.931950701293887 \\
0.188495559215388	0.966338678030516 \\
0.251327412287183	1.00072665476715 \\
0.314159265358979	1.03511463150377 \\
0.376991118430775	1.0695026082404 \\
0.439822971502571	1.10389058497703 \\
0.502654824574367	1.13827856171366 \\
0.565486677646163	1.17266653845029 \\
0.628318530717959	1.20705451518692 \\
0.691150383789754	1.08898209825456 \\
0.75398223686155	0.970909681321856 \\
0.816814089933346	0.852837264389149 \\
0.879645943005142	0.734764847456441 \\
0.942477796076938	0.616692430523733 \\
1.00530964914873	0.498620013591026 \\
1.06814150222053	0.380547596658319 \\
1.13097335529233	0.26247517972561 \\
1.19380520836412	0.144402762792903 \\
1.25663706143592	0.0263303458601946 \\
1.31946891450771	0.0265882176546385 \\
1.38230076757951	0.0268460894494331 \\
1.4451326206513	0.0271039612442277 \\
1.5079644737231	0.0273618330390223 \\
1.5707963267949	0.0276197048338169 \\
1.63362817986669	0.0278775766286115 \\
1.69646003293849	0.0281354484234062 \\
1.75929188601028	0.0283933202182008 \\
1.82212373908208	0.0286511920129954 \\
1.88495559215388	0.02890906380779 \\
1.94778744522567	0.146488043690531 \\
2.01061929829747	0.264067023573699 \\
2.07345115136926	0.381646003456865 \\
2.13628300444106	0.499224983340031 \\
2.19911485751286	0.616803963223199 \\
2.26194671058465	0.734382943106366 \\
2.32477856365645	0.851961922989533 \\
2.38761041672824	0.969540902872699 \\
2.45044226980004	1.08711988275587 \\
2.51327412287183	1.20469886263903 \\
2.57610597594363	1.1705464511578 \\
2.63893782901543	1.13639403967592 \\
2.70176968208722	1.10224162819403 \\
2.76460153515902	1.06808921671215 \\
2.82743338823081	1.03393680523027 \\
2.89026524130261	0.999784393748385 \\
2.95309709437441	0.965631982266502 \\
3.0159289474462	0.931479570784619 \\
3.078760800518	0.897327159302736 \\
3.14159265358979	0.863174747820741 \\
3.20442450666159	0.897562724557258 \\
3.26725635973339	0.931950701293887 \\
3.33008821280518	0.966338678030516 \\
3.39292006587698	1.00072665476715 \\
3.45575191894877	1.03511463150377 \\
3.51858377202057	1.0695026082404 \\
3.58141562509236	1.10389058497703 \\
3.64424747816416	1.13827856171366 \\
3.70707933123596	1.17266653845029 \\
3.76991118430775	1.20705451518692 \\
3.83274303737955	1.08898209825456 \\
3.89557489045134	0.970909681321856 \\
3.95840674352314	0.852837264389149 \\
4.02123859659494	0.734764847456441 \\
4.08407044966673	0.616692430523733 \\
4.14690230273853	0.498620013591026 \\
4.20973415581032	0.380547596658317 \\
4.27256600888212	0.26247517972561 \\
4.33539786195391	0.144402762792902 \\
4.39822971502571	0.0263303458601946 \\
4.46106156809751	0.0265882176546385 \\
4.5238934211693	0.0268460894494331 \\
4.5867252742411	0.0271039612442277 \\
4.64955712731289	0.0273618330390223 \\
4.71238898038469	0.0276197048338169 \\
4.77522083345649	0.0278775766286115 \\
4.83805268652828	0.0281354484234062 \\
4.90088453960008	0.0283933202182008 \\
4.96371639267187	0.0286511920129954 \\
5.02654824574367	0.02890906380779 \\
5.08938009881547	0.146488043690532 \\
5.15221195188726	0.264067023573699 \\
5.21504380495906	0.381646003456865 \\
5.27787565803085	0.499224983340033 \\
5.34070751110265	0.616803963223199 \\
5.40353936417444	0.734382943106367 \\
5.46637121724624	0.851961922989533 \\
5.52920307031804	0.969540902872699 \\
5.59203492338983	1.08711988275587 \\
5.65486677646163	1.20469886263903 \\
5.71769862953342	1.1705464511578 \\
5.78053048260522	1.13639403967592 \\
5.84336233567702	1.10224162819403 \\
5.90619418874881	1.06808921671215 \\
5.96902604182061	1.03393680523027 \\
6.0318578948924	0.999784393748385 \\
6.0946897479642	0.965631982266502 \\
6.157521601036	0.931479570784619 \\
6.22035345410779	0.897327159302736 \\
};
\addplot [very thick, color2, dashed]
table [row sep=\\]{%
0	1.08218392489932 \\
0.0628318530717959	1.04419345578755 \\
0.125663706143592	1.00620298667567 \\
0.188495559215388	0.968212517563779 \\
0.251327412287183	0.930222048451891 \\
0.314159265358979	0.892231579340004 \\
0.376991118430775	0.951543936938535 \\
0.439822971502571	1.01085629453726 \\
0.502654824574367	1.07016865213598 \\
0.565486677646163	1.12948100973471 \\
0.628318530717959	1.18879336733343 \\
0.691150383789754	1.09271982075801 \\
0.75398223686155	0.996646274182232 \\
0.816814089933346	0.900572727606454 \\
0.879645943005142	0.804499181030676 \\
0.942477796076938	0.708425634454898 \\
1.00530964914873	0.56674050806033 \\
1.06814150222053	0.425055381665642 \\
1.13097335529233	0.283370255270954 \\
1.19380520836412	0.141685128876266 \\
1.25663706143592	2.48157721260099e-09 \\
1.31946891450771	2.18248630111254e-09 \\
1.38230076757951	1.88381520067416e-09 \\
1.4451326206513	1.58514410026288e-09 \\
1.5079644737231	1.28647299990582e-09 \\
1.5707963267949	9.87801899440333e-10 \\
1.63362817986669	4.20535144820275e-10 \\
1.69646003293849	1.46731663007004e-10 \\
1.75929188601028	7.13998470888494e-10 \\
1.82212373908208	1.28126527871577e-09 \\
1.88495559215388	1.84853208654305e-09 \\
1.94778744522567	0.141681183982856 \\
2.01061929829747	0.283362369814758 \\
2.07345115136926	0.425043555646659 \\
2.13628300444106	0.566724741478561 \\
2.19911485751286	0.708405927310463 \\
2.26194671058465	0.803925852918479 \\
2.32477856365645	0.899445778526311 \\
2.38761041672824	0.994965704134142 \\
2.45044226980004	1.09048562974197 \\
2.51327412287183	1.18600555534981 \\
2.57610597594363	1.12777681489191 \\
2.63893782901543	1.06954807443335 \\
2.70176968208722	1.0113193339748 \\
2.76460153515902	0.953090593516247 \\
2.82743338823081	0.894861853057693 \\
2.89026524130261	0.932326267425639 \\
2.95309709437441	0.969790681794028 \\
3.0159289474462	1.00725509616242 \\
3.078760800518	1.0447195105308 \\
3.14159265358979	1.08218392489932 \\
3.20442450666159	1.04419345578755 \\
3.26725635973339	1.00620298667567 \\
3.33008821280518	0.968212517563779 \\
3.39292006587698	0.930222048451891 \\
3.45575191894877	0.892231579340004 \\
3.51858377202057	0.951543936938535 \\
3.58141562509236	1.01085629453726 \\
3.64424747816416	1.07016865213598 \\
3.70707933123596	1.12948100973471 \\
3.76991118430775	1.18879336733343 \\
3.83274303737955	1.09271982075801 \\
3.89557489045134	0.996646274182232 \\
3.95840674352314	0.900572727606454 \\
4.02123859659494	0.804499181030675 \\
4.08407044966673	0.708425634454898 \\
4.14690230273853	0.566740508060331 \\
4.20973415581032	0.425055381665641 \\
4.27256600888212	0.283370255270954 \\
4.33539786195391	0.141685128876265 \\
4.39822971502571	2.48157721260099e-09 \\
4.46106156809751	2.18248630116675e-09 \\
4.5238934211693	1.88381520075547e-09 \\
4.5867252742411	1.58514410031709e-09 \\
4.64955712731289	1.28647299990582e-09 \\
4.71238898038469	9.87801899494543e-10 \\
4.77522083345649	4.2053514473896e-10 \\
4.83805268652828	1.46731663088319e-10 \\
4.90088453960008	7.13998470915599e-10 \\
4.96371639267187	1.28126527868867e-09 \\
5.02654824574367	1.84853208654305e-09 \\
5.08938009881547	0.141681183982857 \\
5.15221195188726	0.283362369814758 \\
5.21504380495906	0.425043555646659 \\
5.27787565803085	0.566724741478562 \\
5.34070751110265	0.708405927310463 \\
5.40353936417444	0.80392585291848 \\
5.46637121724624	0.899445778526311 \\
5.52920307031804	0.994965704134142 \\
5.59203492338983	1.09048562974197 \\
5.65486677646163	1.18600555534981 \\
5.71769862953342	1.12777681489191 \\
5.78053048260522	1.06954807443335 \\
5.84336233567702	1.0113193339748 \\
5.90619418874881	0.953090593516246 \\
5.96902604182061	0.894861853057693 \\
6.0318578948924	0.93232626742564 \\
6.0946897479642	0.969790681794028 \\
6.157521601036	1.00725509616242 \\
6.22035345410779	1.0447195105308 \\
};
\addplot [very thick, color3, dashed]
table [row sep=\\]{%
0	1.00168388348922 \\
0.0628318530717959	1.00107971689112 \\
0.125663706143592	1.00047555029302 \\
0.188495559215388	0.999804619189346 \\
0.251327412287183	0.999066923580089 \\
0.314159265358979	0.998329227970833 \\
0.376991118430775	0.997162553165785 \\
0.439822971502571	0.995995878360735 \\
0.502654824574367	1.00163358012971 \\
0.565486677646163	1.01407565847276 \\
0.628318530717959	1.02651773681581 \\
0.691150383789754	0.983283907736955 \\
0.75398223686155	0.940050078657977 \\
0.816814089933346	0.971369126119285 \\
0.879645943005142	1.07724105012162 \\
0.942477796076938	1.18311297412395 \\
1.00530964914873	0.71458842330624 \\
1.06814150222053	0.24606387248702 \\
1.13097335529233	0.00944127869264875 \\
1.19380520836412	0.00472064192573257 \\
1.25663706143592	5.15881635198038e-09 \\
1.31946891450771	1.1694420603333e-09 \\
1.38230076757951	2.81991753109086e-09 \\
1.4451326206513	5.13124058485761e-09 \\
1.5079644737231	5.76452965513719e-09 \\
1.5707963267949	6.39781872168984e-09 \\
1.63362817986669	7.37242292575221e-09 \\
1.69646003293849	8.34702608402027e-09 \\
1.75929188601028	4.407870983104e-09 \\
1.82212373908208	4.44504250571851e-09 \\
1.88495559215388	1.329795599345e-08 \\
1.94778744522567	0.00446063392534878 \\
2.01061929829747	0.008921254552757 \\
2.07345115136926	0.245971733108716 \\
2.13628300444106	0.715612069596757 \\
2.19911485751286	1.1852524060848 \\
2.26194671058465	1.07744004710228 \\
2.32477856365645	0.969627688117463 \\
2.38761041672824	0.938906874522026 \\
2.45044226980004	0.985277606317233 \\
2.51327412287183	1.03164833811244 \\
2.57610597594363	1.01467158331985 \\
2.63893782901543	0.997694828527001 \\
2.70176968208722	0.991546105033281 \\
2.76460153515902	0.996225412838879 \\
2.82743338823081	1.00090472064448 \\
2.89026524130261	1.00004735807913 \\
2.95309709437441	0.999189995513762 \\
3.0159289474462	0.999345828082693 \\
3.078760800518	1.00051485578595 \\
3.14159265358979	1.00168388348922 \\
3.20442450666159	1.00107971689112 \\
3.26725635973339	1.00047555029302 \\
3.33008821280518	0.999804619189346 \\
3.39292006587698	0.999066923580089 \\
3.45575191894877	0.998329227970833 \\
3.51858377202057	0.997162553165785 \\
3.58141562509236	0.995995878360735 \\
3.64424747816416	1.00163358012971 \\
3.70707933123596	1.01407565847276 \\
3.76991118430775	1.02651773681581 \\
3.83274303737955	0.983283907736955 \\
3.89557489045134	0.940050078657977 \\
3.95840674352314	0.971369126119285 \\
4.02123859659494	1.07724105012162 \\
4.08407044966673	1.18311297412395 \\
4.14690230273853	0.714588423306243 \\
4.20973415581032	0.246063872487016 \\
4.27256600888212	0.00944127869264876 \\
4.33539786195391	0.00472064192573252 \\
4.39822971502571	5.15881635198038e-09 \\
4.46106156809751	1.16944206163434e-09 \\
4.5238934211693	2.81991752978982e-09 \\
4.5867252742411	5.13124058575207e-09 \\
4.64955712731289	5.76452965513719e-09 \\
4.71238898038469	6.39781872358719e-09 \\
4.77522083345649	7.372422925183e-09 \\
4.83805268652828	8.34702608347817e-09 \\
4.90088453960008	4.4078709828465e-09 \\
4.96371639267187	4.44504250545424e-09 \\
5.02654824574367	1.32979559937143e-08 \\
5.08938009881547	0.00446063392534882 \\
5.15221195188726	0.008921254552757 \\
5.21504380495906	0.245971733108716 \\
5.27787565803085	0.715612069596762 \\
5.34070751110265	1.1852524060848 \\
5.40353936417444	1.07744004710227 \\
5.46637121724624	0.969627688117463 \\
5.52920307031804	0.938906874522026 \\
5.59203492338983	0.985277606317234 \\
5.65486677646163	1.03164833811244 \\
5.71769862953342	1.01467158331985 \\
5.78053048260522	0.997694828527001 \\
5.84336233567702	0.991546105033281 \\
5.90619418874881	0.996225412838879 \\
5.96902604182061	1.00090472064448 \\
6.0318578948924	1.00004735807913 \\
6.0946897479642	0.999189995513762 \\
6.157521601036	0.999345828082693 \\
6.22035345410779	1.00051485578595 \\
};

\end{axis}

\end{tikzpicture}

%% file: figures/benchmark/benchmarkRBF.tex
\begin{tikzpicture}

\definecolor{color0}{rgb}{0.12156862745098,0.466666666666667,0.705882352941177}
\definecolor{color1}{rgb}{1,0.498039215686275,0.0549019607843137}
\definecolor{color2}{rgb}{0.172549019607843,0.627450980392157,0.172549019607843}
\definecolor{color3}{rgb}{0.83921568627451,0.152941176470588,0.156862745098039}

\begin{axis}[
legend cell align={left},
tick align=outside,
tick pos=left,
x grid style={white!69.01960784313725!black},
xlabel={$x$},
xmin=-0.31101767270539, xmax=6.53137112681318,
y grid style={white!69.01960784313725!black},
ylabel={$y$},
ymin=-0.1, ymax=1.2,
ytick={-0.2,0,0.2,0.4,0.6,0.8,1,1.2},
yticklabels={−0.2,0.0,0.2,0.4,0.6,0.8,1.0,1.2}
]
\addplot [very thick, color0]
table [row sep=\\]{%
0	1 \\
0.0628318530717959	1 \\
0.125663706143592	1 \\
0.188495559215388	1 \\
0.251327412287183	1 \\
0.314159265358979	1 \\
0.376991118430775	1 \\
0.439822971502571	1 \\
0.502654824574367	1 \\
0.565486677646163	1 \\
0.628318530717959	1 \\
0.691150383789754	1 \\
0.75398223686155	1 \\
0.816814089933346	1 \\
0.879645943005142	1 \\
0.942477796076938	1 \\
1.00530964914873	1 \\
1.06814150222053	0 \\
1.13097335529233	0 \\
1.19380520836412	0 \\
1.25663706143592	0 \\
1.31946891450771	0 \\
1.38230076757951	0 \\
1.4451326206513	0 \\
1.5079644737231	0 \\
1.5707963267949	0 \\
1.63362817986669	0 \\
1.69646003293849	0 \\
1.75929188601028	0 \\
1.82212373908208	0 \\
1.88495559215388	0 \\
1.94778744522567	0 \\
2.01061929829747	0 \\
2.07345115136926	0 \\
2.13628300444106	1 \\
2.19911485751286	1 \\
2.26194671058465	1 \\
2.32477856365645	1 \\
2.38761041672824	1 \\
2.45044226980004	1 \\
2.51327412287183	1 \\
2.57610597594363	1 \\
2.63893782901543	1 \\
2.70176968208722	1 \\
2.76460153515902	1 \\
2.82743338823081	1 \\
2.89026524130261	1 \\
2.95309709437441	1 \\
3.0159289474462	1 \\
3.078760800518	1 \\
3.14159265358979	1 \\
3.20442450666159	1 \\
3.26725635973339	1 \\
3.33008821280518	1 \\
3.39292006587698	1 \\
3.45575191894877	1 \\
3.51858377202057	1 \\
3.58141562509236	1 \\
3.64424747816416	1 \\
3.70707933123596	1 \\
3.76991118430775	1 \\
3.83274303737955	1 \\
3.89557489045134	1 \\
3.95840674352314	1 \\
4.02123859659494	1 \\
4.08407044966673	1 \\
4.14690230273853	1 \\
4.20973415581032	0 \\
4.27256600888212	0 \\
4.33539786195391	0 \\
4.39822971502571	0 \\
4.46106156809751	0 \\
4.5238934211693	0 \\
4.5867252742411	0 \\
4.64955712731289	0 \\
4.71238898038469	0 \\
4.77522083345649	0 \\
4.83805268652828	0 \\
4.90088453960008	0 \\
4.96371639267187	0 \\
5.02654824574367	0 \\
5.08938009881547	0 \\
5.15221195188726	0 \\
5.21504380495906	0 \\
5.27787565803085	1 \\
5.34070751110265	1 \\
5.40353936417444	1 \\
5.46637121724624	1 \\
5.52920307031804	1 \\
5.59203492338983	1 \\
5.65486677646163	1 \\
5.71769862953342	1 \\
5.78053048260522	1 \\
5.84336233567702	1 \\
5.90619418874881	1 \\
5.96902604182061	1 \\
6.0318578948924	1 \\
6.0946897479642	1 \\
6.157521601036	1 \\
6.22035345410779	1 \\
};
\addplot [very thick, color1, dashed]
table [row sep=\\]{%
0	0.870777878805583 \\
0.0628318530717959	0.90270635110919 \\
0.125663706143592	0.938758209035443 \\
0.188495559215388	0.977407680127029 \\
0.251327412287183	1.0163624767633 \\
0.314159265358979	1.0526005818335 \\
0.376991118430775	1.08247334380129 \\
0.439822971502571	1.10192321535894 \\
0.502654824574367	1.10686500277822 \\
0.565486677646163	1.09373190093077 \\
0.628318530717959	1.06009644888761 \\
0.691150383789754	1.00519113257018 \\
0.75398223686155	0.930146197047569 \\
0.816814089933346	0.837864000169572 \\
0.879645943005142	0.732601393149801 \\
0.942477796076938	0.619428074671559 \\
1.00530964914873	0.503713466928509 \\
1.06814150222053	0.390704046180599 \\
1.13097335529233	0.285169522005971 \\
1.19380520836412	0.19107811317575 \\
1.25663706143592	0.111310022792704 \\
1.31946891450771	0.0474851181421706 \\
1.38230076757951	2.76736966675628e-12 \\
1.4451326206513	0.0316899824183995 \\
1.5079644737231	0.0486116727109295 \\
1.5707963267949	0.051782082870463 \\
1.63362817986669	0.0418342340230503 \\
1.69646003293849	0.0188621673824403 \\
1.75929188601028	0.0174831850341054 \\
1.82212373908208	0.0676839636778223 \\
1.88495559215388	0.131928752608004 \\
1.94778744522567	0.209705948056326 \\
2.01061929829747	0.299544226112052 \\
2.07345115136926	0.398960267108323 \\
2.13628300444106	0.504585372308278 \\
2.19911485751286	0.612391881315037 \\
2.26194671058465	0.717956231195624 \\
2.32477856365645	0.816751571540115 \\
2.38761041672824	0.904503455474494 \\
2.45044226980004	0.977625837318068 \\
2.51327412287183	1.03368283991926 \\
2.57610597594363	1.07174429074447 \\
2.63893782901543	1.09249023861382 \\
2.70176968208722	1.09800290266784 \\
2.76460153515902	1.09131562089892 \\
2.82743338823081	1.07587640098271 \\
2.89026524130261	1.05507830866363 \\
2.95309709437441	1.03193608400699 \\
3.0159289474462	1.00890952026743 \\
3.078760800518	0.987830852866715 \\
3.14159265358979	0.870777878805584 \\
3.20442450666159	0.90270635110919 \\
3.26725635973339	0.938758209035444 \\
3.33008821280518	0.977407680127029 \\
3.39292006587698	1.0163624767633 \\
3.45575191894877	1.0526005818335 \\
3.51858377202057	1.08247334380129 \\
3.58141562509236	1.10192321535894 \\
3.64424747816416	1.10686500277822 \\
3.70707933123596	1.09373190093077 \\
3.76991118430775	1.06009644888761 \\
3.83274303737955	1.00519113257018 \\
3.89557489045134	0.930146197047569 \\
3.95840674352314	0.837864000169572 \\
4.02123859659494	0.7326013931498 \\
4.08407044966673	0.61942807467156 \\
4.14690230273853	0.503713466928509 \\
4.20973415581032	0.390704046180598 \\
4.27256600888212	0.285169522005972 \\
4.33539786195391	0.191078113175749 \\
4.39822971502571	0.111310022792704 \\
4.46106156809751	0.0474851181421711 \\
4.5238934211693	2.76770273366367e-12 \\
4.5867252742411	0.0316899824183993 \\
4.64955712731289	0.0486116727109295 \\
4.71238898038469	0.051782082870463 \\
4.77522083345649	0.0418342340230507 \\
4.83805268652828	0.0188621673824402 \\
4.90088453960008	0.0174831850341053 \\
4.96371639267187	0.0676839636778222 \\
5.02654824574367	0.131928752608003 \\
5.08938009881547	0.209705948056327 \\
5.15221195188726	0.299544226112052 \\
5.21504380495906	0.398960267108323 \\
5.27787565803085	0.504585372308279 \\
5.34070751110265	0.612391881315037 \\
5.40353936417444	0.717956231195625 \\
5.46637121724624	0.816751571540115 \\
5.52920307031804	0.904503455474494 \\
5.59203492338983	0.977625837318068 \\
5.65486677646163	1.03368283991926 \\
5.71769862953342	1.07174429074447 \\
5.78053048260522	1.09249023861382 \\
5.84336233567702	1.09800290266784 \\
5.90619418874881	1.09131562089892 \\
5.96902604182061	1.07587640098271 \\
6.0318578948924	1.05507830866363 \\
6.0946897479642	1.03193608400699 \\
6.157521601036	1.00890952026743 \\
6.22035345410779	0.987830852866715 \\
};
\addplot [very thick, color2, dashed]
table [row sep=\\]{%
0	0.702125343701208 \\
0.0628318530717959	0.713940816123301 \\
0.125663706143592	0.723859833981633 \\
0.188495559215388	0.748764041622906 \\
0.251327412287183	0.809043763925118 \\
0.314159265358979	0.921002409839422 \\
0.376991118430775	1.08576849811589 \\
0.439822971502571	1.28014454380409 \\
0.502654824574367	1.45551080737678 \\
0.565486677646163	1.54584104037039 \\
0.628318530717959	1.48904613319498 \\
0.691150383789754	1.26065465903926 \\
0.75398223686155	0.894100529264505 \\
0.816814089933346	0.466375203796971 \\
0.879645943005142	0.066442165751068 \\
0.942477796076938	0.233010965109827 \\
1.00530964914873	0.398402679105608 \\
1.06814150222053	0.442198900763698 \\
1.13097335529233	0.404182096368572 \\
1.19380520836412	0.324835893983221 \\
1.25663706143592	0.232061669818243 \\
1.31946891450771	0.141555143439321 \\
1.38230076757951	0.0620486739777849 \\
1.4451326206513	6.67508802576489e-06 \\
1.5079644737231	0.0381527440374876 \\
1.5707963267949	0.0473948645684901 \\
1.63362817986669	0.0285318217142756 \\
1.69646003293849	0.00790445709740684 \\
1.75929188601028	0.0417541288722459 \\
1.82212373908208	0.047040046064006 \\
1.88495559215388	1.49490142486997e-11 \\
1.94778744522567	0.109047032477221 \\
2.01061929829747	0.268679598728412 \\
2.07345115136926	0.451493053895868 \\
2.13628300444106	0.625339638770691 \\
2.19911485751286	0.763718081161188 \\
2.26194671058465	0.854561410555788 \\
2.32477856365645	0.903037159420749 \\
2.38761041672824	0.924405493306498 \\
2.45044226980004	0.933832551990777 \\
2.51327412287183	0.941493736278383 \\
2.57610597594363	0.953042359465916 \\
2.63893782901543	0.97154411106107 \\
2.70176968208722	0.998522187166266 \\
2.76460153515902	1.03420860336415 \\
2.82743338823081	1.07785613906932 \\
2.89026524130261	1.12817111461706 \\
2.95309709437441	1.18297797165209 \\
3.0159289474462	1.23776560065216 \\
3.078760800518	1.28488254064649 \\
3.14159265358979	0.702125343701209 \\
3.20442450666159	0.713940816123301 \\
3.26725635973339	0.723859833981633 \\
3.33008821280518	0.748764041622906 \\
3.39292006587698	0.809043763925118 \\
3.45575191894877	0.921002409839422 \\
3.51858377202057	1.08576849811589 \\
3.58141562509236	1.28014454380409 \\
3.64424747816416	1.45551080737678 \\
3.70707933123596	1.54584104037039 \\
3.76991118430775	1.48904613319498 \\
3.83274303737955	1.26065465903926 \\
3.89557489045134	0.894100529264504 \\
3.95840674352314	0.466375203796971 \\
4.02123859659494	0.0664421657510673 \\
4.08407044966673	0.233010965109826 \\
4.14690230273853	0.398402679105607 \\
4.20973415581032	0.442198900763699 \\
4.27256600888212	0.404182096368572 \\
4.33539786195391	0.32483589398322 \\
4.39822971502571	0.232061669818243 \\
4.46106156809751	0.141555143439321 \\
4.5238934211693	0.0620486739777847 \\
4.5867252742411	6.67508802587591e-06 \\
4.64955712731289	0.0381527440374876 \\
4.71238898038469	0.0473948645684901 \\
4.77522083345649	0.0285318217142754 \\
4.83805268652828	0.00790445709740778 \\
4.90088453960008	0.0417541288722459 \\
4.96371639267187	0.0470400460640059 \\
5.02654824574367	1.4949347315607e-11 \\
5.08938009881547	0.109047032477222 \\
5.15221195188726	0.268679598728412 \\
5.21504380495906	0.451493053895868 \\
5.27787565803085	0.625339638770692 \\
5.34070751110265	0.763718081161188 \\
5.40353936417444	0.854561410555788 \\
5.46637121724624	0.903037159420749 \\
5.52920307031804	0.924405493306498 \\
5.59203492338983	0.933832551990776 \\
5.65486677646163	0.941493736278383 \\
5.71769862953342	0.953042359465917 \\
5.78053048260522	0.97154411106107 \\
5.84336233567702	0.998522187166267 \\
5.90619418874881	1.03420860336415 \\
5.96902604182061	1.07785613906932 \\
6.0318578948924	1.12817111461706 \\
6.0946897479642	1.18297797165209 \\
6.157521601036	1.23776560065217 \\
6.22035345410779	1.28488254064649 \\
};
\addplot [very thick, color3, dashed]
table [row sep=\\]{%
0	0.415292984872412 \\
0.0628318530717959	0.554363392103623 \\
0.125663706143592	0.787272555086925 \\
0.188495559215388	1.08765046126318 \\
0.251327412287183	1.38401770245074 \\
0.314159265358979	1.47186073025686 \\
0.376991118430775	1.17476283695488 \\
0.439822971502571	0.612378579700601 \\
0.502654824574367	0.0738179899144771 \\
0.565486677646163	0.79475296718658 \\
0.628318530717959	1.36917907988468 \\
0.691150383789754	1.64420506081331 \\
0.75398223686155	1.6183830641683 \\
0.816814089933346	1.22326604124454 \\
0.879645943005142	0.542098649677004 \\
0.942477796076938	0.0790765492999952 \\
1.00530964914873	0.417387665156907 \\
1.06814150222053	0.511469230951373 \\
1.13097335529233	0.409841217242174 \\
1.19380520836412	0.184673286956726 \\
1.25663706143592	0.000877021192545124 \\
1.31946891450771	0.0516738044427533 \\
1.38230076757951	0.0213070192602739 \\
1.4451326206513	0.00518943377475889 \\
1.5079644737231	0.00524942806775247 \\
1.5707963267949	0.00110899914179607 \\
1.63362817986669	2.43810027722446e-10 \\
1.69646003293849	0.00437014355527171 \\
1.75929188601028	0.00104115004375124 \\
1.82212373908208	0.00539860348235871 \\
1.88495559215388	0.00572169845540935 \\
1.94778744522567	0.0317644868853246 \\
2.01061929829747	0.00673690453325773 \\
2.07345115136926	0.238183504135984 \\
2.13628300444106	0.637881600479272 \\
2.19911485751286	0.985006232515403 \\
2.26194671058465	1.12221419125249 \\
2.32477856365645	1.10038417826911 \\
2.38761041672824	0.983678254293398 \\
2.45044226980004	0.847632215975384 \\
2.51327412287183	0.847412926217127 \\
2.57610597594363	1.06175381376958 \\
2.63893782901543	1.32456534288315 \\
2.70176968208722	1.34851998567556 \\
2.76460153515902	1.05080885747527 \\
2.82743338823081	0.482104504829824 \\
2.89026524130261	0.335913062061231 \\
2.95309709437441	1.15655279983364 \\
3.0159289474462	1.61716991550185 \\
3.078760800518	1.69306883681684 \\
3.14159265358979	0.415292984872412 \\
3.20442450666159	0.554363392103622 \\
3.26725635973339	0.787272555086927 \\
3.33008821280518	1.08765046126318 \\
3.39292006587698	1.38401770245074 \\
3.45575191894877	1.47186073025686 \\
3.51858377202057	1.17476283695488 \\
3.58141562509236	0.612378579700596 \\
3.64424747816416	0.0738179899144777 \\
3.70707933123596	0.794752967186576 \\
3.76991118430775	1.36917907988468 \\
3.83274303737955	1.64420506081331 \\
3.89557489045134	1.6183830641683 \\
3.95840674352314	1.22326604124454 \\
4.02123859659494	0.542098649677003 \\
4.08407044966673	0.0790765492999942 \\
4.14690230273853	0.417387665156905 \\
4.20973415581032	0.511469230951373 \\
4.27256600888212	0.409841217242174 \\
4.33539786195391	0.184673286956724 \\
4.39822971502571	0.000877021192545124 \\
4.46106156809751	0.0516738044427534 \\
4.5238934211693	0.0213070192602735 \\
4.5867252742411	0.00518943377475856 \\
4.64955712731289	0.00524942806775247 \\
4.71238898038469	0.00110899914179607 \\
4.77522083345649	2.43809972211295e-10 \\
4.83805268652828	0.00437014355527154 \\
4.90088453960008	0.00104115004375135 \\
4.96371639267187	0.00539860348235893 \\
5.02654824574367	0.00572169845540946 \\
5.08938009881547	0.0317644868853247 \\
5.15221195188726	0.00673690453325773 \\
5.21504380495906	0.238183504135984 \\
5.27787565803085	0.637881600479277 \\
5.34070751110265	0.985006232515403 \\
5.40353936417444	1.12221419125249 \\
5.46637121724624	1.10038417826911 \\
5.52920307031804	0.983678254293398 \\
5.59203492338983	0.847632215975383 \\
5.65486677646163	0.847412926217127 \\
5.71769862953342	1.06175381376958 \\
5.78053048260522	1.32456534288315 \\
5.84336233567702	1.34851998567556 \\
5.90619418874881	1.05080885747527 \\
5.96902604182061	0.482104504829824 \\
6.0318578948924	0.335913062061241 \\
6.0946897479642	1.15655279983364 \\
6.157521601036	1.61716991550185 \\
6.22035345410779	1.69306883681684 \\
};

\end{axis}

\end{tikzpicture}

%% file: figures/benchmark/benchmarkNN2.tex
\begin{tikzpicture}

\definecolor{color0}{rgb}{0.12156862745098,0.466666666666667,0.705882352941177}
\definecolor{color1}{rgb}{1,0.498039215686275,0.0549019607843137}
\definecolor{color2}{rgb}{0.172549019607843,0.627450980392157,0.172549019607843}
\definecolor{color3}{rgb}{0.83921568627451,0.152941176470588,0.156862745098039}

\begin{axis}[
legend cell align={left},
legend entries={{Exact},{NN5},{NN10},{NN20}},
legend style={at={(0.03,0.03)}, anchor=south west, draw=white!80.0!black},
tick align=outside,
tick pos=left,
x grid style={white!69.01960784313725!black},
xlabel={$x$},
xmin=-0.31101767270539, xmax=6.53137112681318,
y grid style={white!69.01960784313725!black},
ylabel={$y$},
ymin=-0.1, ymax=1.2,
ytick={-0.2,0,0.2,0.4,0.6,0.8,1,1.2},
yticklabels={−0.2,0.0,0.2,0.4,0.6,0.8,1.0,1.2}
]
\addlegendimage{very thick, color0}
\addlegendimage{very thick, dashed, color1}
\addlegendimage{very thick, dashed,color2}
\addlegendimage{very thick, dashed,color3}
\addplot [very thick, color0]
table [row sep=\\]{%
0	1 \\
0.0628318530717959	1 \\
0.125663706143592	1 \\
0.188495559215388	1 \\
0.251327412287183	1 \\
0.314159265358979	1 \\
0.376991118430775	1 \\
0.439822971502571	1 \\
0.502654824574367	1 \\
0.565486677646163	1 \\
0.628318530717959	1 \\
0.691150383789754	1 \\
0.75398223686155	1 \\
0.816814089933346	1 \\
0.879645943005142	1 \\
0.942477796076938	1 \\
1.00530964914873	1 \\
1.06814150222053	1 \\
1.13097335529233	1 \\
1.19380520836412	1 \\
1.25663706143592	1 \\
1.31946891450771	1 \\
1.38230076757951	1 \\
1.4451326206513	1 \\
1.5079644737231	1 \\
1.5707963267949	1 \\
1.63362817986669	1 \\
1.69646003293849	1 \\
1.75929188601028	1 \\
1.82212373908208	1 \\
1.88495559215388	1 \\
1.94778744522567	1 \\
2.01061929829747	1 \\
2.07345115136926	1 \\
2.13628300444106	1 \\
2.19911485751286	1 \\
2.26194671058465	1 \\
2.32477856365645	1 \\
2.38761041672824	1 \\
2.45044226980004	1 \\
2.51327412287183	1 \\
2.57610597594363	1 \\
2.63893782901543	1 \\
2.70176968208722	1 \\
2.76460153515902	1 \\
2.82743338823081	1 \\
2.89026524130261	1 \\
2.95309709437441	1 \\
3.0159289474462	1 \\
3.078760800518	1 \\
3.14159265358979	1 \\
3.20442450666159	1 \\
3.26725635973339	1 \\
3.33008821280518	1 \\
3.39292006587698	1 \\
3.45575191894877	1 \\
3.51858377202057	1 \\
3.58141562509236	1 \\
3.64424747816416	1 \\
3.70707933123596	1 \\
3.76991118430775	1 \\
3.83274303737955	1 \\
3.89557489045134	1 \\
3.95840674352314	1 \\
4.02123859659494	1 \\
4.08407044966673	1 \\
4.14690230273853	1 \\
4.20973415581032	1 \\
4.27256600888212	1 \\
4.33539786195391	1 \\
4.39822971502571	1 \\
4.46106156809751	1 \\
4.5238934211693	1 \\
4.5867252742411	1 \\
4.64955712731289	1 \\
4.71238898038469	1 \\
4.77522083345649	1 \\
4.83805268652828	1 \\
4.90088453960008	1 \\
4.96371639267187	1 \\
5.02654824574367	1 \\
5.08938009881547	1 \\
5.15221195188726	1 \\
5.21504380495906	1 \\
5.27787565803085	1 \\
5.34070751110265	1 \\
5.40353936417444	1 \\
5.46637121724624	1 \\
5.52920307031804	1 \\
5.59203492338983	1 \\
5.65486677646163	1 \\
5.71769862953342	1 \\
5.78053048260522	1 \\
5.84336233567702	1 \\
5.90619418874881	1 \\
5.96902604182061	1 \\
6.0318578948924	1 \\
6.0946897479642	1 \\
6.157521601036	1 \\
6.22035345410779	1 \\
};
\addplot [very thick, color1, dashed]
table [row sep=\\]{%
0	1.00036307110875 \\
0.0628318530717959	0.999951347339842 \\
0.125663706143592	0.999512568280117 \\
0.188495559215388	0.999092107172621 \\
0.251327412287183	0.998915877507586 \\
0.314159265358979	0.998917101028292 \\
0.376991118430775	0.998891836576418 \\
0.439822971502571	1.0002461778902 \\
0.502654824574367	1.0016098588465 \\
0.565486677646163	1.0013022058797 \\
0.628318530717959	1.00063057417435 \\
0.691150383789754	1.0001202146946 \\
0.75398223686155	0.999959272857956 \\
0.816814089933346	0.999659152635565 \\
0.879645943005142	0.998823590340258 \\
0.942477796076938	0.998557113348461 \\
1.00530964914873	0.998968785004057 \\
1.06814150222053	0.999502677929466 \\
1.13097335529233	1.0000516267385 \\
1.19380520836412	1.00061224918467 \\
1.25663706143592	1.00121299643705 \\
1.31946891450771	1.00159339411092 \\
1.38230076757951	1.00142455821887 \\
1.4451326206513	1.00059226522861 \\
1.5079644737231	0.999803604855131 \\
1.5707963267949	0.999332402207217 \\
1.63362817986669	0.99894764552207 \\
1.69646003293849	0.998969264357171 \\
1.75929188601028	0.99933536923497 \\
1.82212373908208	0.999646300442581 \\
1.88495559215388	0.999934152761532 \\
1.94778744522567	1.00012800583101 \\
2.01061929829747	1.00027750486768 \\
2.07345115136926	1.00036842091831 \\
2.13628300444106	1.00041754191444 \\
2.19911485751286	1.00067220599096 \\
2.26194671058465	1.00056376071734 \\
2.32477856365645	1.00009762505585 \\
2.38761041672824	0.999741906438649 \\
2.45044226980004	0.999731290255762 \\
2.51327412287183	0.999648302713396 \\
2.57610597594363	0.999638998264657 \\
2.63893782901543	0.999493086547585 \\
2.70176968208722	0.998614096450936 \\
2.76460153515902	0.9979336429044 \\
2.82743338823081	0.999673452505611 \\
2.89026524130261	1.00133167133614 \\
2.95309709437441	1.00217751455821 \\
3.0159289474462	1.00145445930546 \\
3.078760800518	1.00082261302026 \\
3.14159265358979	1.00036307110875 \\
3.20442450666159	0.999951347339842 \\
3.26725635973339	0.999512568280117 \\
3.33008821280518	0.999092107172621 \\
3.39292006587698	0.998915877507586 \\
3.45575191894877	0.998917101028292 \\
3.51858377202057	0.998891836576418 \\
3.58141562509236	1.0002461778902 \\
3.64424747816416	1.0016098588465 \\
3.70707933123596	1.0013022058797 \\
3.76991118430775	1.00063057417435 \\
3.83274303737955	1.0001202146946 \\
3.89557489045134	0.999959272857956 \\
3.95840674352314	0.999659152635565 \\
4.02123859659494	0.998823590340258 \\
4.08407044966673	0.998557113348461 \\
4.14690230273853	0.998968785004057 \\
4.20973415581032	0.999502677929466 \\
4.27256600888212	1.0000516267385 \\
4.33539786195391	1.00061224918467 \\
4.39822971502571	1.00121299643705 \\
4.46106156809751	1.00159339411092 \\
4.5238934211693	1.00142455821887 \\
4.5867252742411	1.00059226522861 \\
4.64955712731289	0.999803604855131 \\
4.71238898038469	0.999332402207217 \\
4.77522083345649	0.99894764552207 \\
4.83805268652828	0.998969264357171 \\
4.90088453960008	0.99933536923497 \\
4.96371639267187	0.999646300442581 \\
5.02654824574367	0.999934152761532 \\
5.08938009881547	1.00012800583101 \\
5.15221195188726	1.00027750486768 \\
5.21504380495906	1.00036842091831 \\
5.27787565803085	1.00041754191444 \\
5.34070751110265	1.00067220599096 \\
5.40353936417444	1.00056376071734 \\
5.46637121724624	1.00009762505585 \\
5.52920307031804	0.999741906438649 \\
5.59203492338983	0.999731290255762 \\
5.65486677646163	0.999648302713396 \\
5.71769862953342	0.999638998264657 \\
5.78053048260522	0.999493086547585 \\
5.84336233567702	0.998614096450936 \\
5.90619418874881	0.9979336429044 \\
5.96902604182061	0.999673452505611 \\
6.0318578948924	1.00133167133614 \\
6.0946897479642	1.00217751455821 \\
6.157521601036	1.00145445930546 \\
6.22035345410779	1.00082261302026 \\
};
\addplot [very thick, color2, dashed]
table [row sep=\\]{%
0	0.999818997937643 \\
0.0628318530717959	0.999709918740253 \\
0.125663706143592	0.999891454678227 \\
0.188495559215388	1.00003499908227 \\
0.251327412287183	1.00014508778811 \\
0.314159265358979	1.00020555088234 \\
0.376991118430775	1.00001487226578 \\
0.439822971502571	0.999916866650882 \\
0.502654824574367	0.999770266584396 \\
0.565486677646163	0.999633892872345 \\
0.628318530717959	0.999544319764186 \\
0.691150383789754	0.999713707941855 \\
0.75398223686155	0.999914793154761 \\
0.816814089933346	1.0000939562706 \\
0.879645943005142	1.00019828017072 \\
0.942477796076938	1.0003159481289 \\
1.00530964914873	1.00039820589552 \\
1.06814150222053	1.00043187038392 \\
1.13097335529233	1.00043850927105 \\
1.19380520836412	1.00041734244215 \\
1.25663706143592	1.00000957793612 \\
1.31946891450771	0.99975568155979 \\
1.38230076757951	0.999765162771199 \\
1.4451326206513	0.999928211022467 \\
1.5079644737231	1.00006259481413 \\
1.5707963267949	0.99998624806572 \\
1.63362817986669	0.999936008649118 \\
1.69646003293849	0.9998353397658 \\
1.75929188601028	0.999742387902909 \\
1.82212373908208	0.999667616451501 \\
1.88495559215388	0.999620975569166 \\
1.94778744522567	0.999654418395286 \\
2.01061929829747	0.999884047258403 \\
2.07345115136926	0.999973978972966 \\
2.13628300444106	1.00004058763912 \\
2.19911485751286	1.00023393297429 \\
2.26194671058465	1.0002969984495 \\
2.32477856365645	1.00029684857075 \\
2.38761041672824	1.00015311296111 \\
2.45044226980004	1.00002128109002 \\
2.51327412287183	1.00007408299492 \\
2.57610597594363	0.999927409690783 \\
2.63893782901543	0.999900651136652 \\
2.70176968208722	0.999926150205883 \\
2.76460153515902	1.00005647139086 \\
2.82743338823081	1.00015915910675 \\
2.89026524130261	1.00007479502717 \\
2.95309709437441	0.999945176757667 \\
3.0159289474462	0.999933639164018 \\
3.078760800518	0.999902362956116 \\
3.14159265358979	0.999818997937643 \\
3.20442450666159	0.999709918740253 \\
3.26725635973339	0.999891454678227 \\
3.33008821280518	1.00003499908227 \\
3.39292006587698	1.00014508778811 \\
3.45575191894877	1.00020555088234 \\
3.51858377202057	1.00001487226578 \\
3.58141562509236	0.999916866650882 \\
3.64424747816416	0.999770266584396 \\
3.70707933123596	0.999633892872345 \\
3.76991118430775	0.999544319764186 \\
3.83274303737955	0.999713707941855 \\
3.89557489045134	0.999914793154761 \\
3.95840674352314	1.0000939562706 \\
4.02123859659494	1.00019828017072 \\
4.08407044966673	1.0003159481289 \\
4.14690230273853	1.00039820589552 \\
4.20973415581032	1.00043187038392 \\
4.27256600888212	1.00043850927105 \\
4.33539786195391	1.00041734244215 \\
4.39822971502571	1.00000957793612 \\
4.46106156809751	0.99975568155979 \\
4.5238934211693	0.999765162771199 \\
4.5867252742411	0.999928211022467 \\
4.64955712731289	1.00006259481413 \\
4.71238898038469	0.99998624806572 \\
4.77522083345649	0.999936008649118 \\
4.83805268652828	0.9998353397658 \\
4.90088453960008	0.999742387902909 \\
4.96371639267187	0.999667616451501 \\
5.02654824574367	0.999620975569166 \\
5.08938009881547	0.999654418395286 \\
5.15221195188726	0.999884047258403 \\
5.21504380495906	0.999973978972966 \\
5.27787565803085	1.00004058763912 \\
5.34070751110265	1.00023393297429 \\
5.40353936417444	1.0002969984495 \\
5.46637121724624	1.00029684857075 \\
5.52920307031804	1.00015311296111 \\
5.59203492338983	1.00002128109002 \\
5.65486677646163	1.00007408299492 \\
5.71769862953342	0.999927409690783 \\
5.78053048260522	0.999900651136652 \\
5.84336233567702	0.999926150205883 \\
5.90619418874881	1.00005647139086 \\
5.96902604182061	1.00015915910675 \\
6.0318578948924	1.00007479502717 \\
6.0946897479642	0.999945176757667 \\
6.157521601036	0.999933639164018 \\
6.22035345410779	0.999902362956116 \\
};
\addplot [very thick, color3, dashed]
table [row sep=\\]{%
0	0.999994733436491 \\
0.0628318530717959	0.999994481610356 \\
0.125663706143592	0.999994707268703 \\
0.188495559215388	0.999995151395364 \\
0.251327412287183	0.999994771543161 \\
0.314159265358979	0.999993324520772 \\
0.376991118430775	0.999991907885727 \\
0.439822971502571	0.999989920118606 \\
0.502654824574367	0.999989186796249 \\
0.565486677646163	0.999988269802644 \\
0.628318530717959	0.999987169234968 \\
0.691150383789754	0.999986158267429 \\
0.75398223686155	0.999985362044858 \\
0.816814089933346	0.999984360598343 \\
0.879645943005142	0.999983912265156 \\
0.942477796076938	0.99998321876873 \\
1.00530964914873	0.999982309422747 \\
1.06814150222053	0.999981691945567 \\
1.13097335529233	0.999981394106666 \\
1.19380520836412	0.999979765653343 \\
1.25663706143592	0.999978832701484 \\
1.31946891450771	0.999978366473364 \\
1.38230076757951	0.999978702069329 \\
1.4451326206513	0.999979294959302 \\
1.5079644737231	0.999980224663542 \\
1.5707963267949	0.999981634554237 \\
1.63362817986669	0.999982321150155 \\
1.69646003293849	0.999982465139518 \\
1.75929188601028	0.999982365195731 \\
1.82212373908208	0.999985304442874 \\
1.88495559215388	0.999983965030576 \\
1.94778744522567	0.999982768442017 \\
2.01061929829747	0.999981179640874 \\
2.07345115136926	0.999981085080508 \\
2.13628300444106	0.999982760879824 \\
2.19911485751286	0.999984757306531 \\
2.26194671058465	0.999986104381363 \\
2.32477856365645	0.999986511142592 \\
2.38761041672824	0.999986349119875 \\
2.45044226980004	0.999986434685873 \\
2.51327412287183	0.999986677558111 \\
2.57610597594363	0.999986544626741 \\
2.63893782901543	0.999986294938942 \\
2.70176968208722	0.999987924000272 \\
2.76460153515902	0.999990414950891 \\
2.82743338823081	0.99999246749903 \\
2.89026524130261	0.999994303602822 \\
2.95309709437441	0.999994930876404 \\
3.0159289474462	0.99999524810425 \\
3.078760800518	0.999994784020824 \\
3.14159265358979	0.999994733436491 \\
3.20442450666159	0.999994481610356 \\
3.26725635973339	0.999994707268703 \\
3.33008821280518	0.999995151395364 \\
3.39292006587698	0.999994771543161 \\
3.45575191894877	0.999993324520772 \\
3.51858377202057	0.999991907885727 \\
3.58141562509236	0.999989920118606 \\
3.64424747816416	0.999989186796249 \\
3.70707933123596	0.999988269802644 \\
3.76991118430775	0.999987169234968 \\
3.83274303737955	0.999986158267429 \\
3.89557489045134	0.999985362044858 \\
3.95840674352314	0.999984360598343 \\
4.02123859659494	0.999983912265156 \\
4.08407044966673	0.99998321876873 \\
4.14690230273853	0.999982309422747 \\
4.20973415581032	0.999981691945567 \\
4.27256600888212	0.999981394106666 \\
4.33539786195391	0.999979765653343 \\
4.39822971502571	0.999978832701484 \\
4.46106156809751	0.999978366473364 \\
4.5238934211693	0.999978702069329 \\
4.5867252742411	0.999979294959302 \\
4.64955712731289	0.999980224663542 \\
4.71238898038469	0.999981634554237 \\
4.77522083345649	0.999982321150155 \\
4.83805268652828	0.999982465139518 \\
4.90088453960008	0.999982365195731 \\
4.96371639267187	0.999985304442874 \\
5.02654824574367	0.999983965030576 \\
5.08938009881547	0.999982768442017 \\
5.15221195188726	0.999981179640874 \\
5.21504380495906	0.999981085080508 \\
5.27787565803085	0.999982760879824 \\
5.34070751110265	0.999984757306531 \\
5.40353936417444	0.999986104381363 \\
5.46637121724624	0.999986511142592 \\
5.52920307031804	0.999986349119875 \\
5.59203492338983	0.999986434685873 \\
5.65486677646163	0.999986677558111 \\
5.71769862953342	0.999986544626741 \\
5.78053048260522	0.999986294938942 \\
5.84336233567702	0.999987924000272 \\
5.90619418874881	0.999990414950891 \\
5.96902604182061	0.99999246749903 \\
6.0318578948924	0.999994303602822 \\
6.0946897479642	0.999994930876404 \\
6.157521601036	0.99999524810425 \\
6.22035345410779	0.999994784020824 \\
};

\end{axis}

\end{tikzpicture}

%% file: figures/benchmark/benchmarkPL2.tex
\begin{tikzpicture}

\definecolor{color0}{rgb}{0.12156862745098,0.466666666666667,0.705882352941177}
\definecolor{color1}{rgb}{1,0.498039215686275,0.0549019607843137}
\definecolor{color2}{rgb}{0.172549019607843,0.627450980392157,0.172549019607843}
\definecolor{color3}{rgb}{0.83921568627451,0.152941176470588,0.156862745098039}

\begin{axis}[
legend cell align={left},
legend entries={{Exact},{PL10},{PL20},{PL40}},
legend style={at={(0.03,0.03)}, anchor=south west, draw=white!80.0!black},
tick align=outside,
tick pos=left,
x grid style={white!69.01960784313725!black},
xlabel={$x$},
xmin=-0.31101767270539, xmax=6.53137112681318,
y grid style={white!69.01960784313725!black},
ylabel={$y$},
ymin=-0.1, ymax=1.2,
ytick={-0.2,0,0.2,0.4,0.6,0.8,1,1.2},
yticklabels={−0.2,0.0,0.2,0.4,0.6,0.8,1.0,1.2}
]
\addlegendimage{very thick, color0}
\addlegendimage{very thick, dashed, color1}
\addlegendimage{very thick, dashed,color2}
\addlegendimage{very thick, dashed,color3}
\addplot [very thick, color0]
table [row sep=\\]{%
0	1 \\
0.0628318530717959	1 \\
0.125663706143592	1 \\
0.188495559215388	1 \\
0.251327412287183	1 \\
0.314159265358979	1 \\
0.376991118430775	1 \\
0.439822971502571	1 \\
0.502654824574367	1 \\
0.565486677646163	1 \\
0.628318530717959	1 \\
0.691150383789754	1 \\
0.75398223686155	1 \\
0.816814089933346	1 \\
0.879645943005142	1 \\
0.942477796076938	1 \\
1.00530964914873	1 \\
1.06814150222053	1 \\
1.13097335529233	1 \\
1.19380520836412	1 \\
1.25663706143592	1 \\
1.31946891450771	1 \\
1.38230076757951	1 \\
1.4451326206513	1 \\
1.5079644737231	1 \\
1.5707963267949	1 \\
1.63362817986669	1 \\
1.69646003293849	1 \\
1.75929188601028	1 \\
1.82212373908208	1 \\
1.88495559215388	1 \\
1.94778744522567	1 \\
2.01061929829747	1 \\
2.07345115136926	1 \\
2.13628300444106	1 \\
2.19911485751286	1 \\
2.26194671058465	1 \\
2.32477856365645	1 \\
2.38761041672824	1 \\
2.45044226980004	1 \\
2.51327412287183	1 \\
2.57610597594363	1 \\
2.63893782901543	1 \\
2.70176968208722	1 \\
2.76460153515902	1 \\
2.82743338823081	1 \\
2.89026524130261	1 \\
2.95309709437441	1 \\
3.0159289474462	1 \\
3.078760800518	1 \\
3.14159265358979	1 \\
3.20442450666159	1 \\
3.26725635973339	1 \\
3.33008821280518	1 \\
3.39292006587698	1 \\
3.45575191894877	1 \\
3.51858377202057	1 \\
3.58141562509236	1 \\
3.64424747816416	1 \\
3.70707933123596	1 \\
3.76991118430775	1 \\
3.83274303737955	1 \\
3.89557489045134	1 \\
3.95840674352314	1 \\
4.02123859659494	1 \\
4.08407044966673	1 \\
4.14690230273853	1 \\
4.20973415581032	1 \\
4.27256600888212	1 \\
4.33539786195391	1 \\
4.39822971502571	1 \\
4.46106156809751	1 \\
4.5238934211693	1 \\
4.5867252742411	1 \\
4.64955712731289	1 \\
4.71238898038469	1 \\
4.77522083345649	1 \\
4.83805268652828	1 \\
4.90088453960008	1 \\
4.96371639267187	1 \\
5.02654824574367	1 \\
5.08938009881547	1 \\
5.15221195188726	1 \\
5.21504380495906	1 \\
5.27787565803085	1 \\
5.34070751110265	1 \\
5.40353936417444	1 \\
5.46637121724624	1 \\
5.52920307031804	1 \\
5.59203492338983	1 \\
5.65486677646163	1 \\
5.71769862953342	1 \\
5.78053048260522	1 \\
5.84336233567702	1 \\
5.90619418874881	1 \\
5.96902604182061	1 \\
6.0318578948924	1 \\
6.0946897479642	1 \\
6.157521601036	1 \\
6.22035345410779	1 \\
};
\addplot [very thick, color1, dashed]
table [row sep=\\]{%
0	1.00005208493978 \\
0.0628318530717959	1.00003264640308 \\
0.125663706143592	1.00001320786637 \\
0.188495559215388	0.999993769329673 \\
0.251327412287183	0.999974330792972 \\
0.314159265358979	0.999954892256271 \\
0.376991118430775	0.99993545371957 \\
0.439822971502571	0.999916015182869 \\
0.502654824574367	0.999896576646168 \\
0.565486677646163	0.999877138109467 \\
0.628318530717959	0.999857699572765 \\
0.691150383789754	0.999886789201362 \\
0.75398223686155	0.99991587882996 \\
0.816814089933346	0.999944968458557 \\
0.879645943005142	0.999974058087154 \\
0.942477796076938	1.00000314771575 \\
1.00530964914873	1.00003223734435 \\
1.06814150222053	1.00006132697295 \\
1.13097335529233	1.00009041660154 \\
1.19380520836412	1.00011950623014 \\
1.25663706143592	1.00014859585874 \\
1.31946891450771	1.00012318558323 \\
1.38230076757951	1.00009777530772 \\
1.4451326206513	1.00007236503221 \\
1.5079644737231	1.0000469547567 \\
1.5707963267949	1.00002154448119 \\
1.63362817986669	0.999996134205683 \\
1.69646003293849	0.999970723930174 \\
1.75929188601028	0.999945313654665 \\
1.82212373908208	0.999919903379155 \\
1.88495559215388	0.999894493103646 \\
1.94778744522567	0.999902945534398 \\
2.01061929829747	0.999911397965151 \\
2.07345115136926	0.999919850395903 \\
2.13628300444106	0.999928302826655 \\
2.19911485751286	0.999936755257408 \\
2.26194671058465	0.99994520768816 \\
2.32477856365645	0.999953660118913 \\
2.38761041672824	0.999962112549665 \\
2.45044226980004	0.999970564980418 \\
2.51327412287183	0.99997901741117 \\
2.57610597594363	0.99998632416403 \\
2.63893782901543	0.999993630916891 \\
2.70176968208722	1.00000093766975 \\
2.76460153515902	1.00000824442261 \\
2.82743338823081	1.00001555117547 \\
2.89026524130261	1.00002285792833 \\
2.95309709437441	1.00003016468119 \\
3.0159289474462	1.00003747143405 \\
3.078760800518	1.00004477818692 \\
3.14159265358979	1.00005208493978 \\
3.20442450666159	1.00003264640308 \\
3.26725635973339	1.00001320786637 \\
3.33008821280518	0.999993769329673 \\
3.39292006587698	0.999974330792972 \\
3.45575191894877	0.999954892256271 \\
3.51858377202057	0.99993545371957 \\
3.58141562509236	0.999916015182869 \\
3.64424747816416	0.999896576646168 \\
3.70707933123596	0.999877138109467 \\
3.76991118430775	0.999857699572765 \\
3.83274303737955	0.999886789201362 \\
3.89557489045134	0.99991587882996 \\
3.95840674352314	0.999944968458557 \\
4.02123859659494	0.999974058087154 \\
4.08407044966673	1.00000314771575 \\
4.14690230273853	1.00003223734435 \\
4.20973415581032	1.00006132697295 \\
4.27256600888212	1.00009041660154 \\
4.33539786195391	1.00011950623014 \\
4.39822971502571	1.00014859585874 \\
4.46106156809751	1.00012318558323 \\
4.5238934211693	1.00009777530772 \\
4.5867252742411	1.00007236503221 \\
4.64955712731289	1.0000469547567 \\
4.71238898038469	1.00002154448119 \\
4.77522083345649	0.999996134205683 \\
4.83805268652828	0.999970723930174 \\
4.90088453960008	0.999945313654665 \\
4.96371639267187	0.999919903379155 \\
5.02654824574367	0.999894493103646 \\
5.08938009881547	0.999902945534398 \\
5.15221195188726	0.999911397965151 \\
5.21504380495906	0.999919850395903 \\
5.27787565803085	0.999928302826655 \\
5.34070751110265	0.999936755257408 \\
5.40353936417444	0.99994520768816 \\
5.46637121724624	0.999953660118913 \\
5.52920307031804	0.999962112549665 \\
5.59203492338983	0.999970564980418 \\
5.65486677646163	0.99997901741117 \\
5.71769862953342	0.99998632416403 \\
5.78053048260522	0.999993630916891 \\
5.84336233567702	1.00000093766975 \\
5.90619418874881	1.00000824442261 \\
5.96902604182061	1.00001555117547 \\
6.0318578948924	1.00002285792833 \\
6.0946897479642	1.00003016468119 \\
6.157521601036	1.00003747143405 \\
6.22035345410779	1.00004477818692 \\
};
\addplot [very thick, color2, dashed]
table [row sep=\\]{%
0	0.999584029300593 \\
0.0628318530717959	0.999789546578925 \\
0.125663706143592	0.999995063857257 \\
0.188495559215388	1.00020058113559 \\
0.251327412287183	1.00040609841392 \\
0.314159265358979	1.00061161569225 \\
0.376991118430775	1.00033145635471 \\
0.439822971502571	1.00005129701717 \\
0.502654824574367	0.999771137679632 \\
0.565486677646163	0.999490978342091 \\
0.628318530717959	0.999210819004551 \\
0.691150383789754	0.99976082931632 \\
0.75398223686155	1.00031083962809 \\
0.816814089933346	1.00086084993986 \\
0.879645943005142	1.00141086025164 \\
0.942477796076938	1.00196087056341 \\
1.00530964914873	1.00056420167825 \\
1.06814150222053	0.999167532793089 \\
1.13097335529233	0.997770863907927 \\
1.19380520836412	0.996374195022765 \\
1.25663706143592	0.994977526137603 \\
1.31946891450771	0.997548453361423 \\
1.38230076757951	1.00011938058525 \\
1.4451326206513	1.00269030780908 \\
1.5079644737231	1.00526123503291 \\
1.5707963267949	1.00783216225674 \\
1.63362817986669	1.00478597956099 \\
1.69646003293849	1.00173979686521 \\
1.75929188601028	0.998693614169439 \\
1.82212373908208	0.995647431473664 \\
1.88495559215388	0.99260124877789 \\
1.94778744522567	0.994940009712934 \\
2.01061929829747	0.997278770647998 \\
2.07345115136926	0.999617531583062 \\
2.13628300444106	1.00195629251813 \\
2.19911485751286	1.00429505345319 \\
2.26194671058465	1.00312230949588 \\
2.32477856365645	1.00194956553856 \\
2.38761041672824	1.00077682158123 \\
2.45044226980004	0.999604077623907 \\
2.51327412287183	0.998431333666582 \\
2.57610597594363	0.998817025096493 \\
2.63893782901543	0.999202716526412 \\
2.70176968208722	0.999588407956331 \\
2.76460153515902	0.99997409938625 \\
2.82743338823081	1.00035979081617 \\
2.89026524130261	1.00020463851306 \\
2.95309709437441	1.00004948620994 \\
3.0159289474462	0.999894333906825 \\
3.078760800518	0.999739181603709 \\
3.14159265358979	0.999584029300593 \\
3.20442450666159	0.999789546578925 \\
3.26725635973339	0.999995063857257 \\
3.33008821280518	1.00020058113559 \\
3.39292006587698	1.00040609841392 \\
3.45575191894877	1.00061161569225 \\
3.51858377202057	1.00033145635471 \\
3.58141562509236	1.00005129701717 \\
3.64424747816416	0.999771137679632 \\
3.70707933123596	0.999490978342091 \\
3.76991118430775	0.999210819004551 \\
3.83274303737955	0.99976082931632 \\
3.89557489045134	1.00031083962809 \\
3.95840674352314	1.00086084993986 \\
4.02123859659494	1.00141086025164 \\
4.08407044966673	1.00196087056341 \\
4.14690230273853	1.00056420167825 \\
4.20973415581032	0.999167532793089 \\
4.27256600888212	0.997770863907927 \\
4.33539786195391	0.996374195022765 \\
4.39822971502571	0.994977526137603 \\
4.46106156809751	0.997548453361423 \\
4.5238934211693	1.00011938058525 \\
4.5867252742411	1.00269030780908 \\
4.64955712731289	1.00526123503291 \\
4.71238898038469	1.00783216225674 \\
4.77522083345649	1.00478597956099 \\
4.83805268652828	1.00173979686521 \\
4.90088453960008	0.998693614169439 \\
4.96371639267187	0.995647431473664 \\
5.02654824574367	0.99260124877789 \\
5.08938009881547	0.994940009712934 \\
5.15221195188726	0.997278770647998 \\
5.21504380495906	0.999617531583062 \\
5.27787565803085	1.00195629251813 \\
5.34070751110265	1.00429505345319 \\
5.40353936417444	1.00312230949588 \\
5.46637121724624	1.00194956553856 \\
5.52920307031804	1.00077682158123 \\
5.59203492338983	0.999604077623907 \\
5.65486677646163	0.998431333666582 \\
5.71769862953342	0.998817025096493 \\
5.78053048260522	0.999202716526412 \\
5.84336233567702	0.999588407956331 \\
5.90619418874881	0.99997409938625 \\
5.96902604182061	1.00035979081617 \\
6.0318578948924	1.00020463851306 \\
6.0946897479642	1.00004948620994 \\
6.157521601036	0.999894333906825 \\
6.22035345410779	0.999739181603709 \\
};
\addplot [very thick, color3, dashed]
table [row sep=\\]{%
0	1.00328266864632 \\
0.0628318530717959	1.00029068159536 \\
0.125663706143592	0.997298694544399 \\
0.188495559215388	0.9962563041043 \\
0.251327412287183	0.997163510275085 \\
0.314159265358979	0.99807071644587 \\
0.376991118430775	1.00241574539875 \\
0.439822971502571	1.00676077435165 \\
0.502654824574367	1.00610317274108 \\
0.565486677646163	1.00044294056702 \\
0.628318530717959	0.994782708392958 \\
0.691150383789754	0.99552647383057 \\
0.75398223686155	0.996270239268199 \\
0.816814089933346	0.998499455071 \\
0.879645943005142	1.00221412123899 \\
0.942477796076938	1.00592878740698 \\
1.00530964914873	1.00415320190621 \\
1.06814150222053	1.00237761640543 \\
1.13097335529233	0.998365339693012 \\
1.19380520836412	0.992116371768926 \\
1.25663706143592	0.98586740384484 \\
1.31946891450771	0.997790162148329 \\
1.38230076757951	1.00971292045187 \\
1.4451326206513	1.01096909646381 \\
1.5079644737231	1.00155869018402 \\
1.5707963267949	0.99214828390422 \\
1.63362817986669	1.0023454173757 \\
1.69646003293849	1.01254255084724 \\
1.75929188601028	1.00982592333913 \\
1.82212373908208	0.994195534851182 \\
1.88495559215388	0.978565146363235 \\
1.94778744522567	0.990150695806554 \\
2.01061929829747	1.00173624524997 \\
2.07345115136926	1.00696448499528 \\
2.13628300444106	1.00583541504238 \\
2.19911485751286	1.00470634508949 \\
2.26194671058465	0.999314344161279 \\
2.32477856365645	0.993922343233051 \\
2.38761041672824	0.995509097624303 \\
2.45044226980004	1.00407460733515 \\
2.51327412287183	1.012640117046 \\
2.57610597594363	1.00214118250299 \\
2.63893782901543	0.991642247959902 \\
2.70176968208722	0.990241854382877 \\
2.76460153515902	0.997940001772067 \\
2.82743338823081	1.00563814916126 \\
2.89026524130261	1.00268204652778 \\
2.95309709437441	0.999725943894257 \\
3.0159289474462	0.999254847791236 \\
3.078760800518	1.00126875821877 \\
3.14159265358979	1.00328266864632 \\
3.20442450666159	1.00029068159536 \\
3.26725635973339	0.997298694544399 \\
3.33008821280518	0.9962563041043 \\
3.39292006587698	0.997163510275085 \\
3.45575191894877	0.99807071644587 \\
3.51858377202057	1.00241574539875 \\
3.58141562509236	1.00676077435165 \\
3.64424747816416	1.00610317274108 \\
3.70707933123596	1.00044294056702 \\
3.76991118430775	0.994782708392958 \\
3.83274303737955	0.99552647383057 \\
3.89557489045134	0.996270239268199 \\
3.95840674352314	0.998499455071 \\
4.02123859659494	1.00221412123899 \\
4.08407044966673	1.00592878740698 \\
4.14690230273853	1.00415320190621 \\
4.20973415581032	1.00237761640543 \\
4.27256600888212	0.998365339693012 \\
4.33539786195391	0.992116371768926 \\
4.39822971502571	0.98586740384484 \\
4.46106156809751	0.997790162148329 \\
4.5238934211693	1.00971292045187 \\
4.5867252742411	1.01096909646381 \\
4.64955712731289	1.00155869018402 \\
4.71238898038469	0.992148283904221 \\
4.77522083345649	1.0023454173757 \\
4.83805268652828	1.01254255084724 \\
4.90088453960008	1.00982592333913 \\
4.96371639267187	0.994195534851182 \\
5.02654824574367	0.978565146363235 \\
5.08938009881547	0.990150695806554 \\
5.15221195188726	1.00173624524997 \\
5.21504380495906	1.00696448499528 \\
5.27787565803085	1.00583541504238 \\
5.34070751110265	1.00470634508949 \\
5.40353936417444	0.999314344161278 \\
5.46637121724624	0.993922343233051 \\
5.52920307031804	0.995509097624303 \\
5.59203492338983	1.00407460733515 \\
5.65486677646163	1.012640117046 \\
5.71769862953342	1.00214118250299 \\
5.78053048260522	0.991642247959902 \\
5.84336233567702	0.990241854382877 \\
5.90619418874881	0.997940001772067 \\
5.96902604182061	1.00563814916126 \\
6.0318578948924	1.00268204652778 \\
6.0946897479642	0.999725943894257 \\
6.157521601036	0.999254847791236 \\
6.22035345410779	1.00126875821877 \\
};

\end{axis}

\end{tikzpicture}

%% file: figures/benchmark/benchmarkRBF2.tex
\begin{tikzpicture}

\definecolor{color0}{rgb}{0.12156862745098,0.466666666666667,0.705882352941177}
\definecolor{color1}{rgb}{1,0.498039215686275,0.0549019607843137}
\definecolor{color2}{rgb}{0.172549019607843,0.627450980392157,0.172549019607843}
\definecolor{color3}{rgb}{0.83921568627451,0.152941176470588,0.156862745098039}

\begin{axis}[
legend cell align={left},
legend entries={{Exact},{RBF10},{RBF20},{RBF40}},
legend style={at={(0.97,0.03)}, anchor=south east, draw=white!80.0!black},
tick align=outside,
tick pos=left,
x grid style={white!69.01960784313725!black},
xlabel={$x$},
xmin=-0.31101767270539, xmax=6.53137112681318,
y grid style={white!69.01960784313725!black},
ylabel={$y$},
ymin=-0.1, ymax=1.2,
ytick={-0.2,0,0.2,0.4,0.6,0.8,1,1.2},
yticklabels={−0.2,0.0,0.2,0.4,0.6,0.8,1.0,1.2}
]
\addlegendimage{very thick, color0}
\addlegendimage{very thick, dashed, color1}
\addlegendimage{very thick, dashed,color2}
\addlegendimage{very thick, dashed,color3}
\addplot [very thick, color0]
table [row sep=\\]{%
0	1 \\
0.0628318530717959	1 \\
0.125663706143592	1 \\
0.188495559215388	1 \\
0.251327412287183	1 \\
0.314159265358979	1 \\
0.376991118430775	1 \\
0.439822971502571	1 \\
0.502654824574367	1 \\
0.565486677646163	1 \\
0.628318530717959	1 \\
0.691150383789754	1 \\
0.75398223686155	1 \\
0.816814089933346	1 \\
0.879645943005142	1 \\
0.942477796076938	1 \\
1.00530964914873	1 \\
1.06814150222053	1 \\
1.13097335529233	1 \\
1.19380520836412	1 \\
1.25663706143592	1 \\
1.31946891450771	1 \\
1.38230076757951	1 \\
1.4451326206513	1 \\
1.5079644737231	1 \\
1.5707963267949	1 \\
1.63362817986669	1 \\
1.69646003293849	1 \\
1.75929188601028	1 \\
1.82212373908208	1 \\
1.88495559215388	1 \\
1.94778744522567	1 \\
2.01061929829747	1 \\
2.07345115136926	1 \\
2.13628300444106	1 \\
2.19911485751286	1 \\
2.26194671058465	1 \\
2.32477856365645	1 \\
2.38761041672824	1 \\
2.45044226980004	1 \\
2.51327412287183	1 \\
2.57610597594363	1 \\
2.63893782901543	1 \\
2.70176968208722	1 \\
2.76460153515902	1 \\
2.82743338823081	1 \\
2.89026524130261	1 \\
2.95309709437441	1 \\
3.0159289474462	1 \\
3.078760800518	1 \\
3.14159265358979	1 \\
3.20442450666159	1 \\
3.26725635973339	1 \\
3.33008821280518	1 \\
3.39292006587698	1 \\
3.45575191894877	1 \\
3.51858377202057	1 \\
3.58141562509236	1 \\
3.64424747816416	1 \\
3.70707933123596	1 \\
3.76991118430775	1 \\
3.83274303737955	1 \\
3.89557489045134	1 \\
3.95840674352314	1 \\
4.02123859659494	1 \\
4.08407044966673	1 \\
4.14690230273853	1 \\
4.20973415581032	1 \\
4.27256600888212	1 \\
4.33539786195391	1 \\
4.39822971502571	1 \\
4.46106156809751	1 \\
4.5238934211693	1 \\
4.5867252742411	1 \\
4.64955712731289	1 \\
4.71238898038469	1 \\
4.77522083345649	1 \\
4.83805268652828	1 \\
4.90088453960008	1 \\
4.96371639267187	1 \\
5.02654824574367	1 \\
5.08938009881547	1 \\
5.15221195188726	1 \\
5.21504380495906	1 \\
5.27787565803085	1 \\
5.34070751110265	1 \\
5.40353936417444	1 \\
5.46637121724624	1 \\
5.52920307031804	1 \\
5.59203492338983	1 \\
5.65486677646163	1 \\
5.71769862953342	1 \\
5.78053048260522	1 \\
5.84336233567702	1 \\
5.90619418874881	1 \\
5.96902604182061	1 \\
6.0318578948924	1 \\
6.0946897479642	1 \\
6.157521601036	1 \\
6.22035345410779	1 \\
};
\addplot [very thick, color1, dashed]
table [row sep=\\]{%
0	0.986786410623024 \\
0.0628318530717959	0.994474649531507 \\
0.125663706143592	0.999584029337792 \\
0.188495559215388	1.00251206278809 \\
0.251327412287183	1.00379794220765 \\
0.314159265358979	1.00399569460843 \\
0.376991118430775	1.00356945895886 \\
0.439822971502571	1.002837500658 \\
0.502654824574367	1.00197102240724 \\
0.565486677646163	1.00103698354527 \\
0.628318530717959	1.00006267178414 \\
0.691150383789754	0.999094525003227 \\
0.75398223686155	0.998227520611155 \\
0.816814089933346	0.997595187763469 \\
0.879645943005142	0.997328250267276 \\
0.942477796076938	0.997502651505333 \\
1.00530964914873	0.998099974602704 \\
1.06814150222053	0.998996981516872 \\
1.13097335529233	0.999990776111344 \\
1.19380520836412	1.00085445164794 \\
1.25663706143592	1.00140620705397 \\
1.31946891450771	1.00156658187717 \\
1.38230076757951	1.00138000661041 \\
1.4451326206513	1.00099089867582 \\
1.5079644737231	1.00058448587687 \\
1.5707963267949	1.00031673650903 \\
1.63362817986669	1.00025958517954 \\
1.69646003293849	1.00037950894156 \\
1.75929188601028	1.00055549874691 \\
1.82212373908208	1.00063009914254 \\
1.88495559215388	1.00047569331219 \\
1.94778744522567	1.00005069467502 \\
2.01061929829747	0.999422278045174 \\
2.07345115136926	0.998745938997992 \\
2.13628300444106	0.998211243040733 \\
2.19911485751286	0.997976633835061 \\
2.26194671058465	0.998117905453058 \\
2.32477856365645	0.99860725435483 \\
2.38761041672824	0.999328470464337 \\
2.45044226980004	1.00012214604661 \\
2.51327412287183	1.00084396676312 \\
2.57610597594363	1.00141214181788 \\
2.63893782901543	1.00182205080155 \\
2.70176968208722	1.00211939148727 \\
2.76460153515902	1.00234175184779 \\
2.82743338823081	1.00245235947636 \\
2.89026524130261	1.00229275839459 \\
2.95309709437441	1.00157480751823 \\
3.0159289474462	0.999920791733448 \\
3.078760800518	0.996945484237364 \\
3.14159265358979	0.986786410623024 \\
3.20442450666159	0.994474649531507 \\
3.26725635973339	0.999584029337792 \\
3.33008821280518	1.00251206278809 \\
3.39292006587698	1.00379794220765 \\
3.45575191894877	1.00399569460843 \\
3.51858377202057	1.00356945895886 \\
3.58141562509236	1.002837500658 \\
3.64424747816416	1.00197102240724 \\
3.70707933123596	1.00103698354527 \\
3.76991118430775	1.00006267178414 \\
3.83274303737955	0.999094525003227 \\
3.89557489045134	0.998227520611155 \\
3.95840674352314	0.997595187763469 \\
4.02123859659494	0.997328250267277 \\
4.08407044966673	0.997502651505333 \\
4.14690230273853	0.998099974602704 \\
4.20973415581032	0.998996981516872 \\
4.27256600888212	0.999990776111344 \\
4.33539786195391	1.00085445164794 \\
4.39822971502571	1.00140620705397 \\
4.46106156809751	1.00156658187717 \\
4.5238934211693	1.00138000661041 \\
4.5867252742411	1.00099089867582 \\
4.64955712731289	1.00058448587687 \\
4.71238898038469	1.00031673650903 \\
4.77522083345649	1.00025958517954 \\
4.83805268652828	1.00037950894156 \\
4.90088453960008	1.00055549874691 \\
4.96371639267187	1.00063009914254 \\
5.02654824574367	1.00047569331219 \\
5.08938009881547	1.00005069467502 \\
5.15221195188726	0.999422278045174 \\
5.21504380495906	0.998745938997992 \\
5.27787565803085	0.998211243040733 \\
5.34070751110265	0.997976633835061 \\
5.40353936417444	0.998117905453058 \\
5.46637121724624	0.99860725435483 \\
5.52920307031804	0.999328470464337 \\
5.59203492338983	1.00012214604661 \\
5.65486677646163	1.00084396676312 \\
5.71769862953342	1.00141214181788 \\
5.78053048260522	1.00182205080155 \\
5.84336233567702	1.00211939148727 \\
5.90619418874881	1.00234175184779 \\
5.96902604182061	1.00245235947636 \\
6.0318578948924	1.00229275839459 \\
6.0946897479642	1.00157480751823 \\
6.157521601036	0.999920791733448 \\
6.22035345410779	0.996945484237364 \\
};
\addplot [very thick, color2, dashed]
table [row sep=\\]{%
0	0.995046967119037 \\
0.0628318530717959	1.00228912955761 \\
0.125663706143592	1.00377446026265 \\
0.188495559215388	1.00260275391797 \\
0.251327412287183	1.00097719491626 \\
0.314159265358979	0.999528150298785 \\
0.376991118430775	0.998411704747487 \\
0.439822971502571	0.998216856888918 \\
0.502654824574367	0.99941464004717 \\
0.565486677646163	1.00137862255862 \\
0.628318530717959	1.00262422726253 \\
0.691150383789754	1.00227914344179 \\
0.75398223686155	1.00097614906122 \\
0.816814089933346	0.999927749723008 \\
0.879645943005142	0.999468175584223 \\
0.942477796076938	0.998932662163888 \\
1.00530964914873	0.997882402277023 \\
1.06814150222053	0.996934743889707 \\
1.13097335529233	0.99703086588226 \\
1.19380520836412	0.998234560729002 \\
1.25663706143592	0.999702951053221 \\
1.31946891450771	1.00082994693108 \\
1.38230076757951	1.00194841706107 \\
1.4451326206513	1.00352469018027 \\
1.5079644737231	1.00499035307313 \\
1.5707963267949	1.00493120770851 \\
1.63362817986669	1.00262708542374 \\
1.69646003293849	0.999040756652668 \\
1.75929188601028	0.99593308194548 \\
1.82212373908208	0.994349508774266 \\
1.88495559215388	0.994275054168145 \\
1.94778744522567	0.995471962190336 \\
2.01061929829747	0.998044050153565 \\
2.07345115136926	1.0018086709041 \\
2.13628300444106	1.00544347170155 \\
2.19911485751286	1.006928635898 \\
2.26194671058465	1.00526530969347 \\
2.32477856365645	1.00150925930294 \\
2.38761041672824	0.997734417307675 \\
2.45044226980004	0.995310731174549 \\
2.51327412287183	0.994455772697025 \\
2.57610597594363	0.995073855744138 \\
2.63893782901543	0.997351575208856 \\
2.70176968208722	1.00114413434041 \\
2.76460153515902	1.00511231537836 \\
2.82743338823081	1.00713275835189 \\
2.89026524130261	1.00594569679279 \\
2.95309709437441	1.00204684813698 \\
3.0159289474462	0.996464364659655 \\
3.078760800518	0.989011402422449 \\
3.14159265358979	0.995046967119037 \\
3.20442450666159	1.00228912955761 \\
3.26725635973339	1.00377446026265 \\
3.33008821280518	1.00260275391797 \\
3.39292006587698	1.00097719491626 \\
3.45575191894877	0.999528150298785 \\
3.51858377202057	0.998411704747487 \\
3.58141562509236	0.998216856888918 \\
3.64424747816416	0.99941464004717 \\
3.70707933123596	1.00137862255862 \\
3.76991118430775	1.00262422726253 \\
3.83274303737955	1.00227914344179 \\
3.89557489045134	1.00097614906122 \\
3.95840674352314	0.999927749723008 \\
4.02123859659494	0.999468175584223 \\
4.08407044966673	0.998932662163888 \\
4.14690230273853	0.997882402277023 \\
4.20973415581032	0.996934743889707 \\
4.27256600888212	0.99703086588226 \\
4.33539786195391	0.998234560729002 \\
4.39822971502571	0.999702951053221 \\
4.46106156809751	1.00082994693108 \\
4.5238934211693	1.00194841706107 \\
4.5867252742411	1.00352469018027 \\
4.64955712731289	1.00499035307313 \\
4.71238898038469	1.00493120770851 \\
4.77522083345649	1.00262708542374 \\
4.83805268652828	0.999040756652668 \\
4.90088453960008	0.99593308194548 \\
4.96371639267187	0.994349508774266 \\
5.02654824574367	0.994275054168145 \\
5.08938009881547	0.995471962190336 \\
5.15221195188726	0.998044050153565 \\
5.21504380495906	1.0018086709041 \\
5.27787565803085	1.00544347170155 \\
5.34070751110265	1.006928635898 \\
5.40353936417444	1.00526530969347 \\
5.46637121724624	1.00150925930294 \\
5.52920307031804	0.997734417307675 \\
5.59203492338983	0.995310731174549 \\
5.65486677646163	0.994455772697025 \\
5.71769862953342	0.995073855744138 \\
5.78053048260522	0.997351575208856 \\
5.84336233567702	1.00114413434041 \\
5.90619418874881	1.00511231537836 \\
5.96902604182061	1.00713275835189 \\
6.0318578948924	1.00594569679279 \\
6.0946897479642	1.00204684813698 \\
6.157521601036	0.996464364659654 \\
6.22035345410779	0.989011402422449 \\
};
\addplot [very thick, color3, dashed]
table [row sep=\\]{%
0	1.00460962702877 \\
0.0628318530717959	0.946973577594579 \\
0.125663706143592	0.95517274536991 \\
0.188495559215388	1.11419863106132 \\
0.251327412287183	1.28543587453709 \\
0.314159265358979	0.983438195250538 \\
0.376991118430775	0.0933720788670801 \\
0.439822971502571	1.29765484151093 \\
0.502654824574367	1.63918853490783 \\
0.565486677646163	1.10108965328942 \\
0.628318530717959	0.594415612596514 \\
0.691150383789754	0.68808865965832 \\
0.75398223686155	1.08607817979172 \\
0.816814089933346	1.24847985298492 \\
0.879645943005142	1.09010708954791 \\
0.942477796076938	0.908878947520054 \\
1.00530964914873	0.895225935648196 \\
1.06814150222053	0.989953783733897 \\
1.13097335529233	1.05827380875845 \\
1.19380520836412	1.05991975084017 \\
1.25663706143592	1.01598742924614 \\
1.31946891450771	0.941319810578764 \\
1.38230076757951	0.904749357426103 \\
1.4451326206513	0.986701723053111 \\
1.5079644737231	1.14279094596876 \\
1.5707963267949	1.16415246492477 \\
1.63362817986669	0.926002494665665 \\
1.69646003293849	0.684660118649991 \\
1.75929188601028	0.815467322198538 \\
1.82212373908208	1.26912547581701 \\
1.88495559215388	1.51112639048208 \\
1.94778744522567	1.15094426060451 \\
2.01061929829747	0.389331849715646 \\
2.07345115136926	0.4639911283613 \\
2.13628300444106	1.21261669718543 \\
2.19911485751286	1.54207314121035 \\
2.26194671058465	1.24995960466879 \\
2.32477856365645	0.757100171058913 \\
2.38761041672824	0.643288503124793 \\
2.45044226980004	0.947806381814965 \\
2.51327412287183	1.22098157771183 \\
2.57610597594363	1.16102456652254 \\
2.63893782901543	0.945337024652845 \\
2.70176968208722	0.86435075697197 \\
2.76460153515902	0.959725347844326 \\
2.82743338823081	1.06258493054541 \\
2.89026524130261	1.05251518966818 \\
2.95309709437441	0.990524989233057 \\
3.0159289474462	0.972052471661996 \\
3.078760800518	1.01071698551972 \\
3.14159265358979	1.00460962702877 \\
3.20442450666159	0.94697357759458 \\
3.26725635973339	0.955172745369911 \\
3.33008821280518	1.11419863106132 \\
3.39292006587698	1.28543587453709 \\
3.45575191894877	0.983438195250536 \\
3.51858377202057	0.0933720788670742 \\
3.58141562509236	1.29765484151094 \\
3.64424747816416	1.63918853490783 \\
3.70707933123596	1.10108965328943 \\
3.76991118430775	0.594415612596513 \\
3.83274303737955	0.68808865965832 \\
3.89557489045134	1.08607817979172 \\
3.95840674352314	1.24847985298492 \\
4.02123859659494	1.09010708954791 \\
4.08407044966673	0.908878947520054 \\
4.14690230273853	0.895225935648196 \\
4.20973415581032	0.989953783733898 \\
4.27256600888212	1.05827380875845 \\
4.33539786195391	1.05991975084017 \\
4.39822971502571	1.01598742924614 \\
4.46106156809751	0.941319810578765 \\
4.5238934211693	0.904749357426103 \\
4.5867252742411	0.98670172305311 \\
4.64955712731289	1.14279094596876 \\
4.71238898038469	1.16415246492478 \\
4.77522083345649	0.926002494665666 \\
4.83805268652828	0.684660118649991 \\
4.90088453960008	0.815467322198537 \\
4.96371639267187	1.26912547581701 \\
5.02654824574367	1.51112639048208 \\
5.08938009881547	1.15094426060451 \\
5.15221195188726	0.389331849715646 \\
5.21504380495906	0.4639911283613 \\
5.27787565803085	1.21261669718544 \\
5.34070751110265	1.54207314121035 \\
5.40353936417444	1.24995960466879 \\
5.46637121724624	0.757100171058913 \\
5.52920307031804	0.643288503124793 \\
5.59203492338983	0.947806381814969 \\
5.65486677646163	1.22098157771183 \\
5.71769862953342	1.16102456652254 \\
5.78053048260522	0.945337024652845 \\
5.84336233567702	0.86435075697197 \\
5.90619418874881	0.959725347844326 \\
5.96902604182061	1.06258493054541 \\
6.0318578948924	1.05251518966818 \\
6.0946897479642	0.990524989233057 \\
6.157521601036	0.972052471661995 \\
6.22035345410779	1.01071698551972 \\
};

\end{axis}

\end{tikzpicture}

%% file: figures/experiment/experimentNN.tex
\begin{tikzpicture}

\definecolor{color0}{rgb}{0.12156862745098,0.466666666666667,0.705882352941177}
\definecolor{color1}{rgb}{1,0.498039215686275,0.0549019607843137}
\definecolor{color2}{rgb}{0.172549019607843,0.627450980392157,0.172549019607843}
\definecolor{color3}{rgb}{0.83921568627451,0.152941176470588,0.156862745098039}

\begin{axis}[
legend cell align={left},
tick align=outside,
tick pos=left,
x grid style={white!69.01960784313725!black},
xlabel={$x$},
xmin=-0.314159265358979, xmax=6.59734457253857,
y grid style={white!69.01960784313725!black},
ylabel={$y$},
ymin=-0.1, ymax=1.6,
ytick={-0.2,0,0.2,0.4,0.6,0.8,1,1.2,1.4,1.6},
yticklabels={−0.2,0.0,0.2,0.4,0.6,0.8,1.0,1.2,1.4,1.6}
]
\addplot [very thick, color0]
table [row sep=\\]{%
0	1 \\
0.0315737955134653	1 \\
0.0631475910269305	1 \\
0.0947213865403958	1 \\
0.126295182053861	1 \\
0.157868977567326	1 \\
0.189442773080792	1 \\
0.221016568594257	1 \\
0.252590364107722	1 \\
0.284164159621187	1 \\
0.315737955134653	1 \\
0.347311750648118	1 \\
0.378885546161583	1 \\
0.410459341675048	1 \\
0.442033137188514	1 \\
0.473606932701979	1 \\
0.505180728215444	1 \\
0.536754523728909	1 \\
0.568328319242375	1 \\
0.59990211475584	1 \\
0.631475910269305	1 \\
0.66304970578277	1 \\
0.694623501296236	1 \\
0.726197296809701	1 \\
0.757771092323166	1 \\
0.789344887836631	1 \\
0.820918683350097	1 \\
0.852492478863562	1 \\
0.884066274377027	1 \\
0.915640069890492	1 \\
0.947213865403958	1 \\
0.978787660917423	1 \\
1.01036145643089	1 \\
1.04193525194435	1 \\
1.07350904745782	0 \\
1.10508284297128	0 \\
1.13665663848475	0 \\
1.16823043399821	0 \\
1.19980422951168	0 \\
1.23137802502514	0 \\
1.26295182053861	0 \\
1.29452561605208	0 \\
1.32609941156554	0 \\
1.35767320707901	0 \\
1.38924700259247	0 \\
1.42082079810594	0 \\
1.4523945936194	0 \\
1.48396838913287	0 \\
1.51554218464633	0 \\
1.5471159801598	0 \\
1.57868977567326	0 \\
1.61026357118673	0 \\
1.64183736670019	0 \\
1.67341116221366	0 \\
1.70498495772712	0 \\
1.73655875324059	0 \\
1.76813254875405	0 \\
1.79970634426752	0 \\
1.83128013978098	0 \\
1.86285393529445	0 \\
1.89442773080792	0 \\
1.92600152632138	0 \\
1.95757532183485	0 \\
1.98914911734831	0 \\
2.02072291286178	0 \\
2.05229670837524	0 \\
2.08387050388871	0 \\
2.11544429940217	1 \\
2.14701809491564	1 \\
2.1785918904291	1 \\
2.21016568594257	1 \\
2.24173948145603	1 \\
2.2733132769695	1 \\
2.30488707248296	1 \\
2.33646086799643	1 \\
2.36803466350989	1 \\
2.39960845902336	1 \\
2.43118225453682	1 \\
2.46275605005029	1 \\
2.49432984556376	1 \\
2.52590364107722	1 \\
2.55747743659069	1 \\
2.58905123210415	1 \\
2.62062502761762	1 \\
2.65219882313108	1 \\
2.68377261864455	1 \\
2.71534641415801	1 \\
2.74692020967148	1 \\
2.77849400518494	1 \\
2.81006780069841	1 \\
2.84164159621187	1 \\
2.87321539172534	1 \\
2.9047891872388	1 \\
2.93636298275227	1 \\
2.96793677826573	1 \\
2.9995105737792	1 \\
3.03108436929266	1 \\
3.06265816480613	1 \\
3.0942319603196	1 \\
3.12580575583306	1 \\
3.15737955134653	1 \\
3.18895334685999	1 \\
3.22052714237346	1 \\
3.25210093788692	1 \\
3.28367473340039	1 \\
3.31524852891385	1 \\
3.34682232442732	1 \\
3.37839611994078	1 \\
3.40996991545425	1 \\
3.44154371096771	1 \\
3.47311750648118	1 \\
3.50469130199464	1 \\
3.53626509750811	1 \\
3.56783889302157	1 \\
3.59941268853504	1 \\
3.6309864840485	1 \\
3.66256027956197	1 \\
3.69413407507544	1 \\
3.7257078705889	1 \\
3.75728166610237	1 \\
3.78885546161583	1 \\
3.8204292571293	1 \\
3.85200305264276	1 \\
3.88357684815623	1 \\
3.91515064366969	1 \\
3.94672443918316	1 \\
3.97829823469662	1 \\
4.00987203021009	1 \\
4.04144582572355	1 \\
4.07301962123702	1 \\
4.10459341675048	1 \\
4.13616721226395	1 \\
4.16774100777741	1 \\
4.19931480329088	0 \\
4.23088859880434	0 \\
4.26246239431781	0 \\
4.29403618983127	0 \\
4.32560998534474	0 \\
4.35718378085821	0 \\
4.38875757637167	0 \\
4.42033137188514	0 \\
4.4519051673986	0 \\
4.48347896291207	0 \\
4.51505275842553	0 \\
4.546626553939	0 \\
4.57820034945246	0 \\
4.60977414496593	0 \\
4.64134794047939	0 \\
4.67292173599286	0 \\
4.70449553150632	0 \\
4.73606932701979	0 \\
4.76764312253325	0 \\
4.79921691804672	0 \\
4.83079071356018	0 \\
4.86236450907365	0 \\
4.89393830458711	0 \\
4.92551210010058	0 \\
4.95708589561405	0 \\
4.98865969112751	0 \\
5.02023348664098	0 \\
5.05180728215444	0 \\
5.08338107766791	0 \\
5.11495487318137	0 \\
5.14652866869484	0 \\
5.1781024642083	0 \\
5.20967625972177	0 \\
5.24125005523523	1 \\
5.2728238507487	1 \\
5.30439764626216	1 \\
5.33597144177563	1 \\
5.36754523728909	1 \\
5.39911903280256	1 \\
5.43069282831602	1 \\
5.46226662382949	1 \\
5.49384041934295	1 \\
5.52541421485642	1 \\
5.55698801036989	1 \\
5.58856180588335	1 \\
5.62013560139682	1 \\
5.65170939691028	1 \\
5.68328319242375	1 \\
5.71485698793721	1 \\
5.74643078345068	1 \\
5.77800457896414	1 \\
5.80957837447761	1 \\
5.84115216999107	1 \\
5.87272596550454	1 \\
5.904299761018	1 \\
5.93587355653147	1 \\
5.96744735204493	1 \\
5.9990211475584	1 \\
6.03059494307186	1 \\
6.06216873858533	1 \\
6.09374253409879	1 \\
6.12531632961226	1 \\
6.15689012512573	1 \\
6.18846392063919	1 \\
6.22003771615266	1 \\
6.25161151166612	1 \\
6.28318530717959	1 \\
};
\addplot [very thick, color1, dashed]
table [row sep=\\]{%
0	0.946280318578664 \\
0.0315737955134653	0.941744610740138 \\
0.0631475910269305	0.9380041610744 \\
0.0947213865403958	0.935136672365191 \\
0.126295182053861	0.93404841152745 \\
0.157868977567326	0.932759139302874 \\
0.189442773080792	0.930127091017847 \\
0.221016568594257	0.927428089204903 \\
0.252590364107722	0.92466482428775 \\
0.284164159621187	0.922220344144351 \\
0.315737955134653	0.92065147768123 \\
0.347311750648118	0.921268338671058 \\
0.378885546161583	0.922144371783573 \\
0.410459341675048	0.92306394063774 \\
0.442033137188514	0.923953506155245 \\
0.473606932701979	0.92461539482403 \\
0.505180728215444	0.927857998655146 \\
0.536754523728909	0.943382570433602 \\
0.568328319242375	0.979028328652316 \\
0.59990211475584	1.01289845799512 \\
0.631475910269305	1.05115727778506 \\
0.66304970578277	1.08728353025175 \\
0.694623501296236	1.1208233703481 \\
0.726197296809701	1.15281564155818 \\
0.757771092323166	1.17900808794537 \\
0.789344887836631	1.20291535450023 \\
0.820918683350097	1.22527571162685 \\
0.852492478863562	1.24650829098826 \\
0.884066274377027	1.26781156952672 \\
0.915640069890492	1.2679749710416 \\
0.947213865403958	1.02577047825644 \\
0.978787660917423	0.779596536968644 \\
1.01036145643089	0.53318321188151 \\
1.04193525194435	0.297771150072866 \\
1.07350904745782	0.0778705777925459 \\
1.10508284297128	0.0281256103171393 \\
1.13665663848475	0.00679396789563819 \\
1.16823043399821	0.0054099108905431 \\
1.19980422951168	0.0131682680505761 \\
1.23137802502514	0.0159197019434882 \\
1.26295182053861	0.0160815332507772 \\
1.29452561605208	0.0150785050833689 \\
1.32609941156554	0.0129116172815544 \\
1.35767320707901	0.00958302984622111 \\
1.38924700259247	0.0063874642790781 \\
1.42082079810594	0.00225321759009484 \\
1.4523945936194	0.00296865758677839 \\
1.48396838913287	0.00483618160640759 \\
1.51554218464633	0.00251110779719133 \\
1.5471159801598	0.0012465632204261 \\
1.57868977567326	0.000423039924213719 \\
1.61026357118673	0.00265277946257464 \\
1.64183736670019	0.00448189370015273 \\
1.67341116221366	0.00511615108075358 \\
1.70498495772712	0.00456377900585481 \\
1.73655875324059	0.00212338588057548 \\
1.76813254875405	0.00169860712206138 \\
1.79970634426752	0.000659266184525276 \\
1.83128013978098	9.18329407593177e-05 \\
1.86285393529445	0.00038092290736913 \\
1.89442773080792	0.000516003996740899 \\
1.92600152632138	0.000569547892168565 \\
1.95757532183485	4.88429735756579e-06 \\
1.98914911734831	0.00401388677907144 \\
2.02072291286178	0.00596032540533042 \\
2.05229670837524	0.000609589441912695 \\
2.08387050388871	0.0332084580230755 \\
2.11544429940217	0.116311691453385 \\
2.14701809491564	0.278855541610007 \\
2.1785918904291	0.511363241081779 \\
2.21016568594257	0.752280343082898 \\
2.24173948145603	0.991072865430808 \\
2.2733132769695	1.23296819956019 \\
2.30488707248296	1.41847116623712 \\
2.33646086799643	1.48014591946278 \\
2.36803466350989	1.5248115631987 \\
2.39960845902336	1.54228050662343 \\
2.43118225453682	1.48920416088511 \\
2.46275605005029	1.43388770043012 \\
2.49432984556376	1.376709084598 \\
2.52590364107722	1.3226049241083 \\
2.55747743659069	1.26950344982999 \\
2.58905123210415	1.21624559122379 \\
2.62062502761762	1.16317025107506 \\
2.65219882313108	1.13828447490434 \\
2.68377261864455	1.1090715481301 \\
2.71534641415801	1.08082777106582 \\
2.74692020967148	1.05288279713275 \\
2.77849400518494	1.02387542792716 \\
2.81006780069841	1.00047707846367 \\
2.84164159621187	0.994673703991993 \\
2.87321539172534	0.990729081868393 \\
2.9047891872388	0.98642923966985 \\
2.93636298275227	0.981147148945796 \\
2.96793677826573	0.97563300054344 \\
2.9995105737792	0.970226914713877 \\
3.03108436929266	0.964718563791402 \\
3.06265816480613	0.959113438620008 \\
3.0942319603196	0.953417126510362 \\
3.12580575583306	0.948535958066423 \\
3.15737955134653	0.944005413170282 \\
3.18895334685999	0.939835617007298 \\
3.22052714237346	0.936154492567164 \\
3.25210093788692	0.934584819728515 \\
3.28367473340039	0.933507936604443 \\
3.31524852891385	0.931451649926752 \\
3.34682232442732	0.928785792684009 \\
3.37839611994078	0.926054318948701 \\
3.40996991545425	0.923259951513249 \\
3.44154371096771	0.921336889558967 \\
3.47311750648118	0.920779448819593 \\
3.50469130199464	0.92171616686209 \\
3.53626509750811	0.92258881291541 \\
3.56783889302157	0.92351887617751 \\
3.59941268853504	0.924367722252043 \\
3.6309864840485	0.926247827674784 \\
3.66256027956197	0.929445506476751 \\
3.69413407507544	0.962002374084691 \\
3.7257078705889	0.99599955948933 \\
3.75728166610237	1.03066025584195 \\
3.78885546161583	1.06939887680195 \\
3.8204292571293	1.10410873728919 \\
3.85200305264276	1.13743556202434 \\
3.88357684815623	1.16687530231525 \\
3.91515064366969	1.19102243674956 \\
3.94672443918316	1.21444022537419 \\
3.97829823469662	1.23598076078125 \\
4.00987203021009	1.25739420886001 \\
4.04144582572355	1.27773294264106 \\
4.07301962123702	1.14903063258337 \\
4.10459341675048	0.903086563412912 \\
4.13616721226395	0.656059008355537 \\
4.16774100777741	0.412286231714815 \\
4.19931480329088	0.184039221024442 \\
4.23088859880434	0.0475319083693858 \\
4.26246239431781	0.0162198953623097 \\
4.29403618983127	0.00159592681591664 \\
4.32560998534474	0.00909375282262376 \\
4.35718378085821	0.0154019690371642 \\
4.38875757637167	0.0161462421675504 \\
4.42033137188514	0.0157255913199849 \\
4.4519051673986	0.0141404358083814 \\
4.48347896291207	0.011392355750226 \\
4.51505275842553	0.00807121960794327 \\
4.546626553939	0.00442312539408479 \\
4.57820034945246	0.000210914774674553 \\
4.60977414496593	0.00502431593280595 \\
4.64134794047939	0.00358001838679944 \\
4.67292173599286	0.00173322478681504 \\
4.70449553150632	0.00105124438428683 \\
4.73606932701979	0.00129067431977137 \\
4.76764312253325	0.00371660763212944 \\
4.79921691804672	0.00494844694130969 \\
4.83079071356018	0.00498496432309659 \\
4.86236450907365	0.00360318280200989 \\
4.89393830458711	0.000355975747332815 \\
4.92551210010058	0.00149357503251302 \\
4.95708589561405	0.000130752954575764 \\
4.98865969112751	8.43602840872038e-06 \\
5.02023348664098	0.000106596205161624 \\
5.05180728215444	0.000705574346808291 \\
5.08338107766791	0.00014154239117059 \\
5.11495487318137	0.00212735368450434 \\
5.14652866869484	0.00566401341746103 \\
5.1781024642083	0.00495709487436113 \\
5.20967625972177	0.0100370172731803 \\
5.24125005523523	0.0708088781766336 \\
5.2728238507487	0.170730283650164 \\
5.30439764626216	0.392016601082298 \\
5.33597144177563	0.632488369933473 \\
5.36754523728909	0.871167618010421 \\
5.39911903280256	1.11172652583202 \\
5.43069282831602	1.3547676706484 \\
5.46226662382949	1.45111429153108 \\
5.49384041934295	1.50420840267969 \\
5.52541421485642	1.54509396448236 \\
5.55698801036989	1.51635170283531 \\
5.58856180588335	1.461630304129 \\
5.62013560139682	1.40560884574866 \\
5.65170939691028	1.34893395575431 \\
5.68328319242375	1.29607023269461 \\
5.71485698793721	1.24292783435851 \\
5.74643078345068	1.18946337020021 \\
5.77800457896414	1.15180534917966 \\
5.80957837447761	1.12411495958467 \\
5.84115216999107	1.09474970294324 \\
5.87272596550454	1.06687097635371 \\
5.904299761018	1.03839234525982 \\
5.93587355653147	1.00924834039624 \\
5.96744735204493	0.996721357546238 \\
5.9990211475584	0.992712361949718 \\
6.03059494307186	0.988696664827883 \\
6.06216873858533	0.9839833444209 \\
6.09374253409879	0.978296005886267 \\
6.12531632961226	0.972943080024449 \\
6.15689012512573	0.967485181537067 \\
6.18846392063919	0.961927750976028 \\
6.22003771615266	0.95627632810898 \\
6.25161151166612	0.950774889817414 \\
6.28318530717959	0.946280318578664 \\
};
\addplot [very thick, color2, dashed]
table [row sep=\\]{%
0	0.862100526905573 \\
0.0315737955134653	0.871460925697497 \\
0.0631475910269305	0.883655960853772 \\
0.0947213865403958	0.895748780300894 \\
0.126295182053861	0.907727329653445 \\
0.157868977567326	0.919579668432926 \\
0.189442773080792	0.931293981970284 \\
0.221016568594257	0.942858593183021 \\
0.252590364107722	0.954576166476097 \\
0.284164159621187	0.966587387267129 \\
0.315737955134653	0.97841052489984 \\
0.347311750648118	0.990553737767926 \\
0.378885546161583	1.00344447676762 \\
0.410459341675048	1.01871194012475 \\
0.442033137188514	1.03493714474781 \\
0.473606932701979	1.0510513759312 \\
0.505180728215444	1.06142169570379 \\
0.536754523728909	1.06873349460623 \\
0.568328319242375	1.07457468076316 \\
0.59990211475584	1.07911483269782 \\
0.631475910269305	1.08141394797035 \\
0.66304970578277	1.08150926876842 \\
0.694623501296236	1.08109449591044 \\
0.726197296809701	1.08008458037181 \\
0.757771092323166	1.07848052885832 \\
0.789344887836631	1.07365935624736 \\
0.820918683350097	1.04086018358256 \\
0.852492478863562	0.978501661580011 \\
0.884066274377027	0.913621976559671 \\
0.915640069890492	0.835225348663186 \\
0.947213865403958	0.747845630220825 \\
0.978787660917423	0.659208729563087 \\
1.01036145643089	0.568045520536607 \\
1.04193525194435	0.476328495155597 \\
1.07350904745782	0.384149078945588 \\
1.10508284297128	0.291599158353344 \\
1.13665663848475	0.198770989152456 \\
1.16823043399821	0.104887933196235 \\
1.19980422951168	0.0405304391976422 \\
1.23137802502514	0.00829236239421782 \\
1.26295182053861	0.0156537598083493 \\
1.29452561605208	0.00260103163392172 \\
1.32609941156554	0.00322682383581668 \\
1.35767320707901	0.00296001818270397 \\
1.38924700259247	0.00365093207214223 \\
1.42082079810594	0.011157592103992 \\
1.4523945936194	0.00799320441172024 \\
1.48396838913287	0.00365040652162868 \\
1.51554218464633	0.000702899541981938 \\
1.5471159801598	0.00298561400426689 \\
1.57868977567326	0.00436382354716519 \\
1.61026357118673	0.00225804073460315 \\
1.64183736670019	0.00183705628471714 \\
1.67341116221366	0.00323782538485357 \\
1.70498495772712	0.0093529511114015 \\
1.73655875324059	0.0153824443501593 \\
1.76813254875405	0.00418274927246226 \\
1.79970634426752	0.0114188464435708 \\
1.83128013978098	0.028097427799497 \\
1.86285393529445	0.033400619190857 \\
1.89442773080792	0.0278593229929061 \\
1.92600152632138	0.0157212120749546 \\
1.95757532183485	0.00872816569851287 \\
1.98914911734831	0.0405796376732239 \\
2.02072291286178	0.0923702253348192 \\
2.05229670837524	0.168469638357102 \\
2.08387050388871	0.261372952939353 \\
2.11544429940217	0.364777830821038 \\
2.14701809491564	0.468790856316202 \\
2.1785918904291	0.58138093775834 \\
2.21016568594257	0.693269679171029 \\
2.24173948145603	0.803764983547074 \\
2.2733132769695	0.913351271159174 \\
2.30488707248296	1.02238170516273 \\
2.33646086799643	1.13370464598504 \\
2.36803466350989	1.24207733066251 \\
2.39960845902336	1.31814302087684 \\
2.43118225453682	1.38002816521217 \\
2.46275605005029	1.39640864961916 \\
2.49432984556376	1.39379535325878 \\
2.52590364107722	1.37446794335587 \\
2.55747743659069	1.34640492032277 \\
2.58905123210415	1.31331368632613 \\
2.62062502761762	1.27513480652659 \\
2.65219882313108	1.2421961873918 \\
2.68377261864455	1.21316768584462 \\
2.71534641415801	1.19106667783388 \\
2.74692020967148	1.16681678016423 \\
2.77849400518494	1.1387632189641 \\
2.81006780069841	1.11021878761854 \\
2.84164159621187	1.08121193983751 \\
2.87321539172534	1.04802074605087 \\
2.9047891872388	1.01379165302696 \\
2.93636298275227	0.979640755516379 \\
2.96793677826573	0.945646291487644 \\
2.9995105737792	0.912511822353451 \\
3.03108436929266	0.887330812963341 \\
3.06265816480613	0.875012199415183 \\
3.0942319603196	0.866337918473606 \\
3.12580575583306	0.859736041968144 \\
3.15737955134653	0.865342029444766 \\
3.18895334685999	0.877570462718543 \\
3.22052714237346	0.88971590346947 \\
3.25210093788692	0.901753087828656 \\
3.28367473340039	0.913670016868948 \\
3.31524852891385	0.925454811536314 \\
3.34682232442732	0.93709572449114 \\
3.37839611994078	0.948581151818229 \\
3.40996991545425	0.960604543206232 \\
3.44154371096771	0.972523207608623 \\
3.47311750648118	0.984258175860057 \\
3.50469130199464	0.996872623180497 \\
3.53626509750811	1.01090315888215 \\
3.56783889302157	1.02681644673759 \\
3.59941268853504	1.04301612562948 \\
3.6309864840485	1.05742995479499 \\
3.66256027956197	1.06518221052422 \\
3.69413407507544	1.07213718142059 \\
3.7257078705889	1.07690056209759 \\
3.75728166610237	1.08121694072122 \\
3.78885546161583	1.08149332429841 \\
3.8204292571293	1.08137631492746 \\
3.85200305264276	1.08066388195254 \\
3.88357684815623	1.07935673554234 \\
3.91515064366969	1.07745617868882 \\
3.94672443918316	1.06199223446155 \\
3.97829823469662	1.01004232674955 \\
4.00987203021009	0.946173255175564 \\
4.04144582572355	0.879458695870824 \\
4.07301962123702	0.79183045486666 \\
4.10459341675048	0.703587099772709 \\
4.13616721226395	0.613702045689262 \\
4.16774100777741	0.522250532668368 \\
4.19931480329088	0.430290852732785 \\
4.23088859880434	0.33791467329138 \\
4.26246239431781	0.24521407692879 \\
4.29403618983127	0.150674691545865 \\
4.32560998534474	0.0651121510437724 \\
4.35718378085821	0.0243396113875879 \\
4.38875757637167	0.00781341069206615 \\
4.42033137188514	0.0115724387389289 \\
4.4519051673986	0.00331427418913688 \\
4.48347896291207	0.00313378680674131 \\
4.51505275842553	0.000229394981863418 \\
4.546626553939	0.00709066679824472 \\
4.57820034945246	0.00998254195630038 \\
4.60977414496593	0.00588231724105054 \\
4.64134794047939	0.00129802849239546 \\
4.67292173599286	0.00108851429577733 \\
4.70449553150632	0.0042762060914977 \\
4.73606932701979	0.00356498688648932 \\
4.76764312253325	0.000738332958502064 \\
4.79921691804672	0.00429979719783719 \\
4.83079071356018	0.00432645444069485 \\
4.86236450907365	0.0137771035875859 \\
4.89393830458711	0.0112473740174724 \\
4.92551210010058	0.00369308324560019 \\
4.95708589561405	0.0191126621575912 \\
4.98865969112751	0.0307979563843008 \\
5.02023348664098	0.0334536039161097 \\
5.05180728215444	0.0219510042444588 \\
5.08338107766791	0.00535276451261613 \\
5.11495487318137	0.0229790908439027 \\
5.14652866869484	0.0635993277194229 \\
5.1781024642083	0.126922619882797 \\
5.20967625972177	0.211541459552445 \\
5.24125005523523	0.313201205658942 \\
5.2728238507487	0.416262603122823 \\
5.30439764626216	0.524946542182682 \\
5.33597144177563	0.637759830529934 \\
5.36754523728909	0.748609327443585 \\
5.39911903280256	0.85876306832342 \\
5.43069282831602	0.96794815916137 \\
5.46226662382949	1.0781771427272 \\
5.49384041934295	1.18895037631831 \\
5.52541421485642	1.28017986797683 \\
5.55698801036989	1.35141054795502 \\
5.58856180588335	1.3963076017885 \\
5.62013560139682	1.39623212615857 \\
5.65170939691028	1.38445539978347 \\
5.68328319242375	1.36205094822134 \\
5.71485698793721	1.33083727466763 \\
5.74643078345068	1.29428939593457 \\
5.77800457896414	1.25777645306944 \\
5.80957837447761	1.22649554107412 \\
5.84115216999107	1.20217704247821 \\
5.87272596550454	1.1794499464866 \\
5.904299761018	1.15285311790466 \\
5.93587355653147	1.12455059484076 \\
5.96744735204493	1.09577136908426 \\
5.9990211475584	1.064990018098 \\
6.03059494307186	1.03095320649763 \\
6.06216873858533	0.99654036265777 \\
6.09374253409879	0.962681954579605 \\
6.12531632961226	0.92897742325222 \\
6.15689012512573	0.89898496210579 \\
6.18846392063919	0.879646090848514 \\
6.22003771615266	0.870372062559498 \\
6.25161151166612	0.862565685153822 \\
6.28318530717959	0.862100526905572 \\
};
\addplot [very thick, color3, dashed]
table [row sep=\\]{%
0	0.993190398285914 \\
0.0315737955134653	0.986993760213074 \\
0.0631475910269305	0.982431424150011 \\
0.0947213865403958	0.977392236575674 \\
0.126295182053861	0.971973983292591 \\
0.157868977567326	0.963741475841554 \\
0.189442773080792	0.953851632472355 \\
0.221016568594257	0.941734289841514 \\
0.252590364107722	0.928377337492648 \\
0.284164159621187	0.914241457704594 \\
0.315737955134653	0.898804034903838 \\
0.347311750648118	0.883018591574084 \\
0.378885546161583	0.86662911231905 \\
0.410459341675048	0.849240890339769 \\
0.442033137188514	0.832445690291148 \\
0.473606932701979	0.81777578064606 \\
0.505180728215444	0.803882728768327 \\
0.536754523728909	0.788236838944964 \\
0.568328319242375	0.772047403418275 \\
0.59990211475584	0.752866530273102 \\
0.631475910269305	0.735910214359815 \\
0.66304970578277	0.817035806197183 \\
0.694623501296236	0.972742855910417 \\
0.726197296809701	1.15465845332669 \\
0.757771092323166	1.43614850683977 \\
0.789344887836631	1.7750645318954 \\
0.820918683350097	2.06300830204022 \\
0.852492478863562	2.26352182254529 \\
0.884066274377027	2.07049134394622 \\
0.915640069890492	1.11758626192455 \\
0.947213865403958	0.445344477201028 \\
0.978787660917423	0.023027204635334 \\
1.01036145643089	0.000578272175684645 \\
1.04193525194435	0.00369689848630891 \\
1.07350904745782	0.0028285781915059 \\
1.10508284297128	0.00190042942535004 \\
1.13665663848475	0.00116137153008034 \\
1.16823043399821	0.000902783322122326 \\
1.19980422951168	0.00066196296987641 \\
1.23137802502514	0.00044990945389034 \\
1.26295182053861	0.000386844213408913 \\
1.29452561605208	0.000296102811971087 \\
1.32609941156554	0.000202395260451634 \\
1.35767320707901	0.00012530765832186 \\
1.38924700259247	8.93032418929282e-05 \\
1.42082079810594	5.751201514545e-05 \\
1.4523945936194	3.10396257961337e-05 \\
1.48396838913287	8.15674188532622e-06 \\
1.51554218464633	2.09499227964219e-05 \\
1.5471159801598	6.5983496473368e-05 \\
1.57868977567326	8.28508321865096e-05 \\
1.61026357118673	3.22153018855534e-05 \\
1.64183736670019	1.9451678757304e-06 \\
1.67341116221366	1.29021888937222e-05 \\
1.70498495772712	5.17341155605727e-06 \\
1.73655875324059	6.9293870059231e-06 \\
1.76813254875405	8.8920786351504e-06 \\
1.79970634426752	8.77003019451178e-06 \\
1.83128013978098	3.15111394877735e-05 \\
1.86285393529445	4.79439154519937e-05 \\
1.89442773080792	0.000151355808252589 \\
1.92600152632138	0.00021224180366549 \\
1.95757532183485	0.000102022796370657 \\
1.98914911734831	2.28267570692997e-05 \\
2.02072291286178	0.000532103001217241 \\
2.05229670837524	0.00148353073944537 \\
2.08387050388871	0.233180602123498 \\
2.11544429940217	1.13841687693913 \\
2.14701809491564	0.70549404724624 \\
2.1785918904291	0.360796502026766 \\
2.21016568594257	0.410313750322362 \\
2.24173948145603	0.498813453667742 \\
2.2733132769695	0.822409649755533 \\
2.30488707248296	1.34359433201247 \\
2.33646086799643	1.85367141795345 \\
2.36803466350989	1.85054686153894 \\
2.39960845902336	1.68537605690494 \\
2.43118225453682	1.41746791485619 \\
2.46275605005029	1.21086403306936 \\
2.49432984556376	1.14960667453848 \\
2.52590364107722	1.1443404548075 \\
2.55747743659069	1.12904974062985 \\
2.58905123210415	1.11429355777786 \\
2.62062502761762	1.09919476272872 \\
2.65219882313108	1.08394860277791 \\
2.68377261864455	1.07352032314678 \\
2.71534641415801	1.06549347259027 \\
2.74692020967148	1.06181176009766 \\
2.77849400518494	1.05164574170654 \\
2.81006780069841	1.04248728543186 \\
2.84164159621187	1.03974423263407 \\
2.87321539172534	1.03692860143218 \\
2.9047891872388	1.03381857578485 \\
2.93636298275227	1.03030197278114 \\
2.96793677826573	1.02600872755388 \\
2.9995105737792	1.02090058977976 \\
3.03108436929266	1.01526986948372 \\
3.06265816480613	1.00949541678581 \\
3.0942319603196	1.00453775662432 \\
3.12580575583306	0.997221583084421 \\
3.15737955134653	0.989068225191717 \\
3.18895334685999	0.984797877649371 \\
3.22052714237346	0.979968964199037 \\
3.25210093788692	0.974725402599037 \\
3.28367473340039	0.968320326766684 \\
3.31524852891385	0.958881251862344 \\
3.34682232442732	0.948080001260954 \\
3.37839611994078	0.935135788496911 \\
3.40996991545425	0.921460621176404 \\
3.44154371096771	0.906465301870626 \\
3.47311750648118	0.890987815483075 \\
3.50469130199464	0.874898349274536 \\
3.53626509750811	0.858071779561766 \\
3.56783889302157	0.840837134894283 \\
3.59941268853504	0.824458718617442 \\
3.6309864840485	0.81130672694036 \\
3.66256027956197	0.796107050756392 \\
3.69413407507544	0.780256609281816 \\
3.7257078705889	0.762500864490903 \\
3.75728166610237	0.741663769909579 \\
3.78885546161583	0.763469310456242 \\
3.8204292571293	0.891243743663419 \\
3.85200305264276	1.06021541428228 \\
3.88357684815623	1.26945579823235 \\
3.91515064366969	1.62389922489916 \\
3.94672443918316	1.92057381454178 \\
3.97829823469662	2.18285678818186 \\
4.00987203021009	2.31326880547179 \\
4.04144582572355	1.5788617634564 \\
4.07301962123702	0.742300255306429 \\
4.10459341675048	0.164097069181119 \\
4.13616721226395	0.00378770261568898 \\
4.16774100777741	0.00359956302879434 \\
4.19931480329088	0.00329227045308282 \\
4.23088859880434	0.00236459263728549 \\
4.26246239431781	0.00143593439088459 \\
4.29403618983127	0.00102668307188686 \\
4.32560998534474	0.000770072526857526 \\
4.35718378085821	0.000559034093010793 \\
4.38875757637167	0.00040950473282713 \\
4.42033137188514	0.000341850005131782 \\
4.4519051673986	0.000249614035084791 \\
4.48347896291207	0.000154458255980192 \\
4.51505275842553	0.000107394614564153 \\
4.546626553939	7.22099091554212e-05 \\
4.57820034945246	4.29153306788915e-05 \\
4.60977414496593	2.01802257534314e-05 \\
4.64134794047939	4.0132570538684e-06 \\
4.67292173599286	4.34105162015891e-05 \\
4.70449553150632	8.86632379517791e-05 \\
4.73606932701979	5.88980129274178e-05 \\
4.76764312253325	3.51720277941975e-06 \\
4.79921691804672	7.41159286821169e-06 \\
4.83079071356018	1.11541123882092e-05 \\
4.86236450907365	8.54922371298916e-07 \\
4.89393830458711	8.12883290108452e-06 \\
4.92551210010058	2.17097936617422e-06 \\
4.95708589561405	1.96910444079934e-05 \\
4.98865969112751	3.12947109374073e-05 \\
5.02023348664098	8.70990771496272e-05 \\
5.05180728215444	0.000227459399162522 \\
5.08338107766791	0.000165071184382706 \\
5.11495487318137	4.22751527062726e-05 \\
5.14652866869484	9.25128239023035e-05 \\
5.1781024642083	0.00103893594910027 \\
5.20967625972177	0.00264269188596885 \\
5.24125005523523	0.801746995892292 \\
5.2728238507487	0.97075158194384 \\
5.30439764626216	0.497496092853063 \\
5.33597144177563	0.382811427336769 \\
5.36754523728909	0.450905307743091 \\
5.39911903280256	0.632387094485372 \\
5.43069282831602	1.06620086942369 \\
5.46226662382949	1.63822051097315 \\
5.49384041934295	1.86101293634868 \\
5.52541421485642	1.78400128471307 \\
5.55698801036989	1.56333458783875 \\
5.58856180588335	1.30327874555308 \\
5.62013560139682	1.1607575389944 \\
5.65170939691028	1.14871164806617 \\
5.68328319242375	1.13791025314301 \\
5.71485698793721	1.12147539397831 \\
5.74643078345068	1.10683055077311 \\
5.77800457896414	1.09169950962641 \\
5.80957837447761	1.07884708021387 \\
5.84115216999107	1.06922167589097 \\
5.87272596550454	1.06165466326813 \\
5.904299761018	1.05754742123485 \\
5.93587355653147	1.04602090243368 \\
5.96744735204493	1.04084702545835 \\
5.9990211475584	1.03843798277707 \\
6.03059494307186	1.03526526459988 \\
6.06216873858533	1.03215612292828 \\
6.09374253409879	1.0282494886732 \\
6.12531632961226	1.02357631918329 \\
6.15689012512573	1.01813076763079 \\
6.18846392063919	1.0123889906966 \\
6.22003771615266	1.0069733135934 \\
6.25161151166612	1.00106806403615 \\
6.28318530717959	0.993190398285912 \\
};

\end{axis}

\end{tikzpicture}

%% file: figures/experiment/experimentPL.tex
\begin{tikzpicture}

\definecolor{color0}{rgb}{0.12156862745098,0.466666666666667,0.705882352941177}
\definecolor{color1}{rgb}{1,0.498039215686275,0.0549019607843137}
\definecolor{color2}{rgb}{0.172549019607843,0.627450980392157,0.172549019607843}
\definecolor{color3}{rgb}{0.83921568627451,0.152941176470588,0.156862745098039}

\begin{axis}[
legend cell align={left},
tick align=outside,
tick pos=left,
x grid style={white!69.01960784313725!black},
xlabel={$x$},
xmin=-0.314159265358979, xmax=6.59734457253857,
y grid style={white!69.01960784313725!black},
ylabel={$y$},
ymin=-0.1, ymax=1.6,
ytick={-0.2,0,0.2,0.4,0.6,0.8,1,1.2,1.4,1.6},
yticklabels={−0.2,0.0,0.2,0.4,0.6,0.8,1.0,1.2,1.4,1.6}
]

\addplot [very thick, color0]
table [row sep=\\]{%
0	1 \\
0.0315737955134653	1 \\
0.0631475910269305	1 \\
0.0947213865403958	1 \\
0.126295182053861	1 \\
0.157868977567326	1 \\
0.189442773080792	1 \\
0.221016568594257	1 \\
0.252590364107722	1 \\
0.284164159621187	1 \\
0.315737955134653	1 \\
0.347311750648118	1 \\
0.378885546161583	1 \\
0.410459341675048	1 \\
0.442033137188514	1 \\
0.473606932701979	1 \\
0.505180728215444	1 \\
0.536754523728909	1 \\
0.568328319242375	1 \\
0.59990211475584	1 \\
0.631475910269305	1 \\
0.66304970578277	1 \\
0.694623501296236	1 \\
0.726197296809701	1 \\
0.757771092323166	1 \\
0.789344887836631	1 \\
0.820918683350097	1 \\
0.852492478863562	1 \\
0.884066274377027	1 \\
0.915640069890492	1 \\
0.947213865403958	1 \\
0.978787660917423	1 \\
1.01036145643089	1 \\
1.04193525194435	1 \\
1.07350904745782	0 \\
1.10508284297128	0 \\
1.13665663848475	0 \\
1.16823043399821	0 \\
1.19980422951168	0 \\
1.23137802502514	0 \\
1.26295182053861	0 \\
1.29452561605208	0 \\
1.32609941156554	0 \\
1.35767320707901	0 \\
1.38924700259247	0 \\
1.42082079810594	0 \\
1.4523945936194	0 \\
1.48396838913287	0 \\
1.51554218464633	0 \\
1.5471159801598	0 \\
1.57868977567326	0 \\
1.61026357118673	0 \\
1.64183736670019	0 \\
1.67341116221366	0 \\
1.70498495772712	0 \\
1.73655875324059	0 \\
1.76813254875405	0 \\
1.79970634426752	0 \\
1.83128013978098	0 \\
1.86285393529445	0 \\
1.89442773080792	0 \\
1.92600152632138	0 \\
1.95757532183485	0 \\
1.98914911734831	0 \\
2.02072291286178	0 \\
2.05229670837524	0 \\
2.08387050388871	0 \\
2.11544429940217	1 \\
2.14701809491564	1 \\
2.1785918904291	1 \\
2.21016568594257	1 \\
2.24173948145603	1 \\
2.2733132769695	1 \\
2.30488707248296	1 \\
2.33646086799643	1 \\
2.36803466350989	1 \\
2.39960845902336	1 \\
2.43118225453682	1 \\
2.46275605005029	1 \\
2.49432984556376	1 \\
2.52590364107722	1 \\
2.55747743659069	1 \\
2.58905123210415	1 \\
2.62062502761762	1 \\
2.65219882313108	1 \\
2.68377261864455	1 \\
2.71534641415801	1 \\
2.74692020967148	1 \\
2.77849400518494	1 \\
2.81006780069841	1 \\
2.84164159621187	1 \\
2.87321539172534	1 \\
2.9047891872388	1 \\
2.93636298275227	1 \\
2.96793677826573	1 \\
2.9995105737792	1 \\
3.03108436929266	1 \\
3.06265816480613	1 \\
3.0942319603196	1 \\
3.12580575583306	1 \\
3.15737955134653	1 \\
3.18895334685999	1 \\
3.22052714237346	1 \\
3.25210093788692	1 \\
3.28367473340039	1 \\
3.31524852891385	1 \\
3.34682232442732	1 \\
3.37839611994078	1 \\
3.40996991545425	1 \\
3.44154371096771	1 \\
3.47311750648118	1 \\
3.50469130199464	1 \\
3.53626509750811	1 \\
3.56783889302157	1 \\
3.59941268853504	1 \\
3.6309864840485	1 \\
3.66256027956197	1 \\
3.69413407507544	1 \\
3.7257078705889	1 \\
3.75728166610237	1 \\
3.78885546161583	1 \\
3.8204292571293	1 \\
3.85200305264276	1 \\
3.88357684815623	1 \\
3.91515064366969	1 \\
3.94672443918316	1 \\
3.97829823469662	1 \\
4.00987203021009	1 \\
4.04144582572355	1 \\
4.07301962123702	1 \\
4.10459341675048	1 \\
4.13616721226395	1 \\
4.16774100777741	1 \\
4.19931480329088	0 \\
4.23088859880434	0 \\
4.26246239431781	0 \\
4.29403618983127	0 \\
4.32560998534474	0 \\
4.35718378085821	0 \\
4.38875757637167	0 \\
4.42033137188514	0 \\
4.4519051673986	0 \\
4.48347896291207	0 \\
4.51505275842553	0 \\
4.546626553939	0 \\
4.57820034945246	0 \\
4.60977414496593	0 \\
4.64134794047939	0 \\
4.67292173599286	0 \\
4.70449553150632	0 \\
4.73606932701979	0 \\
4.76764312253325	0 \\
4.79921691804672	0 \\
4.83079071356018	0 \\
4.86236450907365	0 \\
4.89393830458711	0 \\
4.92551210010058	0 \\
4.95708589561405	0 \\
4.98865969112751	0 \\
5.02023348664098	0 \\
5.05180728215444	0 \\
5.08338107766791	0 \\
5.11495487318137	0 \\
5.14652866869484	0 \\
5.1781024642083	0 \\
5.20967625972177	0 \\
5.24125005523523	1 \\
5.2728238507487	1 \\
5.30439764626216	1 \\
5.33597144177563	1 \\
5.36754523728909	1 \\
5.39911903280256	1 \\
5.43069282831602	1 \\
5.46226662382949	1 \\
5.49384041934295	1 \\
5.52541421485642	1 \\
5.55698801036989	1 \\
5.58856180588335	1 \\
5.62013560139682	1 \\
5.65170939691028	1 \\
5.68328319242375	1 \\
5.71485698793721	1 \\
5.74643078345068	1 \\
5.77800457896414	1 \\
5.80957837447761	1 \\
5.84115216999107	1 \\
5.87272596550454	1 \\
5.904299761018	1 \\
5.93587355653147	1 \\
5.96744735204493	1 \\
5.9990211475584	1 \\
6.03059494307186	1 \\
6.06216873858533	1 \\
6.09374253409879	1 \\
6.12531632961226	1 \\
6.15689012512573	1 \\
6.18846392063919	1 \\
6.22003771615266	1 \\
6.25161151166612	1 \\
6.28318530717959	1 \\
};
\addplot [very thick, color1, dashed]
table [row sep=\\]{%
0	0.805203233531251 \\
0.0315737955134653	0.827199016341983 \\
0.0631475910269305	0.849194799152881 \\
0.0947213865403958	0.871190581963779 \\
0.126295182053861	0.893186364774677 \\
0.157868977567326	0.915182147585575 \\
0.189442773080792	0.937177930396473 \\
0.221016568594257	0.959173713207371 \\
0.252590364107722	0.981169496018269 \\
0.284164159621187	1.00316527882917 \\
0.315737955134653	1.02516106164006 \\
0.347311750648118	1.04715684445096 \\
0.378885546161583	1.06915262726186 \\
0.410459341675048	1.09114841007276 \\
0.442033137188514	1.11314419288366 \\
0.473606932701979	1.13513997569455 \\
0.505180728215444	1.15713575850545 \\
0.536754523728909	1.17913154131635 \\
0.568328319242375	1.20112732412725 \\
0.59990211475584	1.22312310693815 \\
0.631475910269305	1.2366734857831 \\
0.66304970578277	1.17421522893063 \\
0.694623501296236	1.11175697207816 \\
0.726197296809701	1.04929871522569 \\
0.757771092323166	0.986840458373228 \\
0.789344887836631	0.924382201520761 \\
0.820918683350097	0.861923944668294 \\
0.852492478863562	0.799465687815827 \\
0.884066274377027	0.737007430963361 \\
0.915640069890492	0.674549174110894 \\
0.947213865403958	0.612090917258427 \\
0.978787660917423	0.54963266040596 \\
1.01036145643089	0.487174403553494 \\
1.04193525194435	0.424716146701027 \\
1.07350904745782	0.36225788984856 \\
1.10508284297128	0.299799632996094 \\
1.13665663848475	0.237341376143627 \\
1.16823043399821	0.17488311929116 \\
1.19980422951168	0.112424862438694 \\
1.23137802502514	0.0499666055862265 \\
1.26295182053861	1.02726425049758e-10 \\
1.29452561605208	9.69364963461766e-11 \\
1.32609941156554	9.11465678865402e-11 \\
1.35767320707901	8.53566392100633e-11 \\
1.38924700259247	7.95667107233219e-11 \\
1.42082079810594	7.37767820739501e-11 \\
1.4523945936194	6.79868536414188e-11 \\
1.48396838913287	6.21969251546773e-11 \\
1.51554218464633	5.64069965053055e-11 \\
1.5471159801598	5.06170680185641e-11 \\
1.57868977567326	4.48271393962973e-11 \\
1.61026357118673	3.90372108824508e-11 \\
1.64183736670019	3.32472823686043e-11 \\
1.67341116221366	2.74573538005477e-11 \\
1.70498495772712	2.16674252324911e-11 \\
1.73655875324059	1.58774966373294e-11 \\
1.76813254875405	1.0087568150588e-11 \\
1.79970634426752	4.29763960963642e-12 \\
1.83128013978098	1.49228893131514e-12 \\
1.86285393529445	7.28221743160912e-12 \\
1.89442773080792	0.0207008697415554 \\
1.92600152632138	0.0897037689195144 \\
1.95757532183485	0.158706668097475 \\
1.98914911734831	0.227709567275433 \\
2.02072291286178	0.296712466453393 \\
2.05229670837524	0.365715365631352 \\
2.08387050388871	0.434718264809312 \\
2.11544429940217	0.503721163987273 \\
2.14701809491564	0.572724063165231 \\
2.1785918904291	0.641726962343192 \\
2.21016568594257	0.710729861521151 \\
2.24173948145603	0.77973276069911 \\
2.2733132769695	0.848735659877071 \\
2.30488707248296	0.917738559055029 \\
2.33646086799643	0.98674145823299 \\
2.36803466350989	1.05574435741095 \\
2.39960845902336	1.12474725658891 \\
2.43118225453682	1.19375015576687 \\
2.46275605005029	1.26275305494483 \\
2.49432984556376	1.33175595412279 \\
2.52590364107722	1.36174152357815 \\
2.55747743659069	1.33320109844755 \\
2.58905123210415	1.30466067331695 \\
2.62062502761762	1.27612024818635 \\
2.65219882313108	1.24757982305575 \\
2.68377261864455	1.21903939792515 \\
2.71534641415801	1.19049897279455 \\
2.74692020967148	1.16195854766395 \\
2.77849400518494	1.13341812253335 \\
2.81006780069841	1.10487769740275 \\
2.84164159621187	1.07633727227215 \\
2.87321539172534	1.04779684714154 \\
2.9047891872388	1.01925642201094 \\
2.93636298275227	0.990715996880343 \\
2.96793677826573	0.962175571749742 \\
2.9995105737792	0.933635146619141 \\
3.03108436929266	0.90509472148854 \\
3.06265816480613	0.87655429635794 \\
3.0942319603196	0.848013871227339 \\
3.12580575583306	0.819473446096738 \\
3.15737955134653	0.816201124936534 \\
3.18895334685999	0.838196907747432 \\
3.22052714237346	0.86019269055833 \\
3.25210093788692	0.882188473369228 \\
3.28367473340039	0.904184256180126 \\
3.31524852891385	0.926180038991024 \\
3.34682232442732	0.948175821801921 \\
3.37839611994078	0.97017160461282 \\
3.40996991545425	0.992167387423718 \\
3.44154371096771	1.01416317023462 \\
3.47311750648118	1.03615895304551 \\
3.50469130199464	1.05815473585641 \\
3.53626509750811	1.08015051866731 \\
3.56783889302157	1.10214630147821 \\
3.59941268853504	1.12414208428911 \\
3.6309864840485	1.1461378671 \\
3.66256027956197	1.1681336499109 \\
3.69413407507544	1.1901294327218 \\
3.7257078705889	1.2121252155327 \\
3.75728166610237	1.2341209983436 \\
3.78885546161583	1.20544435735686 \\
3.8204292571293	1.1429861005044 \\
3.85200305264276	1.08052784365193 \\
3.88357684815623	1.01806958679946 \\
3.91515064366969	0.955611329946993 \\
3.94672443918316	0.893153073094527 \\
3.97829823469662	0.830694816242062 \\
4.00987203021009	0.768236559389596 \\
4.04144582572355	0.705778302537127 \\
4.07301962123702	0.64332004568466 \\
4.10459341675048	0.580861788832194 \\
4.13616721226395	0.518403531979728 \\
4.16774100777741	0.455945275127261 \\
4.19931480329088	0.393487018274794 \\
4.23088859880434	0.331028761422326 \\
4.26246239431781	0.26857050456986 \\
4.29403618983127	0.206112247717394 \\
4.32560998534474	0.143653990864927 \\
4.35718378085821	0.0811957340124593 \\
4.38875757637167	0.0187374771599919 \\
4.42033137188514	9.98314605624422e-11 \\
4.4519051673986	9.40415321028058e-11 \\
4.48347896291207	8.82516036160644e-11 \\
4.51505275842553	8.24616750209027e-11 \\
4.546626553939	7.66717465341613e-11 \\
4.57820034945246	7.08818180474198e-11 \\
4.60977414496593	6.50918894522581e-11 \\
4.64134794047939	5.93019608842015e-11 \\
4.67292173599286	5.3512032370355e-11 \\
4.70449553150632	4.77221036396681e-11 \\
4.73606932701979	4.19321751800317e-11 \\
4.76764312253325	3.61422466661852e-11 \\
4.79921691804672	3.03523180981286e-11 \\
4.83079071356018	2.4562389557177e-11 \\
4.86236450907365	1.87724610162254e-11 \\
4.89393830458711	1.29825324752739e-11 \\
4.92551210010058	7.19260393432231e-12 \\
4.95708589561405	1.40267533916064e-12 \\
4.98865969112751	4.38725317468587e-12 \\
5.02023348664098	1.01771817427425e-11 \\
5.05180728215444	0.0552023193305349 \\
5.08338107766791	0.124205218508493 \\
5.11495487318137	0.193208117686454 \\
5.14652866869484	0.262211016864414 \\
5.1781024642083	0.331213916042373 \\
5.20967625972177	0.400216815220333 \\
5.24125005523523	0.469219714398292 \\
5.2728238507487	0.538222613576252 \\
5.30439764626216	0.607225512754212 \\
5.33597144177563	0.676228411932171 \\
5.36754523728909	0.745231311110131 \\
5.39911903280256	0.81423421028809 \\
5.43069282831602	0.88323710946605 \\
5.46226662382949	0.952240008644011 \\
5.49384041934295	1.02124290782197 \\
5.52541421485642	1.09024580699993 \\
5.55698801036989	1.15924870617789 \\
5.58856180588335	1.22825160535585 \\
5.62013560139682	1.29725450453381 \\
5.65170939691028	1.36625740371177 \\
5.68328319242375	1.34747131101285 \\
5.71485698793721	1.31893088588225 \\
5.74643078345068	1.29039046075165 \\
5.77800457896414	1.26185003562105 \\
5.80957837447761	1.23330961049045 \\
5.84115216999107	1.20476918535985 \\
5.87272596550454	1.17622876022925 \\
5.904299761018	1.14768833509865 \\
5.93587355653147	1.11914790996805 \\
5.96744735204493	1.09060748483745 \\
5.9990211475584	1.06206705970684 \\
6.03059494307186	1.03352663457624 \\
6.06216873858533	1.00498620944564 \\
6.09374253409879	0.976445784315043 \\
6.12531632961226	0.947905359184442 \\
6.15689012512573	0.919364934053841 \\
6.18846392063919	0.890824508923239 \\
6.22003771615266	0.862284083792639 \\
6.25161151166612	0.833743658662039 \\
6.28318530717959	0.805203233531437 \\
};
\addplot [very thick, color2, dashed]
table [row sep=\\]{%
0	1.72588491079557 \\
0.0315737955134653	1.56917413027797 \\
0.0631475910269305	1.41246334975936 \\
0.0947213865403958	1.25575256924075 \\
0.126295182053861	1.09904178872213 \\
0.157868977567326	0.942331008203519 \\
0.189442773080792	0.785620227684906 \\
0.221016568594257	0.628909447166293 \\
0.252590364107722	0.47219866664768 \\
0.284164159621187	0.315487886129068 \\
0.315737955134653	0.17425062563424 \\
0.347311750648118	0.327010245615702 \\
0.378885546161583	0.479769865597165 \\
0.410459341675048	0.632529485578627 \\
0.442033137188514	0.78528910556009 \\
0.473606932701979	0.938048725541553 \\
0.505180728215444	1.09080834552302 \\
0.536754523728909	1.24356796550448 \\
0.568328319242375	1.39632758548594 \\
0.59990211475584	1.5490872054674 \\
0.631475910269305	1.67556359597308 \\
0.66304970578277	1.56549092118462 \\
0.694623501296236	1.45541824639616 \\
0.726197296809701	1.3453455716077 \\
0.757771092323166	1.23527289681924 \\
0.789344887836631	1.12520022203078 \\
0.820918683350097	1.01512754724232 \\
0.852492478863562	0.90505487245386 \\
0.884066274377027	0.7949821976654 \\
0.915640069890492	0.68490952287694 \\
0.947213865403958	0.58243295908186 \\
0.978787660917423	0.523001024251043 \\
1.01036145643089	0.463569089420226 \\
1.04193525194435	0.40413715458941 \\
1.07350904745782	0.344705219758593 \\
1.10508284297128	0.285273284927777 \\
1.13665663848475	0.22584135009696 \\
1.16823043399821	0.166409415266143 \\
1.19980422951168	0.106977480435327 \\
1.23137802502514	0.0475455456045097 \\
1.26295182053861	2.20834242071252e-09 \\
1.29452561605208	1.9475829075822e-09 \\
1.32609941156554	1.68682339437057e-09 \\
1.35767320707901	1.42606388126736e-09 \\
1.38924700259247	1.16530436808284e-09 \\
1.42082079810594	9.04544854979632e-10 \\
1.4523945936194	6.43785341768004e-10 \\
1.48396838913287	3.83025828502165e-10 \\
1.51554218464633	1.22266315426061e-10 \\
1.5471159801598	1.38493197731357e-10 \\
1.57868977567326	3.23261230039977e-10 \\
1.61026357118673	2.80054607298573e-10 \\
1.64183736670019	2.36847984475853e-10 \\
1.67341116221366	1.93641361734449e-10 \\
1.70498495772712	1.50434738938835e-10 \\
1.73655875324059	1.07228116224536e-10 \\
1.76813254875405	6.40214934018168e-11 \\
1.79970634426752	2.08148706333077e-11 \\
1.83128013978098	2.23917521352014e-11 \\
1.86285393529445	6.55983749308156e-11 \\
1.89442773080792	0.0094440047327813 \\
1.92600152632138	0.0409240208289513 \\
1.95757532183485	0.0724040369251218 \\
1.98914911734831	0.103884053021292 \\
2.02072291286178	0.135364069117462 \\
2.05229670837524	0.166844085213632 \\
2.08387050388871	0.198324101309803 \\
2.11544429940217	0.229804117405973 \\
2.14701809491564	0.261284133502143 \\
2.1785918904291	0.292764149598314 \\
2.21016568594257	0.378347999012108 \\
2.24173948145603	0.564410396019244 \\
2.2733132769695	0.750472793026383 \\
2.30488707248296	0.936535190033517 \\
2.33646086799643	1.12259758704066 \\
2.36803466350989	1.30865998404779 \\
2.39960845902336	1.49472238105493 \\
2.43118225453682	1.68078477806206 \\
2.46275605005029	1.8668471750692 \\
2.49432984556376	2.05290957207634 \\
2.52590364107722	2.08556908215276 \\
2.55747743659069	1.88812426182492 \\
2.58905123210415	1.69067944149709 \\
2.62062502761762	1.49323462116925 \\
2.65219882313108	1.29578980084141 \\
2.68377261864455	1.09834498051358 \\
2.71534641415801	0.900900160185741 \\
2.74692020967148	0.703455339857907 \\
2.77849400518494	0.506010519530069 \\
2.81006780069841	0.308565699202237 \\
2.84164159621187	0.268982227742233 \\
2.87321539172534	0.422340404905637 \\
2.9047891872388	0.575698582069038 \\
2.93636298275227	0.729056759232443 \\
2.96793677826573	0.882414936395844 \\
2.9995105737792	1.03577311355925 \\
3.03108436929266	1.18913129072265 \\
3.06265816480613	1.34248946788605 \\
3.0942319603196	1.49584764504946 \\
3.12580575583306	1.64920582221286 \\
3.15737955134653	1.64752952053728 \\
3.18895334685999	1.49081874001867 \\
3.22052714237346	1.33410795950005 \\
3.25210093788692	1.17739717898144 \\
3.28367473340039	1.02068639846282 \\
3.31524852891385	0.863975617944214 \\
3.34682232442732	0.707264837425603 \\
3.37839611994078	0.550554056906988 \\
3.40996991545425	0.393843276388375 \\
3.44154371096771	0.237132495869759 \\
3.47311750648118	0.25063043562497 \\
3.50469130199464	0.403390055606431 \\
3.53626509750811	0.556149675587896 \\
3.56783889302157	0.708909295569359 \\
3.59941268853504	0.861668915550823 \\
3.6309864840485	1.01442853553228 \\
3.66256027956197	1.16718815551374 \\
3.69413407507544	1.31994777549521 \\
3.7257078705889	1.47270739547667 \\
3.75728166610237	1.62546701545814 \\
3.78885546161583	1.62052725857885 \\
3.8204292571293	1.51045458379039 \\
3.85200305264276	1.40038190900193 \\
3.88357684815623	1.29030923421347 \\
3.91515064366969	1.18023655942501 \\
3.94672443918316	1.07016388463655 \\
3.97829823469662	0.960091209848091 \\
4.00987203021009	0.850018535059632 \\
4.04144582572355	0.739945860271169 \\
4.07301962123702	0.629873185482708 \\
4.10459341675048	0.552716991666452 \\
4.13616721226395	0.493285056835636 \\
4.16774100777741	0.433853122004819 \\
4.19931480329088	0.374421187174002 \\
4.23088859880434	0.314989252343184 \\
4.26246239431781	0.255557317512368 \\
4.29403618983127	0.196125382681552 \\
4.32560998534474	0.136693447850735 \\
4.35718378085821	0.0772615130199175 \\
4.38875757637167	0.0178295781891001 \\
4.42033137188514	2.07796266424223e-09 \\
4.4519051673986	1.81720315094928e-09 \\
4.48347896291207	1.55644363776476e-09 \\
4.51505275842553	1.29568412463445e-09 \\
4.546626553939	1.03492461150413e-09 \\
4.57820034945246	7.74165098265398e-10 \\
4.60977414496593	5.13405585107979e-10 \\
4.64134794047939	2.52646071977666e-10 \\
4.67292173599286	8.11344123396313e-12 \\
4.70449553150632	2.68872954310067e-10 \\
4.73606932701979	3.0165791864217e-10 \\
4.76764312253325	2.5845129587366e-10 \\
4.79921691804672	2.15244673078046e-10 \\
4.83079071356018	1.72038050309537e-10 \\
4.86236450907365	1.28831427568133e-10 \\
4.89393830458711	8.56248047454138e-11 \\
4.92551210010058	4.24181820311148e-11 \\
4.95708589561405	7.88440737394328e-13 \\
4.98865969112751	4.39950635601136e-11 \\
5.02023348664098	8.72016863286227e-11 \\
5.05180728215444	0.0251840127808663 \\
5.08338107766791	0.056664028877036 \\
5.11495487318137	0.0881440449732067 \\
5.14652866869484	0.119624061069377 \\
5.1781024642083	0.151104077165547 \\
5.20967625972177	0.182584093261718 \\
5.24125005523523	0.214064109357887 \\
5.2728238507487	0.245544125454058 \\
5.30439764626216	0.277024141550229 \\
5.33597144177563	0.308504157646398 \\
5.36754523728909	0.471379197515677 \\
5.39911903280256	0.657441594522811 \\
5.43069282831602	0.84350399152995 \\
5.46226662382949	1.02956638853709 \\
5.49384041934295	1.21562878554422 \\
5.52541421485642	1.40169118255136 \\
5.55698801036989	1.5877535795585 \\
5.58856180588335	1.77381597656563 \\
5.62013560139682	1.95987837357277 \\
5.65170939691028	2.14594077057991 \\
5.68328319242375	1.98684667198884 \\
5.71485698793721	1.789401851661 \\
5.74643078345068	1.59195703133317 \\
5.77800457896414	1.39451221100533 \\
5.80957837447761	1.1970673906775 \\
5.84115216999107	0.999622570349658 \\
5.87272596550454	0.80217775002182 \\
5.904299761018	0.604732929693987 \\
5.93587355653147	0.407288109366155 \\
5.96744735204493	0.209843289038316 \\
5.9990211475584	0.345661316323937 \\
6.03059494307186	0.499019493487342 \\
6.06216873858533	0.652377670650742 \\
6.09374253409879	0.805735847814142 \\
6.12531632961226	0.959094024977547 \\
6.15689012512573	1.11245220214095 \\
6.18846392063919	1.26581037930436 \\
6.22003771615266	1.41916855646776 \\
6.25161151166612	1.57252673363116 \\
6.28318530717959	1.72588491079456 \\
};
\addplot [very thick, color3, dashed]
table [row sep=\\]{%
0	1.36148562732505 \\
0.0315737955134653	1.33807356630238 \\
0.0631475910269305	1.31466150527925 \\
0.0947213865403958	1.29124944425612 \\
0.126295182053861	1.26783738323299 \\
0.157868977567326	1.24184085003104 \\
0.189442773080792	1.11504990184067 \\
0.221016568594257	0.988258953650287 \\
0.252590364107722	0.861468005459909 \\
0.284164159621187	0.734677057269531 \\
0.315737955134653	0.612008947481619 \\
0.347311750648118	0.567674767347072 \\
0.378885546161583	0.523340587212525 \\
0.410459341675048	0.479006407077977 \\
0.442033137188514	0.43467222694343 \\
0.473606932701979	0.406205041296717 \\
0.505180728215444	0.573430787678638 \\
0.536754523728909	0.740656534060558 \\
0.568328319242375	0.907882280442479 \\
0.59990211475584	1.0751080268244 \\
0.631475910269305	1.23780903668335 \\
0.66304970578277	1.35978741783352 \\
0.694623501296236	1.48176579898369 \\
0.726197296809701	1.60374418013385 \\
0.757771092323166	1.72572256128402 \\
0.789344887836631	1.80503999431809 \\
0.820918683350097	1.58573079052609 \\
0.852492478863562	1.36642158673408 \\
0.884066274377027	1.14711238294208 \\
0.915640069890492	0.927803179150069 \\
0.947213865403958	0.719036880085688 \\
0.978787660917423	0.57001370781365 \\
1.01036145643089	0.420990535541611 \\
1.04193525194435	0.271967363269574 \\
1.07350904745782	0.122944190997534 \\
1.10508284297128	7.055481424296e-08 \\
1.13665663848475	5.15979153830747e-08 \\
1.16823043399821	3.26410165228913e-08 \\
1.19980422951168	1.36841176639547e-08 \\
1.23137802502514	5.27278119624231e-09 \\
1.26295182053861	1.82885835520801e-08 \\
1.29452561605208	7.53999633757753e-09 \\
1.32609941156554	3.20859087828704e-09 \\
1.35767320707901	1.39571780927964e-08 \\
1.38924700259247	2.47057653086067e-08 \\
1.42082079810594	2.99985012112611e-08 \\
1.4523945936194	1.6498854692236e-08 \\
1.48396838913287	2.99920817323804e-09 \\
1.51554218464633	1.05004383411385e-08 \\
1.5471159801598	2.40000848592556e-08 \\
1.57868977567326	3.26468076268429e-08 \\
1.61026357118673	2.67347549630652e-08 \\
1.64183736670019	2.08227022991791e-08 \\
1.67341116221366	1.49106496357538e-08 \\
1.70498495772712	8.99859697219289e-09 \\
1.73655875324059	4.31498229839069e-09 \\
1.76813254875405	2.86997663521027e-09 \\
1.79970634426752	1.42497097230091e-09 \\
1.83128013978098	2.00346906220009e-11 \\
1.86285393529445	1.46504035380114e-09 \\
1.89442773080792	3.53668637713246e-09 \\
1.92600152632138	7.07049075605632e-09 \\
1.95757532183485	1.06042951334047e-08 \\
1.98914911734831	1.41380995117831e-08 \\
2.02072291286178	1.76719038891314e-08 \\
2.05229670837524	0.0202935506884928 \\
2.08387050388871	0.0827351834015494 \\
2.11544429940217	0.145176816114606 \\
2.14701809491564	0.207618448827662 \\
2.1785918904291	0.270060081540719 \\
2.21016568594257	0.420979654364836 \\
2.24173948145603	0.73621540168659 \\
2.2733132769695	1.05145114900835 \\
2.30488707248296	1.3666868963301 \\
2.33646086799643	1.68192264365186 \\
2.36803466350989	1.86320116670112 \\
2.39960845902336	1.8212176492884 \\
2.43118225453682	1.77923413187568 \\
2.46275605005029	1.73725061446296 \\
2.49432984556376	1.69526709705023 \\
2.52590364107722	1.61046340215362 \\
2.55747743659069	1.46142944102891 \\
2.58905123210415	1.3123954799042 \\
2.62062502761762	1.16336151877948 \\
2.65219882313108	1.01432755765477 \\
2.68377261864455	0.91207984079973 \\
2.71534641415801	0.873131160311834 \\
2.74692020967148	0.834182479823939 \\
2.77849400518494	0.795233799336042 \\
2.81006780069841	0.756285118848147 \\
2.84164159621187	0.738808609517549 \\
2.87321539172534	0.747575864935737 \\
2.9047891872388	0.756343120353925 \\
2.93636298275227	0.765110375772113 \\
2.96793677826573	0.773877631190301 \\
2.9995105737792	0.834144252111799 \\
3.03108436929266	0.95133122438124 \\
3.06265816480613	1.06851819665068 \\
3.0942319603196	1.18570516892012 \\
3.12580575583306	1.30289214118956 \\
3.15737955134653	1.34977959681395 \\
3.18895334685999	1.32636753579082 \\
3.22052714237346	1.30295547476768 \\
3.25210093788692	1.27954341374455 \\
3.28367473340039	1.25613135272142 \\
3.31524852891385	1.17844537593586 \\
3.34682232442732	1.05165442774548 \\
3.37839611994078	0.924863479555099 \\
3.40996991545425	0.798072531364721 \\
3.44154371096771	0.67128158317434 \\
3.47311750648118	0.589841857414346 \\
3.50469130199464	0.545507677279799 \\
3.53626509750811	0.501173497145251 \\
3.56783889302157	0.456839317010704 \\
3.59941268853504	0.412505136876156 \\
3.6309864840485	0.489817914487674 \\
3.66256027956197	0.657043660869595 \\
3.69413407507544	0.824269407251518 \\
3.7257078705889	0.99149515363344 \\
3.75728166610237	1.15872090001536 \\
3.78885546161583	1.29879822725844 \\
3.8204292571293	1.4207766084086 \\
3.85200305264276	1.54275498955877 \\
3.88357684815623	1.66473337070894 \\
3.91515064366969	1.7867117518591 \\
3.94672443918316	1.69538539242209 \\
3.97829823469662	1.47607618863009 \\
4.00987203021009	1.25676698483808 \\
4.04144582572355	1.03745778104607 \\
4.07301962123702	0.818148577254062 \\
4.10459341675048	0.64452529394967 \\
4.13616721226395	0.495502121677633 \\
4.16774100777741	0.346478949405593 \\
4.19931480329088	0.197455777133555 \\
4.23088859880434	0.0484326048615122 \\
4.26246239431781	6.10763648119196e-08 \\
4.29403618983127	4.2119465952739e-08 \\
4.32560998534474	2.31625670930977e-08 \\
4.35718378085821	4.20566823352414e-09 \\
4.38875757637167	1.47512306260359e-08 \\
4.42033137188514	1.29142899461603e-08 \\
4.4519051673986	2.1657027294284e-09 \\
4.48347896291207	8.58288448600249e-09 \\
4.51505275842553	1.93314717013521e-08 \\
4.546626553939	3.00800589167287e-08 \\
4.57820034945246	2.32486779489839e-08 \\
4.60977414496593	9.74903143182901e-09 \\
4.64134794047939	3.75061508439177e-09 \\
4.67292173599286	1.72502616024807e-08 \\
4.70449553150632	3.07499081159492e-08 \\
4.73606932701979	2.96907812945475e-08 \\
4.76764312253325	2.37787286309866e-08 \\
4.79921691804672	1.78666759673715e-08 \\
4.83079071356018	1.19546233037836e-08 \\
4.86236450907365	6.04257064016852e-09 \\
4.89393830458711	3.59247946641423e-09 \\
4.92551210010058	2.14747380349132e-09 \\
4.95708589561405	7.02468140568405e-10 \\
4.98865969112751	7.42537522340956e-10 \\
5.02023348664098	2.18754318526048e-09 \\
5.05180728215444	5.30358856659608e-09 \\
5.08338107766791	8.83739294443235e-09 \\
5.11495487318137	1.23711973222822e-08 \\
5.14652866869484	1.59050017001726e-08 \\
5.1781024642083	1.94388060780902e-08 \\
5.20967625972177	0.0515143670450218 \\
5.24125005523523	0.113955999758077 \\
5.2728238507487	0.176397632471134 \\
5.30439764626216	0.238839265184191 \\
5.33597144177563	0.301280897897246 \\
5.36754523728909	0.578597528025714 \\
5.39911903280256	0.893833275347466 \\
5.43069282831602	1.20906902266923 \\
5.46226662382949	1.52430476999099 \\
5.49384041934295	1.83954051731274 \\
5.52541421485642	1.84220940799476 \\
5.55698801036989	1.80022589058204 \\
5.58856180588335	1.75824237316932 \\
5.62013560139682	1.7162588557566 \\
5.65170939691028	1.67427533834387 \\
5.68328319242375	1.53594642159126 \\
5.71485698793721	1.38691246046655 \\
5.74643078345068	1.23787849934184 \\
5.77800457896414	1.08884453821713 \\
5.80957837447761	0.939810577092409 \\
5.84115216999107	0.892605500555782 \\
5.87272596550454	0.853656820067886 \\
5.904299761018	0.81470813957999 \\
5.93587355653147	0.775759459092095 \\
5.96744735204493	0.736810778604199 \\
5.9990211475584	0.743192237226643 \\
6.03059494307186	0.751959492644831 \\
6.06216873858533	0.760726748063019 \\
6.09374253409879	0.769494003481207 \\
6.12531632961226	0.778261258899395 \\
6.15689012512573	0.89273773824652 \\
6.18846392063919	1.00992471051596 \\
6.22003771615266	1.1271116827854 \\
6.25161151166612	1.24429865505484 \\
6.28318530717959	1.36148562732428 \\
};

\end{axis}

\end{tikzpicture}

%% file: figures/experiment/experimentRBF.tex
\begin{tikzpicture}

\definecolor{color0}{rgb}{0.12156862745098,0.466666666666667,0.705882352941177}
\definecolor{color1}{rgb}{1,0.498039215686275,0.0549019607843137}
\definecolor{color2}{rgb}{0.172549019607843,0.627450980392157,0.172549019607843}
\definecolor{color3}{rgb}{0.83921568627451,0.152941176470588,0.156862745098039}

\begin{axis}[
legend cell align={left},
tick align=outside,
tick pos=left,
x grid style={white!69.01960784313725!black},
xlabel={$x$},
xmin=-0.314159265358979, xmax=6.59734457253857,
y grid style={white!69.01960784313725!black},
ylabel={$y$},
ymin=-0.1, ymax=1.6,
ytick={-0.2,0,0.2,0.4,0.6,0.8,1,1.2,1.4,1.6},
yticklabels={−0.2,0.0,0.2,0.4,0.6,0.8,1.0,1.2,1.4,1.6}
]

\addplot [very thick, color0]
table [row sep=\\]{%
0	1 \\
0.0315737955134653	1 \\
0.0631475910269305	1 \\
0.0947213865403958	1 \\
0.126295182053861	1 \\
0.157868977567326	1 \\
0.189442773080792	1 \\
0.221016568594257	1 \\
0.252590364107722	1 \\
0.284164159621187	1 \\
0.315737955134653	1 \\
0.347311750648118	1 \\
0.378885546161583	1 \\
0.410459341675048	1 \\
0.442033137188514	1 \\
0.473606932701979	1 \\
0.505180728215444	1 \\
0.536754523728909	1 \\
0.568328319242375	1 \\
0.59990211475584	1 \\
0.631475910269305	1 \\
0.66304970578277	1 \\
0.694623501296236	1 \\
0.726197296809701	1 \\
0.757771092323166	1 \\
0.789344887836631	1 \\
0.820918683350097	1 \\
0.852492478863562	1 \\
0.884066274377027	1 \\
0.915640069890492	1 \\
0.947213865403958	1 \\
0.978787660917423	1 \\
1.01036145643089	1 \\
1.04193525194435	1 \\
1.07350904745782	0 \\
1.10508284297128	0 \\
1.13665663848475	0 \\
1.16823043399821	0 \\
1.19980422951168	0 \\
1.23137802502514	0 \\
1.26295182053861	0 \\
1.29452561605208	0 \\
1.32609941156554	0 \\
1.35767320707901	0 \\
1.38924700259247	0 \\
1.42082079810594	0 \\
1.4523945936194	0 \\
1.48396838913287	0 \\
1.51554218464633	0 \\
1.5471159801598	0 \\
1.57868977567326	0 \\
1.61026357118673	0 \\
1.64183736670019	0 \\
1.67341116221366	0 \\
1.70498495772712	0 \\
1.73655875324059	0 \\
1.76813254875405	0 \\
1.79970634426752	0 \\
1.83128013978098	0 \\
1.86285393529445	0 \\
1.89442773080792	0 \\
1.92600152632138	0 \\
1.95757532183485	0 \\
1.98914911734831	0 \\
2.02072291286178	0 \\
2.05229670837524	0 \\
2.08387050388871	0 \\
2.11544429940217	1 \\
2.14701809491564	1 \\
2.1785918904291	1 \\
2.21016568594257	1 \\
2.24173948145603	1 \\
2.2733132769695	1 \\
2.30488707248296	1 \\
2.33646086799643	1 \\
2.36803466350989	1 \\
2.39960845902336	1 \\
2.43118225453682	1 \\
2.46275605005029	1 \\
2.49432984556376	1 \\
2.52590364107722	1 \\
2.55747743659069	1 \\
2.58905123210415	1 \\
2.62062502761762	1 \\
2.65219882313108	1 \\
2.68377261864455	1 \\
2.71534641415801	1 \\
2.74692020967148	1 \\
2.77849400518494	1 \\
2.81006780069841	1 \\
2.84164159621187	1 \\
2.87321539172534	1 \\
2.9047891872388	1 \\
2.93636298275227	1 \\
2.96793677826573	1 \\
2.9995105737792	1 \\
3.03108436929266	1 \\
3.06265816480613	1 \\
3.0942319603196	1 \\
3.12580575583306	1 \\
3.15737955134653	1 \\
3.18895334685999	1 \\
3.22052714237346	1 \\
3.25210093788692	1 \\
3.28367473340039	1 \\
3.31524852891385	1 \\
3.34682232442732	1 \\
3.37839611994078	1 \\
3.40996991545425	1 \\
3.44154371096771	1 \\
3.47311750648118	1 \\
3.50469130199464	1 \\
3.53626509750811	1 \\
3.56783889302157	1 \\
3.59941268853504	1 \\
3.6309864840485	1 \\
3.66256027956197	1 \\
3.69413407507544	1 \\
3.7257078705889	1 \\
3.75728166610237	1 \\
3.78885546161583	1 \\
3.8204292571293	1 \\
3.85200305264276	1 \\
3.88357684815623	1 \\
3.91515064366969	1 \\
3.94672443918316	1 \\
3.97829823469662	1 \\
4.00987203021009	1 \\
4.04144582572355	1 \\
4.07301962123702	1 \\
4.10459341675048	1 \\
4.13616721226395	1 \\
4.16774100777741	1 \\
4.19931480329088	0 \\
4.23088859880434	0 \\
4.26246239431781	0 \\
4.29403618983127	0 \\
4.32560998534474	0 \\
4.35718378085821	0 \\
4.38875757637167	0 \\
4.42033137188514	0 \\
4.4519051673986	0 \\
4.48347896291207	0 \\
4.51505275842553	0 \\
4.546626553939	0 \\
4.57820034945246	0 \\
4.60977414496593	0 \\
4.64134794047939	0 \\
4.67292173599286	0 \\
4.70449553150632	0 \\
4.73606932701979	0 \\
4.76764312253325	0 \\
4.79921691804672	0 \\
4.83079071356018	0 \\
4.86236450907365	0 \\
4.89393830458711	0 \\
4.92551210010058	0 \\
4.95708589561405	0 \\
4.98865969112751	0 \\
5.02023348664098	0 \\
5.05180728215444	0 \\
5.08338107766791	0 \\
5.11495487318137	0 \\
5.14652866869484	0 \\
5.1781024642083	0 \\
5.20967625972177	0 \\
5.24125005523523	1 \\
5.2728238507487	1 \\
5.30439764626216	1 \\
5.33597144177563	1 \\
5.36754523728909	1 \\
5.39911903280256	1 \\
5.43069282831602	1 \\
5.46226662382949	1 \\
5.49384041934295	1 \\
5.52541421485642	1 \\
5.55698801036989	1 \\
5.58856180588335	1 \\
5.62013560139682	1 \\
5.65170939691028	1 \\
5.68328319242375	1 \\
5.71485698793721	1 \\
5.74643078345068	1 \\
5.77800457896414	1 \\
5.80957837447761	1 \\
5.84115216999107	1 \\
5.87272596550454	1 \\
5.904299761018	1 \\
5.93587355653147	1 \\
5.96744735204493	1 \\
5.9990211475584	1 \\
6.03059494307186	1 \\
6.06216873858533	1 \\
6.09374253409879	1 \\
6.12531632961226	1 \\
6.15689012512573	1 \\
6.18846392063919	1 \\
6.22003771615266	1 \\
6.25161151166612	1 \\
6.28318530717959	1 \\
};
\addplot [very thick, color1, dashed]
table [row sep=\\]{%
0	0.937277097218375 \\
0.0315737955134653	0.937139722407416 \\
0.0631475910269305	0.935793504209761 \\
0.0947213865403958	0.933274146769427 \\
0.126295182053861	0.929639098440975 \\
0.157868977567326	0.924963149023889 \\
0.189442773080792	0.919333000479768 \\
0.221016568594257	0.912841382869173 \\
0.252590364107722	0.905581281336046 \\
0.284164159621187	0.897640766771853 \\
0.315737955134653	0.889098807270104 \\
0.347311750648118	0.880022305534636 \\
0.378885546161583	0.870464479176391 \\
0.410459341675048	0.86046458766799 \\
0.442033137188514	0.850048914786922 \\
0.473606932701979	0.83923283678471 \\
0.505180728215444	0.828023741388527 \\
0.536754523728909	0.816424510610063 \\
0.568328319242375	0.804437244801724 \\
0.59990211475584	0.792066893414715 \\
0.631475910269305	0.779324476924026 \\
0.66304970578277	0.766229638345498 \\
0.694623501296236	0.752812348678016 \\
0.726197296809701	0.739113697240443 \\
0.757771092323166	0.725185807639385 \\
0.789344887836631	0.711091013511251 \\
0.820918683350097	0.696900489305887 \\
0.852492478863562	0.682692552145778 \\
0.884066274377027	0.668550832576122 \\
0.915640069890492	0.654562464287177 \\
0.947213865403958	0.640816379869819 \\
0.978787660917423	0.627401736494314 \\
1.01036145643089	0.614406444996707 \\
1.04193525194435	0.60191574705215 \\
1.07350904745782	0.590010781949337 \\
1.10508284297128	0.578767106181726 \\
1.13665663848475	0.568253170369367 \\
1.16823043399821	0.558528809749217 \\
1.19980422951168	0.54964385467498 \\
1.23137802502514	0.54163700347459 \\
1.26295182053861	0.534535110776947 \\
1.29452561605208	0.528353024123862 \\
1.32609941156554	0.523094051549016 \\
1.35767320707901	0.518751071513503 \\
1.38924700259247	0.515308218223376 \\
1.42082079810594	0.51274300549381 \\
1.4523945936194	0.511028703658335 \\
1.48396838913287	0.51013676317748 \\
1.51554218464633	0.510039085236778 \\
1.5471159801598	0.510709967838184 \\
1.57868977567326	0.51212759706093 \\
1.61026357118673	0.514274998956037 \\
1.64183736670019	0.517140411987444 \\
1.67341116221366	0.520717080201882 \\
1.70498495772712	0.52500250311771 \\
1.73655875324059	0.529997210572475 \\
1.76813254875405	0.535703159971145 \\
1.79970634426752	0.54212187846851 \\
1.83128013978098	0.549252490623931 \\
1.86285393529445	0.557089778803688 \\
1.89442773080792	0.565622415293844 \\
1.92600152632138	0.574831480292559 \\
1.95757532183485	0.584689341077913 \\
1.98914911734831	0.595158921192498 \\
2.02072291286178	0.60619334375665 \\
2.05229670837524	0.617735899658308 \\
2.08387050388871	0.629720276572545 \\
2.11544429940217	0.642070991259074 \\
2.14701809491564	0.654703992933689 \\
2.1785918904291	0.667527442822111 \\
2.21016568594257	0.680442714557567 \\
2.24173948145603	0.693345691202367 \\
2.2733132769695	0.706128447479371 \\
2.30488707248296	0.718681392732323 \\
2.33646086799643	0.730895907403876 \\
2.36803466350989	0.742667434856332 \\
2.39960845902336	0.75389889902263 \\
2.43118225453682	0.764504221272975 \\
2.46275605005029	0.774411626964634 \\
2.49432984556376	0.783566385027642 \\
2.52590364107722	0.791932630337799 \\
2.55747743659069	0.799493986369978 \\
2.58905123210415	0.806252828626301 \\
2.62062502761762	0.812228187623448 \\
2.65219882313108	0.817452454696422 \\
2.68377261864455	0.821967194356491 \\
2.71534641415801	0.825818460516405 \\
2.74692020967148	0.829052050333451 \\
2.77849400518494	0.831709111874172 \\
2.81006780069841	0.833822463156647 \\
2.84164159621187	0.835413897014448 \\
2.87321539172534	0.836492653314956 \\
2.9047891872388	0.837055146521157 \\
2.93636298275227	0.837085945523032 \\
2.96793677826573	0.836559913191216 \\
2.9995105737792	0.835445323885025 \\
3.03108436929266	0.833707690046526 \\
3.06265816480613	0.831313951427613 \\
3.0942319603196	0.828236625247749 \\
3.12580575583306	0.824457497951326 \\
3.15737955134653	0.937360806418529 \\
3.18895334685999	0.93661603907795 \\
3.22052714237346	0.934677341756587 \\
3.25210093788692	0.931591756071768 \\
3.28367473340039	0.927426028524954 \\
3.31524852891385	0.922261625792672 \\
3.34682232442732	0.916189005130337 \\
3.37839611994078	0.90930171836955 \\
3.40996991545425	0.901690885710163 \\
3.44154371096771	0.893440477778366 \\
3.47311750648118	0.884623717692009 \\
3.50469130199464	0.875300782793401 \\
3.53626509750811	0.865517864157471 \\
3.56783889302157	0.855307537943327 \\
3.59941268853504	0.844690316283356 \\
3.6309864840485	0.833677173814661 \\
3.66256027956197	0.822272787142475 \\
3.69413407507544	0.810479179987898 \\
3.7257078705889	0.79829944202491 \\
3.75728166610237	0.785741192115378 \\
3.78885546161583	0.772819493014065 \\
3.8204292571293	0.759558995443788 \\
3.85200305264276	0.745995187594121 \\
3.88357684815623	0.732174736743557 \\
3.91515064366969	0.718155013629405 \\
3.94672443918316	0.704002969028824 \\
3.97829823469662	0.689793573379162 \\
4.00987203021009	0.675608030800449 \\
4.04144582572355	0.661531944333768 \\
4.07301962123702	0.647653551910403 \\
4.10459341675048	0.634062087658511 \\
4.13616721226395	0.620846264978011 \\
4.16774100777741	0.608092837398196 \\
4.19931480329088	0.595885177047935 \\
4.23088859880434	0.584301820224281 \\
4.26246239431781	0.573414961966168 \\
4.29403618983127	0.563288929405711 \\
4.32560998534474	0.553978716234822 \\
4.35718378085821	0.545528705228052 \\
4.38875757637167	0.537971730212459 \\
4.42033137188514	0.531328624436786 \\
4.4519051673986	0.525608366438395 \\
4.48347896291207	0.520808872295207 \\
4.51505275842553	0.516918406362309 \\
4.546626553939	0.513917506617823 \\
4.57820034945246	0.511781260220883 \\
4.60977414496593	0.510481729745029 \\
4.64134794047939	0.509990323892074 \\
4.67292173599286	0.510279924898643 \\
4.70449553150632	0.511326620664467 \\
4.73606932701979	0.513110934039279 \\
4.76764312253325	0.51561848737374 \\
4.79921691804672	0.51884008296838 \\
4.83079071356018	0.522771218016671 \\
4.86236450907365	0.527411086543831 \\
4.89393830458711	0.532761151583788 \\
4.92551210010058	0.538823398240134 \\
4.95708589561405	0.545598400301231 \\
4.98865969112751	0.5530833459933 \\
5.02023348664098	0.56127016803817 \\
5.05180728215444	0.57014390659198 \\
5.08338107766791	0.579681401239473 \\
5.11495487318137	0.58985036451945 \\
5.14652866869484	0.600608842613882 \\
5.1781024642083	0.611905028672903 \\
5.20967625972177	0.623677369514614 \\
5.24125005523523	0.635854902272871 \\
5.2728238507487	0.648357774100939 \\
5.30439764626216	0.661097930451041 \\
5.33597144177563	0.673979997201243 \\
5.36754523728909	0.686902418559495 \\
5.39911903280256	0.699758935760812 \\
5.43069282831602	0.712440492186505 \\
5.46226662382949	0.724837622816795 \\
5.49384041934295	0.736843328522565 \\
5.52541421485642	0.748356353138216 \\
5.55698801036989	0.759284684819417 \\
5.58856180588335	0.76954901046861 \\
5.62013560139682	0.779085784391235 \\
5.65170939691028	0.78784955026179 \\
5.68328319242375	0.795814192384881 \\
5.71485698793721	0.802972889294107 \\
5.74643078345068	0.809336686606795 \\
5.77800457896414	0.81493177134826 \\
5.80957837447761	0.819795685928727 \\
5.84115216999107	0.823972839130926 \\
5.87272596550454	0.827509736873803 \\
5.904299761018	0.830450363835306 \\
5.93587355653147	0.832832106925663 \\
5.96744735204493	0.834682538651107 \\
5.9990211475584	0.836017288858444 \\
6.03059494307186	0.836839139377354 \\
6.06216873858533	0.837138383707473 \\
6.09374253409879	0.836894403859984 \\
6.12531632961226	0.836078327159812 \\
6.15689012512573	0.83465653710525 \\
6.18846392063919	0.832594728739785 \\
6.22003771615266	0.829862130715874 \\
6.25161151166612	0.826435477912858 \\
6.28318530717959	0.822302324763963 \\
};
\addplot [very thick, color2, dashed]
table [row sep=\\]{%
0	0.688222338248625 \\
0.0315737955134653	0.702647000324963 \\
0.0631475910269305	0.722050557557158 \\
0.0947213865403958	0.746064830978114 \\
0.126295182053861	0.774114883567164 \\
0.157868977567326	0.805488944577544 \\
0.189442773080792	0.83940413611213 \\
0.221016568594257	0.87506169761439 \\
0.252590364107722	0.911695688388686 \\
0.284164159621187	0.948618507913427 \\
0.315737955134653	0.985255423007463 \\
0.347311750648118	1.02114793764481 \\
0.378885546161583	1.05590487971711 \\
0.410459341675048	1.08909640001834 \\
0.442033137188514	1.12011264132675 \\
0.473606932701979	1.14803152678618 \\
0.505180728215444	1.17154952721849 \\
0.536754523728909	1.189023589205 \\
0.568328319242375	1.19865020019809 \\
0.59990211475584	1.19876510699177 \\
0.631475910269305	1.18818984551401 \\
0.66304970578277	1.1665065715482 \\
0.694623501296236	1.1341499577973 \\
0.726197296809701	1.09227676141395 \\
0.757771092323166	1.04247084267208 \\
0.789344887836631	0.986401715124604 \\
0.820918683350097	0.925548930980423 \\
0.852492478863562	0.861054195036009 \\
0.884066274377027	0.793707807454988 \\
0.915640069890492	0.724039008814424 \\
0.947213865403958	0.652462950183836 \\
0.978787660917423	0.579435456702469 \\
1.01036145643089	0.505578006097504 \\
1.04193525194435	0.431755633784789 \\
1.07350904745782	0.359109462963552 \\
1.10508284297128	0.289051301153328 \\
1.13665663848475	0.223217823289093 \\
1.16823043399821	0.163367449197919 \\
1.19980422951168	0.111201925569269 \\
1.23137802502514	0.068119008779083 \\
1.26295182053861	0.0349468732035757 \\
1.29452561605208	0.0117469770462767 \\
1.32609941156554	0.00223443229605574 \\
1.35767320707901	0.00844312704701802 \\
1.38924700259247	0.0087717237553024 \\
1.42082079810594	0.00528113955025127 \\
1.4523945936194	3.64160772011735e-05 \\
1.48396838913287	0.00542215788584495 \\
1.51554218464633	0.0094520133337293 \\
1.5471159801598	0.0110990879524542 \\
1.57868977567326	0.00977582269513934 \\
1.61026357118673	0.00536043964033533 \\
1.64183736670019	0.00179574590630534 \\
1.67341116221366	0.0108816053904994 \\
1.70498495772712	0.0206568764977488 \\
1.73655875324059	0.0294999959909138 \\
1.76813254875405	0.0354853342149383 \\
1.79970634426752	0.0365141491979869 \\
1.83128013978098	0.0305195626837187 \\
1.86285393529445	0.015735949375864 \\
1.89442773080792	0.00902429392673476 \\
1.92600152632138	0.0441780803541597 \\
1.95757532183485	0.0893416327020972 \\
1.98914911734831	0.143477657545782 \\
2.02072291286178	0.205158743503145 \\
2.05229670837524	0.272849136587941 \\
2.08387050388871	0.345123950328759 \\
2.11544429940217	0.420790657844004 \\
2.14701809491564	0.498918474336878 \\
2.1785918904291	0.5788005020585 \\
2.21016568594257	0.659874002023049 \\
2.24173948145603	0.741616474346428 \\
2.2733132769695	0.823428580067547 \\
2.30488707248296	0.904514005408629 \\
2.33646086799643	0.983771908954566 \\
2.36803466350989	1.05972777462519 \\
2.39960845902336	1.13053882253711 \\
2.43118225453682	1.1941117985072 \\
2.46275605005029	1.24835144987178 \\
2.49432984556376	1.29150975484825 \\
2.52590364107722	1.32254234223402 \\
2.55747743659069	1.34133992592453 \\
2.58905123210415	1.34873114477101 \\
2.62062502761762	1.34624740839249 \\
2.65219882313108	1.33574297093041 \\
2.68377261864455	1.31901014024873 \\
2.71534641415801	1.29750499568977 \\
2.74692020967148	1.27223692684788 \\
2.77849400518494	1.24381533157295 \\
2.81006780069841	1.21260565420404 \\
2.84164159621187	1.17892483990044 \\
2.87321539172534	1.14320440912453 \\
2.9047891872388	1.10607169933302 \\
2.93636298275227	1.06834026508398 \\
2.96793677826573	1.03093800838554 \\
2.9995105737792	0.9948158811035 \\
3.03108436929266	0.960868783028771 \\
3.06265816480613	0.929878554362259 \\
3.0942319603196	0.902473525563336 \\
3.12580575583306	0.879097002961998 \\
3.15737955134653	0.694799087694604 \\
3.18895334685999	0.711743805163978 \\
3.22052714237346	0.733513247124589 \\
3.25210093788692	0.759627484828117 \\
3.28367473340039	0.789434371301729 \\
3.31524852891385	0.822179036131001 \\
3.34682232442732	0.857064313512959 \\
3.37839611994078	0.893301948059064 \\
3.40996991545425	0.93015981219634 \\
3.44154371096771	0.967003803771758 \\
3.47311750648118	1.00331976967221 \\
3.50469130199464	1.03869274034223 \\
3.53626509750811	1.07272849123081 \\
3.56783889302157	1.10492549220398 \\
3.59941268853504	1.13453163522765 \\
3.6309864840485	1.16043678304546 \\
3.66256027956197	1.18115404866299 \\
3.69413407507544	1.19492919871251 \\
3.7257078705889	1.19998554397283 \\
3.75728166610237	1.19485986159017 \\
3.78885546161583	1.17872903777429 \\
3.8204292571293	1.15160414626241 \\
3.85200305264276	1.11430986280041 \\
3.88357684815623	1.0682592876217 \\
3.91515064366969	1.01511983125129 \\
3.94672443918316	0.956493237271539 \\
3.97829823469662	0.89369983514583 \\
4.00987203021009	0.827699832078002 \\
4.04144582572355	0.759136974387933 \\
4.07301962123702	0.688463527730841 \\
4.10459341675048	0.616096827081324 \\
4.13616721226395	0.542562700901353 \\
4.16774100777741	0.468597696619099 \\
4.19931480329088	0.395203318428317 \\
4.23088859880434	0.323658985389984 \\
4.26246239431781	0.255497730914885 \\
4.29403618983127	0.192434427415814 \\
4.32560998534474	0.136226384469665 \\
4.35718378085821	0.0884571531269839 \\
4.38875757637167	0.0502708496488128 \\
4.42033137188514	0.0221290550458644 \\
4.4519051673986	0.00368058478963232 \\
4.48347896291207	0.00619922938190779 \\
4.51505275842553	0.00921446730131736 \\
4.546626553939	0.00737541288304821 \\
4.57820034945246	0.00273389907051554 \\
4.60977414496593	0.00281847479201083 \\
4.64134794047939	0.0076806289030745 \\
4.67292173599286	0.0106208099474762 \\
4.70449553150632	0.0108274646976565 \\
4.73606932701979	0.00794371683163939 \\
4.76764312253325	0.00208474401798435 \\
4.79921691804672	0.00616538861045997 \\
4.83079071356018	0.0157761856883018 \\
4.86236450907365	0.0253093859254629 \\
4.89393830458711	0.0329790196896309 \\
4.92551210010058	0.0367521621490351 \\
4.95708589561405	0.0345156060016456 \\
4.98865969112751	0.0243170632834694 \\
5.02023348664098	0.00464829612745257 \\
5.05180728215444	0.0253081528828226 \\
5.08338107766791	0.065560899605801 \\
5.11495487318137	0.115371608489056 \\
5.14652866869484	0.173471526447245 \\
5.1781024642083	0.238346350884265 \\
5.20967625972177	0.308494211350711 \\
5.24125005523523	0.382597442219307 \\
5.2728238507487	0.459595573554654 \\
5.30439764626216	0.538677630464431 \\
5.33597144177563	0.61922058758354 \\
5.36754523728909	0.700695852402189 \\
5.39911903280256	0.782557603624429 \\
5.43069282831602	0.864122716057986 \\
5.46226662382949	0.944454397337212 \\
5.49384041934295	1.0222698900716 \\
5.52541421485642	1.09590363895528 \\
5.55698801036989	1.16336473798392 \\
5.58856180588335	1.22252013107347 \\
5.62013560139682	1.27140118483388 \\
5.65170939691028	1.30857180584648 \\
5.68328319242375	1.33343897229085 \\
5.71485698793721	1.34637800978567 \\
5.74643078345068	1.34861049688051 \\
5.77800457896414	1.34188034976326 \\
5.80957837447761	1.32805408638725 \\
5.84115216999107	1.30878041445283 \\
5.87272596550454	1.2852953160762 \\
5.904299761018	1.25839404574487 \\
5.93587355653147	1.22854028437408 \\
5.96744735204493	1.19605127606443 \\
5.9990211475584	1.16128524975456 \\
6.03059494307186	1.1247674630368 \\
6.06216873858533	1.08722443431803 \\
6.09374253409879	1.04953806155589 \\
6.12531632961226	1.01265891161896 \\
6.15689012512573	0.977518407163329 \\
6.18846392063919	0.944960789411073 \\
6.22003771615266	0.915695524703745 \\
6.25161151166612	0.890261930366756 \\
6.28318530717959	0.869001496088679 \\
};
\addplot [very thick, color3, dashed]
table [row sep=\\]{%
0	1.42781768680763 \\
0.0315737955134653	1.42949182672494 \\
0.0631475910269305	1.41054656964755 \\
0.0947213865403958	1.37369082816136 \\
0.126295182053861	1.31578994782572 \\
0.157868977567326	1.22873284841795 \\
0.189442773080792	1.10791854735806 \\
0.221016568594257	0.957815251982719 \\
0.252590364107722	0.787545783090774 \\
0.284164159621187	0.603191555619767 \\
0.315737955134653	0.405869241591217 \\
0.347311750648118	0.195737779773873 \\
0.378885546161583	0.0238844806453591 \\
0.410459341675048	0.246117595066059 \\
0.442033137188514	0.462568576571962 \\
0.473606932701979	0.666158761981789 \\
0.505180728215444	0.853952844571133 \\
0.536754523728909	1.02690231771334 \\
0.568328319242375	1.18495214732126 \\
0.59990211475584	1.32260019112678 \\
0.631475910269305	1.43117505745497 \\
0.66304970578277	1.50641078109961 \\
0.694623501296236	1.55128687721061 \\
0.726197296809701	1.56893039171118 \\
0.757771092323166	1.55482338193797 \\
0.789344887836631	1.49937162168107 \\
0.820918683350097	1.39904009960337 \\
0.852492478863562	1.26244066734469 \\
0.884066274377027	1.104093366092 \\
0.915640069890492	0.935132009674899 \\
0.947213865403958	0.760934755685863 \\
0.978787660917423	0.585045080160369 \\
1.01036145643089	0.413610132626144 \\
1.04193525194435	0.256925547391652 \\
1.07350904745782	0.128040883200902 \\
1.10508284297128	0.0380877293422996 \\
1.13665663848475	0.0103484907059104 \\
1.16823043399821	0.0254446654170414 \\
1.19980422951168	0.0208863580886131 \\
1.23137802502514	0.00908576463208759 \\
1.26295182053861	0.00189885686674771 \\
1.29452561605208	0.00836202076432654 \\
1.32609941156554	0.00982741189207035 \\
1.35767320707901	0.00771707487307354 \\
1.38924700259247	0.00421097049644125 \\
1.42082079810594	0.00126590466165294 \\
1.4523945936194	0.000218353178375374 \\
1.48396838913287	0.0005273815450022 \\
1.51554218464633	0.000383825502168318 \\
1.5471159801598	0.00018237617892744 \\
1.57868977567326	6.9998939411002e-05 \\
1.61026357118673	0.000334723560579515 \\
1.64183736670019	0.000369520127700487 \\
1.67341116221366	5.25767894407267e-05 \\
1.70498495772712	0.000326030952233408 \\
1.73655875324059	0.00031059266349398 \\
1.76813254875405	0.000154148575490634 \\
1.79970634426752	0.000588004942796587 \\
1.83128013978098	0.000498051376196511 \\
1.86285393529445	0.000112871206623705 \\
1.89442773080792	0.000806103486227241 \\
1.92600152632138	0.00102686697559941 \\
1.95757532183485	0.000126980034742125 \\
1.98914911734831	0.00478466904250779 \\
2.02072291286178	0.0167256566014361 \\
2.05229670837524	0.0403281000231778 \\
2.08387050388871	0.0790071500754871 \\
2.11544429940217	0.135786433329352 \\
2.14701809491564	0.216148738230371 \\
2.1785918904291	0.329135917407984 \\
2.21016568594257	0.483135961993753 \\
2.24173948145603	0.678625393948864 \\
2.2733132769695	0.904886272528567 \\
2.30488707248296	1.14306409148342 \\
2.33646086799643	1.37186480378127 \\
2.36803466350989	1.57428571741456 \\
2.39960845902336	1.74289658636874 \\
2.43118225453682	1.87642936372387 \\
2.46275605005029	1.96838729831541 \\
2.49432984556376	2.00132885729931 \\
2.52590364107722	1.95698583410641 \\
2.55747743659069	1.83329386719502 \\
2.58905123210415	1.6470265085234 \\
2.62062502761762	1.41883080815225 \\
2.65219882313108	1.16018392728758 \\
2.68377261864455	0.874653622477767 \\
2.71534641415801	0.567465338376012 \\
2.74692020967148	0.251426543044148 \\
2.77849400518494	0.0544683641103288 \\
2.81006780069841	0.330080343280303 \\
2.84164159621187	0.561243669291761 \\
2.87321539172534	0.745898120551009 \\
2.9047891872388	0.891641265688679 \\
2.93636298275227	1.00550072788776 \\
2.96793677826573	1.08751904923034 \\
2.9995105737792	1.13494252029887 \\
3.03108436929266	1.15072839141619 \\
3.06265816480613	1.14470531719811 \\
3.0942319603196	1.12644233884385 \\
3.12580575583306	1.09977919901332 \\
3.15737955134653	1.43145608550318 \\
3.18895334685999	1.42235119378944 \\
3.22052714237346	1.39437460984781 \\
3.25210093788692	1.34784253932205 \\
3.28367473340039	1.27638673950463 \\
3.31524852891385	1.17247221593482 \\
3.34682232442732	1.03596314339558 \\
3.37839611994078	0.874688370885979 \\
3.40996991545425	0.696973158509422 \\
3.44154371096771	0.506182279172721 \\
3.47311750648118	0.302294999550797 \\
3.50469130199464	0.0867476815664079 \\
3.53626509750811	0.135193756709527 \\
3.56783889302157	0.355576769154462 \\
3.59941268853504	0.566274161399809 \\
3.6309864840485	0.762021536464801 \\
3.66256027956197	0.942189191224516 \\
3.69413407507544	1.10799144466657 \\
3.7257078705889	1.25687603247875 \\
3.75728166610237	1.380976490486 \\
3.78885546161583	1.47289975372717 \\
3.8204292571293	1.53232794363209 \\
3.85200305264276	1.56358211682171 \\
3.88357684815623	1.5664419892368 \\
3.91515064366969	1.53276185274266 \\
3.94672443918316	1.45454002616884 \\
3.97829823469662	1.33437555290278 \\
4.00987203021009	1.18513967641384 \\
4.04144582572355	1.02049724485228 \\
4.07301962123702	0.848486382957984 \\
4.10459341675048	0.672912194421713 \\
4.13616721226395	0.498221710233547 \\
4.16774100777741	0.332636531571927 \\
4.19931480329088	0.188188003658369 \\
4.23088859880434	0.0777698837915272 \\
4.26246239431781	0.00897186852448255 \\
4.29403618983127	0.0212456939902188 \\
4.32560998534474	0.0247596412273753 \\
4.35718378085821	0.0152766059742763 \\
4.38875757637167	0.0031686300822126 \\
4.42033137188514	0.00578516604447578 \\
4.4519051673986	0.00966032906834604 \\
4.48347896291207	0.00908906418502082 \\
4.51505275842553	0.00599959501008969 \\
4.546626553939	0.00258063514602523 \\
4.57820034945246	0.000337100568830032 \\
4.60977414496593	0.000473247374081089 \\
4.64134794047939	0.00047523461165527 \\
4.67292173599286	0.000285124965934466 \\
4.70449553150632	6.57345877273507e-05 \\
4.73606932701979	0.000213303517031159 \\
4.76764312253325	0.000396680155662313 \\
4.79921691804672	0.000246676604386327 \\
4.83079071356018	0.00016039070057329 \\
4.86236450907365	0.000385857820132951 \\
4.89393830458711	0.000112951888124727 \\
4.92551210010058	0.000413614805326576 \\
4.95708589561405	0.000622266855589429 \\
4.98865969112751	0.000236052735760689 \\
5.02023348664098	0.000482850718826075 \\
5.05180728215444	0.00101489963929793 \\
5.08338107766791	0.000715242995885185 \\
5.11495487318137	0.0018158156238107 \\
5.14652866869484	0.00956311544657487 \\
5.1781024642083	0.0268237065600324 \\
5.20967625972177	0.0576140017810619 \\
5.24125005523523	0.104883337147274 \\
5.2728238507487	0.172517257341574 \\
5.30439764626216	0.267938409937385 \\
5.33597144177563	0.400720064033137 \\
5.36754523728909	0.576114402877442 \\
5.39911903280256	0.788959657985345 \\
5.43069282831602	1.02383228973019 \\
5.46226662382949	1.25988491673925 \\
5.49384041934295	1.47708011737019 \\
5.52541421485642	1.66292118819301 \\
5.55698801036989	1.8141966944324 \\
5.58856180588335	1.92847706180677 \\
5.62013560139682	1.99357607659066 \\
5.65170939691028	1.98947163649235 \\
5.68328319242375	1.90432402913659 \\
5.71485698793721	1.74656754604628 \\
5.74643078345068	1.53718144001981 \\
5.77800457896414	1.29300914123325 \\
5.80957837447761	1.02060123783888 \\
5.84115216999107	0.723155645360234 \\
5.87272596550454	0.409460645270243 \\
5.904299761018	0.0959060836904494 \\
5.93587355653147	0.197179936941364 \\
5.96744735204493	0.45165914087009 \\
5.9990211475584	0.659030028179901 \\
6.03059494307186	0.823057128960793 \\
6.06216873858533	0.952389084018724 \\
6.09374253409879	1.05070677257057 \\
6.12531632961226	1.11557449456146 \\
6.15689012512573	1.1462729510113 \\
6.18846392063919	1.14974135207796 \\
6.22003771615266	1.13671651485583 \\
6.25161151166612	1.11413722010064 \\
6.28318530717959	1.08326565571715 \\
};

\end{axis}

\end{tikzpicture}

%% file: figures/experiment/experimentNN2.tex
\begin{tikzpicture}

\definecolor{color0}{rgb}{0.12156862745098,0.466666666666667,0.705882352941177}
\definecolor{color1}{rgb}{1,0.498039215686275,0.0549019607843137}
\definecolor{color2}{rgb}{0.172549019607843,0.627450980392157,0.172549019607843}
\definecolor{color3}{rgb}{0.83921568627451,0.152941176470588,0.156862745098039}

\begin{axis}[
legend cell align={left},
legend entries={{Reference},{NN5},{NN10},{NN20}},
legend style={at={(0.03,0.03)}, anchor=south west, draw=white!80.0!black},
tick align=outside,
tick pos=left,
x grid style={white!69.01960784313725!black},
xlabel={$x$},
xmin=-0.314159265358979, xmax=6.59734457253857,
y grid style={white!69.01960784313725!black},
ylabel={$y$},
ymin=-0.1, ymax=1.6,
ytick={-0.2,0,0.2,0.4,0.6,0.8,1,1.2,1.4,1.6},
yticklabels={−0.2,0.0,0.2,0.4,0.6,0.8,1.0,1.2,1.4,1.6}
]
\addlegendimage{very thick, color0}
\addlegendimage{very thick, dashed, color1}
\addlegendimage{very thick, dashed,color2}
\addlegendimage{very thick, dashed,color3}
\addplot [very thick, color0]
table [row sep=\\]{%
0	1 \\
0.0315737955134653	1 \\
0.0631475910269305	1 \\
0.0947213865403958	1 \\
0.126295182053861	1 \\
0.157868977567326	1 \\
0.189442773080792	1 \\
0.221016568594257	1 \\
0.252590364107722	1 \\
0.284164159621187	1 \\
0.315737955134653	1 \\
0.347311750648118	1 \\
0.378885546161583	1 \\
0.410459341675048	1 \\
0.442033137188514	1 \\
0.473606932701979	1 \\
0.505180728215444	1 \\
0.536754523728909	1 \\
0.568328319242375	1 \\
0.59990211475584	1 \\
0.631475910269305	1 \\
0.66304970578277	1 \\
0.694623501296236	1 \\
0.726197296809701	1 \\
0.757771092323166	1 \\
0.789344887836631	1 \\
0.820918683350097	1 \\
0.852492478863562	1 \\
0.884066274377027	1 \\
0.915640069890492	1 \\
0.947213865403958	1 \\
0.978787660917423	1 \\
1.01036145643089	1 \\
1.04193525194435	1 \\
1.07350904745782	1 \\
1.10508284297128	1 \\
1.13665663848475	1 \\
1.16823043399821	1 \\
1.19980422951168	1 \\
1.23137802502514	1 \\
1.26295182053861	1 \\
1.29452561605208	1 \\
1.32609941156554	1 \\
1.35767320707901	1 \\
1.38924700259247	1 \\
1.42082079810594	1 \\
1.4523945936194	1 \\
1.48396838913287	1 \\
1.51554218464633	1 \\
1.5471159801598	1 \\
1.57868977567326	1 \\
1.61026357118673	1 \\
1.64183736670019	1 \\
1.67341116221366	1 \\
1.70498495772712	1 \\
1.73655875324059	1 \\
1.76813254875405	1 \\
1.79970634426752	1 \\
1.83128013978098	1 \\
1.86285393529445	1 \\
1.89442773080792	1 \\
1.92600152632138	1 \\
1.95757532183485	1 \\
1.98914911734831	1 \\
2.02072291286178	1 \\
2.05229670837524	1 \\
2.08387050388871	1 \\
2.11544429940217	1 \\
2.14701809491564	1 \\
2.1785918904291	1 \\
2.21016568594257	1 \\
2.24173948145603	1 \\
2.2733132769695	1 \\
2.30488707248296	1 \\
2.33646086799643	1 \\
2.36803466350989	1 \\
2.39960845902336	1 \\
2.43118225453682	1 \\
2.46275605005029	1 \\
2.49432984556376	1 \\
2.52590364107722	1 \\
2.55747743659069	1 \\
2.58905123210415	1 \\
2.62062502761762	1 \\
2.65219882313108	1 \\
2.68377261864455	1 \\
2.71534641415801	1 \\
2.74692020967148	1 \\
2.77849400518494	1 \\
2.81006780069841	1 \\
2.84164159621187	1 \\
2.87321539172534	1 \\
2.9047891872388	1 \\
2.93636298275227	1 \\
2.96793677826573	1 \\
2.9995105737792	1 \\
3.03108436929266	1 \\
3.06265816480613	1 \\
3.0942319603196	1 \\
3.12580575583306	1 \\
3.15737955134653	1 \\
3.18895334685999	1 \\
3.22052714237346	1 \\
3.25210093788692	1 \\
3.28367473340039	1 \\
3.31524852891385	1 \\
3.34682232442732	1 \\
3.37839611994078	1 \\
3.40996991545425	1 \\
3.44154371096771	1 \\
3.47311750648118	1 \\
3.50469130199464	1 \\
3.53626509750811	1 \\
3.56783889302157	1 \\
3.59941268853504	1 \\
3.6309864840485	1 \\
3.66256027956197	1 \\
3.69413407507544	1 \\
3.7257078705889	1 \\
3.75728166610237	1 \\
3.78885546161583	1 \\
3.8204292571293	1 \\
3.85200305264276	1 \\
3.88357684815623	1 \\
3.91515064366969	1 \\
3.94672443918316	1 \\
3.97829823469662	1 \\
4.00987203021009	1 \\
4.04144582572355	1 \\
4.07301962123702	1 \\
4.10459341675048	1 \\
4.13616721226395	1 \\
4.16774100777741	1 \\
4.19931480329088	1 \\
4.23088859880434	1 \\
4.26246239431781	1 \\
4.29403618983127	1 \\
4.32560998534474	1 \\
4.35718378085821	1 \\
4.38875757637167	1 \\
4.42033137188514	1 \\
4.4519051673986	1 \\
4.48347896291207	1 \\
4.51505275842553	1 \\
4.546626553939	1 \\
4.57820034945246	1 \\
4.60977414496593	1 \\
4.64134794047939	1 \\
4.67292173599286	1 \\
4.70449553150632	1 \\
4.73606932701979	1 \\
4.76764312253325	1 \\
4.79921691804672	1 \\
4.83079071356018	1 \\
4.86236450907365	1 \\
4.89393830458711	1 \\
4.92551210010058	1 \\
4.95708589561405	1 \\
4.98865969112751	1 \\
5.02023348664098	1 \\
5.05180728215444	1 \\
5.08338107766791	1 \\
5.11495487318137	1 \\
5.14652866869484	1 \\
5.1781024642083	1 \\
5.20967625972177	1 \\
5.24125005523523	1 \\
5.2728238507487	1 \\
5.30439764626216	1 \\
5.33597144177563	1 \\
5.36754523728909	1 \\
5.39911903280256	1 \\
5.43069282831602	1 \\
5.46226662382949	1 \\
5.49384041934295	1 \\
5.52541421485642	1 \\
5.55698801036989	1 \\
5.58856180588335	1 \\
5.62013560139682	1 \\
5.65170939691028	1 \\
5.68328319242375	1 \\
5.71485698793721	1 \\
5.74643078345068	1 \\
5.77800457896414	1 \\
5.80957837447761	1 \\
5.84115216999107	1 \\
5.87272596550454	1 \\
5.904299761018	1 \\
5.93587355653147	1 \\
5.96744735204493	1 \\
5.9990211475584	1 \\
6.03059494307186	1 \\
6.06216873858533	1 \\
6.09374253409879	1 \\
6.12531632961226	1 \\
6.15689012512573	1 \\
6.18846392063919	1 \\
6.22003771615266	1 \\
6.25161151166612	1 \\
6.28318530717959	1 \\
};
\addplot [very thick, color1, dashed]
table [row sep=\\]{%
0	0.887552927662295 \\
0.0315737955134653	0.875434990480954 \\
0.0631475910269305	0.89041586881357 \\
0.0947213865403958	0.92505872744772 \\
0.126295182053861	0.967303308454441 \\
0.157868977567326	1.01756461902744 \\
0.189442773080792	1.04239285589884 \\
0.221016568594257	1.06348330024913 \\
0.252590364107722	1.08308060409267 \\
0.284164159621187	1.09787153843352 \\
0.315737955134653	1.08903763970942 \\
0.347311750648118	1.04237865135811 \\
0.378885546161583	0.997632106698812 \\
0.410459341675048	0.952319483151136 \\
0.442033137188514	0.906379112102976 \\
0.473606932701979	0.844054071610035 \\
0.505180728215444	0.795400137788588 \\
0.536754523728909	0.764969102184043 \\
0.568328319242375	0.75751693521678 \\
0.59990211475584	0.774571899180224 \\
0.631475910269305	0.835851217904133 \\
0.66304970578277	0.895635851796796 \\
0.694623501296236	0.961998427716393 \\
0.726197296809701	1.03261664533649 \\
0.757771092323166	1.09951545258657 \\
0.789344887836631	1.16429021143625 \\
0.820918683350097	1.17623904813243 \\
0.852492478863562	1.17870643325696 \\
0.884066274377027	1.17415554406611 \\
0.915640069890492	1.16888259239864 \\
0.947213865403958	1.16289283444743 \\
0.978787660917423	1.15619224093347 \\
1.01036145643089	1.14875326609232 \\
1.04193525194435	1.14035605823175 \\
1.07350904745782	1.1269837296462 \\
1.10508284297128	1.11255068248476 \\
1.13665663848475	1.09705816170529 \\
1.16823043399821	1.08021481300045 \\
1.19980422951168	1.06712579309951 \\
1.23137802502514	1.05738939927094 \\
1.26295182053861	1.05113878917524 \\
1.29452561605208	1.04572574777659 \\
1.32609941156554	1.0396128519182 \\
1.35767320707901	1.03307472331933 \\
1.38924700259247	1.02610614776039 \\
1.42082079810594	1.01859813548199 \\
1.4523945936194	1.01691867018584 \\
1.48396838913287	1.01674369958275 \\
1.51554218464633	1.01612540983863 \\
1.5471159801598	1.01512027133804 \\
1.57868977567326	1.01597710499514 \\
1.61026357118673	1.01648083199016 \\
1.64183736670019	1.01567915168964 \\
1.67341116221366	1.01415347894586 \\
1.70498495772712	1.0121606933116 \\
1.73655875324059	1.00970278123891 \\
1.76813254875405	1.00678219282799 \\
1.79970634426752	0.99003618616897 \\
1.83128013978098	0.951651609418176 \\
1.86285393529445	0.912789275460339 \\
1.89442773080792	0.863167643991435 \\
1.92600152632138	0.80218543168112 \\
1.95757532183485	0.732230649835011 \\
1.98914911734831	0.649257054319607 \\
2.02072291286178	0.551578337250595 \\
2.05229670837524	0.437208829887221 \\
2.08387050388871	0.344222305215898 \\
2.11544429940217	0.568845052279114 \\
2.14701809491564	0.881878889957544 \\
2.1785918904291	1.27873673366643 \\
2.21016568594257	1.70909255539504 \\
2.24173948145603	2.08079044910433 \\
2.2733132769695	2.20904775606401 \\
2.30488707248296	2.03671690904199 \\
2.33646086799643	1.81986084499884 \\
2.36803466350989	1.58886044347301 \\
2.39960845902336	1.22492125564605 \\
2.43118225453682	0.869510474538355 \\
2.46275605005029	0.680675263589435 \\
2.49432984556376	0.718639988200009 \\
2.52590364107722	0.777483351324186 \\
2.55747743659069	0.836928427198298 \\
2.58905123210415	0.889925156603346 \\
2.62062502761762	0.941602221734997 \\
2.65219882313108	0.980239428019796 \\
2.68377261864455	1.00158733298613 \\
2.71534641415801	1.01770155660855 \\
2.74692020967148	1.03126823150532 \\
2.77849400518494	1.04328304578661 \\
2.81006780069841	1.05066567828222 \\
2.84164159621187	1.05688838682562 \\
2.87321539172534	1.04209940169597 \\
2.9047891872388	1.03240318669276 \\
2.93636298275227	1.01751840546424 \\
2.96793677826573	1.00154808974202 \\
2.9995105737792	0.984508159084124 \\
3.03108436929266	0.966415599264022 \\
3.06265816480613	0.946187743930452 \\
3.0942319603196	0.923812242820049 \\
3.12580575583306	0.899883069954659 \\
3.15737955134653	0.875584748425143 \\
3.18895334685999	0.883201293345615 \\
3.22052714237346	0.907683948855356 \\
3.25210093788692	0.944473719551111 \\
3.28367473340039	0.989885975059815 \\
3.31524852891385	1.03046483006997 \\
3.34682232442732	1.0532432328634 \\
3.37839611994078	1.07342897108803 \\
3.40996991545425	1.09193900282234 \\
3.44154371096771	1.09765877552785 \\
3.47311750648118	1.06560035536658 \\
3.50469130199464	1.02003887229071 \\
3.53626509750811	0.975057101597436 \\
3.56783889302157	0.929424918051116 \\
3.59941268853504	0.881345476073845 \\
3.6309864840485	0.810642120335305 \\
3.66256027956197	0.777983493431361 \\
3.69413407507544	0.761271947965568 \\
3.7257078705889	0.76019680670773 \\
3.75728166610237	0.805606827361257 \\
3.78885546161583	0.865863385936468 \\
3.8204292571293	0.92516119556245 \\
3.85200305264276	0.998577094288881 \\
3.88357684815623	1.06622933560484 \\
3.91515064366969	1.1324667006835 \\
3.94672443918316	1.17042578881444 \\
3.97829823469662	1.18070965854624 \\
4.00987203021009	1.17652154692376 \\
4.04144582572355	1.17160901434154 \\
4.07301962123702	1.16597695771891 \\
4.10459341675048	1.15963099121238 \\
4.13616721226395	1.15257744061937 \\
4.16774100777741	1.14464093088507 \\
4.19931480329088	1.13397934467775 \\
4.23088859880434	1.11984024881347 \\
4.26246239431781	1.10511684737281 \\
4.29403618983127	1.08870679972761 \\
4.32560998534474	1.07180296995631 \\
4.35718378085821	1.06232087037523 \\
4.38875757637167	1.05361246177622 \\
4.42033137188514	1.04855480265787 \\
4.4519051673986	1.04272284478965 \\
4.48347896291207	1.03639654423766 \\
4.51505275842553	1.02964821703064 \\
4.546626553939	1.02245769765711 \\
4.57820034945246	1.01685369313592 \\
4.60977414496593	1.01688709273161 \\
4.64134794047939	1.01648295837257 \\
4.67292173599286	1.01567114308952 \\
4.70449553150632	1.01556303522072 \\
4.73606932701979	1.01628306266864 \\
4.76764312253325	1.01626632723693 \\
4.79921691804672	1.01497481041375 \\
4.83079071356018	1.01321536197901 \\
4.86236450907365	1.01098973578918 \\
4.89393830458711	1.00830015039684 \\
4.92551210010058	1.00252570224457 \\
4.95708589561405	0.97090602046698 \\
4.98865969112751	0.932277751625452 \\
5.02023348664098	0.892709902211565 \\
5.05180728215444	0.832946038876186 \\
5.08338107766791	0.767443180969042 \\
5.11495487318137	0.689057359657164 \\
5.14652866869484	0.608308688627986 \\
5.1781024642083	0.494560615079346 \\
5.20967625972177	0.379452543714413 \\
5.24125005523523	0.434006176940607 \\
5.2728238507487	0.706400379128076 \\
5.30439764626216	1.06185263138789 \\
5.33597144177563	1.49495770632894 \\
5.36754523728909	1.91816419547819 \\
5.39911903280256	2.19324523312546 \\
5.43069282831602	2.14759052796348 \\
5.46226662382949	1.92628017960068 \\
5.49384041934295	1.71367236391025 \\
5.52541421485642	1.42141587910245 \\
5.55698801036989	1.03481778834413 \\
5.58856180588335	0.703720842840462 \\
5.62013560139682	0.700122195734256 \\
5.65170939691028	0.746851240470274 \\
5.68328319242375	0.807758699858352 \\
5.71485698793721	0.863616463619456 \\
5.74643078345068	0.915847949472125 \\
5.77800457896414	0.962590886837664 \\
5.80957837447761	0.993533514955224 \\
5.84115216999107	1.00956780024226 \\
5.87272596550454	1.02467806751724 \\
5.904299761018	1.03747040616576 \\
5.93587355653147	1.04744010056404 \\
5.96744735204493	1.0535076280542 \\
5.9990211475584	1.04964715292128 \\
6.03059494307186	1.03778759720995 \\
6.06216873858533	1.02509744077229 \\
6.09374253409879	1.00966796962308 \\
6.12531632961226	0.993160789465343 \\
6.15689012512573	0.975592355015254 \\
6.18846392063919	0.956816531328889 \\
6.22003771615266	0.935225050778942 \\
6.25161151166612	0.911970512488608 \\
6.28318530717959	0.887552927662295 \\
};
\addplot [very thick, color2, dashed]
table [row sep=\\]{%
0	0.962385556904124 \\
0.0315737955134653	0.956542347127082 \\
0.0631475910269305	0.951737612120735 \\
0.0947213865403958	0.947366287390707 \\
0.126295182053861	0.942688336258228 \\
0.157868977567326	0.938123924428181 \\
0.189442773080792	0.933263652005197 \\
0.221016568594257	0.930882029066742 \\
0.252590364107722	0.931792021978809 \\
0.284164159621187	0.932173753492172 \\
0.315737955134653	0.931320380415502 \\
0.347311750648118	0.93019162648593 \\
0.378885546161583	0.928938480677561 \\
0.410459341675048	0.927613256705626 \\
0.442033137188514	0.92790891404346 \\
0.473606932701979	0.936347513229995 \\
0.505180728215444	0.945371979920935 \\
0.536754523728909	0.958547668743596 \\
0.568328319242375	0.971855537923432 \\
0.59990211475584	0.984645241564169 \\
0.631475910269305	0.997354168486564 \\
0.66304970578277	1.00945372971699 \\
0.694623501296236	1.02010653585165 \\
0.726197296809701	1.03034090852789 \\
0.757771092323166	1.0424948544813 \\
0.789344887836631	1.05453839090871 \\
0.820918683350097	1.06793756642851 \\
0.852492478863562	1.08094956371801 \\
0.884066274377027	1.09340069170979 \\
0.915640069890492	1.10601383178469 \\
0.947213865403958	1.11428708196196 \\
0.978787660917423	1.12095624801339 \\
1.01036145643089	1.12709453448251 \\
1.04193525194435	1.13246199136534 \\
1.07350904745782	1.13443670234594 \\
1.10508284297128	1.13080183793893 \\
1.13665663848475	1.12521862141365 \\
1.16823043399821	1.11891492764564 \\
1.19980422951168	1.11189704029395 \\
1.23137802502514	1.10413219855186 \\
1.26295182053861	1.09211926070142 \\
1.29452561605208	1.07724327667576 \\
1.32609941156554	1.06157394104096 \\
1.35767320707901	1.04523243597761 \\
1.38924700259247	1.02981940706336 \\
1.42082079810594	1.01493848558236 \\
1.4523945936194	0.999537812780638 \\
1.48396838913287	0.984184605064176 \\
1.51554218464633	0.970766680190547 \\
1.5471159801598	0.957380290735432 \\
1.57868977567326	0.944388651464703 \\
1.61026357118673	0.931278264579453 \\
1.64183736670019	0.923556293543344 \\
1.67341116221366	0.917586366600347 \\
1.70498495772712	0.913368897134414 \\
1.73655875324059	0.912153639950051 \\
1.76813254875405	0.911132421418727 \\
1.79970634426752	0.910011761025761 \\
1.83128013978098	0.908352850324977 \\
1.86285393529445	0.905872150905886 \\
1.89442773080792	0.906617714561976 \\
1.92600152632138	0.907914948399374 \\
1.95757532183485	0.919275236082138 \\
1.98914911734831	0.932721919878664 \\
2.02072291286178	0.945127315203757 \\
2.05229670837524	0.959053306653131 \\
2.08387050388871	0.976257646241102 \\
2.11544429940217	1.03622360290773 \\
2.14701809491564	1.10326116322443 \\
2.1785918904291	1.1729253131763 \\
2.21016568594257	1.23880180638061 \\
2.24173948145603	1.28558558645511 \\
2.2733132769695	1.3072731059888 \\
2.30488707248296	1.28919747193992 \\
2.33646086799643	1.27422988306767 \\
2.36803466350989	1.25784269745595 \\
2.39960845902336	1.23792200807581 \\
2.43118225453682	1.20637394934712 \\
2.46275605005029	1.15171724897726 \\
2.49432984556376	1.09318538912513 \\
2.52590364107722	1.06762744757361 \\
2.55747743659069	1.05001935111201 \\
2.58905123210415	1.04210275243766 \\
2.62062502761762	1.03642886618804 \\
2.65219882313108	1.03061907297756 \\
2.68377261864455	1.02509627958031 \\
2.71534641415801	1.01992488486974 \\
2.74692020967148	1.01359762983376 \\
2.77849400518494	1.00670856970801 \\
2.81006780069841	0.999401660398753 \\
2.84164159621187	0.991605047356401 \\
2.87321539172534	0.985467784156629 \\
2.9047891872388	0.97927310641317 \\
2.93636298275227	0.972897655169221 \\
2.96793677826573	0.966780678071148 \\
2.9995105737792	0.968911477071082 \\
3.03108436929266	0.971880488595891 \\
3.06265816480613	0.97270306580014 \\
3.0942319603196	0.972171205286151 \\
3.12580575583306	0.965974912969014 \\
3.15737955134653	0.959315944925228 \\
3.18895334685999	0.953786516062581 \\
3.22052714237346	0.949594518642116 \\
3.25210093788692	0.945053473688453 \\
3.28367473340039	0.94043279456782 \\
3.31524852891385	0.935734011701854 \\
3.34682232442732	0.930713461003636 \\
3.37839611994078	0.931417608579136 \\
3.40996991545425	0.931964432242487 \\
3.44154371096771	0.931834867180597 \\
3.47311750648118	0.930721840454529 \\
3.50469130199464	0.929590457092848 \\
3.53626509750811	0.928425497701221 \\
3.56783889302157	0.926694471584868 \\
3.59941268853504	0.932187712226745 \\
3.6309864840485	0.940387280343591 \\
3.66256027956197	0.951659303304822 \\
3.69413407507544	0.965280306254604 \\
3.7257078705889	0.978271725064501 \\
3.75728166610237	0.991056270998216 \\
3.78885546161583	1.00348716421652 \\
3.8204292571293	1.01488680549533 \\
3.85200305264276	1.02512836649169 \\
3.88357684815623	1.03646728344356 \\
3.91515064366969	1.04853848977708 \\
3.94672443918316	1.06122527363315 \\
3.97829823469662	1.07451287602513 \\
4.00987203021009	1.08724602534965 \\
4.04144582572355	1.09964383079999 \\
4.07301962123702	1.11048756823829 \\
4.10459341675048	1.11769812757768 \\
4.13616721226395	1.12410339576741 \\
4.16774100777741	1.12990845429418 \\
4.19931480329088	1.13385724142941 \\
4.23088859880434	1.13332151773855 \\
4.26246239431781	1.12810065407465 \\
4.29403618983127	1.12215645821889 \\
4.32560998534474	1.11549483755128 \\
4.35718378085821	1.10812243252015 \\
4.38875757637167	1.09912738668302 \\
4.42033137188514	1.08482097324937 \\
4.4519051673986	1.06949362214749 \\
4.48347896291207	1.05348620710671 \\
4.51505275842553	1.03706922332077 \\
4.546626553939	1.02241745963147 \\
4.57820034945246	1.00729382104976 \\
4.60977414496593	0.991940831194779 \\
4.64134794047939	0.977436718634184 \\
4.67292173599286	0.96394798118534 \\
4.70449553150632	0.950938273588221 \\
4.73606932701979	0.937755704731068 \\
4.76764312253325	0.926384150432027 \\
4.79921691804672	0.92062345529119 \\
4.83079071356018	0.914652613227516 \\
4.86236450907365	0.912811328695574 \\
4.89393830458711	0.911398958000976 \\
4.92551210010058	0.910622686047814 \\
4.95708589561405	0.909299798607889 \\
4.98865969112751	0.906983442691698 \\
5.02023348664098	0.906096391053041 \\
5.05180728215444	0.907248004309699 \\
5.08338107766791	0.912732568095328 \\
5.11495487318137	0.925743154219974 \\
5.14652866869484	0.938916898519405 \\
5.1781024642083	0.952144512574247 \\
5.20967625972177	0.965738493701553 \\
5.24125005523523	1.00429538850818 \\
5.2728238507487	1.06809467261591 \\
5.30439764626216	1.13878393401896 \\
5.33597144177563	1.2066554065741 \\
5.36754523728909	1.2622918837079 \\
5.39911903280256	1.30786527701297 \\
5.43069282831602	1.29947613138492 \\
5.46226662382949	1.28041514947933 \\
5.49384041934295	1.26616650358936 \\
5.52541421485642	1.2492982129833 \\
5.55698801036989	1.22645289992307 \\
5.58856180588335	1.18007539802304 \\
5.62013560139682	1.12324128730703 \\
5.65170939691028	1.07892309293862 \\
5.68328319242375	1.0566497250038 \\
5.71485698793721	1.04554488028458 \\
5.74643078345068	1.03915706741565 \\
5.77800457896414	1.0335826315949 \\
5.80957837447761	1.02765288719615 \\
5.84115216999107	1.02255410711692 \\
5.87272596550454	1.01688330400337 \\
5.904299761018	1.01020576996111 \\
5.93587355653147	1.00310690065008 \\
5.96744735204493	0.995129768937807 \\
5.9990211475584	0.988586561795281 \\
6.03059494307186	0.982249491705218 \\
6.06216873858533	0.976415911033351 \\
6.09374253409879	0.969353677472258 \\
6.12531632961226	0.967249565533751 \\
6.15689012512573	0.970455246338351 \\
6.18846392063919	0.972569543927033 \\
6.22003771615266	0.972627759067563 \\
6.25161151166612	0.969476088872837 \\
6.28318530717959	0.962385556904125 \\
};
\addplot [very thick, color3, dashed]
table [row sep=\\]{%
0	0.954562493417518 \\
0.0315737955134653	0.952083262642697 \\
0.0631475910269305	0.949973914391654 \\
0.0947213865403958	0.948254514953581 \\
0.126295182053861	0.947538470402625 \\
0.157868977567326	0.94746790757756 \\
0.189442773080792	0.947815834410932 \\
0.221016568594257	0.948601995129172 \\
0.252590364107722	0.949189582965238 \\
0.284164159621187	0.950078180372164 \\
0.315737955134653	0.950960579847872 \\
0.347311750648118	0.952012477061557 \\
0.378885546161583	0.953271524729554 \\
0.410459341675048	0.95548054637193 \\
0.442033137188514	0.95788869073525 \\
0.473606932701979	0.960519181967772 \\
0.505180728215444	0.963100577773768 \\
0.536754523728909	0.964897679497405 \\
0.568328319242375	0.967372039515077 \\
0.59990211475584	0.970102014586843 \\
0.631475910269305	0.973044008758443 \\
0.66304970578277	0.976217672906274 \\
0.694623501296236	0.982531264595975 \\
0.726197296809701	0.989211885397699 \\
0.757771092323166	0.996702204592233 \\
0.789344887836631	1.00530024977565 \\
0.820918683350097	1.01500747043809 \\
0.852492478863562	1.0303863944275 \\
0.884066274377027	1.04683208361602 \\
0.915640069890492	1.06614914135708 \\
0.947213865403958	1.08676328501056 \\
0.978787660917423	1.10898033691879 \\
1.01036145643089	1.13258243008852 \\
1.04193525194435	1.15679107037722 \\
1.07350904745782	1.18818708477261 \\
1.10508284297128	1.21817060711495 \\
1.13665663848475	1.21590256962918 \\
1.16823043399821	1.19451640365635 \\
1.19980422951168	1.16980445444118 \\
1.23137802502514	1.1448546817816 \\
1.26295182053861	1.1213066080317 \\
1.29452561605208	1.09838716512599 \\
1.32609941156554	1.07303791630031 \\
1.35767320707901	1.04588645549423 \\
1.38924700259247	1.01847964296637 \\
1.42082079810594	0.989551011933606 \\
1.4523945936194	0.968422972598032 \\
1.48396838913287	0.956822486400271 \\
1.51554218464633	0.946908684192854 \\
1.5471159801598	0.935888487570078 \\
1.57868977567326	0.926365306388363 \\
1.61026357118673	0.921455751527242 \\
1.64183736670019	0.916616488573803 \\
1.67341116221366	0.912486928165493 \\
1.70498495772712	0.906966110245036 \\
1.73655875324059	0.898773345205603 \\
1.76813254875405	0.889618332657526 \\
1.79970634426752	0.87873195572129 \\
1.83128013978098	0.870010692220134 \\
1.86285393529445	0.87282732899782 \\
1.89442773080792	0.88414497886246 \\
1.92600152632138	0.902885046716196 \\
1.95757532183485	0.928332704249352 \\
1.98914911734831	0.956018894809631 \\
2.02072291286178	0.983245931057879 \\
2.05229670837524	1.0171650006969 \\
2.08387050388871	1.05237341874451 \\
2.11544429940217	1.08946731836467 \\
2.14701809491564	1.12610581937144 \\
2.1785918904291	1.16274917377443 \\
2.21016568594257	1.19676896353017 \\
2.24173948145603	1.22906017558044 \\
2.2733132769695	1.26128110308586 \\
2.30488707248296	1.29310546034513 \\
2.33646086799643	1.3102691459619 \\
2.36803466350989	1.27835971415789 \\
2.39960845902336	1.22870448535184 \\
2.43118225453682	1.18387112058859 \\
2.46275605005029	1.14247345493457 \\
2.49432984556376	1.10652070978567 \\
2.52590364107722	1.08462931984206 \\
2.55747743659069	1.06934859299398 \\
2.58905123210415	1.05489591308817 \\
2.62062502761762	1.04290517041468 \\
2.65219882313108	1.03154682654539 \\
2.68377261864455	1.02166787846254 \\
2.71534641415801	1.01295389369725 \\
2.74692020967148	1.0050638675179 \\
2.77849400518494	0.997448560525429 \\
2.81006780069841	0.990958763847412 \\
2.84164159621187	0.985356700199856 \\
2.87321539172534	0.979999084120424 \\
2.9047891872388	0.975638585527954 \\
2.93636298275227	0.971671738226966 \\
2.96793677826573	0.967949454115119 \\
2.9995105737792	0.964823013776147 \\
3.03108436929266	0.961747935062371 \\
3.06265816480613	0.959518609314635 \\
3.0942319603196	0.957734464179979 \\
3.12580575583306	0.955910992845282 \\
3.15737955134653	0.95328647734667 \\
3.18895334685999	0.950953149171927 \\
3.22052714237346	0.9490781467928 \\
3.25210093788692	0.947712977708102 \\
3.28367473340039	0.947432818366002 \\
3.31524852891385	0.947608509851801 \\
3.34682232442732	0.948169948867067 \\
3.37839611994078	0.948846350706789 \\
3.40996991545425	0.949600454028728 \\
3.44154371096771	0.950503901385973 \\
3.47311750648118	0.951463480184169 \\
3.50469130199464	0.952623448611881 \\
3.53626509750811	0.954354930384521 \\
3.56783889302157	0.956656669940454 \\
3.59941268853504	0.959176301710906 \\
3.6309864840485	0.961916996831936 \\
3.66256027956197	0.963964787733476 \\
3.69413407507544	0.966102711996757 \\
3.7257078705889	0.968705345709336 \\
3.75728166610237	0.971532386876915 \\
3.78885546161583	0.974577047846595 \\
3.8204292571293	0.979322808514647 \\
3.85200305264276	0.985825410826994 \\
3.88357684815623	0.992912584297351 \\
3.91515064366969	1.00076775756106 \\
3.94672443918316	1.00991086267106 \\
3.97829823469662	1.02237915189576 \\
4.00987203021009	1.03853974120923 \\
4.04144582572355	1.05621195288918 \\
4.07301962123702	1.07624019180664 \\
4.10459341675048	1.0976263488689 \\
4.13616721226395	1.1207040802227 \\
4.16774100777741	1.14461242618244 \\
4.19931480329088	1.17182510044728 \\
4.23088859880434	1.20476550513747 \\
4.26246239431781	1.21871249218748 \\
4.29403618983127	1.20654755298873 \\
4.32560998534474	1.18220297424502 \\
4.35718378085821	1.15733244913291 \\
4.38875757637167	1.13287362381192 \\
4.42033137188514	1.10998000806009 \\
4.4519051673986	1.08622949555571 \\
4.48347896291207	1.05955254137993 \\
4.51505275842553	1.03220240613141 \\
4.546626553939	1.00459029655425 \\
4.57820034945246	0.976080685464532 \\
4.60977414496593	0.962637514910769 \\
4.64134794047939	0.951597895826508 \\
4.67292173599286	0.942152293668081 \\
4.70449553150632	0.930064406176233 \\
4.73606932701979	0.923826970794674 \\
4.76764312253325	0.918463392463702 \\
4.79921691804672	0.914515859556233 \\
4.83079071356018	0.91028893280196 \\
4.86236450907365	0.902992696238791 \\
4.89393830458711	0.894491841719357 \\
4.92551210010058	0.884043776316513 \\
4.95708589561405	0.874148583731045 \\
4.98865969112751	0.869601610775272 \\
5.02023348664098	0.877700584852925 \\
5.05180728215444	0.89223505204597 \\
5.08338107766791	0.914718099833773 \\
5.11495487318137	0.942456531749399 \\
5.14652866869484	0.968139189014155 \\
5.1781024642083	0.99941479530265 \\
5.20967625972177	1.03521831251906 \\
5.24125005523523	1.07072825313905 \\
5.2728238507487	1.10789584530052 \\
5.30439764626216	1.14448837750216 \\
5.33597144177563	1.18049671949677 \\
5.36754523728909	1.21295147161087 \\
5.39911903280256	1.2452012080207 \\
5.43069282831602	1.27727430477045 \\
5.46226662382949	1.30691873019848 \\
5.49384041934295	1.30128764600091 \\
5.52541421485642	1.25313364152863 \\
5.55698801036989	1.20608424835192 \\
5.58856180588335	1.16265434082192 \\
5.62013560139682	1.12371203536845 \\
5.65170939691028	1.09362051305187 \\
5.68328319242375	1.07689932997363 \\
5.71485698793721	1.06173459411545 \\
5.74643078345068	1.0489530609165 \\
5.77800457896414	1.03695212277361 \\
5.80957837447761	1.02622771768445 \\
5.84115216999107	1.01728503154637 \\
5.87272596550454	1.00910477015556 \\
5.904299761018	1.00116248915436 \\
5.93587355653147	0.994122849824423 \\
5.96744735204493	0.98828543244912 \\
5.9990211475584	0.982567775426878 \\
6.03059494307186	0.977761674841163 \\
6.06216873858533	0.973598679107389 \\
6.09374253409879	0.969723297403449 \\
6.12531632961226	0.966361765526849 \\
6.15689012512573	0.963247603162333 \\
6.18846392063919	0.960468595095355 \\
6.22003771615266	0.958595390470838 \\
6.25161151166612	0.956898208154251 \\
6.28318530717959	0.954562493417517 \\
};

\end{axis}

\end{tikzpicture}

%% file: figures/experiment/experimentPL2.tex
\begin{tikzpicture}

\definecolor{color0}{rgb}{0.12156862745098,0.466666666666667,0.705882352941177}
\definecolor{color1}{rgb}{1,0.498039215686275,0.0549019607843137}
\definecolor{color2}{rgb}{0.172549019607843,0.627450980392157,0.172549019607843}
\definecolor{color3}{rgb}{0.83921568627451,0.152941176470588,0.156862745098039}

\begin{axis}[
legend cell align={left},
legend entries={{Reference},{PL10},{PL20},{PL40}},
legend style={at={(0.03,0.03)}, anchor=south west, draw=white!80.0!black},
tick align=outside,
tick pos=left,
x grid style={white!69.01960784313725!black},
xlabel={$x$},
xmin=-0.314159265358979, xmax=6.59734457253857,
y grid style={white!69.01960784313725!black},
ylabel={$y$},
ymin=-0.1, ymax=1.6,
ytick={-0.2,0,0.2,0.4,0.6,0.8,1,1.2,1.4,1.6},
yticklabels={−0.2,0.0,0.2,0.4,0.6,0.8,1.0,1.2,1.4,1.6}
]
\addlegendimage{very thick, color0}
\addlegendimage{very thick, dashed, color1}
\addlegendimage{very thick, dashed,color2}
\addlegendimage{very thick, dashed,color3}
\addplot [very thick, color0]
table [row sep=\\]{%
0	1 \\
0.0315737955134653	1 \\
0.0631475910269305	1 \\
0.0947213865403958	1 \\
0.126295182053861	1 \\
0.157868977567326	1 \\
0.189442773080792	1 \\
0.221016568594257	1 \\
0.252590364107722	1 \\
0.284164159621187	1 \\
0.315737955134653	1 \\
0.347311750648118	1 \\
0.378885546161583	1 \\
0.410459341675048	1 \\
0.442033137188514	1 \\
0.473606932701979	1 \\
0.505180728215444	1 \\
0.536754523728909	1 \\
0.568328319242375	1 \\
0.59990211475584	1 \\
0.631475910269305	1 \\
0.66304970578277	1 \\
0.694623501296236	1 \\
0.726197296809701	1 \\
0.757771092323166	1 \\
0.789344887836631	1 \\
0.820918683350097	1 \\
0.852492478863562	1 \\
0.884066274377027	1 \\
0.915640069890492	1 \\
0.947213865403958	1 \\
0.978787660917423	1 \\
1.01036145643089	1 \\
1.04193525194435	1 \\
1.07350904745782	1 \\
1.10508284297128	1 \\
1.13665663848475	1 \\
1.16823043399821	1 \\
1.19980422951168	1 \\
1.23137802502514	1 \\
1.26295182053861	1 \\
1.29452561605208	1 \\
1.32609941156554	1 \\
1.35767320707901	1 \\
1.38924700259247	1 \\
1.42082079810594	1 \\
1.4523945936194	1 \\
1.48396838913287	1 \\
1.51554218464633	1 \\
1.5471159801598	1 \\
1.57868977567326	1 \\
1.61026357118673	1 \\
1.64183736670019	1 \\
1.67341116221366	1 \\
1.70498495772712	1 \\
1.73655875324059	1 \\
1.76813254875405	1 \\
1.79970634426752	1 \\
1.83128013978098	1 \\
1.86285393529445	1 \\
1.89442773080792	1 \\
1.92600152632138	1 \\
1.95757532183485	1 \\
1.98914911734831	1 \\
2.02072291286178	1 \\
2.05229670837524	1 \\
2.08387050388871	1 \\
2.11544429940217	1 \\
2.14701809491564	1 \\
2.1785918904291	1 \\
2.21016568594257	1 \\
2.24173948145603	1 \\
2.2733132769695	1 \\
2.30488707248296	1 \\
2.33646086799643	1 \\
2.36803466350989	1 \\
2.39960845902336	1 \\
2.43118225453682	1 \\
2.46275605005029	1 \\
2.49432984556376	1 \\
2.52590364107722	1 \\
2.55747743659069	1 \\
2.58905123210415	1 \\
2.62062502761762	1 \\
2.65219882313108	1 \\
2.68377261864455	1 \\
2.71534641415801	1 \\
2.74692020967148	1 \\
2.77849400518494	1 \\
2.81006780069841	1 \\
2.84164159621187	1 \\
2.87321539172534	1 \\
2.9047891872388	1 \\
2.93636298275227	1 \\
2.96793677826573	1 \\
2.9995105737792	1 \\
3.03108436929266	1 \\
3.06265816480613	1 \\
3.0942319603196	1 \\
3.12580575583306	1 \\
3.15737955134653	1 \\
3.18895334685999	1 \\
3.22052714237346	1 \\
3.25210093788692	1 \\
3.28367473340039	1 \\
3.31524852891385	1 \\
3.34682232442732	1 \\
3.37839611994078	1 \\
3.40996991545425	1 \\
3.44154371096771	1 \\
3.47311750648118	1 \\
3.50469130199464	1 \\
3.53626509750811	1 \\
3.56783889302157	1 \\
3.59941268853504	1 \\
3.6309864840485	1 \\
3.66256027956197	1 \\
3.69413407507544	1 \\
3.7257078705889	1 \\
3.75728166610237	1 \\
3.78885546161583	1 \\
3.8204292571293	1 \\
3.85200305264276	1 \\
3.88357684815623	1 \\
3.91515064366969	1 \\
3.94672443918316	1 \\
3.97829823469662	1 \\
4.00987203021009	1 \\
4.04144582572355	1 \\
4.07301962123702	1 \\
4.10459341675048	1 \\
4.13616721226395	1 \\
4.16774100777741	1 \\
4.19931480329088	1 \\
4.23088859880434	1 \\
4.26246239431781	1 \\
4.29403618983127	1 \\
4.32560998534474	1 \\
4.35718378085821	1 \\
4.38875757637167	1 \\
4.42033137188514	1 \\
4.4519051673986	1 \\
4.48347896291207	1 \\
4.51505275842553	1 \\
4.546626553939	1 \\
4.57820034945246	1 \\
4.60977414496593	1 \\
4.64134794047939	1 \\
4.67292173599286	1 \\
4.70449553150632	1 \\
4.73606932701979	1 \\
4.76764312253325	1 \\
4.79921691804672	1 \\
4.83079071356018	1 \\
4.86236450907365	1 \\
4.89393830458711	1 \\
4.92551210010058	1 \\
4.95708589561405	1 \\
4.98865969112751	1 \\
5.02023348664098	1 \\
5.05180728215444	1 \\
5.08338107766791	1 \\
5.11495487318137	1 \\
5.14652866869484	1 \\
5.1781024642083	1 \\
5.20967625972177	1 \\
5.24125005523523	1 \\
5.2728238507487	1 \\
5.30439764626216	1 \\
5.33597144177563	1 \\
5.36754523728909	1 \\
5.39911903280256	1 \\
5.43069282831602	1 \\
5.46226662382949	1 \\
5.49384041934295	1 \\
5.52541421485642	1 \\
5.55698801036989	1 \\
5.58856180588335	1 \\
5.62013560139682	1 \\
5.65170939691028	1 \\
5.68328319242375	1 \\
5.71485698793721	1 \\
5.74643078345068	1 \\
5.77800457896414	1 \\
5.80957837447761	1 \\
5.84115216999107	1 \\
5.87272596550454	1 \\
5.904299761018	1 \\
5.93587355653147	1 \\
5.96744735204493	1 \\
5.9990211475584	1 \\
6.03059494307186	1 \\
6.06216873858533	1 \\
6.09374253409879	1 \\
6.12531632961226	1 \\
6.15689012512573	1 \\
6.18846392063919	1 \\
6.22003771615266	1 \\
6.25161151166612	1 \\
6.28318530717959	1 \\
};
\addplot [very thick, color1, dashed]
table [row sep=\\]{%
0	0.85139769278522 \\
0.0315737955134653	0.861214510788442 \\
0.0631475910269305	0.871031328791757 \\
0.0947213865403958	0.880848146795073 \\
0.126295182053861	0.890664964798388 \\
0.157868977567326	0.900481782801704 \\
0.189442773080792	0.910298600805019 \\
0.221016568594257	0.920115418808335 \\
0.252590364107722	0.92993223681165 \\
0.284164159621187	0.939749054814966 \\
0.315737955134653	0.949565872818281 \\
0.347311750648118	0.959382690821597 \\
0.378885546161583	0.969199508824912 \\
0.410459341675048	0.979016326828228 \\
0.442033137188514	0.988833144831543 \\
0.473606932701979	0.998649962834859 \\
0.505180728215444	1.00846678083817 \\
0.536754523728909	1.01828359884149 \\
0.568328319242375	1.02810041684481 \\
0.59990211475584	1.03791723484812 \\
0.631475910269305	1.04692562449476 \\
0.66304970578277	1.0486581589309 \\
0.694623501296236	1.05039069336704 \\
0.726197296809701	1.05212322780319 \\
0.757771092323166	1.05385576223933 \\
0.789344887836631	1.05558829667547 \\
0.820918683350097	1.05732083111162 \\
0.852492478863562	1.05905336554776 \\
0.884066274377027	1.0607858999839 \\
0.915640069890492	1.06251843442005 \\
0.947213865403958	1.06425096885619 \\
0.978787660917423	1.06598350329233 \\
1.01036145643089	1.06771603772848 \\
1.04193525194435	1.06944857216462 \\
1.07350904745782	1.07118110660076 \\
1.10508284297128	1.07291364103691 \\
1.13665663848475	1.07464617547305 \\
1.16823043399821	1.07637870990919 \\
1.19980422951168	1.07811124434534 \\
1.23137802502514	1.07984377878148 \\
1.26295182053861	1.07953079848132 \\
1.29452561605208	1.07103575923563 \\
1.32609941156554	1.06254071998993 \\
1.35767320707901	1.05404568074424 \\
1.38924700259247	1.04555064149855 \\
1.42082079810594	1.03705560225286 \\
1.4523945936194	1.02856056300717 \\
1.48396838913287	1.02006552376147 \\
1.51554218464633	1.01157048451578 \\
1.5471159801598	1.00307544527009 \\
1.57868977567326	0.994580406024399 \\
1.61026357118673	0.986085366778707 \\
1.64183736670019	0.977590327533015 \\
1.67341116221366	0.969095288287323 \\
1.70498495772712	0.960600249041631 \\
1.73655875324059	0.952105209795939 \\
1.76813254875405	0.943610170550247 \\
1.79970634426752	0.935115131304555 \\
1.83128013978098	0.926620092058863 \\
1.86285393529445	0.918125052813171 \\
1.89442773080792	0.916854481495036 \\
1.92600152632138	0.932441002008445 \\
1.95757532183485	0.948027522521854 \\
1.98914911734831	0.963614043035263 \\
2.02072291286178	0.979200563548672 \\
2.05229670837524	0.994787084062081 \\
2.08387050388871	1.01037360457549 \\
2.11544429940217	1.0259601250889 \\
2.14701809491564	1.04154664560231 \\
2.1785918904291	1.05713316611572 \\
2.21016568594257	1.07271968662913 \\
2.24173948145603	1.08830620714254 \\
2.2733132769695	1.10389272765594 \\
2.30488707248296	1.11947924816935 \\
2.33646086799643	1.13506576868276 \\
2.36803466350989	1.15065228919617 \\
2.39960845902336	1.16623880970958 \\
2.43118225453682	1.18182533022299 \\
2.46275605005029	1.1974118507364 \\
2.49432984556376	1.21299837124981 \\
2.52590364107722	1.21489395007527 \\
2.55747743659069	1.1962531163681 \\
2.58905123210415	1.17761228266092 \\
2.62062502761762	1.15897144895375 \\
2.65219882313108	1.14033061524657 \\
2.68377261864455	1.12168978153939 \\
2.71534641415801	1.10304894783222 \\
2.74692020967148	1.08440811412504 \\
2.77849400518494	1.06576728041787 \\
2.81006780069841	1.04712644671069 \\
2.84164159621187	1.02848561300351 \\
2.87321539172534	1.00984477929634 \\
2.9047891872388	0.991203945589162 \\
2.93636298275227	0.972563111881986 \\
2.96793677826573	0.95392227817481 \\
2.9995105737792	0.935281444467634 \\
3.03108436929266	0.916640610760457 \\
3.06265816480613	0.897999777053281 \\
3.0942319603196	0.879358943346105 \\
3.12580575583306	0.860718109638929 \\
3.15737955134653	0.856306101786784 \\
3.18895334685999	0.866122919790099 \\
3.22052714237346	0.875939737793415 \\
3.25210093788692	0.88575655579673 \\
3.28367473340039	0.895573373800046 \\
3.31524852891385	0.905390191803361 \\
3.34682232442732	0.915207009806677 \\
3.37839611994078	0.925023827809992 \\
3.40996991545425	0.934840645813308 \\
3.44154371096771	0.944657463816624 \\
3.47311750648118	0.954474281819939 \\
3.50469130199464	0.964291099823254 \\
3.53626509750811	0.97410791782657 \\
3.56783889302157	0.983924735829885 \\
3.59941268853504	0.993741553833201 \\
3.6309864840485	1.00355837183652 \\
3.66256027956197	1.01337518983983 \\
3.69413407507544	1.02319200784315 \\
3.7257078705889	1.03300882584646 \\
3.75728166610237	1.04282564384978 \\
3.78885546161583	1.04779189171283 \\
3.8204292571293	1.04952442614897 \\
3.85200305264276	1.05125696058511 \\
3.88357684815623	1.05298949502126 \\
3.91515064366969	1.0547220294574 \\
3.94672443918316	1.05645456389354 \\
3.97829823469662	1.05818709832969 \\
4.00987203021009	1.05991963276583 \\
4.04144582572355	1.06165216720197 \\
4.07301962123702	1.06338470163812 \\
4.10459341675048	1.06511723607426 \\
4.13616721226395	1.06684977051041 \\
4.16774100777741	1.06858230494655 \\
4.19931480329088	1.07031483938269 \\
4.23088859880434	1.07204737381884 \\
4.26246239431781	1.07377990825498 \\
4.29403618983127	1.07551244269112 \\
4.32560998534474	1.07724497712727 \\
4.35718378085821	1.07897751156341 \\
4.38875757637167	1.08071004599955 \\
4.42033137188514	1.07528327885847 \\
4.4519051673986	1.06678823961278 \\
4.48347896291207	1.05829320036709 \\
4.51505275842553	1.0497981611214 \\
4.546626553939	1.0413031218757 \\
4.57820034945246	1.03280808263001 \\
4.60977414496593	1.02431304338432 \\
4.64134794047939	1.01581800413863 \\
4.67292173599286	1.00732296489294 \\
4.70449553150632	0.998827925647245 \\
4.73606932701979	0.990332886401553 \\
4.76764312253325	0.981837847155861 \\
4.79921691804672	0.973342807910169 \\
4.83079071356018	0.964847768664477 \\
4.86236450907365	0.956352729418785 \\
4.89393830458711	0.947857690173093 \\
4.92551210010058	0.939362650927401 \\
4.95708589561405	0.930867611681709 \\
4.98865969112751	0.922372572436017 \\
5.02023348664098	0.913877533190325 \\
5.05180728215444	0.92464774175174 \\
5.08338107766791	0.940234262265149 \\
5.11495487318137	0.955820782778558 \\
5.14652866869484	0.971407303291967 \\
5.1781024642083	0.986993823805376 \\
5.20967625972177	1.00258034431879 \\
5.24125005523523	1.01816686483219 \\
5.2728238507487	1.0337533853456 \\
5.30439764626216	1.04933990585901 \\
5.33597144177563	1.06492642637242 \\
5.36754523728909	1.08051294688583 \\
5.39911903280256	1.09609946739924 \\
5.43069282831602	1.11168598791265 \\
5.46226662382949	1.12727250842606 \\
5.49384041934295	1.14285902893947 \\
5.52541421485642	1.15844554945288 \\
5.55698801036989	1.17403206996629 \\
5.58856180588335	1.18961859047969 \\
5.62013560139682	1.2052051109931 \\
5.65170939691028	1.22079163150651 \\
5.68328319242375	1.20557353322169 \\
5.71485698793721	1.18693269951451 \\
5.74643078345068	1.16829186580733 \\
5.77800457896414	1.14965103210016 \\
5.80957837447761	1.13101019839298 \\
5.84115216999107	1.11236936468581 \\
5.87272596550454	1.09372853097863 \\
5.904299761018	1.07508769727145 \\
5.93587355653147	1.05644686356428 \\
5.96744735204493	1.0378060298571 \\
5.9990211475584	1.01916519614993 \\
6.03059494307186	1.00052436244275 \\
6.06216873858533	0.981883528735574 \\
6.09374253409879	0.963242695028398 \\
6.12531632961226	0.944601861321221 \\
6.15689012512573	0.925961027614045 \\
6.18846392063919	0.907320193906869 \\
6.22003771615266	0.888679360199693 \\
6.25161151166612	0.870038526492517 \\
6.28318530717959	0.851397692785341 \\
};
\addplot [very thick, color2, dashed]
table [row sep=\\]{%
0	0.586787957088387 \\
0.0315737955134653	0.665959908892616 \\
0.0631475910269305	0.745131860697343 \\
0.0947213865403958	0.824303812502069 \\
0.126295182053861	0.903475764306795 \\
0.157868977567326	0.982647716111522 \\
0.189442773080792	1.06181966791625 \\
0.221016568594257	1.14099161972098 \\
0.252590364107722	1.2201635715257 \\
0.284164159621187	1.29933552333043 \\
0.315737955134653	1.37010146174805 \\
0.347311750648118	1.28115314579742 \\
0.378885546161583	1.19220482984678 \\
0.410459341675048	1.10325651389615 \\
0.442033137188514	1.01430819794552 \\
0.473606932701979	0.925359881994889 \\
0.505180728215444	0.836411566044257 \\
0.536754523728909	0.747463250093625 \\
0.568328319242375	0.658514934142993 \\
0.59990211475584	0.569566618192361 \\
0.631475910269305	0.500519447975003 \\
0.66304970578277	0.610582589366271 \\
0.694623501296236	0.720645730757538 \\
0.726197296809701	0.830708872148806 \\
0.757771092323166	0.940772013540073 \\
0.789344887836631	1.05083515493134 \\
0.820918683350097	1.16089829632261 \\
0.852492478863562	1.27096143771388 \\
0.884066274377027	1.38102457910514 \\
0.915640069890492	1.49108772049641 \\
0.947213865403958	1.57089509358505 \\
0.978787660917423	1.47925311295173 \\
1.01036145643089	1.3876111323184 \\
1.04193525194435	1.29596915168508 \\
1.07350904745782	1.20432717105176 \\
1.10508284297128	1.11268519041844 \\
1.13665663848475	1.02104320978512 \\
1.16823043399821	0.929401229151798 \\
1.19980422951168	0.837759248518477 \\
1.23137802502514	0.746117267885156 \\
1.26295182053861	0.687509494044053 \\
1.29452561605208	0.761038547376691 \\
1.32609941156554	0.834567600709329 \\
1.35767320707901	0.908096654041967 \\
1.38924700259247	0.981625707374605 \\
1.42082079810594	1.05515476070724 \\
1.4523945936194	1.12868381403988 \\
1.48396838913287	1.20221286737252 \\
1.51554218464633	1.27574192070516 \\
1.5471159801598	1.3492709740378 \\
1.57868977567326	1.3787284062368 \\
1.61026357118673	1.27597097503026 \\
1.64183736670019	1.17321354382373 \\
1.67341116221366	1.07045611261719 \\
1.70498495772712	0.967698681410656 \\
1.73655875324059	0.864941250204123 \\
1.76813254875405	0.762183818997587 \\
1.79970634426752	0.659426387791051 \\
1.83128013978098	0.556668956584518 \\
1.86285393529445	0.453911525377983 \\
1.89442773080792	0.420252533060419 \\
1.92600152632138	0.547823231489325 \\
1.95757532183485	0.675393929918232 \\
1.98914911734831	0.802964628347137 \\
2.02072291286178	0.930535326776044 \\
2.05229670837524	1.05810602520495 \\
2.08387050388871	1.18567672363386 \\
2.11544429940217	1.31324742206276 \\
2.14701809491564	1.44081812049167 \\
2.1785918904291	1.56838881892058 \\
2.21016568594257	1.62032926519354 \\
2.24173948145603	1.5318135288863 \\
2.2733132769695	1.44329779257907 \\
2.30488707248296	1.35478205627184 \\
2.33646086799643	1.26626631996461 \\
2.36803466350989	1.17775058365738 \\
2.39960845902336	1.08923484735014 \\
2.43118225453682	1.00071911104291 \\
2.46275605005029	0.912203374735679 \\
2.49432984556376	0.823687638428446 \\
2.52590364107722	0.792323633453456 \\
2.55747743659069	0.846687225479882 \\
2.58905123210415	0.901050817506306 \\
2.62062502761762	0.955414409532732 \\
2.65219882313108	1.00977800155916 \\
2.68377261864455	1.06414159358558 \\
2.71534641415801	1.11850518561201 \\
2.74692020967148	1.17286877763843 \\
2.77849400518494	1.22723236966486 \\
2.81006780069841	1.28159596169128 \\
2.84164159621187	1.27872019950817 \\
2.87321539172534	1.20588522662193 \\
2.9047891872388	1.13305025373569 \\
2.93636298275227	1.06021528084944 \\
2.96793677826573	0.987380307963203 \\
2.9995105737792	0.914545335076959 \\
3.03108436929266	0.841710362190716 \\
3.06265816480613	0.768875389304473 \\
3.0942319603196	0.696040416418229 \\
3.12580575583306	0.623205443531987 \\
3.15737955134653	0.626373932990252 \\
3.18895334685999	0.705545884794978 \\
3.22052714237346	0.784717836599705 \\
3.25210093788692	0.863889788404432 \\
3.28367473340039	0.94306174020916 \\
3.31524852891385	1.02223369201388 \\
3.34682232442732	1.10140564381861 \\
3.37839611994078	1.18057759562334 \\
3.40996991545425	1.25974954742806 \\
3.44154371096771	1.33892149923279 \\
3.47311750648118	1.32562730377273 \\
3.50469130199464	1.2366789878221 \\
3.53626509750811	1.14773067187147 \\
3.56783889302157	1.05878235592084 \\
3.59941268853504	0.969834039970204 \\
3.6309864840485	0.880885724019574 \\
3.66256027956197	0.791937408068942 \\
3.69413407507544	0.702989092118309 \\
3.7257078705889	0.614040776167677 \\
3.75728166610237	0.525092460217044 \\
3.78885546161583	0.555551018670637 \\
3.8204292571293	0.665614160061903 \\
3.85200305264276	0.775677301453171 \\
3.88357684815623	0.885740442844439 \\
3.91515064366969	0.995803584235708 \\
3.94672443918316	1.10586672562697 \\
3.97829823469662	1.21592986701824 \\
4.00987203021009	1.32599300840951 \\
4.04144582572355	1.43605614980078 \\
4.07301962123702	1.54611929119204 \\
4.10459341675048	1.52507410326839 \\
4.13616721226395	1.43343212263507 \\
4.16774100777741	1.34179014200174 \\
4.19931480329088	1.25014816136842 \\
4.23088859880434	1.1585061807351 \\
4.26246239431781	1.06686420010178 \\
4.29403618983127	0.97522221946846 \\
4.32560998534474	0.883580238835138 \\
4.35718378085821	0.791938258201816 \\
4.38875757637167	0.700296277568493 \\
4.42033137188514	0.724274020710371 \\
4.4519051673986	0.797803074043009 \\
4.48347896291207	0.871332127375648 \\
4.51505275842553	0.944861180708287 \\
4.546626553939	1.01839023404093 \\
4.57820034945246	1.09191928737356 \\
4.60977414496593	1.1654483407062 \\
4.64134794047939	1.23897739403884 \\
4.67292173599286	1.31250644737148 \\
4.70449553150632	1.38603550070411 \\
4.73606932701979	1.32734969063353 \\
4.76764312253325	1.22459225942699 \\
4.79921691804672	1.12183482822046 \\
4.83079071356018	1.01907739701392 \\
4.86236450907365	0.91631996580739 \\
4.89393830458711	0.813562534600854 \\
4.92551210010058	0.710805103394321 \\
4.95708589561405	0.608047672187785 \\
4.98865969112751	0.505290240981249 \\
5.02023348664098	0.402532809774716 \\
5.05180728215444	0.484037882274872 \\
5.08338107766791	0.611608580703776 \\
5.11495487318137	0.739179279132684 \\
5.14652866869484	0.866749977561592 \\
5.1781024642083	0.994320675990496 \\
5.20967625972177	1.1218913744194 \\
5.24125005523523	1.24946207284831 \\
5.2728238507487	1.37703277127722 \\
5.30439764626216	1.50460346970612 \\
5.33597144177563	1.63217416813503 \\
5.36754523728909	1.57607139703992 \\
5.39911903280256	1.48755566073269 \\
5.43069282831602	1.39903992442546 \\
5.46226662382949	1.31052418811822 \\
5.49384041934295	1.22200845181099 \\
5.52541421485642	1.13349271550376 \\
5.55698801036989	1.04497697919653 \\
5.58856180588335	0.956461242889295 \\
5.62013560139682	0.867945506582064 \\
5.65170939691028	0.779429770274831 \\
5.68328319242375	0.81950542946667 \\
5.71485698793721	0.873869021493095 \\
5.74643078345068	0.928232613519519 \\
5.77800457896414	0.982596205545944 \\
5.80957837447761	1.03695979757237 \\
5.84115216999107	1.0913233895988 \\
5.87272596550454	1.14568698162522 \\
5.904299761018	1.20005057365164 \\
5.93587355653147	1.25441416567807 \\
5.96744735204493	1.30877775770449 \\
5.9990211475584	1.24230271306505 \\
6.03059494307186	1.16946774017881 \\
6.06216873858533	1.09663276729257 \\
6.09374253409879	1.02379779440632 \\
6.12531632961226	0.95096282152008 \\
6.15689012512573	0.878127848633836 \\
6.18846392063919	0.805292875747593 \\
6.22003771615266	0.732457902861351 \\
6.25161151166612	0.659622929975109 \\
6.28318530717959	0.586787957088865 \\
};
\addplot [very thick, color3, dashed]
table [row sep=\\]{%
0	1.16533953430462 \\
0.0315737955134653	1.12329314443101 \\
0.0631475910269305	1.08124675455628 \\
0.0947213865403958	1.03920036468155 \\
0.126295182053861	0.997153974806824 \\
0.157868977567326	0.958497946431015 \\
0.189442773080792	1.05206601653269 \\
0.221016568594257	1.14563408663437 \\
0.252590364107722	1.23920215673604 \\
0.284164159621187	1.33277022683772 \\
0.315737955134653	1.4144821667835 \\
0.347311750648118	1.27092763374872 \\
0.378885546161583	1.12737310071393 \\
0.410459341675048	0.983818567679143 \\
0.442033137188514	0.840264034644357 \\
0.473606932701979	0.696779767028013 \\
0.505180728215444	0.554162106239182 \\
0.536754523728909	0.411544445450351 \\
0.568328319242375	0.26892678466152 \\
0.59990211475584	0.12630912387269 \\
0.631475910269305	0.0501723868533306 \\
0.66304970578277	0.572363963789802 \\
0.694623501296236	1.09455554072627 \\
0.726197296809701	1.61674711766273 \\
0.757771092323166	2.1389386945992 \\
0.789344887836631	2.54911688813593 \\
0.820918683350097	2.17520139783918 \\
0.852492478863562	1.80128590754244 \\
0.884066274377027	1.42737041724569 \\
0.915640069890492	1.05345492694894 \\
0.947213865403958	0.710243075193318 \\
0.978787660917423	0.541018508511211 \\
1.01036145643089	0.371793941829101 \\
1.04193525194435	0.202569375146995 \\
1.07350904745782	0.0333448084648853 \\
1.10508284297128	0.00221914540933276 \\
1.13665663848475	0.622131170904162 \\
1.16823043399821	1.24204319639899 \\
1.19980422951168	1.86195522189381 \\
1.23137802502514	2.48186724738863 \\
1.26295182053861	2.84083388550151 \\
1.29452561605208	2.15601897404809 \\
1.32609941156554	1.47120406259466 \\
1.35767320707901	0.786389151141239 \\
1.38924700259247	0.101574239687808 \\
1.42082079810594	0.394681756591893 \\
1.4523945936194	0.241457045027783 \\
1.48396838913287	0.0882323334636719 \\
1.51554218464633	0.0649923781004371 \\
1.5471159801598	0.218217089664548 \\
1.57868977567326	0.453570737518502 \\
1.61026357118673	0.935311194250644 \\
1.64183736670019	1.41705165098279 \\
1.67341116221366	1.89879210771492 \\
1.70498495772712	2.38053256444706 \\
1.73655875324059	2.58370392780738 \\
1.76813254875405	2.05246586316206 \\
1.79970634426752	1.52122779851674 \\
1.83128013978098	0.989989733871426 \\
1.86285393529445	0.458751669226103 \\
1.89442773080792	0.068213001922067 \\
1.92600152632138	0.00597292842563477 \\
1.95757532183485	0.0562671450707986 \\
1.98914911734831	0.118507218567231 \\
2.02072291286178	0.180747292063664 \\
2.05229670837524	0.0652213143760335 \\
2.08387050388871	0.419511077321981 \\
2.11544429940217	0.904243469019999 \\
2.14701809491564	1.388975860718 \\
2.1785918904291	1.87370825241602 \\
2.21016568594257	2.16913579874476 \\
2.24173948145603	2.11299720366125 \\
2.2733132769695	2.05685860857774 \\
2.30488707248296	2.00072001349423 \\
2.33646086799643	1.94458141841071 \\
2.36803466350989	1.76556302790605 \\
2.39960845902336	1.38174497835895 \\
2.43118225453682	0.997926928811863 \\
2.46275605005029	0.614108879264768 \\
2.49432984556376	0.230290829717671 \\
2.52590364107722	0.0672224003948781 \\
2.55747743659069	0.235278401420308 \\
2.58905123210415	0.403334402445735 \\
2.62062502761762	0.571390403471165 \\
2.65219882313108	0.739446404496595 \\
2.68377261864455	0.968432567828077 \\
2.71534641415801	1.27985365663543 \\
2.74692020967148	1.59127474544278 \\
2.77849400518494	1.90269583425014 \\
2.81006780069841	2.21411692305749 \\
2.84164159621187	2.14143584847389 \\
2.87321539172534	1.59929657417279 \\
2.9047891872388	1.0571572998717 \\
2.93636298275227	0.515018025570597 \\
2.96793677826573	0.0271212487304894 \\
2.9995105737792	0.170716256495625 \\
3.03108436929266	0.126185030348443 \\
3.06265816480613	0.423086317192506 \\
3.0942319603196	0.719987604036575 \\
3.12580575583306	1.01688889088064 \\
3.15737955134653	1.14431633936837 \\
3.18895334685999	1.10226994949364 \\
3.22052714237346	1.06022355961892 \\
3.25210093788692	1.01817716974419 \\
3.28367473340039	0.97613077986946 \\
3.31524852891385	1.00528198148185 \\
3.34682232442732	1.09885005158353 \\
3.37839611994078	1.1924181216852 \\
3.40996991545425	1.28598619178688 \\
3.44154371096771	1.37955426188856 \\
3.47311750648118	1.34270490026611 \\
3.50469130199464	1.19915036723133 \\
3.53626509750811	1.05559583419654 \\
3.56783889302157	0.91204130116175 \\
3.59941268853504	0.768486768126961 \\
3.6309864840485	0.625470936633599 \\
3.66256027956197	0.482853275844769 \\
3.69413407507544	0.340235615055936 \\
3.7257078705889	0.197617954267105 \\
3.75728166610237	0.0550002934782721 \\
3.78885546161583	0.311268175321564 \\
3.8204292571293	0.833459752258026 \\
3.85200305264276	1.3556513291945 \\
3.88357684815623	1.87784290613097 \\
3.91515064366969	2.40003448306744 \\
3.94672443918316	2.36215914298755 \\
3.97829823469662	1.98824365269081 \\
4.00987203021009	1.61432816239407 \\
4.04144582572355	1.24041267209732 \\
4.07301962123702	0.866497181800565 \\
4.10459341675048	0.625630791852265 \\
4.13616721226395	0.456406225170159 \\
4.16774100777741	0.287181658488049 \\
4.19931480329088	0.117957091805941 \\
4.23088859880434	0.0512674748761723 \\
4.26246239431781	0.312175158156748 \\
4.29403618983127	0.932087183651564 \\
4.32560998534474	1.5519992091464 \\
4.35718378085821	2.17191123464123 \\
4.38875757637167	2.79182326013606 \\
4.42033137188514	2.49842642977481 \\
4.4519051673986	1.81361151832138 \\
4.48347896291207	1.12879660686795 \\
4.51505275842553	0.443981695414517 \\
4.546626553939	0.240833216038916 \\
4.57820034945246	0.318069400809837 \\
4.60977414496593	0.16484468924573 \\
4.64134794047939	0.0116199776816177 \\
4.67292173599286	0.141604733882494 \\
4.70449553150632	0.294829445446603 \\
4.73606932701979	0.694440965884578 \\
4.76764312253325	1.17618142261671 \\
4.79921691804672	1.65792187934885 \\
4.83079071356018	2.13966233608099 \\
4.86236450907365	2.62140279281313 \\
4.89393830458711	2.31808489548472 \\
4.92551210010058	1.78684683083941 \\
4.95708589561405	1.25560876619408 \\
4.98865969112751	0.724370701548758 \\
5.02023348664098	0.193132636903449 \\
5.05180728215444	0.0370929651738509 \\
5.08338107766791	0.0251471083225809 \\
5.11495487318137	0.0873871818190145 \\
5.14652866869484	0.149627255315448 \\
5.1781024642083	0.21186732881188 \\
5.20967625972177	0.177144881472979 \\
5.24125005523523	0.661877273170983 \\
5.2728238507487	1.146609664869 \\
5.30439764626216	1.63134205656702 \\
5.33597144177563	2.11607444826502 \\
5.36754523728909	2.141066501203 \\
5.39911903280256	2.08492790611949 \\
5.43069282831602	2.02878931103598 \\
5.46226662382949	1.97265071595247 \\
5.49384041934295	1.91651212086896 \\
5.52541421485642	1.5736540031325 \\
5.55698801036989	1.1898359535854 \\
5.58856180588335	0.806017904038312 \\
5.62013560139682	0.422199854491225 \\
5.65170939691028	0.0383818049441273 \\
5.68328319242375	0.151250400907595 \\
5.71485698793721	0.319306401933025 \\
5.74643078345068	0.487362402958452 \\
5.77800457896414	0.655418403983878 \\
5.80957837447761	0.823474405009308 \\
5.84115216999107	1.12414311223176 \\
5.87272596550454	1.43556420103912 \\
5.904299761018	1.74698528984647 \\
5.93587355653147	2.05840637865381 \\
5.96744735204493	2.36982746746117 \\
5.9990211475584	1.87036621132333 \\
6.03059494307186	1.32822693702223 \\
6.06216873858533	0.786087662721144 \\
6.09374253409879	0.243948388420057 \\
6.12531632961226	0.298190885881044 \\
6.15689012512573	0.0222656130735879 \\
6.18846392063919	0.274635673770481 \\
6.22003771615266	0.571536960614543 \\
6.25161151166612	0.868438247458604 \\
6.28318530717959	1.16533953430267 \\
};

\end{axis}

\end{tikzpicture}

%% file: figures/experiment/experimentRBF2.tex
\begin{tikzpicture}

\definecolor{color0}{rgb}{0.12156862745098,0.466666666666667,0.705882352941177}
\definecolor{color1}{rgb}{1,0.498039215686275,0.0549019607843137}
\definecolor{color2}{rgb}{0.172549019607843,0.627450980392157,0.172549019607843}
\definecolor{color3}{rgb}{0.83921568627451,0.152941176470588,0.156862745098039}

\begin{axis}[
legend cell align={left},
legend entries={{Reference},{RBF10},{RBF20},{RBF40}},
legend style={at={(0.03,0.03)}, anchor=south west, draw=white!80.0!black},
tick align=outside,
tick pos=left,
x grid style={white!69.01960784313725!black},
xlabel={$x$},
xmin=-0.314159265358979, xmax=6.59734457253857,
y grid style={white!69.01960784313725!black},
ylabel={$y$},
ymin=-0.1, ymax=1.6,
ytick={-0.2,0,0.2,0.4,0.6,0.8,1,1.2,1.4,1.6},
yticklabels={−0.2,0.0,0.2,0.4,0.6,0.8,1.0,1.2,1.4,1.6}
]
\addlegendimage{very thick, color0}
\addlegendimage{very thick, dashed, color1}
\addlegendimage{very thick, dashed,color2}
\addlegendimage{very thick, dashed,color3}
\addplot [very thick, color0]
table [row sep=\\]{%
0	1 \\
0.0315737955134653	1 \\
0.0631475910269305	1 \\
0.0947213865403958	1 \\
0.126295182053861	1 \\
0.157868977567326	1 \\
0.189442773080792	1 \\
0.221016568594257	1 \\
0.252590364107722	1 \\
0.284164159621187	1 \\
0.315737955134653	1 \\
0.347311750648118	1 \\
0.378885546161583	1 \\
0.410459341675048	1 \\
0.442033137188514	1 \\
0.473606932701979	1 \\
0.505180728215444	1 \\
0.536754523728909	1 \\
0.568328319242375	1 \\
0.59990211475584	1 \\
0.631475910269305	1 \\
0.66304970578277	1 \\
0.694623501296236	1 \\
0.726197296809701	1 \\
0.757771092323166	1 \\
0.789344887836631	1 \\
0.820918683350097	1 \\
0.852492478863562	1 \\
0.884066274377027	1 \\
0.915640069890492	1 \\
0.947213865403958	1 \\
0.978787660917423	1 \\
1.01036145643089	1 \\
1.04193525194435	1 \\
1.07350904745782	1 \\
1.10508284297128	1 \\
1.13665663848475	1 \\
1.16823043399821	1 \\
1.19980422951168	1 \\
1.23137802502514	1 \\
1.26295182053861	1 \\
1.29452561605208	1 \\
1.32609941156554	1 \\
1.35767320707901	1 \\
1.38924700259247	1 \\
1.42082079810594	1 \\
1.4523945936194	1 \\
1.48396838913287	1 \\
1.51554218464633	1 \\
1.5471159801598	1 \\
1.57868977567326	1 \\
1.61026357118673	1 \\
1.64183736670019	1 \\
1.67341116221366	1 \\
1.70498495772712	1 \\
1.73655875324059	1 \\
1.76813254875405	1 \\
1.79970634426752	1 \\
1.83128013978098	1 \\
1.86285393529445	1 \\
1.89442773080792	1 \\
1.92600152632138	1 \\
1.95757532183485	1 \\
1.98914911734831	1 \\
2.02072291286178	1 \\
2.05229670837524	1 \\
2.08387050388871	1 \\
2.11544429940217	1 \\
2.14701809491564	1 \\
2.1785918904291	1 \\
2.21016568594257	1 \\
2.24173948145603	1 \\
2.2733132769695	1 \\
2.30488707248296	1 \\
2.33646086799643	1 \\
2.36803466350989	1 \\
2.39960845902336	1 \\
2.43118225453682	1 \\
2.46275605005029	1 \\
2.49432984556376	1 \\
2.52590364107722	1 \\
2.55747743659069	1 \\
2.58905123210415	1 \\
2.62062502761762	1 \\
2.65219882313108	1 \\
2.68377261864455	1 \\
2.71534641415801	1 \\
2.74692020967148	1 \\
2.77849400518494	1 \\
2.81006780069841	1 \\
2.84164159621187	1 \\
2.87321539172534	1 \\
2.9047891872388	1 \\
2.93636298275227	1 \\
2.96793677826573	1 \\
2.9995105737792	1 \\
3.03108436929266	1 \\
3.06265816480613	1 \\
3.0942319603196	1 \\
3.12580575583306	1 \\
3.15737955134653	1 \\
3.18895334685999	1 \\
3.22052714237346	1 \\
3.25210093788692	1 \\
3.28367473340039	1 \\
3.31524852891385	1 \\
3.34682232442732	1 \\
3.37839611994078	1 \\
3.40996991545425	1 \\
3.44154371096771	1 \\
3.47311750648118	1 \\
3.50469130199464	1 \\
3.53626509750811	1 \\
3.56783889302157	1 \\
3.59941268853504	1 \\
3.6309864840485	1 \\
3.66256027956197	1 \\
3.69413407507544	1 \\
3.7257078705889	1 \\
3.75728166610237	1 \\
3.78885546161583	1 \\
3.8204292571293	1 \\
3.85200305264276	1 \\
3.88357684815623	1 \\
3.91515064366969	1 \\
3.94672443918316	1 \\
3.97829823469662	1 \\
4.00987203021009	1 \\
4.04144582572355	1 \\
4.07301962123702	1 \\
4.10459341675048	1 \\
4.13616721226395	1 \\
4.16774100777741	1 \\
4.19931480329088	1 \\
4.23088859880434	1 \\
4.26246239431781	1 \\
4.29403618983127	1 \\
4.32560998534474	1 \\
4.35718378085821	1 \\
4.38875757637167	1 \\
4.42033137188514	1 \\
4.4519051673986	1 \\
4.48347896291207	1 \\
4.51505275842553	1 \\
4.546626553939	1 \\
4.57820034945246	1 \\
4.60977414496593	1 \\
4.64134794047939	1 \\
4.67292173599286	1 \\
4.70449553150632	1 \\
4.73606932701979	1 \\
4.76764312253325	1 \\
4.79921691804672	1 \\
4.83079071356018	1 \\
4.86236450907365	1 \\
4.89393830458711	1 \\
4.92551210010058	1 \\
4.95708589561405	1 \\
4.98865969112751	1 \\
5.02023348664098	1 \\
5.05180728215444	1 \\
5.08338107766791	1 \\
5.11495487318137	1 \\
5.14652866869484	1 \\
5.1781024642083	1 \\
5.20967625972177	1 \\
5.24125005523523	1 \\
5.2728238507487	1 \\
5.30439764626216	1 \\
5.33597144177563	1 \\
5.36754523728909	1 \\
5.39911903280256	1 \\
5.43069282831602	1 \\
5.46226662382949	1 \\
5.49384041934295	1 \\
5.52541421485642	1 \\
5.55698801036989	1 \\
5.58856180588335	1 \\
5.62013560139682	1 \\
5.65170939691028	1 \\
5.68328319242375	1 \\
5.71485698793721	1 \\
5.74643078345068	1 \\
5.77800457896414	1 \\
5.80957837447761	1 \\
5.84115216999107	1 \\
5.87272596550454	1 \\
5.904299761018	1 \\
5.93587355653147	1 \\
5.96744735204493	1 \\
5.9990211475584	1 \\
6.03059494307186	1 \\
6.06216873858533	1 \\
6.09374253409879	1 \\
6.12531632961226	1 \\
6.15689012512573	1 \\
6.18846392063919	1 \\
6.22003771615266	1 \\
6.25161151166612	1 \\
6.28318530717959	1 \\
};
\addplot [very thick, color1, dashed]
table [row sep=\\]{%
0	1.08040104372994 \\
0.0315737955134653	1.06495089378139 \\
0.0631475910269305	1.05055200791558 \\
0.0947213865403958	1.03723323608178 \\
0.126295182053861	1.02501138923972 \\
0.157868977567326	1.01389402019987 \\
0.189442773080792	1.00388224203447 \\
0.221016568594257	0.994973428687908 \\
0.252590364107722	0.987163642454584 \\
0.284164159621187	0.980449647695079 \\
0.315737955134653	0.974830394888988 \\
0.347311750648118	0.970307889191066 \\
0.378885546161583	0.966887389959894 \\
0.410459341675048	0.964576921296853 \\
0.442033137188514	0.963386109097003 \\
0.473606932701979	0.96332439837966 \\
0.505180728215444	0.964398745349475 \\
0.536754523728909	0.966610918858699 \\
0.568328319242375	0.969954580062064 \\
0.59990211475584	0.974412329721732 \\
0.631475910269305	0.979952913015853 \\
0.66304970578277	0.986528748406616 \\
0.694623501296236	0.994073902444997 \\
0.726197296809701	1.00250257507063 \\
0.757771092323166	1.01170810358937 \\
0.789344887836631	1.02156245295672 \\
0.820918683350097	1.03191614712692 \\
0.852492478863562	1.04259861634656 \\
0.884066274377027	1.05341898543691 \\
0.915640069890492	1.06416739756667 \\
0.947213865403958	1.07461704020709 \\
0.978787660917423	1.0845270952764 \\
1.01036145643089	1.0936468542129 \\
1.04193525194435	1.10172120414458 \\
1.07350904745782	1.10849759316828 \\
1.10508284297128	1.11373442105926 \\
1.13665663848475	1.11721059051936 \\
1.16823043399821	1.1187357231568 \\
1.19980422951168	1.11816033731539 \\
1.23137802502514	1.11538515139001 \\
1.26295182053861	1.11036865973402 \\
1.29452561605208	1.10313225111705 \\
1.32609941156554	1.09376239236252 \\
1.35767320707901	1.08240973861482 \\
1.38924700259247	1.06928538827151 \\
1.42082079810594	1.05465480044888 \\
1.4523945936194	1.03883007636298 \\
1.48396838913287	1.02216134434707 \\
1.51554218464633	1.00502788842297 \\
1.5471159801598	0.987829459558921 \\
1.57868977567326	0.970977961451874 \\
1.61026357118673	0.954889466327104 \\
1.64183736670019	0.939976339793124 \\
1.67341116221366	0.92663917067568 \\
1.70498495772712	0.915258226799742 \\
1.73655875324059	0.906184286185181 \\
1.76813254875405	0.899728901145072 \\
1.79970634426752	0.89615439906634 \\
1.83128013978098	0.895664154546767 \\
1.86285393529445	0.898393826309228 \\
1.89442773080792	0.90440429240039 \\
1.92600152632138	0.913676916571037 \\
1.95757532183485	0.926111549776874 \\
1.98914911734831	0.941527360609497 \\
2.02072291286178	0.959666268122359 \\
2.05229670837524	0.980198494996994 \\
2.08387050388871	1.00272962601459 \\
2.11544429940217	1.02680857133274 \\
2.14701809491564	1.05193598374257 \\
2.1785918904291	1.07757292030078 \\
2.21016568594257	1.10314980951895 \\
2.24173948145603	1.12807601958585 \\
2.2733132769695	1.15175046273847 \\
2.30488707248296	1.17357367421266 \\
2.33646086799643	1.19296165213505 \\
2.36803466350989	1.20936144567972 \\
2.39960845902336	1.22226807211476 \\
2.43118225453682	1.23124189890393 \\
2.46275605005029	1.23592523781858 \\
2.49432984556376	1.23605666317566 \\
2.52590364107722	1.23148156598978 \\
2.55747743659069	1.2221577246209 \\
2.58905123210415	1.20815518178484 \\
2.62062502761762	1.18965037623381 \\
2.65219882313108	1.16691515275624 \\
2.68377261864455	1.1403018299637 \\
2.71534641415801	1.11022583922601 \\
2.74692020967148	1.07714751713138 \\
2.77849400518494	1.04155445907336 \\
2.81006780069841	1.00394549318587 \\
2.84164159621187	0.964816905310648 \\
2.87321539172534	0.924651127701959 \\
2.9047891872388	0.883907766329535 \\
2.93636298275227	0.843016624187198 \\
2.96793677826573	0.802372292145209 \\
2.9995105737792	0.76232991174315 \\
3.03108436929266	0.723201835540495 \\
3.06265816480613	0.685255078859998 \\
3.0942319603196	0.648709626176206 \\
3.12580575583306	0.613737783040342 \\
3.15737955134653	1.07254690709128 \\
3.18895334685999	1.05761792586695 \\
3.22052714237346	1.04375630693526 \\
3.25210093788692	1.03098454121347 \\
3.28367473340039	1.01931445710719 \\
3.31524852891385	1.00875003889682 \\
3.34682232442732	0.999290206055422 \\
3.37839611994078	0.990931396085073 \\
3.40996991545425	0.983669801322394 \\
3.44154371096771	0.977503130509927 \\
3.47311750648118	0.972431793841716 \\
3.50469130199464	0.968459441630905 \\
3.53626509750811	0.965592819679932 \\
3.56783889302157	0.963840938817253 \\
3.59941268853504	0.963213592861392 \\
3.6309864840485	0.96371929893218 \\
3.66256027956197	0.965362775020883 \\
3.69413407507544	0.968142107753533 \\
3.7257078705889	0.972045791625217 \\
3.75728166610237	0.977049832221023 \\
3.78885546161583	0.983115094598563 \\
3.8204292571293	0.990185043318465 \\
3.85200305264276	0.998183968151037 \\
3.88357684815623	1.00701573076162 \\
3.91515064366969	1.0165630174414 \\
3.94672443918316	1.02668705510114 \\
3.97829823469662	1.03722775121011 \\
4.00987203021009	1.04800425439249 \\
4.04144582572355	1.05881599397236 \\
4.07301962123702	1.06944432991062 \\
4.10459341675048	1.07965501085737 \\
4.13616721226395	1.08920167741853 \\
4.16774100777741	1.09783064159103 \\
4.19931480329088	1.1052871076907 \\
4.23088859880434	1.11132286936874 \\
4.26246239431781	1.11570532812784 \\
4.29403618983127	1.11822745299205 \\
4.32560998534474	1.11871807591421 \\
4.35718378085821	1.11705174106698 \\
4.38875757637167	1.11315724671582 \\
4.42033137188514	1.10702407055169 \\
4.4519051673986	1.09870606053383 \\
4.48347896291207	1.08832207639413 \\
4.51505275842553	1.07605362409365 \\
4.546626553939	1.06213986257861 \\
4.57820034945246	1.04687060964309 \\
4.60977414496593	1.0305780859782 \\
4.64134794047939	1.01362810274916 \\
4.67292173599286	0.996411241675643 \\
4.70449553150632	0.979334345237357 \\
4.73606932701979	0.962812385939049 \\
4.76764312253325	0.947260571908697 \\
4.79921691804672	0.933086413261695 \\
4.83079071356018	0.920681443992513 \\
4.86236450907365	0.910412372736058 \\
4.89393830458711	0.902611608146298 \\
4.92551210010058	0.897567337892331 \\
4.95708589561405	0.895513586039001 \\
4.98865969112751	0.896620875198706 \\
5.02023348664098	0.900988223853476 \\
5.05180728215444	0.908637179731852 \\
5.08338107766791	0.919508421494986 \\
5.11495487318137	0.933461183672743 \\
5.14652866869484	0.950275434907005 \\
5.1781024642083	0.969656442864401 \\
5.20967625972177	0.991241159515925 \\
5.24125005523523	1.01460580050074 \\
5.2728238507487	1.0392740781431 \\
5.30439764626216	1.06472575007914 \\
5.33597144177563	1.0904054097766 \\
5.36754523728909	1.11573170577459 \\
5.39911903280256	1.14010736984188 \\
5.43069282831602	1.16293050965028 \\
5.46226662382949	1.18360754795074 \\
5.49384041934295	1.20156796175611 \\
5.52541421485642	1.21628061526631 \\
5.55698801036989	1.22727104422902 \\
5.58856180588335	1.23413862007266 \\
5.62013560139682	1.23657219827027 \\
5.65170939691028	1.23436273006425 \\
5.68328319242375	1.22741145030386 \\
5.71485698793721	1.21573265095444 \\
5.74643078345068	1.19945064841332 \\
5.77800457896414	1.17879123740557 \\
5.80957837447761	1.15406855554543 \\
5.84115216999107	1.12566873695918 \\
5.87272596550454	1.09403193603532 \\
5.904299761018	1.05963424325292 \\
5.93587355653147	1.02297074257698 \\
5.96744735204493	0.984540559394615 \\
5.9990211475584	0.944834314377909 \\
6.03059494307186	0.904324013806466 \\
6.06216873858533	0.863455126099235 \\
6.09374253409879	0.822640442895229 \\
6.12531632961226	0.782255299374596 \\
6.15689012512573	0.742633810184355 \\
6.18846392063919	0.704065927659064 \\
6.22003771615266	0.66679530357393 \\
6.25161151166612	0.631018089083962 \\
6.28318530717959	0.596882901188537 \\
};
\addplot [very thick, color2, dashed]
table [row sep=\\]{%
0	0.99483816210402 \\
0.0315737955134653	0.993907544667959 \\
0.0631475910269305	1.01071880304362 \\
0.0947213865403958	1.0421640367811 \\
0.126295182053861	1.08359230093074 \\
0.157868977567326	1.12926878536967 \\
0.189442773080792	1.17285419495903 \\
0.221016568594257	1.20789837959499 \\
0.252590364107722	1.22840246727053 \\
0.284164159621187	1.22948327425841 \\
0.315737955134653	1.20807604299645 \\
0.347311750648118	1.16350640669156 \\
0.378885546161583	1.09774471899452 \\
0.410459341675048	1.0152625521899 \\
0.442033137188514	0.922570143971492 \\
0.473606932701979	0.827600853792656 \\
0.505180728215444	0.739066186098395 \\
0.536754523728909	0.665788257565989 \\
0.568328319242375	0.615936094683583 \\
0.59990211475584	0.596124868231477 \\
0.631475910269305	0.610473731617414 \\
0.66304970578277	0.659858611070934 \\
0.694623501296236	0.741611768484063 \\
0.726197296809701	0.849765773698224 \\
0.757771092323166	0.975720691865928 \\
0.789344887836631	1.10909863865402 \\
0.820918683350097	1.23861198631013 \\
0.852492478863562	1.35292996789476 \\
0.884066274377027	1.44163552065098 \\
0.915640069890492	1.496322459121 \\
0.947213865403958	1.51171024512777 \\
0.978787660917423	1.48648129925819 \\
1.01036145643089	1.4235364636312 \\
1.04193525194435	1.32956691372131 \\
1.07350904745782	1.2141182335679 \\
1.10508284297128	1.08846599601252 \\
1.13665663848475	0.964548211475334 \\
1.16823043399821	0.854007758379093 \\
1.19980422951168	0.76725742767941 \\
1.23137802502514	0.712495904224746 \\
1.26295182053861	0.694753624851266 \\
1.29452561605208	0.715203256166501 \\
1.32609941156554	0.770980529459429 \\
1.35767320707901	0.855587970685855 \\
1.38924700259247	0.959723618430797 \\
1.42082079810594	1.07227106578765 \\
1.4523945936194	1.18127029643139 \\
1.48396838913287	1.27486257490455 \\
1.51554218464633	1.34230978108391 \\
1.5471159801598	1.3751357859012 \\
1.57868977567326	1.36825575734218 \\
1.61026357118673	1.32079234275948 \\
1.64183736670019	1.23628681112476 \\
1.67341116221366	1.12223025928345 \\
1.70498495772712	0.98911027864498 \\
1.73655875324059	0.849285561166936 \\
1.76813254875405	0.715900723970254 \\
1.79970634426752	0.601854603398592 \\
1.83128013978098	0.518714383755102 \\
1.86285393529445	0.475523857713407 \\
1.89442773080792	0.477646995431864 \\
1.92600152632138	0.525959722152975 \\
1.95757532183485	0.616684499532256 \\
1.98914911734831	0.741929135886054 \\
2.02072291286178	0.890712564153096 \\
2.05229670837524	1.05014952982299 \\
2.08387050388871	1.20658027982669 \\
2.11544429940217	1.34663963095008 \\
2.14701809491564	1.45837824230159 \\
2.1785918904291	1.53248085043399 \\
2.21016568594257	1.56342099217969 \\
2.24173948145603	1.55021914853314 \\
2.2733132769695	1.49650483918023 \\
2.30488707248296	1.40983953505129 \\
2.33646086799643	1.30055298568679 \\
2.36803466350989	1.18046150130239 \\
2.39960845902336	1.06172054609724 \\
2.43118225453682	0.95584970064282 \\
2.46275605005029	0.872832197859157 \\
2.49432984556376	0.820219552251593 \\
2.52590364107722	0.802319784516564 \\
2.55747743659069	0.819678402397806 \\
2.58905123210415	0.869043919588731 \\
2.62062502761762	0.943836724039341 \\
2.65219882313108	1.03494834055487 \\
2.68377261864455	1.13164182103777 \\
2.71534641415801	1.22242935359316 \\
2.74692020967148	1.29595910399582 \\
2.77849400518494	1.34200877218322 \\
2.81006780069841	1.35259277540982 \\
2.84164159621187	1.32299899955465 \\
2.87321539172534	1.25243353497821 \\
2.9047891872388	1.14401209338756 \\
2.93636298275227	1.00408877361664 \\
2.96793677826573	0.841178541202727 \\
2.9995105737792	0.664819838584267 \\
3.03108436929266	0.48460905580679 \\
3.06265816480613	0.309447039980532 \\
3.0942319603196	0.146918947113761 \\
3.12580575583306	0.0027453244653115 \\
3.15737955134653	0.99203201909102 \\
3.18895334685999	1.00024857933576 \\
3.22052714237346	1.0248718333487 \\
3.25210093788692	1.06196894233216 \\
3.28367473340039	1.10628716566573 \\
3.31524852891385	1.15172950861429 \\
3.34682232442732	1.19183675753947 \\
3.37839611994078	1.22030765779255 \\
3.40996991545425	1.23161279640244 \\
3.44154371096771	1.22169373522495 \\
3.47311750648118	1.18862559204698 \\
3.50469130199464	1.13304947350706 \\
3.53626509750811	1.05822770851805 \\
3.56783889302157	0.969722584265295 \\
3.59941268853504	0.874836309808411 \\
3.6309864840485	0.781972126850307 \\
3.66256027956197	0.699984319395526 \\
3.69413407507544	0.637472784059237 \\
3.7257078705889	0.601949012395191 \\
3.75728166610237	0.598892348260564 \\
3.78885546161583	0.630870629950512 \\
3.8204292571293	0.696991085469551 \\
3.85200305264276	0.792874137835712 \\
3.88357684815623	0.911135387323038 \\
3.91515064366969	1.04217592666997 \\
3.94672443918316	1.17505606480665 \\
3.97829823469662	1.29835512431859 \\
4.00987203021009	1.40107038171022 \\
4.04144582572355	1.47364663243967 \\
4.07301962123702	1.50911036942096 \\
4.10459341675048	1.50408859159057 \\
4.13616721226395	1.45938476971987 \\
4.16774100777741	1.37988704012669 \\
4.19931480329088	1.27385015407425 \\
4.23088859880434	1.15182533690877 \\
4.26246239431781	1.02554491788454 \\
4.29403618983127	0.906914908445113 \\
4.32560998534474	0.807082735515734 \\
4.35718378085821	0.735479336046124 \\
4.38875757637167	0.69882861889739 \\
4.42033137188514	0.700290906365426 \\
4.4519051673986	0.739005151511921 \\
4.48347896291207	0.81020755256205 \\
4.51505275842553	0.905878690808749 \\
4.546626553939	1.01568458844547 \\
4.57820034945246	1.12796906966649 \\
4.60977414496593	1.23070257797577 \\
4.64134794047939	1.31245099797036 \\
4.67292173599286	1.36345966308524 \\
4.70449553150632	1.37681840060943 \\
4.73606932701979	1.34947680254327 \\
4.76764312253325	1.28278465175703 \\
4.79921691804672	1.18235409774256 \\
4.83079071356018	1.05731147795658 \\
4.86236450907365	0.919220762177739 \\
4.89393830458711	0.780964045406264 \\
4.92551210010058	0.655692613584362 \\
4.95708589561405	0.555782502038074 \\
4.98865969112751	0.491691869626996 \\
5.02023348664098	0.470755725022204 \\
5.05180728215444	0.496159847976474 \\
5.08338107766791	0.566427888671177 \\
5.11495487318137	0.675619605556142 \\
5.14652866869484	0.814152934753001 \\
5.1781024642083	0.969947519212622 \\
5.20967625972177	1.12959550943077 \\
5.24125005523523	1.27945034024525 \\
5.2728238507487	1.40670751917435 \\
5.30439764626216	1.50057998142306 \\
5.33597144177563	1.5535186072578 \\
5.36754523728909	1.56221431620198 \\
5.39911903280256	1.52803343907814 \\
5.43069282831602	1.45669505022155 \\
5.46226662382949	1.35730581969529 \\
5.49384041934295	1.24109400162697 \\
5.52541421485642	1.1201767032065 \\
5.55698801036989	1.00650587830323 \\
5.58856180588335	0.910944969197916 \\
5.62013560139682	0.842371896689039 \\
5.65170939691028	0.80680522462323 \\
5.68328319242375	0.806709320598209 \\
5.71485698793721	0.840701874867985 \\
5.74643078345068	0.90378245763225 \\
5.77800457896414	0.987996037602882 \\
5.80957837447761	1.08330783586431 \\
5.84115216999107	1.17849743376615 \\
5.87272596550454	1.2620294684673 \\
5.904299761018	1.32298368420181 \\
5.93587355653147	1.35211551204309 \\
5.96744735204493	1.34296300757187 \\
5.9990211475584	1.29272959384684 \\
6.03059494307186	1.20262205970061 \\
6.06216873858533	1.07749282067643 \\
6.09374253409879	0.924922409855037 \\
6.12531632961226	0.754072103422379 \\
6.15689012512573	0.574615674152905 \\
6.18846392063919	0.395885286333884 \\
6.22003771615266	0.226196604145246 \\
6.25161151166612	0.0722662894131889 \\
6.28318530717959	0.0612915341325858 \\
};
\addplot [very thick, color3, dashed]
table [row sep=\\]{%
0	1.04347893592467 \\
0.0315737955134653	1.09060344544631 \\
0.0631475910269305	1.14677266667443 \\
0.0947213865403958	1.2058330818623 \\
0.126295182053861	1.25976858965825 \\
0.157868977567326	1.30044394282602 \\
0.189442773080792	1.32270193782531 \\
0.221016568594257	1.32502576579607 \\
0.252590364107722	1.30547607455324 \\
0.284164159621187	1.2576188325732 \\
0.315737955134653	1.17294349332786 \\
0.347311750648118	1.0494429029457 \\
0.378885546161583	0.89783669654103 \\
0.410459341675048	0.739390946062655 \\
0.442033137188514	0.599172420499159 \\
0.473606932701979	0.498793360712476 \\
0.505180728215444	0.449669902567859 \\
0.536754523728909	0.452167049712529 \\
0.568328319242375	0.503647251201142 \\
0.59990211475584	0.606558602911675 \\
0.631475910269305	0.766111218581524 \\
0.66304970578277	0.978409101590385 \\
0.694623501296236	1.22046490913996 \\
0.726197296809701	1.45019423701953 \\
0.757771092323166	1.61492655155205 \\
0.789344887836631	1.66891633094623 \\
0.820918683350097	1.5970661703458 \\
0.852492478863562	1.42494943310724 \\
0.884066274377027	1.20343197436174 \\
0.915640069890492	0.985374661055225 \\
0.947213865403958	0.810537271330109 \\
0.978787660917423	0.698827838165464 \\
1.01036145643089	0.653107536982069 \\
1.04193525194435	0.671402358082252 \\
1.07350904745782	0.756935643091926 \\
1.10508284297128	0.914623599568135 \\
1.13665663848475	1.13651779268742 \\
1.16823043399821	1.39033282284505 \\
1.19980422951168	1.62003780314723 \\
1.23137802502514	1.75711973154579 \\
1.26295182053861	1.74446871323538 \\
1.29452561605208	1.56795128533614 \\
1.32609941156554	1.26853034031396 \\
1.35767320707901	0.921838231116156 \\
1.38924700259247	0.607751332189445 \\
1.42082079810594	0.386713147348467 \\
1.4523945936194	0.283876383915603 \\
1.48396838913287	0.290552314345193 \\
1.51554218464633	0.385230411801876 \\
1.5471159801598	0.553084494676193 \\
1.57868977567326	0.786555024735262 \\
1.61026357118673	1.07161022225146 \\
1.64183736670019	1.37566544952606 \\
1.67341116221366	1.64599111723248 \\
1.70498495772712	1.81870899037088 \\
1.73655875324059	1.84137804630354 \\
1.76813254875405	1.70212518700795 \\
1.79970634426752	1.43737045753074 \\
1.83128013978098	1.10873795615772 \\
1.86285393529445	0.775195595998555 \\
1.89442773080792	0.47863042812081 \\
1.92600152632138	0.241254131986441 \\
1.95757532183485	0.0740665679439972 \\
1.98914911734831	0.00739833437032611 \\
2.02072291286178	0.0243682825200384 \\
2.05229670837524	0.194846365961107 \\
2.08387050388871	0.501164371407633 \\
2.11544429940217	0.901292960511146 \\
2.14701809491564	1.32815971585433 \\
2.1785918904291	1.71156937350792 \\
2.21016568594257	1.9984646107528 \\
2.24173948145603	2.1677291453755 \\
2.2733132769695	2.22467298039608 \\
2.30488707248296	2.17574643527047 \\
2.33646086799643	2.01188678625889 \\
2.36803466350989	1.72121387886451 \\
2.39960845902336	1.31858230628185 \\
2.43118225453682	0.860106910693585 \\
2.46275605005029	0.431960170652347 \\
2.49432984556376	0.126695231272197 \\
2.52590364107722	0.0115605098513495 \\
2.55747743659069	0.09766761648618 \\
2.58905123210415	0.338826391387154 \\
2.62062502761762	0.663163916770354 \\
2.65219882313108	1.00496215490734 \\
2.68377261864455	1.31924260497385 \\
2.71534641415801	1.58258420663254 \\
2.74692020967148	1.78167541520952 \\
2.77849400518494	1.89577592898809 \\
2.81006780069841	1.89117481831549 \\
2.84164159621187	1.74004079246554 \\
2.87321539172534	1.44916029967585 \\
2.9047891872388	1.06651778688493 \\
2.93636298275227	0.660644994118552 \\
2.96793677826573	0.295950244195123 \\
2.9995105737792	0.0146639690697578 \\
3.03108436929266	0.173349022661772 \\
3.06265816480613	0.28444397519485 \\
3.0942319603196	0.341774472624045 \\
3.12580575583306	0.361002369498596 \\
3.15737955134653	1.06560423297162 \\
3.18895334685999	1.11789286032077 \\
3.22052714237346	1.17640820813135 \\
3.25210093788692	1.23398281865005 \\
3.28367473340039	1.282187787012 \\
3.31524852891385	1.3140265097712 \\
3.34682232442732	1.32639980179616 \\
3.37839611994078	1.31826868569789 \\
3.40996991545425	1.28565447029117 \\
3.44154371096771	1.22027172432196 \\
3.47311750648118	1.11568888251189 \\
3.50469130199464	0.975997538171273 \\
3.53626509750811	0.817907634802605 \\
3.56783889302157	0.665483280305087 \\
3.59941268853504	0.542992524954042 \\
3.6309864840485	0.467603745115438 \\
3.66256027956197	0.444688500565206 \\
3.69413407507544	0.471787316182991 \\
3.7257078705889	0.548298980304969 \\
3.75728166610237	0.679125812446245 \\
3.78885546161583	0.866605666743757 \\
3.8204292571293	1.09796479751041 \\
3.85200305264276	1.34006595393613 \\
3.88357684815623	1.5439862024687 \\
3.91515064366969	1.65767332489971 \\
3.94672443918316	1.64799997008957 \\
3.97829823469662	1.52065780238773 \\
4.00987203021009	1.31688332178253 \\
4.04144582572355	1.09106399582321 \\
4.07301962123702	0.890818929987027 \\
4.10459341675048	0.746328055987753 \\
4.13616721226395	0.667895803509068 \\
4.16774100777741	0.65421032633707 \\
4.19931480329088	0.705351211510364 \\
4.23088859880434	0.826767985904137 \\
4.26246239431781	1.01894001252667 \\
4.29403618983127	1.26246424527323 \\
4.32560998534474	1.5123943419395 \\
4.35718378085821	1.70437424907748 \\
4.38875757637167	1.77171747005199 \\
4.42033137188514	1.6753113823281 \\
4.4519051673986	1.42930984176199 \\
4.48347896291207	1.09588772838674 \\
4.51505275842553	0.756251987796047 \\
4.546626553939	0.48310022230864 \\
4.57820034945246	0.320454920977988 \\
4.60977414496593	0.27487060822886 \\
4.64134794047939	0.328084671804328 \\
4.67292173599286	0.460524688633546 \\
4.70449553150632	0.662156036393418 \\
4.73606932701979	0.924164835032189 \\
4.76764312253325	1.22413774766181 \\
4.79921691804672	1.51896386675946 \\
4.83079071356018	1.74846198913289 \\
4.86236450907365	1.85076085151259 \\
4.89393830458711	1.79068307520131 \\
4.92551210010058	1.58179510012535 \\
4.95708589561405	1.27705255654749 \\
4.98865969112751	0.939482810631972 \\
5.02023348664098	0.620482436449075 \\
5.05180728215444	0.351770718092131 \\
5.08338107766791	0.148196971145609 \\
5.11495487318137	0.0211301260198898 \\
5.14652866869484	0.00762032494666365 \\
5.1781024642083	0.0915158363878633 \\
5.20967625972177	0.332809866569476 \\
5.24125005523523	0.693371580089652 \\
5.2728238507487	1.11594129457447 \\
5.30439764626216	1.52924342348601 \\
5.33597144177563	1.86922195838178 \\
5.36754523728909	2.09783518698807 \\
5.39911903280256	2.20954608144043 \\
5.43069282831602	2.21364402652229 \\
5.46226662382949	2.10920112468847 \\
5.49384041934295	1.88241269522375 \\
5.52541421485642	1.53131931313892 \\
5.55698801036989	1.09139802908575 \\
5.58856180588335	0.636282011329632 \\
5.62013560139682	0.258743102484796 \\
5.65170939691028	0.043126137481903 \\
5.68328319242375	0.0312816099337593 \\
5.71485698793721	0.203208292739714 \\
5.74643078345068	0.495101159968425 \\
5.77800457896414	0.835244742954462 \\
5.80957837447761	1.1674375558339 \\
5.84115216999107	1.45813740673082 \\
5.87272596550454	1.69110961669264 \\
5.904299761018	1.8512534059442 \\
5.93587355653147	1.91055970373019 \\
5.96744735204493	1.83456061644086 \\
5.9990211475584	1.60987469549248 \\
6.03059494307186	1.26518528218875 \\
6.06216873858533	0.862123731493636 \\
6.09374253409879	0.469816578771174 \\
6.12531632961226	0.143478870996758 \\
6.15689012512573	0.090389671122886 \\
6.18846392063919	0.236987684035201 \\
6.22003771615266	0.318609708806547 \\
6.25161151166612	0.355547668998811 \\
6.28318530717959	0.358939277836687 \\
};

\end{axis}

\end{tikzpicture}

%% file: figures/stock/91/theta.tex
\begin{tikzpicture}

\definecolor{color0}{rgb}{0.12156862745098,0.466666666666667,0.705882352941177}

\begin{axis}[
tick align=outside,
tick pos=left,
title={EOG vs MSFT},
x grid style={white!69.01960784313725!black},
xlabel={$\Theta$},
xmin=-0.314159265358979, xmax=6.59734457253857,
y grid style={white!69.01960784313725!black},
ylabel={$\Gamma(\Theta)$},
ymin=-0.728636242860817, ymax=15.3024971719396
]
\addplot [semithick, color0, forget plot]
table [row sep=\\]{%
0	0.258606540972753 \\
0.00628947478196155	0.291743250328467 \\
0.0125789495639231	0.348304719420732 \\
0.0188684243458846	0.496218691431 \\
0.0251578991278462	0.678818725835873 \\
0.0314473739098077	1.06605734823831 \\
0.0377368486917693	1.36114890156564 \\
0.0440263234737308	2.24585009077416 \\
0.0503157982556924	3.16144153749495 \\
0.0566052730376539	4.28372462371625 \\
0.0628947478196155	5.27474863171515 \\
0.069184222601577	6.19052451344243 \\
0.0754736973835386	7.10782275030391 \\
0.0817631721655001	8.01386044350166 \\
0.0880526469474617	8.92697688542689 \\
0.0943421217294232	9.83658762581811 \\
0.100631596511385	10.7331478247877 \\
0.106921071293346	11.6175225392709 \\
0.113210546075308	12.4825278442039 \\
0.119500020857269	13.3350400847912 \\
0.125789495639231	14.1672944380216 \\
0.132078970421193	14.4485952988229 \\
0.138368445203154	14.0797079609622 \\
0.144657919985116	13.7002754171484 \\
0.150947394767077	13.299278789096 \\
0.157236869549039	12.8776634064931 \\
0.163526344331	12.4441246918839 \\
0.169815819112962	11.9953720576362 \\
0.176105293894923	11.5325947537494 \\
0.182394768676885	11.0348355628187 \\
0.188684243458846	10.2053732929964 \\
0.194973718240808	9.37607345875869 \\
0.20126319302277	8.53885574713375 \\
0.207552667804731	7.69190643808945 \\
0.213842142586693	6.75964087824672 \\
0.220131617368654	5.4435331358945 \\
0.226421092150616	4.10489336384719 \\
0.232710566932577	2.77614123823869 \\
0.239000041714539	1.7260598836444 \\
0.2452895164965	0.783347970110973 \\
0.251578991278462	0.43649653887445 \\
0.257868466060423	0.195152005370909 \\
0.264157940842385	0.109146668672262 \\
0.270447415624347	5.16396301115662e-05 \\
0.276736890406308	0.0813710900055411 \\
0.28302636518827	0.135538996708298 \\
0.289315839970231	0.110659194616916 \\
0.295605314752193	0.0853892340000217 \\
0.301894789534154	0.0602837420259954 \\
0.308184264316116	0.0338266362151156 \\
0.314473739098077	0.00201699496394037 \\
0.320763213880039	0.0427481901967344 \\
0.327052688662	0.166134230796139 \\
0.333342163443962	0.301897217773687 \\
0.339631638225924	0.441508312432022 \\
0.345921113007885	0.59131568492621 \\
0.352210587789847	0.724199570943009 \\
0.358500062571808	0.855720361147794 \\
0.36478953735377	0.981270795119845 \\
0.371079012135731	1.10641974204592 \\
0.377368486917693	1.2316209669377 \\
0.383657961699654	1.35832896435986 \\
0.389947436481616	1.48785841356644 \\
0.396236911263577	1.61695885559083 \\
0.402526386045539	1.74562518356004 \\
0.408815860827501	1.87089465658322 \\
0.415105335609462	1.98010619846433 \\
0.421394810391424	2.07863192728065 \\
0.427684285173385	2.17635237754414 \\
0.433973759955347	2.28275188820926 \\
0.440263234737308	2.38894221665396 \\
0.44655270951927	2.49158737165795 \\
0.452842184301231	2.59236797103442 \\
0.459131659083193	2.68709260532473 \\
0.465421133865154	2.73010849317114 \\
0.471710608647116	2.77166134442461 \\
0.478000083429078	2.80943329454751 \\
0.484289558211039	2.84466167042956 \\
0.490579032993001	2.82301761092485 \\
0.496868507774962	2.78965355537186 \\
0.503157982556924	2.75481159401008 \\
0.509447457338885	2.71849310509565 \\
0.515736932120847	2.67703077011822 \\
0.522026406902808	2.59743528696468 \\
0.52831588168477	2.51498339471123 \\
0.534605356466732	2.43284906102367 \\
0.540894831248693	2.38027667644434 \\
0.547184306030655	2.32089315763008 \\
0.553473780812616	2.21897164135078 \\
0.559763255594578	2.11043842973175 \\
0.566052730376539	1.99903860701278 \\
0.572342205158501	1.88681957537351 \\
0.578631679940462	1.77380174618114 \\
0.584921154722424	1.65998959012292 \\
0.591210629504386	1.54538760930759 \\
0.597500104286347	1.42883853668093 \\
0.603789579068309	1.31095702454461 \\
0.61007905385027	1.19237597986257 \\
0.616368528632232	1.07659197361112 \\
0.622658003414193	0.960528863085887 \\
0.628947478196155	0.844710705201157 \\
0.635236952978116	0.726808029873041 \\
0.641526427760078	0.608115346651223 \\
0.647815902542039	0.544421945219364 \\
0.654105377324001	0.501233119649989 \\
0.660394852105962	0.464281051425332 \\
0.666684326887924	0.429506943673206 \\
0.672973801669886	0.400730988837745 \\
0.679263276451847	0.373311700105435 \\
0.685552751233809	0.346363476477488 \\
0.69184222601577	0.319208293000069 \\
0.698131700797732	0.291819663580128 \\
0.704421175579693	0.264198671639614 \\
0.710710650361655	0.242462486652455 \\
0.717000125143616	0.234406449090564 \\
0.723289599925578	0.226213256230554 \\
0.72957907470754	0.21788323217352 \\
0.735868549489501	0.209416706433251 \\
0.742158024271463	0.200814013923172 \\
0.748447499053424	0.192075494943113 \\
0.754736973835386	0.182346492399784 \\
0.761026448617347	0.172336021916654 \\
0.767315923399309	0.162201808636912 \\
0.77360539818127	0.15192043739366 \\
0.779894872963232	0.141428121588691 \\
0.786184347745194	0.130813805157076 \\
0.792473822527155	0.12007790797318 \\
0.798763297309117	0.10922085472078 \\
0.805052772091078	0.0982430748762697 \\
0.81134224687304	0.0845252337532676 \\
0.817631721655001	0.0675404085082496 \\
0.823921196436963	0.051084445956032 \\
0.830210671218924	0.0371342242042627 \\
0.836500146000886	0.0244034854892914 \\
0.842789620782847	0.0157598210814682 \\
0.849079095564809	0.00701933923796538 \\
0.855368570346771	0.00181761429080263 \\
0.861658045128732	0.0107506899382628 \\
0.867947519910694	0.0197795343355027 \\
0.874236994692655	0.0289037903252565 \\
0.880526469474617	0.0381230969760051 \\
0.886815944256578	0.0474370895963063 \\
0.89310541903854	0.0564232026828089 \\
0.899394893820501	0.0649923259777569 \\
0.905684368602463	0.0736646220737893 \\
0.911973843384424	0.0824397479177437 \\
0.918263318166386	0.0913173563887817 \\
0.924552792948347	0.100297096312131 \\
0.930842267730309	0.109378612472961 \\
0.937131742512271	0.118561545630444 \\
0.943421217294232	0.127907136638412 \\
0.949710692076194	0.13795080629983 \\
0.956000166858155	0.148093393089593 \\
0.962289641640117	0.158334495793714 \\
0.968579116422078	0.168673709301183 \\
0.97486859120404	0.179649348356782 \\
0.981158065986002	0.190489796121721 \\
0.987447540767963	0.201433414539861 \\
0.993737015549924	0.212835083146128 \\
1.00002649033189	0.225808791630852 \\
1.00631596511385	0.238498040683165 \\
1.01260543989581	0.251287466974284 \\
1.01889491467777	0.264176564588236 \\
1.02518438945973	0.277164823666316 \\
1.03147386424169	0.300946691067404 \\
1.03776333902366	0.33000976002872 \\
1.04405281380562	0.358890687801273 \\
1.05034228858758	0.395143900012056 \\
1.05663176336954	0.435403475706992 \\
1.0629212381515	0.475658685351202 \\
1.06921071293346	0.519913665166976 \\
1.07550018771542	0.56424063640452 \\
1.08178966249739	0.614069137604936 \\
1.08807913727935	0.666505566517166 \\
1.09436861206131	0.70883658071847 \\
1.10065808684327	0.747254951355133 \\
1.10694756162523	0.785628183705825 \\
1.11323703640719	0.82396361440996 \\
1.11952651118916	0.862298674875868 \\
1.12581598597112	0.900584134280178 \\
1.13210546075308	0.948921704856366 \\
1.13839493553504	1.02884811673066 \\
1.144684410317	1.14944494281975 \\
1.15097388509896	1.33012540303177 \\
1.15726335988092	1.51043727316876 \\
1.16355283466289	1.69037342056868 \\
1.16984230944485	1.86992672743212 \\
1.17613178422681	2.04909009110378 \\
1.18242125900877	2.23272528298377 \\
1.18871073379073	2.40140493069137 \\
1.19500020857269	2.50196020956376 \\
1.20128968335466	2.60191864684114 \\
1.20757915813662	2.70127628843133 \\
1.21386863291858	2.8000292040081 \\
1.22015810770054	2.86225733125282 \\
1.2264475824825	2.91691208586607 \\
1.23273705726446	2.97097318672068 \\
1.23902653204642	2.98483555051624 \\
1.24531600682839	2.99137739476855 \\
1.25160548161035	2.97605922213863 \\
1.25789495639231	2.95557837843083 \\
1.26418443117427	2.93459102861618 \\
1.27047390595623	2.91306360074662 \\
1.27676338073819	2.89099694639042 \\
1.28305285552016	2.86284314170012 \\
1.28934233030212	2.80180466668922 \\
1.29563180508408	2.74020382135121 \\
1.30192127986604	2.67804304245307 \\
1.308210754648	2.61532478891127 \\
1.31450022942996	2.5520515416945 \\
1.32078970421192	2.48530936357586 \\
1.32707917899389	2.41687863847566 \\
1.33336865377585	2.34760890912359 \\
1.33965812855781	2.27141599948023 \\
1.34594760333977	2.19409672842256 \\
1.35223707812173	2.1161832449921 \\
1.35852655290369	2.03767863124077 \\
1.36481602768566	1.95858599260403 \\
1.37110550246762	1.87890845777808 \\
1.37739497724958	1.79864917859606 \\
1.38368445203154	1.71782966859343 \\
1.3899739268135	1.63663813573953 \\
1.39626340159546	1.55487482514321 \\
1.40255287637743	1.47254297114539 \\
1.40884235115939	1.38964583057708 \\
1.41513182594135	1.30966118815123 \\
1.42142130072331	1.26535434419236 \\
1.42771077550527	1.22230264872678 \\
1.43400025028723	1.18177849942682 \\
1.44028972506919	1.15539418726336 \\
1.44657919985116	1.13484978913154 \\
1.45286867463312	1.11314373480551 \\
1.45915814941508	1.09137290769831 \\
1.46544762419704	1.06953816900646 \\
1.471737098979	1.04764038245462 \\
1.47802657376096	1.02568041426148 \\
1.48431604854293	1.00365913310547 \\
1.49060552332489	0.981061466039867 \\
1.49689499810685	0.958279259716217 \\
1.50318447288881	0.935438120327122 \\
1.50947394767077	0.91361591723732 \\
1.51576342245273	0.893474774391745 \\
1.52205289723469	0.871821700401027 \\
1.52834237201666	0.852210988009277 \\
1.53463184679862	0.833407939311568 \\
1.54092132158058	0.812234387163554 \\
1.54721079636254	0.791026885432812 \\
1.5535002711445	0.770775960628164 \\
1.55978974592646	0.753530104358505 \\
1.56607922070843	0.747794887688817 \\
1.57236869549039	0.740999137766722 \\
1.57865817027235	0.73430218566887 \\
1.58494764505431	0.726853806772906 \\
1.59123711983627	0.718165007559703 \\
1.59752659461823	0.709549049785328 \\
1.60381606940019	0.701027857533839 \\
1.61010554418216	0.692560618020127 \\
1.61639501896412	0.684991411291459 \\
1.62268449374608	0.681732485570535 \\
1.62897396852804	0.678398629366383 \\
1.63526344331	0.674680779780744 \\
1.64155291809196	0.671036615667667 \\
1.64784239287393	0.667466281180677 \\
1.65413186765589	0.663969917552782 \\
1.66042134243785	0.660547663090915 \\
1.66671081721981	0.658204791120926 \\
1.67300029200177	0.656542163072362 \\
1.67928976678373	0.650874650326783 \\
1.68557924156569	0.641533984588794 \\
1.69186871634766	0.632271027963329 \\
1.69815819112962	0.621847551941086 \\
1.70444766591158	0.611425684202186 \\
1.71073714069354	0.601086899727724 \\
1.7170266154755	0.59083160749275 \\
1.72331609025746	0.580660213169575 \\
1.72960556503943	0.570573119111736 \\
1.73589503982139	0.559773451297563 \\
1.74218451460335	0.549005473371858 \\
1.74847398938531	0.538321316311553 \\
1.75476346416727	0.527212863505035 \\
1.76105293894923	0.515191268393688 \\
1.76734241373119	0.504219488411066 \\
1.77363188851316	0.493342631761384 \\
1.77992136329512	0.482561128704402 \\
1.78621083807708	0.47137096594501 \\
1.79250031285904	0.459827051915132 \\
1.798789787641	0.448411317797297 \\
1.80507926242296	0.437124215167842 \\
1.81136873720493	0.425966190514771 \\
1.81765821198689	0.412090363866921 \\
1.82394768676885	0.39814743069214 \\
1.83023716155081	0.384345484148445 \\
1.83652663633277	0.370685070204439 \\
1.84281611111473	0.357166729230071 \\
1.84910558589669	0.343686596158602 \\
1.85539506067866	0.334860094019079 \\
1.86168453546062	0.326165127055187 \\
1.86797401024258	0.317778849581885 \\
1.87426348502454	0.310114677466848 \\
1.8805529598065	0.302558996078758 \\
1.88684243458846	0.295112104300443 \\
1.89313190937043	0.284414458072895 \\
1.89942138415239	0.273661537564707 \\
1.90571085893435	0.263000822561802 \\
1.91200033371631	0.25245055102457 \\
1.91828980849827	0.242011140293927 \\
1.92457928328023	0.231659932404541 \\
1.9308687580622	0.221412623242228 \\
1.93715823284416	0.211277456146766 \\
1.94344770762612	0.201254832038635 \\
1.94973718240808	0.191551221721689 \\
1.95602665719004	0.182155274882742 \\
1.962316131972	0.172873524412132 \\
1.96860560675396	0.163706337471424 \\
1.97489508153593	0.154654076690354 \\
1.98118455631789	0.145717100152489 \\
1.98747403109985	0.136895761381049 \\
1.99376350588181	0.127727883002288 \\
2.00005298066377	0.118605860340282 \\
2.00634245544573	0.109603510670602 \\
2.0126319302277	0.100930911253057 \\
2.01892140500966	0.0936280817989155 \\
2.02521087979162	0.0863935164993319 \\
2.03150035457358	0.0792984544945212 \\
2.03778982935554	0.0723431764464193 \\
2.0440793041375	0.0655279574874896 \\
2.05036877891946	0.0588530672098191 \\
2.05665825370143	0.0523187696544589 \\
2.06294772848339	0.0468268375951961 \\
2.06923720326535	0.0441514051547034 \\
2.07552667804731	0.0417069203961988 \\
2.08181615282927	0.0394286125143499 \\
2.08810562761123	0.0372845539065483 \\
2.0943951023932	0.0352748293861063 \\
2.10068457717516	0.0333995184524132 \\
2.10697405195712	0.0316586952878324 \\
2.11326352673908	0.030052428754737 \\
2.11955300152104	0.0285807823927873 \\
2.125842476303	0.0272438144164426 \\
2.13213195108496	0.0260415777126224 \\
2.13842142586693	0.0255728065602741 \\
2.14471090064889	0.0257046884694825 \\
2.15100037543085	0.0259713621770468 \\
2.15728985021281	0.0263728171340425 \\
2.16357932499477	0.0269090374599776 \\
2.16986879977673	0.0275800019433938 \\
2.1761582745587	0.0283856840427039 \\
2.18244774934066	0.0293260518872467 \\
2.18873722412262	0.030401068278552 \\
2.19502669890458	0.0316106906918048 \\
2.20131617368654	0.0329548712775347 \\
2.2076056484685	0.0344335568635024 \\
2.21389512325046	0.0360466889568052 \\
2.22018459803243	0.037737195205036 \\
2.22647407281439	0.0392515328759502 \\
2.23276354759635	0.0408857023282003 \\
2.23905302237831	0.042639638918355 \\
2.24534249716027	0.0417566918438483 \\
2.25163197194223	0.0377425250365242 \\
2.2579214467242	0.0328072165800792 \\
2.26421092150616	0.0280123161928998 \\
2.27050039628812	0.0233580135485969 \\
2.27678987107008	0.0170147553208833 \\
2.28307934585204	0.0103406741224266 \\
2.289368820634	0.00333976136488845 \\
2.29565829541597	0.00489187282432724 \\
2.30194777019793	0.0129825002862982 \\
2.30823724497989	0.0209318009771393 \\
2.31452671976185	0.0287394604434836 \\
2.32081619454381	0.0364051698349117 \\
2.32710566932577	0.0439286259161791 \\
2.33339514410773	0.0518103035284323 \\
2.3396846188897	0.0598563372921381 \\
2.34597409367166	0.0677602728152515 \\
2.35226356845362	0.0755217974389333 \\
2.35855304323558	0.0837517395431542 \\
2.36484251801754	0.0937066576390926 \\
2.3711319927995	0.104127101864704 \\
2.37742146758147	0.114568759171543 \\
2.38371094236343	0.12485017680029 \\
2.39000041714539	0.13500328596198 \\
2.39628989192735	0.145200082634071 \\
2.40257936670931	0.155235655680441 \\
2.40886884149127	0.165111037513021 \\
2.41515831627323	0.175197471446347 \\
2.4214477910552	0.185121020366469 \\
2.42773726583716	0.194881291723959 \\
2.43402674061912	0.204477899428221 \\
2.44031621540108	0.213910463862771 \\
2.44660569018304	0.224288413027485 \\
2.452895164965	0.236454069882089 \\
2.45918463974697	0.248449570195113 \\
2.46547411452893	0.261556802078314 \\
2.47176358931089	0.274372666250769 \\
2.47805306409285	0.286937174005775 \\
2.48434253887481	0.299314652946102 \\
2.49063201365677	0.312210253851531 \\
2.49692148843873	0.327733210409602 \\
2.5032109632207	0.349669824468494 \\
2.50950043800266	0.377225451541817 \\
2.51578991278462	0.404465886973892 \\
2.52207938756658	0.431390053204938 \\
2.52836886234854	0.457996885185929 \\
2.5346583371305	0.48811829929391 \\
2.54094781191247	0.518032158653032 \\
2.54723728669443	0.547583849635762 \\
2.55352676147639	0.576772203255159 \\
2.55981623625835	0.605596064896905 \\
2.56610571104031	0.634054294365078 \\
2.57239518582227	0.662145765927169 \\
2.57868466060423	0.689869368358645 \\
2.5849741353862	0.71722400498691 \\
2.59126361016816	0.744208593734693 \\
2.59755308495012	0.770822067162827 \\
2.60384255973208	0.797063372512483 \\
2.61013203451404	0.823261823329618 \\
2.616421509296	0.850644677039195 \\
2.62271098407797	0.877629633216602 \\
2.62900045885993	0.904215624408125 \\
2.63528993364189	0.930401598942068 \\
2.64157940842385	0.956186520970353 \\
2.64786888320581	0.981569370509467 \\
2.65415835798777	1.00041543153481 \\
2.66044783276973	1.01700659025136 \\
2.6667373075517	1.03319445347496 \\
2.67302678233366	1.04897838085645 \\
2.67931625711562	1.06435774802526 \\
2.68560573189758	1.07933194661422 \\
2.69189520667954	1.09390038428351 \\
2.6981846814615	1.10806248474416 \\
2.70447415624347	1.12181768778081 \\
2.71076363102543	1.13516544927394 \\
2.71705310580739	1.14810524122128 \\
2.72334258058935	1.1606365517588 \\
2.72963205537131	1.1727588851809 \\
2.73592153015327	1.18447176196004 \\
2.74221100493523	1.19577471876572 \\
2.7485004797172	1.20113609000541 \\
2.75478995449916	1.19700402749678 \\
2.76107942928112	1.192184052618 \\
2.76736890406308	1.18689885520813 \\
2.77365837884504	1.18250988997657 \\
2.779947853627	1.17946409904467 \\
2.78623732840897	1.17594055204674 \\
2.79252680319093	1.17193938836502 \\
2.79881627797289	1.16746076627498 \\
2.80510575275485	1.16250486293911 \\
2.81139522753681	1.15707187439986 \\
2.81768470231877	1.15106880222361 \\
2.82397417710073	1.14372358949387 \\
2.8302636518827	1.13590144220654 \\
2.83655312666466	1.12760266978515 \\
2.84284260144662	1.11882760050724 \\
2.84913207622858	1.1095765814914 \\
2.85542155101054	1.09984997868357 \\
2.8617110257925	1.08964817684248 \\
2.86800050057447	1.073832513654 \\
2.87428997535643	1.05464472157999 \\
2.88057945013839	1.03497171370528 \\
2.88686892492035	1.01481426824213 \\
2.89315839970231	0.994173182565975 \\
2.89944787448427	0.97304927318368 \\
2.90573734926624	0.951443375701406 \\
2.9120268240482	0.930354494066129 \\
2.91831629883016	0.908921507690891 \\
2.92460577361212	0.890512983293249 \\
2.93089524839408	0.874505137762432 \\
2.93718472317604	0.858064135068184 \\
2.943474197958	0.841466799528616 \\
2.94976367273997	0.824675584077883 \\
2.95605314752193	0.807451160987539 \\
2.96234262230389	0.789794211610331 \\
2.96863209708585	0.771705434408613 \\
2.97492157186781	0.753185544926694 \\
2.98121104664977	0.734235275762561 \\
2.98750052143174	0.7148553765389 \\
2.9937899962137	0.68379857092755 \\
3.00007947099566	0.643267757543056 \\
3.00636894577762	0.60227693564726 \\
3.01265842055958	0.561291125566228 \\
3.01894789534154	0.519930488824222 \\
3.0252373701235	0.478120325054207 \\
3.03152684490547	0.438117688351827 \\
3.03781631968743	0.401363405268014 \\
3.04410579446939	0.345630016801389 \\
3.05039526925135	0.285839500037613 \\
3.05668474403331	0.225620989305208 \\
3.06297421881527	0.164976866689666 \\
3.06926369359724	0.111760290527684 \\
3.0755531683792	0.0944907961090804 \\
3.08184264316116	0.0774057527544163 \\
3.08813211794312	0.067129342855762 \\
3.09442159272508	0.0660031272785813 \\
3.10071106750704	0.0799783387946822 \\
3.107000542289	0.100563811221452 \\
3.11329001707097	0.127021508310574 \\
3.11957949185293	0.154336652295093 \\
3.12586896663489	0.181211363362273 \\
3.13215844141685	0.207644578419414 \\
3.13844791619881	0.241387094806021 \\
3.14473739098077	0.275002929985984 \\
3.15102686576274	0.313516704305969 \\
3.1573163405447	0.41506853446356 \\
3.16360581532666	0.57706696644698 \\
3.16989529010862	0.86807781954617 \\
3.17618476489058	1.17122159495345 \\
3.18247423967254	1.78710994796175 \\
3.1887637144545	2.70412902862377 \\
3.19505318923647	3.67711711820216 \\
3.20134266401843	4.8138802855383 \\
3.20763213880039	5.73407730151212 \\
3.21392161358235	6.65057894238566 \\
3.22021108836431	7.56225141534614 \\
3.22650056314627	8.46485671861046 \\
3.23279003792824	9.38385201425709 \\
3.2390795127102	10.2864011685733 \\
3.24536898749216	11.1768001553019 \\
3.25165846227412	12.0519018099069 \\
3.25794793705608	12.9102420124392 \\
3.26423741183804	13.7569178602766 \\
3.27052688662001	14.5738092894487 \\
3.27681636140197	14.2650778425062 \\
3.28310583618393	13.8914398652257 \\
3.28939531096589	13.5028563044645 \\
3.29568478574785	13.0899625677322 \\
3.30197426052981	12.6623834048828 \\
3.30826373531177	12.2222562270076 \\
3.31455321009374	11.7654841420683 \\
3.3208426848757	11.2967061958079 \\
3.32713215965766	10.6210629245594 \\
3.33342163443962	9.7905073382183 \\
3.33971110922158	8.95888996160373 \\
3.34600058400354	8.11597496922052 \\
3.3522900587855	7.27462354425111 \\
3.35857953356747	6.10290219259691 \\
3.36486900834943	4.77620177175358 \\
3.37115848313139	3.44155471043202 \\
3.37744795791335	2.22842249867231 \\
3.38373743269531	1.21245766587513 \\
3.39002690747727	0.576415283630294 \\
3.39631638225924	0.297455190573015 \\
3.4026058570412	0.151144605629129 \\
3.40889533182316	0.0553062531401469 \\
3.41518480660512	0.049468288137819 \\
3.42147428138708	0.109218552811305 \\
3.42776375616904	0.123410118026235 \\
3.43405323095101	0.0979855248299986 \\
3.44034270573297	0.0728704466963142 \\
3.44663218051493	0.0477222000521964 \\
3.45292165529689	0.0184126731560013 \\
3.45921113007885	0.0221634156382762 \\
3.46550060486081	0.0967780983270465 \\
3.47179007964277	0.234746082111558 \\
3.47807955442474	0.369270131224454 \\
3.4843690292067	0.515410273455027 \\
3.49065850398866	0.65779485721272 \\
3.49694797877062	0.790529169416633 \\
3.50323745355258	0.918545453981742 \\
3.50952692833454	1.04389576424829 \\
3.51581640311651	1.16884211019025 \\
3.52210587789847	1.29438710720761 \\
3.52839535268043	1.42314699501607 \\
3.53468482746239	1.55246258005608 \\
3.54097430224435	1.68134660234345 \\
3.54726377702631	1.80933805197161 \\
3.55355325180827	1.93063222360967 \\
3.55984272659024	2.02946947984775 \\
3.5661322013722	2.12759305457759 \\
3.57242167615416	2.22937917488285 \\
3.57871115093612	2.335939744131 \\
3.58500062571808	2.44089885875845 \\
3.59129010050004	2.54207724218753 \\
3.59757957528201	2.64245906085523 \\
3.60386905006397	2.70878332382801 \\
3.61015852484593	2.75106790246176 \\
3.61644799962789	2.79125522292292 \\
3.62273747440985	2.82723550240057 \\
3.62902694919181	2.8391450093023 \\
3.63531642397377	2.80652040523241 \\
3.64160589875574	2.77241722814564 \\
3.6478953735377	2.73683682707375 \\
3.65418484831966	2.69978060948357 \\
3.66047432310162	2.63746676358331 \\
3.66676379788358	2.55637351419702 \\
3.67305327266554	2.47327758466139 \\
3.67934274744751	2.40543794876672 \\
3.68563222222947	2.35519810402049 \\
3.69192169701143	2.27064391788225 \\
3.69821117179339	2.1651471195565 \\
3.70450064657535	2.05484718529559 \\
3.71079012135731	1.94302921907299 \\
3.71707959613927	1.83041023179202 \\
3.72336907092124	1.71699467836283 \\
3.7296585457032	1.60278704520557 \\
3.73594802048516	1.48750687172429 \\
3.74223749526712	1.36997582679531 \\
3.74852697004908	1.25176284769491 \\
3.75481644483104	1.1343456836426 \\
3.76110591961301	1.01865284243181 \\
3.76739539439497	0.902706390868419 \\
3.77368486917693	0.78588173829427 \\
3.77997434395889	0.667552370658419 \\
3.78626381874085	0.569022743741335 \\
3.79255329352281	0.52332347555374 \\
3.79884276830477	0.481931949121055 \\
3.80513224308674	0.446926613158843 \\
3.8114217178687	0.41419759673966 \\
3.81771119265066	0.387205552099503 \\
3.82400066743262	0.359486089482791 \\
3.83029014221458	0.332815132835238 \\
3.83657961699654	0.305543091534977 \\
3.84286909177851	0.278038144851521 \\
3.84915856656047	0.249800618681771 \\
3.85544804134243	0.238451632328309 \\
3.86173751612439	0.230326976943516 \\
3.86802699090635	0.222065327633762 \\
3.87431646568831	0.213667011208102 \\
3.88060594047027	0.205132359881786 \\
3.88689541525224	0.196461711263112 \\
3.8931848900342	0.18730520015479 \\
3.89947436481616	0.177356749871943 \\
3.90576383959812	0.167284358185729 \\
3.91205331438008	0.157088423533394 \\
3.91834278916204	0.146689555623331 \\
3.92463226394401	0.136136187321983 \\
3.93092173872597	0.125461027728954 \\
3.93721121350793	0.114664499125398 \\
3.94350068828989	0.103747028593517 \\
3.94979016307185	0.0925903211607135 \\
3.95607963785381	0.0760472442934361 \\
3.96236911263578	0.0590048105249168 \\
3.96865858741774	0.0441243744843987 \\
3.9749480621997	0.0301140642437767 \\
3.98123753698166	0.020093776924794 \\
3.98752701176362	0.0114016608193785 \\
3.99381648654558	0.00261289967560963 \\
4.00010596132754	0.00627215884629884 \\
4.00639543610951	0.0152531632768649 \\
4.01268491089147	0.0243297583512438 \\
4.01897438567343	0.0335015850232643 \\
4.02526386045539	0.0427682804796561 \\
4.03155333523735	0.052129478154392 \\
4.03784281001931	0.0606948464485892 \\
4.04413228480128	0.0693155987709639 \\
4.05042175958324	0.0780393528771515 \\
4.0567112343652	0.0868657636784484 \\
4.06300070914716	0.0957944820253185 \\
4.06929018392912	0.104825154721218 \\
4.07557965871108	0.11395742453655 \\
4.08186913349304	0.123190930222801 \\
4.08815860827501	0.132916581904868 \\
4.09444808305697	0.143009760038018 \\
4.10073755783893	0.153201655180647 \\
4.10702703262089	0.16349186416826 \\
4.11331650740285	0.174193134194178 \\
4.11960598218481	0.185056649010528 \\
4.12589545696678	0.195948735959981 \\
4.13218493174874	0.206943777621374 \\
4.1384744065307	0.219501890584969 \\
4.14476388131266	0.232140862536971 \\
4.15105335609462	0.244880263200978 \\
4.15734283087658	0.2577195886399 \\
4.16363230565855	0.270658330963895 \\
4.16992178044051	0.286429061491784 \\
4.17621125522247	0.315473638467144 \\
4.18250073000443	0.344554911999117 \\
4.18879020478639	0.375453345020728 \\
4.19507967956835	0.41527413408401 \\
4.20136915435031	0.455531725814597 \\
4.20765862913228	0.497744498979337 \\
4.21394810391424	0.542079117376113 \\
4.2202375786962	0.587838175464458 \\
4.22652705347816	0.64029168773063 \\
4.23281652826012	0.689610943390242 \\
4.23910600304208	0.728051313380607 \\
4.24539547782404	0.766447304730269 \\
4.25168495260601	0.80479739859508 \\
4.25797442738797	0.843137250043782 \\
4.26426390216993	0.881447699411901 \\
4.27055337695189	0.919707790233513 \\
4.27684285173385	0.988220071982741 \\
4.28313232651581	1.06991069226934 \\
4.28942180129778	1.2398308003209 \\
4.29571127607974	1.42032785799606 \\
4.3020007508617	1.60045275742497 \\
4.30829022564366	1.78019837334163 \\
4.31457970042562	1.95955759548362 \\
4.32086917520758	2.13989935565795 \\
4.32715864998954	2.32545013090216 \\
4.33344812477151	2.4517569272721 \\
4.33973759955347	2.55201428108776 \\
4.34602707433543	2.65167281330134 \\
4.35231654911739	2.75072858168417 \\
4.35860602389935	2.8332382762937 \\
4.36489549868131	2.88965878070381 \\
4.37118497346328	2.94401697722175 \\
4.37747444824524	2.9803525820368 \\
4.3837639230272	2.9890434366833 \\
4.39005339780916	2.98612490001983 \\
4.39634287259112	2.96586926379332 \\
4.40263234737308	2.94515226566924 \\
4.40892182215504	2.92389477171557 \\
4.41521129693701	2.90209762282271 \\
4.42150077171897	2.87999655508703 \\
4.42779024650093	2.83239435191564 \\
4.43407972128289	2.77107438853344 \\
4.44036919606485	2.70919327043287 \\
4.44665867084681	2.64675344546778 \\
4.45294814562878	2.58375738359297 \\
4.45923762041074	2.51901412408179 \\
4.4655270951927	2.45129219259157 \\
4.47181656997466	2.38231743005687 \\
4.47810604475662	2.309851848652 \\
4.48439551953858	2.23283083220947 \\
4.49068499432054	2.15521407117518 \\
4.49697446910251	2.07700463586345 \\
4.50326394388447	1.99820562003329 \\
4.50955341866643	1.91882014076595 \\
4.51584289344839	1.83885133834164 \\
4.52213236823035	1.75830237611522 \\
4.52842184301231	1.67730557564857 \\
4.53471131779428	1.59582775104148 \\
4.54100079257624	1.51377976304315 \\
4.5472902673582	1.4311648572556 \\
4.55357974214016	1.34793266536941 \\
4.55986921692212	1.2875305329759 \\
4.56615869170408	1.24327978933748 \\
4.57244816648605	1.20214745640342 \\
4.57873764126801	1.16463501634475 \\
4.58502711604997	1.1456782580774 \\
4.59131659083193	1.12400491229126 \\
4.59760606561389	1.10226636408443 \\
4.60389554039585	1.08046347337653 \\
4.61018501517781	1.05859710263242 \\
4.61647448995978	1.03666811682806 \\
4.62276396474174	1.01467738341629 \\
4.6290534395237	0.992430187583861 \\
4.63534291430566	0.969677785891527 \\
4.64163238908762	0.946866000238118 \\
4.64792186386958	0.923987193253259 \\
4.65421133865155	0.90414130287703 \\
4.66050081343351	0.882654431425897 \\
4.66679028821547	0.86115997145236 \\
4.67307976299743	0.843252485003353 \\
4.67936923777939	0.822825459315387 \\
4.68565871256135	0.801634827595059 \\
4.69194818734332	0.7807544297155 \\
4.69823766212528	0.759883258934339 \\
4.70452713690724	0.75123007528328 \\
4.7108166116892	0.744366344467911 \\
4.71710608647116	0.737644407108325 \\
4.72339556125312	0.730972506500808 \\
4.72968503603508	0.722502749898565 \\
4.73597451081705	0.71384062265384 \\
4.74226398559901	0.705281730584741 \\
4.74855346038097	0.696787472700731 \\
4.75484293516293	0.68834733529294 \\
4.76113240994489	0.683352814436655 \\
4.76742188472685	0.68013044071713 \\
4.77371135950881	0.676530503013017 \\
4.78000083429078	0.672849477962157 \\
4.78629030907274	0.669242210825333 \\
4.7925797838547	0.665708844296513 \\
4.79886925863666	0.662249518146365 \\
4.80515873341862	0.65908228832789 \\
4.81144820820058	0.657579587256203 \\
4.81773768298255	0.653792786122825 \\
4.82402715776451	0.646194626840104 \\
4.83031663254647	0.636892769663648 \\
4.83660610732843	0.627089513330887 \\
4.84289558211039	0.616626258353095 \\
4.84918505689235	0.606245880918741 \\
4.85547453167432	0.595948791648185 \\
4.86176400645628	0.585735397867119 \\
4.86805348123824	0.57560610359047 \\
4.8743429560202	0.565188740103749 \\
4.88063243080216	0.55437901127107 \\
4.88692190558412	0.543652890740805 \\
4.89321138036608	0.53301080280999 \\
4.89950085514805	0.520857191675637 \\
4.90579032993001	0.509693540023875 \\
4.91207980471197	0.498769167690167 \\
4.91836927949393	0.487939934289714 \\
4.92465875427589	0.477190847922417 \\
4.93094822905785	0.465583014862862 \\
4.93723770383982	0.454103134024639 \\
4.94352717862178	0.442751659521557 \\
4.94981665340374	0.431529040388003 \\
4.9561061281857	0.419114528300208 \\
4.96239560296766	0.405101308291794 \\
4.96868507774962	0.391228799837403 \\
4.97497455253158	0.377497551696889 \\
4.98126402731355	0.363908107042214 \\
4.98755350209551	0.350461003435966 \\
4.99384297687747	0.339256924892167 \\
5.00013245165943	0.330496147021198 \\
5.00642192644139	0.321867076952091 \\
5.01271140122335	0.313933221031456 \\
5.01900087600531	0.306323256651035 \\
5.02529035078728	0.298821932976168 \\
5.03157982556924	0.289778319219271 \\
5.0378693003512	0.279033179751686 \\
5.04415877513316	0.268317400884704 \\
5.05044824991512	0.257711855173582 \\
5.05673772469708	0.247216962145729 \\
5.06302719947905	0.236825513759699 \\
5.06931667426101	0.226522285296029 \\
5.07560614904297	0.216330996774453 \\
5.08189562382493	0.206252051335475 \\
5.08818509860689	0.196291902805695 \\
5.09447457338885	0.186838996880404 \\
5.10076404817082	0.177500102047752 \\
5.10705352295278	0.168275587729797 \\
5.11334299773474	0.159165818823993 \\
5.1196324725167	0.150171155688771 \\
5.12592194729866	0.141291954129284 \\
5.13221142208062	0.132333659126453 \\
5.13850089686258	0.123151934989159 \\
5.14479037164455	0.114089704013459 \\
5.15107984642651	0.105100434914911 \\
5.15736932120847	0.0972975889681491 \\
5.16365879599043	0.0899933789943415 \\
5.16994827077239	0.0828285299142397 \\
5.17623774555435	0.0758033251503765 \\
5.18252722033631	0.0689180426013265 \\
5.18881669511828	0.0621729546307046 \\
5.19510616990024	0.0555683280564052 \\
5.2013956446822	0.0491044241400385 \\
5.20768511946416	0.0454722731593291 \\
5.21397459424612	0.0428963898585355 \\
5.22026406902808	0.0405509908043342 \\
5.22655354381005	0.0383397966258561 \\
5.23284301859201	0.0362628947921118 \\
5.23913249337397	0.0343203674598658 \\
5.24542196815593	0.0325122914703653 \\
5.25171144293789	0.0308387383463375 \\
5.25800091771985	0.0292997742891288 \\
5.26429039250182	0.0278954601760932 \\
5.27057986728378	0.0266258515581961 \\
5.27686934206574	0.0255574143686168 \\
5.2831588168477	0.0256218980890552 \\
5.28944829162966	0.0258211768828125 \\
5.29573776641162	0.0261552428669405 \\
5.30202724119358	0.0266240828266739 \\
5.30831671597555	0.0272276782159375 \\
5.31460619075751	0.0279660051580932 \\
5.32089566553947	0.0288390344468743 \\
5.32718514032143	0.0298467315475373 \\
5.33347461510339	0.0309890565982509 \\
5.33976408988535	0.0322659644116476 \\
5.34605356466732	0.0336774044766188 \\
5.35234303944928	0.0352233209603199 \\
5.35863251423124	0.0369036527103721 \\
5.3649219890132	0.0384793812127469 \\
5.37121146379516	0.0400536425585529 \\
5.37750093857712	0.0417477039563532 \\
5.38379041335909	0.0430914693620972 \\
5.39007988814105	0.0402627714987882 \\
5.39636936292301	0.035257331871295 \\
5.40265883770497	0.0303922033929669 \\
5.40894831248693	0.0256675785154554 \\
5.41523778726889	0.0204043375131469 \\
5.42152726205085	0.0136601897991344 \\
5.42781673683282	0.00705624111866676 \\
5.43410621161478	0.00079366134984804 \\
5.44039568639674	0.00895483252984342 \\
5.4466851611787	0.0169748362626194 \\
5.45297463596066	0.024853355297989 \\
5.45926411074262	0.0325900779825132 \\
5.46555358552459	0.0401846982718479 \\
5.47184306030655	0.0477340990495945 \\
5.47813253508851	0.0558510629294631 \\
5.48442200987047	0.0638260870068357 \\
5.49071148465243	0.0716588558107873 \\
5.49700095943439	0.0793490594976438 \\
5.50329043421635	0.0887215823977301 \\
5.50957990899832	0.0989307358029361 \\
5.51586938378028	0.109367934810982 \\
5.52215885856224	0.119729523513539 \\
5.5284483333442	0.129930668391783 \\
5.53473780812616	0.14012181219063 \\
5.54102728290812	0.150238047071431 \\
5.54731675769009	0.160192859037921 \\
5.55360623247205	0.170174590320119 \\
5.55989570725401	0.180179631218595 \\
5.56618518203597	0.190021590022765 \\
5.57247465681793	0.199700077410686 \\
5.57876413159989	0.209214710526934 \\
5.58505360638185	0.218565112997713 \\
5.59134308116382	0.23039248110244 \\
5.59763255594578	0.242473119421173 \\
5.60392203072774	0.254585763352311 \\
5.6102115055097	0.268020431597675 \\
5.61650098029166	0.280678267839601 \\
5.62279045507362	0.293149322852675 \\
5.62907992985559	0.305433103314847 \\
5.63536940463755	0.319995621145637 \\
5.64165887941951	0.335774154214221 \\
5.64794835420147	0.363486969125278 \\
5.65423782898343	0.390885135853221 \\
5.66052730376539	0.417967570598827 \\
5.66681677854735	0.444733202052392 \\
5.67310625332932	0.473025925700279 \\
5.67939572811128	0.503120426398656 \\
5.68568520289324	0.532853348589456 \\
5.6919746776752	0.562223516116668 \\
5.69826415245716	0.591229767173941 \\
5.70455362723912	0.619870954350483 \\
5.71084310202109	0.648145944676468 \\
5.71713257680305	0.676053619667845 \\
5.72342205158501	0.703592875370594 \\
5.72971152636697	0.730762622404388 \\
5.73600100114893	0.757561786005704 \\
5.74229047593089	0.783989306070296 \\
5.74857995071285	0.810044137195175 \\
5.75486942549482	0.837002920044743 \\
5.76115890027678	0.864186959404675 \\
5.76744837505874	0.890972565535456 \\
5.7737378498407	0.917358678869152 \\
5.78002732462266	0.943344255640689 \\
5.78631679940462	0.968928267929205 \\
5.79260627418659	0.991968820838731 \\
5.79889574896855	1.0087613821844 \\
5.80518522375051	1.02515097419581 \\
5.81147469853247	1.04113694854392 \\
5.81776417331443	1.05671867286575 \\
5.82405364809639	1.07189553078938 \\
5.83034312287835	1.08666692195834 \\
5.83663259766032	1.10103226205536 \\
5.84292207244228	1.11499098282545 \\
5.84921154722424	1.12854253209845 \\
5.8555010220062	1.14168637381081 \\
5.86179049678816	1.15442198802681 \\
5.86807997157012	1.16674887095915 \\
5.87436944635209	1.17866653498883 \\
5.88065892113405	1.19017450868452 \\
5.88694839591601	1.20127233682111 \\
5.89323787069797	1.19923949618498 \\
5.89952734547993	1.19465220522178 \\
5.90581682026189	1.18959959409389 \\
5.91210629504385	1.1840818626693 \\
5.91839576982582	1.18104672199209 \\
5.92468524460778	1.17776203678545 \\
5.93097471938974	1.17399966284188 \\
5.9372641941717	1.16975974899097 \\
5.94355366895366	1.16504246295254 \\
5.94984314373562	1.15984799132996 \\
5.95613261851759	1.15417653960285 \\
5.96242209329955	1.14745583128058 \\
5.96871156808151	1.13987211377297 \\
5.97500104286347	1.131811614062 \\
5.98129051764543	1.12327465099977 \\
5.98757999242739	1.11426156228603 \\
5.99386946720935	1.10477270445481 \\
6.00015894199132	1.09480845286029 \\
6.00644841677328	1.08324421582189 \\
6.01273789155524	1.06429931748095 \\
6.0190273663372	1.04486882142895 \\
6.02531684111916	1.02495349628506 \\
6.03160631590112	1.00455412984702 \\
6.03789579068309	0.983671529059961 \\
6.04418526546505	0.962306519984497 \\
6.05047474024701	0.940897647396571 \\
6.05676421502897	0.919695745694812 \\
6.06305368981093	0.898354273568563 \\
6.06934316459289	0.882563245151915 \\
6.07563263937485	0.866338740814356 \\
6.08192211415682	0.849699746151439 \\
6.08821158893878	0.833125384673661 \\
6.09450106372074	0.816117481304435 \\
6.1007905385027	0.798676708831575 \\
6.10708001328466	0.780803757166044 \\
6.11336948806662	0.762499333314659 \\
6.11965896284859	0.743764161352158 \\
6.12594843763055	0.724598982392534 \\
6.13223791241251	0.703890972686735 \\
6.13852738719447	0.663590765932215 \\
6.14481686197643	0.622829746741761 \\
6.15110633675839	0.581802359318547 \\
6.15739581154036	0.540667100748614 \\
6.16368528632232	0.499081494864981 \\
6.16997476110428	0.45704718668455 \\
6.17626423588624	0.419795142030253 \\
6.1825537106682	0.375364037670165 \\
6.18884318545016	0.315788405885685 \\
6.19513266023212	0.255783595432858 \\
6.20142213501409	0.19535197994372 \\
6.20771160979605	0.134495949933624 \\
6.21400108457801	0.103191474639732 \\
6.22029055935997	0.0856039442234837 \\
6.22658003414193	0.0718284780844707 \\
6.23286950892389	0.062302607466012 \\
6.23915898370586	0.0725277534837989 \\
6.24544845848782	0.0896856973301685 \\
6.25173793326978	0.113199110955867 \\
6.25802740805174	0.140734067293149 \\
6.2643168828337	0.167829128795475 \\
6.27060635761566	0.194483223653675 \\
6.27689583239763	0.222311375566591 \\
6.28318530717959	0.25860654097275 \\
};

\end{axis}

\end{tikzpicture}

%% file: figures/stock/91/x0.tex
\begin{tikzpicture}

\definecolor{color0}{rgb}{0.12156862745098,0.466666666666667,0.705882352941177}

\begin{axis}[
tick align=outside,
tick pos=left,
x grid style={white!69.01960784313725!black},
xlabel={EOG},
xmin=-60, xmax=60,
y grid style={white!69.01960784313725!black},
ylabel={MSFT},
ymin=-60, ymax=60
]
\addplot [only marks, draw=color0, fill=color0, colormap/viridis]
table [row sep=\\]{%
x                      y\\ 
+2.827590556994081e+00 +1.794429720381066e+00\\ 
-2.083772105839302e+01 -9.647564428713348e+00\\ 
-1.440012457008313e+01 -1.818236841243388e+01\\ 
+6.701079060020265e-02 +1.049696787932699e+00\\ 
-1.471204720323741e+01 -7.681295435929079e-01\\ 
-6.315782364713557e+00 +4.086575115174787e+00\\ 
+1.074821894910480e+01 -1.139928743534286e+01\\ 
+8.030665256080843e+00 +1.355292934096391e+01\\ 
-1.825150877755391e+01 -2.084925457059679e+01\\ 
-1.644731491556264e+01 -4.715797501203609e+00\\ 
-6.495753479527605e+00 +1.787962600294282e+00\\ 
+2.792522582815169e+01 -3.542538177065355e+00\\ 
+1.710319674796843e+01 +1.713256009656343e+01\\ 
-3.512017780833857e+01 -5.285438769095771e+00\\ 
+3.441375780847108e+01 +3.173863412272595e+00\\ 
-8.827352763843277e-02 -9.670173018427272e+00\\ 
+1.861641099407915e+01 +7.603980690693948e+00\\ 
+1.672481085190888e+01 +2.785227489197044e+01\\ 
-1.416683695241242e+01 -3.942256591416735e+00\\ 
-1.455715959097484e+01 -1.635870214877617e+01\\ 
+2.167293482120256e+01 -8.469402853839966e+00\\ 
-1.301527530445881e+01 -2.017504405459400e+00\\ 
-1.057592843824498e+01 -1.849431601535174e+01\\ 
+9.720717517865934e+00 -8.013936620485504e+00\\ 
-2.158511427467660e+01 -1.798661703172730e+00\\ 
+1.131228450844207e+00 +3.862495553692049e+00\\ 
-2.142123056432049e+00 -6.826121331531907e-01\\ 
+1.566913857600724e+01 +7.603410446562034e+00\\ 
+1.010209752367625e+00 +8.837158112399537e+00\\ 
+2.073527473981304e+01 +1.236830238930036e+01\\ 
-1.008627412921007e+01 -2.680051006942972e+00\\ 
+1.405291645899726e+00 -4.038770147645707e+00\\ 
+1.722630603709760e+01 +7.463626836035900e+00\\ 
-3.027945488731001e+01 -1.464010881114637e+01\\ 
+1.051527624374568e+01 +1.274009438049164e+00\\ 
-7.781442289769097e+00 +6.671239801837306e+00\\ 
-7.744148140136603e+00 -8.218503178271616e+00\\ 
-2.126524913548769e+01 -4.591821459401910e+00\\ 
+1.766758743536573e+00 +1.595154524795943e+01\\ 
+1.460777781230879e+01 +3.024721485857798e+00\\ 
+1.612418839295168e+01 -6.179472252400739e+00\\ 
+1.496631080149433e+01 -3.547135894146856e+00\\ 
+2.725758092258587e+01 -1.018479500013576e+01\\ 
-3.402698174441568e+01 +5.556845357904841e+00\\ 
+1.151258256518670e+01 +1.090773715726894e+01\\ 
-3.218846575936685e-01 -8.013247183548852e+00\\ 
+1.934699197528087e+01 +9.222096002438221e+00\\ 
-3.058458738826958e+00 +4.598604512406259e-01\\ 
-8.275866184468212e+00 +3.452673108280171e+00\\ 
+9.229134868387250e+00 +6.559660143263748e+00\\ 
+9.243643621966251e+00 +2.362378579375093e+00\\ 
-6.109929927566676e+00 -1.130013486535659e+01\\ 
+1.785853592777470e+00 +2.965278533174567e+00\\ 
-7.861752165417013e-01 +1.464303118250345e+00\\ 
-1.077483421153216e+01 -1.406988268659213e+00\\ 
-6.601810179856227e-01 +1.737123433879375e+00\\ 
+1.154314139652807e+00 -6.699582476997376e+00\\ 
+5.743153714831556e+00 +1.032733010478964e+01\\ 
-1.239621154376305e+01 +2.616272659502641e+00\\ 
-7.776329061287140e+00 +1.150805485231101e+00\\ 
-4.125015183879005e+00 +2.587196802392435e+00\\ 
-1.254481649363085e+01 -1.742666817035785e+00\\ 
+1.241869853126437e+00 -8.391379835227816e+00\\ 
+4.405802075564597e+00 +4.624401698125327e+00\\ 
+1.221038817830616e+00 -6.504474616874376e+00\\ 
+1.354855318286471e+01 -8.487818322307581e-01\\ 
+3.270455343303808e+00 -1.493085541148640e+00\\ 
+2.271229609672584e+01 +2.639016400922169e+00\\ 
+2.423206982271359e+00 +5.882317379079119e+00\\ 
-1.982599946706298e-01 -3.910788095931682e-01\\ 
-1.407092720210759e+01 +2.130978319098300e+00\\ 
+1.090552098126604e+01 +6.743767664916181e+00\\ 
+1.040323888109881e+01 -1.101697390102416e+00\\ 
+1.123122295546536e+00 -7.625659313829367e+00\\ 
-1.715745629503634e+00 +1.224621730007527e+00\\ 
+1.060387133729909e+01 -3.768710882495181e+01\\ 
-7.610517768273067e+00 +2.695042411893824e+00\\ 
+1.819110588820758e+01 -6.951799382857013e+00\\ 
+9.950617672608713e+00 -5.365210075998725e+00\\ 
-1.613756808243696e+01 -1.079513828388807e+01\\ 
+4.046374897970412e+00 -7.820964866978672e-01\\ 
+5.762777306881073e-01 +6.883377947007377e+00\\ 
-1.354892325712179e+01 -8.749349030245110e+00\\ 
-8.455261250665453e+00 +4.217564223138372e-01\\ 
+1.156466329141034e+01 +2.199277598250917e-01\\ 
-1.652352296465345e+01 +4.003824033465822e+00\\ 
-6.453332862699501e-01 -3.666763062306686e+00\\ 
+6.491650892453488e+00 +8.916438298929899e+00\\ 
+3.295290269216017e+00 -1.874889384679366e-01\\ 
+7.281897952553240e-01 +4.003804596433172e+00\\ 
-1.201585190540547e+01 -4.672871093548256e+00\\ 
+1.241243673276912e+01 +6.806645631665045e+00\\ 
+5.032373280748209e+00 -9.829291570451268e+00\\ 
-7.552574821507979e+00 +2.479519675864175e+00\\ 
-2.447280221839404e+00 -5.326691493694533e+00\\ 
+4.826364200997552e+00 +2.490666475420891e+00\\ 
+1.766891787488428e+00 -6.343369695405229e+00\\ 
+7.058990108476077e+00 +1.487110999124124e+01\\ 
+3.172388122338194e+00 +4.629043613069086e+00\\ 
-5.107520107736879e+00 -2.692734325474408e+00\\ 
-4.845854630555452e+00 +3.644901997677354e+00\\ 
-1.679972231386523e+00 +5.975198314610289e+00\\ 
-4.153969264064444e+00 -1.898505504907064e+00\\ 
-2.003578943854195e+00 -3.994188727177781e+00\\ 
-5.260026778191734e+00 -7.098936986996327e+00\\ 
+9.299873717343059e+00 +2.790354285130350e+00\\ 
+2.451058587881214e+01 -7.692298973203757e-01\\ 
-7.424036552030888e-01 -1.057552444480756e+00\\ 
-4.980961919870436e+00 -4.533134563897764e+00\\ 
-1.857051761433517e+01 -1.839310668240455e+00\\ 
-1.434339628942266e+00 -1.366854988586559e+01\\ 
+9.282525337591190e+00 -3.582345105390525e+00\\ 
-3.655503886526273e+00 -1.888158226521839e+00\\ 
-1.428939884924121e+01 +6.513113652169158e+00\\ 
+4.016975462457689e+00 -3.067960595522676e+00\\ 
+1.362739618767891e+01 -1.080207353751758e+00\\ 
+9.841287412443139e+00 +1.057968217684418e+01\\ 
-8.165250405602260e+00 -2.438737588191195e+00\\ 
+4.962932690943251e+00 +8.459552499058878e+00\\ 
-2.117203102677070e+01 -2.092852365988180e+01\\ 
-1.586365260633318e+01 -1.473028282606166e+01\\ 
+1.931959800434655e+01 +9.838773480177634e+00\\ 
+9.839926012285643e+00 +1.052123951833200e+01\\ 
-5.109354071531069e-01 +5.712755541498796e+00\\ 
+7.694714132553599e+00 -5.791279401535319e-01\\ 
-1.244848096886855e+01 -3.836818337957325e-01\\ 
-3.910744452924331e-01 +1.566379545396337e+00\\ 
-7.387599143386297e+00 -4.814048869746322e-01\\ 
+5.274535976351941e+00 +8.392284807591846e+00\\ 
+1.784944463111698e+00 +2.283403378212803e+00\\ 
+1.293987468409005e+01 +5.378775616535483e+00\\ 
-7.086514359362167e+00 +2.329696970492392e+00\\ 
+4.298222109343913e+00 +1.663120545092925e+00\\ 
-3.532197796849240e+00 -8.537057232363507e-01\\ 
-5.470293504466622e+00 +1.933610025394097e+00\\ 
-7.711672273477160e+00 -8.608627258960786e+00\\ 
+3.536481296448346e-02 +2.539592933451045e+01\\ 
-4.180823639575753e+00 -1.466097712624690e+00\\ 
-3.530125809349888e+00 +6.371065096926801e+00\\ 
-1.435304089442926e+01 +9.307761047351784e-01\\ 
+1.028772342103831e+01 -2.170644102926310e-01\\ 
-4.572587925862043e+00 -5.527928470841667e+00\\ 
+3.604287021954089e-01 -3.034689420367641e-01\\ 
+1.074800010156915e+01 +3.681964017276209e+00\\ 
-1.909649766659463e+01 -1.364329361509606e+00\\ 
+7.112692090009282e+00 -4.814048869746322e-01\\ 
+1.489422778430580e+01 +2.953218922472070e+00\\ 
+1.083321299099069e+01 +3.191224427003971e+00\\ 
+3.356692650760770e+01 +4.460117779130071e+00\\ 
+2.374113445300114e+00 +3.805156850324367e-01\\ 
-1.937695210077611e+00 +7.227931834751702e-01\\ 
-5.730391033944820e+00 -2.030192909036119e+00\\ 
+4.394716126462778e+00 +1.925752444619613e+00\\ 
+4.678398120181643e+00 -3.578455394601952e+00\\ 
-3.967430820728063e+00 +1.069518293467178e+00\\ 
-1.994969410575892e+00 -3.241936398848131e+00\\ 
-6.127056545943976e-01 +5.620737526258937e-01\\ 
+7.908099325368151e+00 -1.340620456378814e-01\\ 
-2.275777955172142e-01 -3.078239097121325e-01\\ 
-3.616363057443977e+00 -4.771590150138601e-02\\ 
-1.717521476656547e-01 +1.422370344821003e+00\\ 
-3.029312950558332e+00 +3.655096045001265e-02\\ 
-3.043307429567737e+00 +1.413189181479415e+00\\ 
-7.790365553173306e-01 -1.686224747046891e+00\\ 
+4.903481976109986e+00 +1.213679406976610e-01\\ 
-6.295894465335988e+00 -2.291907770453359e+00\\ 
-9.239871703490971e+00 -4.209207156861636e+00\\ 
-5.216797050147659e-02 +6.485390967513796e-01\\ 
+1.637554506814104e+00 +2.126781673907254e-01\\ 
+3.183952193600101e+01 -1.001876828490131e+00\\ 
-3.087353778553653e+00 -4.764065398872071e-02\\ 
+5.458794654731508e+00 -2.479843554006165e+00\\ 
-1.517309529419000e+01 -1.121747381808531e+01\\ 
+1.386657320241652e+00 +6.935294760077307e+00\\ 
-1.446114887744512e+01 -5.059707697411772e+00\\ 
-1.129449784146111e+01 -2.875238699722014e+00\\ 
+1.354516773387646e+01 +7.716222545784460e+00\\ 
-2.811360103731897e-01 +4.288736658846452e-02\\ 
-5.473429551293735e+00 -3.284004622661050e+00\\ 
+1.548523442789412e+00 -1.536443083944248e+00\\ 
+1.657445043754777e+01 +7.810659120789312e+00\\ 
-5.280169436108900e+00 +3.771611484076025e-02\\ 
-1.607562435218604e+01 -3.865365742976865e+00\\ 
-3.128358528925706e+00 -5.117141758934915e+00\\ 
-4.327659845666770e+00 +8.661209439767051e+00\\ 
+2.920336124987719e+01 +2.083693214071560e-01\\ 
+5.375880126565952e+00 -5.939312325480126e+00\\ 
+8.265795951924753e+00 +1.257727301193048e+00\\ 
-3.502794892131150e+00 -2.046351391440668e+00\\ 
-5.463013360818697e+00 -2.051264889310576e+00\\ 
+1.124460542991754e+01 +3.000503334562745e+00\\ 
-1.624881363771175e+00 +3.852966647574420e-01\\ 
-4.007166247609356e+00 +3.789583479745466e-02\\ 
+3.996503391890477e+00 +1.590421281855983e+00\\ 
-6.804734331675971e+00 -7.858089931354348e+00\\ 
+9.086403494655559e-01 -1.181317512205135e+00\\ 
-3.685914919852068e+00 -2.147634614533998e+00\\ 
-5.760112244076493e+00 +3.891545848658742e+00\\ 
-4.109353559129475e+00 -2.225998422447815e+00\\ 
-3.868698009227317e+00 +3.348697407170744e+00\\ 
+9.864877880853856e+00 -1.609975554500329e+00\\ 
-1.180672967024315e+00 -2.920859107401539e+00\\ 
-5.846926025079996e+00 +2.013566343915112e+01\\ 
-1.777742000740795e+00 +1.062463765638116e+01\\ 
-5.490694581954419e+00 -5.633788194535170e-01\\ 
-3.258892490708459e+00 -3.441452999333026e+00\\ 
-2.566409614661397e+00 -4.871394158213979e+00\\ 
-5.540676226183177e+00 -2.398553188930850e+00\\ 
-2.487976315687562e+00 -6.400775111920816e-02\\ 
+3.684286810220455e+00 -1.483744013165042e+00\\ 
-1.599202435604213e+00 -3.584660675984461e+00\\ 
+5.837833563091283e+00 -2.335756304416690e+00\\ 
-1.913807838883647e+00 -4.721593120770093e+00\\ 
+8.409023143918864e+00 +1.387364805424401e+01\\ 
-7.658268585755412e-01 -6.780572367343958e-02\\ 
-1.632203200932492e-02 -2.968147413408515e+00\\ 
-4.690789214760543e+00 -1.284848608607964e+01\\ 
-7.020165245393422e+00 +2.236916453395013e+00\\ 
+6.507834503928417e+00 -8.164670293332518e+00\\ 
+9.702726286331147e+00 +9.230946834698875e+00\\ 
+3.033776924186799e+00 +6.099857864903127e+00\\ 
-1.574736778906008e+01 +7.748891508124236e+00\\ 
+3.975765539087045e+00 -2.878301782270951e+00\\ 
+1.295576187594413e+01 +3.726193575671990e+00\\ 
-2.485976430988385e+00 +1.650095307748225e+00\\ 
+8.606931947057388e+00 -6.406424845729239e+00\\ 
-5.676645481935055e+00 +5.935975944441532e-01\\ 
-1.364072806309594e+01 +1.789881428215147e-01\\ 
-5.931446339446894e+00 +3.462740462265292e+00\\ 
+5.135798359956853e+01 -7.321231366735523e+00\\ 
+1.036825420787110e+01 -9.354900793070728e+00\\ 
-5.235469384762742e+00 -5.928582433407421e-02\\ 
+3.067152597156148e+00 +7.637966632198781e+00\\ 
-2.741987954462371e+00 -2.728225677003435e+00\\ 
-2.511559551215098e+00 +1.122371257676018e+01\\ 
+6.324943250631793e+00 -3.423070505592050e+00\\ 
+6.799890639535206e+00 +7.324901018151023e+00\\ 
-3.311753562237812e+00 +1.129680766641153e+00\\ 
+9.803816717114360e+00 +5.990918808838666e+00\\ 
-1.829893515832229e+01 -2.868803979496912e+00\\ 
-1.784435266311030e+00 -1.279744510517331e+00\\ 
-6.311055783553288e-03 -2.723561110644932e+00\\ 
-2.897345357053354e+00 +1.000182537682208e+01\\ 
-1.939662597612718e+00 -1.110533626785879e+00\\ 
-3.356422692670585e+00 -4.814048869746322e-01\\ 
+2.428847459498424e+00 -4.027204899264645e-01\\ 
-5.752794765708807e+00 -2.926397534073486e+00\\ 
+2.101964888897995e+00 -1.652493279127952e-01\\ 
-4.749787585897974e+00 -2.778074799754402e+00\\ 
-2.214365284554792e+00 -1.196314850694888e+00\\ 
-1.280490629987502e+00 -6.559534757718597e+00\\ 
+1.153286827837549e+01 +3.046512250943036e+00\\ 
-4.899488854431810e-03 -2.723561110644932e+00\\ 
+2.303831019919239e+00 -4.814048869746322e-01\\ 
+4.528015806291669e+00 +3.833788910032767e+00\\ 
-1.090867878667304e+01 -2.075285739412023e+00\\ 
+1.449834523138377e+00 -6.410727791140245e-01\\ 
+6.905522016873195e+00 +4.049267556575467e+00\\ 
+2.545386873820305e+00 -5.091930401046449e+00\\ 
-2.961797781016184e-01 +2.368140538521659e-01\\ 
+3.572882433708346e-02 -1.838907916804919e+00\\ 
-7.704371341968053e-01 -7.213473054051376e-01\\ 
-1.388273123353018e+00 -2.083970361449536e+00\\ 
+9.380823233737512e-01 +3.037483597587867e+00\\ 
-1.001181830183320e+01 +1.268795786337476e+00\\ 
+6.710095689956563e-01 +3.946201527887874e+00\\ 
+6.687583053064106e-01 +7.764574116514902e-01\\ 
+1.961419212238108e+00 +4.129804036559984e+00\\ 
-7.579195735433669e+00 +1.113003580188392e+01\\ 
-8.720517438575632e+00 -5.446688988277890e+00\\ 
+1.482879814548282e+00 -4.179988557195655e+00\\ 
-3.254148711190369e+00 -8.825979754579391e+00\\ 
-6.818346445577800e+00 -3.716130173002272e+00\\ 
+5.739079590415058e+00 +3.539113425788878e+00\\ 
-6.571376417457114e+00 -7.955739202441612e-01\\ 
-1.314000017149766e+01 -2.134038798867375e+00\\ 
+4.472528127895642e+00 -1.191352154629241e+00\\ 
+8.045599573075906e+00 +5.170146648283128e+00\\ 
+7.708481549907304e+00 -9.499352976465956e-01\\ 
-5.792782168563964e+00 +5.112189808313675e+00\\ 
+3.974465738878691e+00 +1.369306664618255e+00\\ 
-2.767551837226795e+00 -7.912422322891663e-01\\ 
-3.277803546865049e+00 -5.588942323487365e-01\\ 
-6.114028490523698e+00 +2.929486826991813e-01\\ 
+1.825553870262925e+00 -1.488298574005962e+00\\ 
-1.023910853549540e+01 -1.490330336024590e+00\\ 
-1.860513773242201e-01 +1.534414249106655e+00\\ 
-6.941949603182423e+00 -4.814048869746322e-01\\ 
+7.899110835660944e+00 -3.508189450302910e+00\\ 
-5.111478925557106e+00 -2.431333261139488e+00\\ 
+1.514001586685129e+01 +6.965211189690004e+00\\ 
-8.350138330308427e+00 -7.693627647364762e+00\\ 
+1.164367635279884e-01 +1.389796423730319e+00\\ 
+7.160418267214669e+00 -3.257870830418262e-01\\ 
-3.096491867490752e+00 +5.289319765801517e-01\\ 
-2.457772012646078e+01 +4.078484511006389e+00\\ 
+4.175167336346347e+00 -2.485724561398348e+00\\ 
-5.168264442269504e+00 +1.061092094517038e+00\\ 
+4.521849513780818e-01 -2.178413612588701e+00\\ 
-2.290411961898081e+00 -2.804829142159468e+00\\ 
+1.249771643070094e+01 +2.150995123447232e+00\\ 
-6.767852255992218e+00 -1.331548070376451e+00\\ 
+2.729709638016970e+00 +1.294521619087321e+00\\ 
-1.481134570782389e+00 -1.915523925772640e-02\\ 
-9.251070362299691e+00 -6.056807244198056e+00\\ 
+2.432421637241500e-01 +5.863465150582828e+00\\ 
+4.534311919794642e-01 -1.713122215025573e+00\\ 
-1.274586855098125e-01 +3.657266210185718e-01\\ 
-8.253539898618351e+00 +4.411045998119946e-01\\ 
+9.732767463361130e+00 +9.757631050339357e-01\\ 
+8.501978126351089e+00 +8.951637402573789e-01\\ 
+4.844472286913148e-01 +1.348144341409776e+00\\ 
+1.560448163592215e+00 +6.586731322448161e-01\\ 
-3.424346819351137e+00 -2.840438256223108e+00\\ 
+6.217289768892548e+00 +8.897111459089027e-01\\ 
-3.812422626129824e+00 -1.776249086901613e+00\\ 
+2.011424682037716e+00 +8.134393129523489e-01\\ 
-3.922126426832123e+00 -8.618934526277242e-01\\ 
+5.784679053744774e+00 -1.624610933356705e+00\\ 
-3.045974944930557e+00 -8.630551612152735e-01\\ 
-2.387289411940579e+00 -2.394038234439686e+00\\ 
-9.970407470818705e+00 -2.632276232669215e+00\\ 
+1.314462418268991e-01 +3.582099806185004e+00\\ 
-5.166258979655308e+00 -1.169110351847033e+00\\ 
-8.577521406677919e+00 -3.164843053791009e+00\\ 
-3.374810050769395e-01 +3.042433813850545e+00\\ 
+1.107214347095126e+00 +6.342052033615866e+00\\ 
+4.630790829509290e-01 +7.956038665205936e+00\\ 
+7.756391830952420e+00 +2.397894321198348e+00\\ 
-5.507422690921332e+00 -1.144388409523975e+00\\ 
-7.384942146559298e+00 +2.751523349125031e+00\\ 
-3.086498762729733e+00 +9.081959489500215e-01\\ 
-2.938091279098637e+00 +6.409253237748372e+00\\ 
-1.261112728991681e+00 -1.274426464789988e+00\\ 
-9.901051764183155e-01 -2.071231345660627e+00\\ 
-1.791338205033391e+01 -2.439489919019719e+00\\ 
+1.231297423561203e+01 +8.973058044598383e-01\\ 
+4.045609099191355e+00 -9.163766412452843e-01\\ 
-6.638066835525092e-01 +2.433379562535777e-01\\ 
+1.518650944460351e+01 +1.470169690386205e+00\\ 
-1.768048093870239e+00 -6.651186300904063e+00\\ 
+1.831386616760716e+00 -1.066029332417904e+00\\ 
-2.327045975880904e-01 -1.159345103521424e-01\\ 
-5.164127238037858e+00 +9.429937514988820e+00\\ 
-5.102534512545081e+00 -1.458125072383682e+01\\ 
-7.215270498109910e+00 +1.219906022915820e+00\\ 
+1.103799459273055e+01 -6.291153754938803e-01\\ 
-4.136009564001314e+00 +5.101141720001470e+00\\ 
-3.515411265042872e+00 +1.195837260430979e+00\\ 
+3.648652235390404e-01 +1.733787642598707e-01\\ 
-9.373840950348795e+00 +5.660734638796841e+00\\ 
+7.063799022419548e-01 +2.025943300283541e+00\\ 
-9.238905408072624e+00 +2.652954999644732e+00\\ 
+3.443769007059422e+00 -4.545598386824176e+00\\ 
+1.069611224228939e-01 +1.376536367626071e+00\\ 
-4.389318762407140e+00 +1.122080696802435e+01\\ 
-3.074444034263629e+00 +3.128719099765911e+00\\ 
+8.663219518787651e+00 +1.176055194337769e+00\\ 
-1.536593378041509e+01 -1.378513677775592e+00\\ 
-6.180119279138030e-01 -3.564519931805352e+00\\ 
+1.066499081107983e+01 -1.190424730620594e+01\\ 
-1.502432982418062e+00 -4.335813696118970e+00\\ 
+3.377509123580341e+00 +5.713931798978214e+00\\ 
-9.472609966955565e+00 -3.177976850189010e+00\\ 
-1.113195459724928e+01 -3.121062869147659e+00\\ 
+8.039291592314173e+00 +2.333915178260228e-01\\ 
-2.939355628029078e+00 +5.694580499426552e+00\\ 
-1.641643537426046e+00 -7.300661036107778e+00\\ 
-4.563242449490606e+00 +2.086727453129702e+00\\ 
+2.869620137986501e+00 -5.525640701911725e-01\\ 
+1.787911281338761e-01 +6.233900691202601e+00\\ 
-2.674444094623834e+00 -5.278958273949483e+00\\ 
+3.086902974084288e+00 -9.927800070128898e+00\\ 
+5.371050109459236e+00 +3.762919570305436e+00\\ 
+5.083228925985867e+00 -9.954535166237852e+00\\ 
+2.489500772239611e+00 +2.720468956899749e+00\\ 
+9.724640474302076e+00 -6.024860648405747e+00\\ 
-1.172346953608264e+01 +6.148929064917199e+00\\ 
-9.724604281610622e+00 -4.186470878077062e+00\\ 
-1.464096825820854e+00 +5.966558122852009e+00\\ 
+6.236002239082356e+00 +3.247814924836214e+00\\ 
+5.212312909053302e+00 -4.099610060247037e-01\\ 
-7.222002852078913e-01 +7.737570030229948e+00\\ 
+7.322118370288261e+00 +3.856713566084904e+00\\ 
-6.441938141035775e-02 +6.505910484045410e+00\\ 
-2.278269061626592e-01 +3.419251315515728e+00\\ 
-1.480399494903112e+00 -8.223527351723467e-01\\ 
+1.364135020139764e+01 +3.323995378936286e+00\\ 
-3.849086845494814e+00 +2.084434639788293e+00\\ 
-4.411034996377766e+00 -3.521338590814447e+00\\ 
+1.756925851061214e+00 -1.770502554692863e+00\\ 
+7.899050634135402e+00 +3.510767346330093e+00\\ 
-1.023919166414783e+00 -1.425819630792644e+00\\ 
+6.057921788488158e+00 -6.527263482545129e+00\\ 
-3.316892616364503e+00 -1.302198504947791e+00\\ 
-1.860428987201318e+00 -2.676798095723696e+00\\ 
-3.977587713375002e+00 -1.444928929121689e+00\\ 
-1.184890529938423e+01 -2.690742919693909e+00\\ 
-1.926540456525741e+01 -1.243125136775111e+00\\ 
+8.251460215431743e+00 +3.178074267064337e+00\\ 
-4.721547883966420e+00 -2.411376955119791e+00\\ 
+7.794043635063836e+00 +2.204736982269020e+00\\ 
-4.798816488507801e+00 -2.684354891459970e+00\\ 
-7.244694874928542e+00 -7.848788667061799e+00\\ 
-3.712223723300295e+00 +7.092918262829587e+00\\ 
-3.450227428036162e+00 +6.979887651963937e+00\\ 
-5.933671627760956e+00 -3.455353209205845e-01\\ 
-1.225209574822220e+01 +2.446362721299324e+00\\ 
-9.894805562301185e+00 -9.040326645457350e+00\\ 
+3.848516250782391e+00 +1.397560735833438e-01\\ 
-2.775153345303122e+00 -2.832072933426419e+00\\ 
+2.884111445174503e+00 +6.469364264419289e+00\\ 
+4.746260386283313e+00 -3.497591713695967e+00\\ 
-3.278020543765941e+00 -6.877180733123132e-01\\ 
-1.041461813056338e+00 +4.120047383972449e-01\\ 
-7.001855041138843e+00 -4.127471457121089e-01\\ 
-1.892126308174175e+00 +1.026685096468574e+00\\ 
+2.726067888995071e+00 +6.046635619359078e+00\\ 
+7.435481656220960e+00 +4.626853014253188e+00\\ 
+3.476736684501630e+00 -6.062795638644156e+00\\ 
+8.915755118430180e-01 -2.717938605914691e+00\\ 
+1.100707052906077e+01 -1.909880112045690e+00\\ 
+3.859447043675440e+00 +5.881209737347673e+00\\ 
-1.076891306570607e+01 -2.908594119375449e+00\\ 
+6.191357065057551e+00 +4.762693204552074e+00\\ 
+2.670159606899175e+00 -1.016736843893648e+00\\ 
+1.425607194001729e+01 +3.054536195421667e+00\\ 
+3.327227905095071e+00 -3.415137800219468e+00\\ 
+7.392866543658644e+00 +3.116691663739670e+00\\ 
-1.303728373457408e+00 -1.478281670968704e+00\\ 
+3.861727736317877e-02 +1.377827002990431e+00\\ 
+5.214664695625898e+00 -3.806327366801929e+00\\ 
-1.933054717768973e+00 -5.375845578722791e+00\\ 
+3.111380971715344e+00 +8.643098265867424e-01\\ 
+8.726763628567030e+00 -8.269196916228802e+00\\ 
+2.611626647758686e+00 -4.814048869746322e-01\\ 
+1.926247793220454e+00 +3.529220926389678e+00\\ 
-2.966541147177787e+00 -3.460136053495353e-01\\ 
-8.043830036446320e-01 +3.697643436796285e+00\\ 
+1.877906346167107e+00 +3.234242606430706e-01\\ 
-3.902034761700951e+00 -2.832453754219542e+00\\ 
-1.272680232999229e+00 +2.405477923051949e+00\\ 
+7.495316072058944e-01 +8.014750686099406e+00\\ 
-2.258877800295239e+00 -2.839975013764849e-01\\ 
+1.217728644918842e+00 +1.422859020933467e+00\\ 
+1.057602064819835e+00 -4.814048869746322e-01\\ 
+3.832791243791600e-01 +3.698820713330854e-01\\ 
-1.837505870376593e+00 +4.077699254794334e+00\\ 
-5.462977093342591e-01 +1.911717849806640e+00\\ 
+3.516202214314229e+00 +5.499220224420200e-01\\ 
+8.418451150038198e-01 -8.679032313899191e-01\\ 
-4.460954213608731e+00 -3.525389068833127e-01\\ 
-5.068377028902199e+00 +1.447609851384415e+00\\ 
+5.216460964777542e+00 +5.261384124425719e+00\\ 
+2.796856780060419e-01 -3.545335338486864e-01\\ 
-2.924034114082180e+00 -2.911581981382503e-01\\ 
-1.948223009971689e+00 -1.941816110004721e+00\\ 
+3.441155604111596e+00 +3.445689967753098e-01\\ 
+7.541944210400525e+00 +3.059213428297196e+01\\ 
+4.962512053838103e+00 -4.362513281128534e-03\\ 
+1.314042553243161e+00 -4.731146158695696e+00\\ 
+1.158021366164405e+01 -4.814048869746322e-01\\ 
+2.824684057148835e+00 +4.721057882955007e+00\\ 
+7.080744520240962e+00 +5.370426690081380e-02\\ 
+1.124285624911866e+01 +1.475776901538486e+00\\ 
-4.474645606546340e+00 -1.666660864543987e+00\\ 
-6.603512690845720e+00 -1.550542856703372e+00\\ 
-1.346367055336889e+00 -1.791241311218281e+00\\ 
-9.651273212083755e-01 -1.238363377640915e-01\\ 
-1.000814954355558e+01 +2.329659929680063e-01\\ 
-5.836842844663598e+00 -4.363180131935993e+00\\ 
-4.588975778199722e-02 +8.424615742456589e-01\\ 
+6.430520867366230e-01 -5.312360342842945e+00\\ 
-1.375908326816139e+00 +3.068084576087117e-01\\ 
+6.921362196281311e-01 +6.676612099628601e+00\\ 
-7.354779029438430e-01 -4.137837893407001e+00\\ 
-4.667098764102318e-02 +4.202002139727569e-01\\ 
-9.523748433939245e+00 +3.168464727770706e+00\\ 
-2.909041554650937e-01 +5.503851767728987e+00\\ 
+2.903336993713896e+00 -9.636339042684414e+00\\ 
+8.582640440436078e+00 +4.473563727614030e+00\\ 
+2.483756843414148e+00 +5.294177671826510e-02\\ 
-9.510762444477711e+00 -1.971687237051063e+01\\ 
-1.133832690325659e+00 +2.653787041480606e+00\\ 
-1.060368418566762e+01 +6.758485552963814e+00\\ 
+3.939700942979768e+00 -2.236111110947745e+00\\ 
+2.957096578746405e+00 +9.539932671405341e+00\\ 
+3.676408303394857e+00 +5.835462885428624e+00\\ 
-5.006720900650824e+00 +1.567658340257674e+00\\ 
-2.089208132309644e+00 -1.826985761085373e+00\\ 
-1.894232206331808e+00 -4.362864040445677e+00\\ 
-5.890902842917961e+00 +1.211107598982879e+01\\ 
+6.461799127806617e+00 -3.194567464254237e+00\\ 
+7.158487876196523e+00 -3.675190788188074e+00\\ 
+9.903478597632825e+00 -2.290569393643980e+00\\ 
+1.299769064095299e+01 -5.983502740119582e-01\\ 
+3.169326619853780e+00 -4.229287744002639e-01\\ 
+1.023190041078756e+01 -1.125018573644285e+00\\ 
-5.212712581194517e+00 +1.330297171158781e+00\\ 
+2.403558270211437e-02 -4.230720396821240e-01\\ 
-2.886919893913015e+00 -1.532438790281374e+00\\ 
+6.008349958132903e+00 +1.909409628073335e+00\\ 
+7.010413838745428e+00 +1.840131494940047e+00\\ 
-3.009356312751177e-01 +3.900036732081771e+00\\ 
+5.076889693971881e+00 +5.679540147669629e+00\\ 
+2.010210853134992e+00 +2.859685724858856e-02\\ 
-4.847034780907561e+00 -8.213482428456962e-01\\ 
+9.498882751534932e-01 -2.753619658585846e+00\\ 
+1.338923866671512e+01 +9.967087534518606e-01\\ 
+2.034853030691258e+00 +7.961889295390259e+00\\ 
-1.047478955323358e+01 -7.394345893698981e+00\\ 
+2.258831510541792e-01 +9.547504361423867e+00\\ 
+2.726115262123602e+00 -7.033312110798582e-01\\ 
+1.775796556333364e-01 -1.036652028988014e+00\\ 
+1.020666646422638e+01 +8.383977961451341e+00\\ 
+3.955667325813400e+00 +1.098891780287816e+00\\ 
+2.274921460853924e+00 -9.168501572574626e-01\\ 
-4.714600876165403e+00 +2.288083594755344e+00\\ 
+9.268974925300266e-01 +8.800475479208083e+00\\ 
-5.426535626873702e+00 -1.226165105532149e+00\\ 
-6.475291913586450e+00 -6.803143445210333e+00\\ 
-1.129659104075281e+00 +1.160973024069694e+01\\ 
+6.174327418699582e+00 -4.444018748511366e+00\\ 
-1.648140991573202e+01 -1.381266115142339e+01\\ 
-1.509603282303143e+01 -1.557960450248281e+01\\ 
-6.313192981327625e+00 +1.215925059698380e+01\\ 
-1.202439528777902e+01 -9.987604790704660e+00\\ 
-2.331356824309141e+01 -2.753671496010030e+01\\ 
-5.214342915228070e+00 +1.807671120915415e+01\\ 
+1.471759931493174e+01 +5.330317713271240e+00\\ 
-3.649682350885613e+00 +3.430103662270303e+00\\ 
+2.011555141705632e+01 +4.943803343335931e+00\\ 
+3.945627956940284e+00 +9.602329256408185e+00\\ 
-6.232761863345371e-01 -4.055556356705578e+00\\ 
+1.093156700256975e+00 +3.416406104253136e+00\\ 
-1.044895206215937e+01 -7.158666057000504e+00\\ 
+5.621210303064589e+00 +8.830013518161175e-01\\ 
+7.667640202300791e+00 +1.200578878696328e+01\\ 
+2.632020143389119e+00 +6.696256898554845e+00\\ 
-1.217372012766640e+01 -6.915414139796303e+00\\ 
-2.688988402214878e+01 -2.769007962366408e+00\\ 
-1.575675196833687e+00 -5.411248703384006e+00\\ 
+9.414042587241045e-01 +5.944427577976776e-01\\ 
+1.380057978348254e+00 +2.678924880250724e+00\\ 
-4.335479084298460e+00 -2.193002621715243e+00\\ 
-1.084555832466610e+00 +2.403526719546514e+00\\ 
-1.780929405541337e+00 +2.545849567378846e+00\\ 
+1.047517749211677e+01 +1.056790059170980e+01\\ 
-1.095917491292456e+01 +7.083944432772407e-01\\ 
-3.380619973321758e+00 -1.282641946118741e+01\\ 
-1.191314688134409e+00 -3.456023242928217e+00\\ 
-6.636748483009463e+00 +1.273636004470438e+00\\ 
+8.649854950867313e+00 +1.743410734237274e+00\\ 
-1.034515344590872e+01 -9.602144216175843e+00\\ 
+1.392421765496518e+01 +8.087796800936766e-01\\ 
+1.634912381463113e+01 -3.983385691724026e+00\\ 
-2.734791368768335e+00 -1.524080208264984e+01\\ 
+7.079918703067650e+00 -1.523073674820969e+01\\ 
+7.621091561743593e+00 +3.600692853862213e+01\\ 
-6.188240835751896e+00 -2.400552493794257e+01\\ 
};

\end{axis}

\end{tikzpicture}